\documentclass[runningheads]{llncs}
\usepackage{graphicx}

\usepackage{tikz}
\usepackage{comment}
\usepackage{amsmath,amssymb} 
\usepackage{color}

\usepackage[accsupp]{axessibility}  

\usepackage{bm}
\usepackage{subcaption}
\usepackage[pagebackref,breaklinks,colorlinks=true,citecolor=black]{hyperref}
\newcommand{\rulesep}{\color{black} \unskip\ \vrule\ }

\newcommand{\OURS}{P2R-Net}

\begin{document}
\pagestyle{headings}
\mainmatter
\def\ECCVSubNumber{3217}  

\title{Pose2Room: Understanding 3D Scenes from Human Activities} 

\titlerunning{Pose2Room: Understanding 3D Scenes from Human Activities}
%
\author{Yinyu Nie\inst{1} \and
Angela Dai\inst{1} \and
Xiaoguang Han\inst{2} \and Matthias Nießner\inst{1}}
\authorrunning{Y. Nie et al.}
%
\institute{Technical University of Munich \and The Chinese University of Hong Kong, Shenzhen}
\maketitle

\begin{abstract}
With wearable IMU sensors, one can estimate human poses from wearable devices without requiring visual input~\cite{von2017sparse}.
In this work, we pose the question: Can we reason about object structure in real-world environments solely from human trajectory information?
Crucially, we observe that human motion and interactions tend to give strong information about the objects in a scene -- for instance a person sitting indicates the likely presence of a chair or sofa.
To this end, we propose \OURS{} to learn a probabilistic 3D model of the objects in a scene characterized by their class categories and oriented 3D bounding boxes, based on an input observed human trajectory in the environment.
\OURS{} models the probability distribution of object class as well as a deep Gaussian mixture model for object boxes, enabling sampling of multiple, diverse, likely modes of object configurations from an observed human trajectory.
In our experiments we show that \OURS{} can effectively learn multi-modal distributions of likely objects for human motions, and produce a variety of plausible object structures of the environment, even without any visual information. The results demonstrate that \OURS{} consistently outperforms the baselines on the PROX dataset and the VirtualHome platform.
\end{abstract}

\section{Introduction}
\label{sec:intro}

Understanding the structure of real-world 3D environments is fundamental to many computer vision tasks, with a well-studied history of research into 3D reconstruction from various visual input mediums, such as RGB video \cite{mur2015orb,engel2014lsd,runz2020frodo,qian2020associative3d}, RGB-D video \cite{newcombe2011kinectfusion,choi2015robust,niessner2013real,whelan2015elasticfusion,dai2017bundlefusion}, or single images \cite{huang2018holistic,Nie_2020_CVPR,popov2020corenet,Zhang_2021_CVPR,engelmann2021points,kuo2020mask2cad,kuo2021patch2cad,dahnert2021panoptic}.
Such approaches with active cameras have shown impressive capture of geometric structures leveraging strong visual signals.
We consider an unconventional view of passive 3D scene perception: in the case of a lack of any visual signal, we look to human pose data, which for instance can be estimated from wearable IMU sensors \cite{von2017sparse,glauser2019interactive,huang2018deep}, and ask ``What can we learn about a 3D environment from only human pose trajectory information?''.
This opens up new possibilities to explore information embedded in wearable devices (e.g., phones, fitness watches, etc.) towards understanding mapping, interactions, and content creation.

\begin{figure*}[!t]
	\centering
	\includegraphics[height=0.13\textheight]{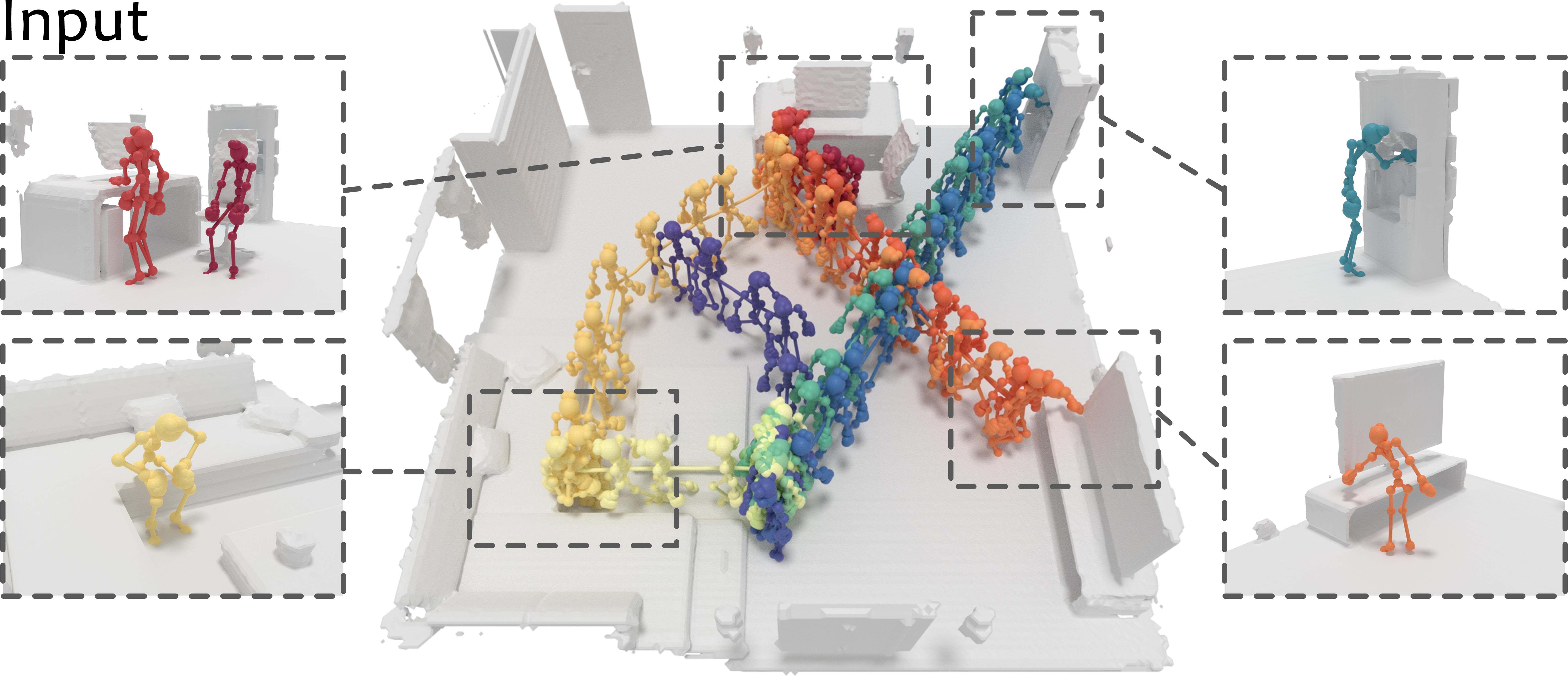}
	\rulesep
	\includegraphics[height=0.13\textheight]{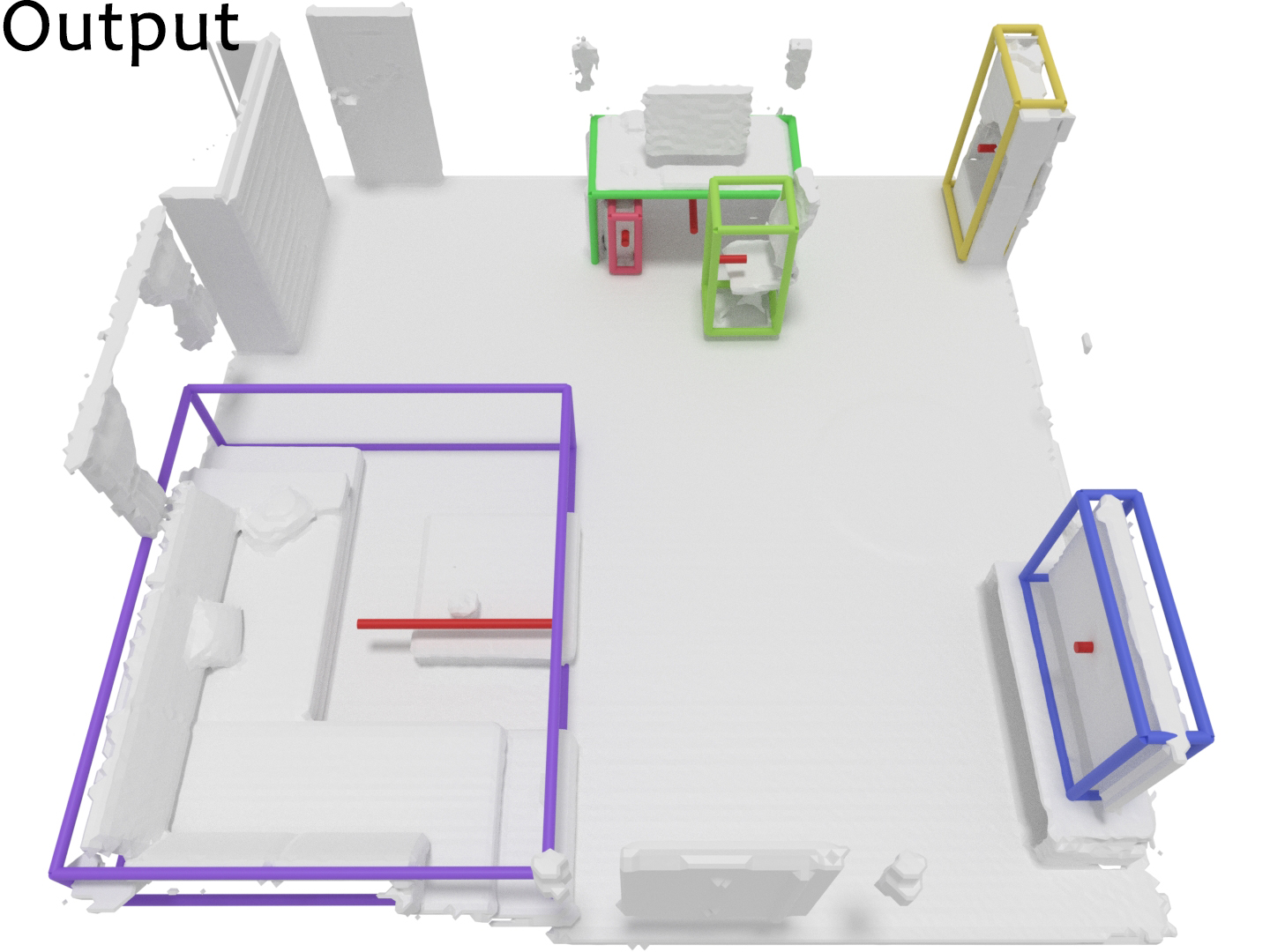}
	\includegraphics[height=0.13\textheight]{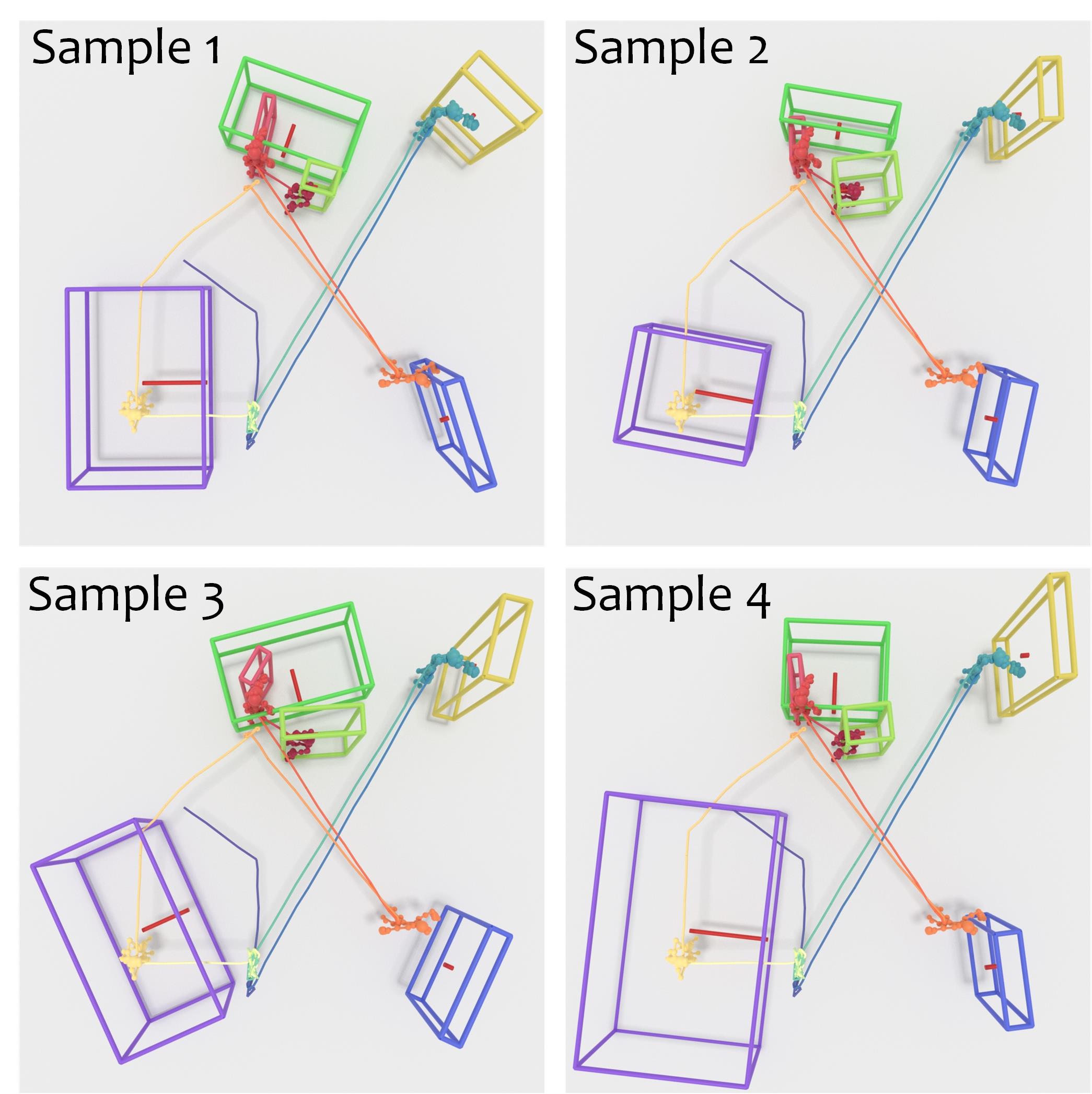}
	\caption{From an observed pose trajectory of a person performing daily activities in an indoor scene (left), we learn to estimate likely object configurations of the scene underlying these interactions, as set of object class labels and oriented 3D bounding boxes (middle). By sampling from our probabilistic decoder, we synthesize multiple plausible object arrangements (right). (Scene geometry is shown only for visualization.)
	}
	\label{fig:teaser}
\end{figure*}

In particular, we observe that human movement in a 3D environment often interacts both passively and actively with objects in the environment, giving strong cues about likely objects and their locations.
For instance, walking around a room indicates where empty floor space is available, a sitting motion indicates high likelihood of a chair or sofa to support the sitting pose, and a single outstretched arm suggests picking up/putting down an object to furniture that supports the object.
We thus propose to address a new scene estimation task: from only a sequence observation of 3D human poses, to estimate the object arrangement in the scene of the objects the person has interacted with, as a set of object class categories and 3D oriented bounding boxes (see Fig.~\ref{fig:teaser}).

As there are inherent ambiguities that lie in 3D object localization from only a human pose trajectory in the scene, we propose \OURS{} to learn a probabilistic model of the most likely modes of object configurations in the scene.
From the sequence of poses, \OURS{} leverages the pose joint locations to vote for potential object centers that participate in the observed pose interactions.
We then introduce a probabilistic decoder that learns a Gaussian mixture model for object box parameters, from which we can sample multiple diverse hypotheses of object arrangements. To enable massive training, we introduce a large-scale dataset with VirtualHome~\cite{puig2018virtualhome} platform to learn object configurations from human motions. Experiments on VirtualHome and the real dataset PROX~\cite{hassan2019resolving,yi2022mover} demonstrate our superiority against the baseline methods.

In summary, we present the following contributions:
\begin{itemize}
	\item We propose a new perspective on 3D scene understanding by studying estimation of 3D object configurations from solely observing 3D human pose sequences of interactions in an environment, without any visual input, and predicting the object class categories and 3D oriented bounding boxes of the interacted objects in the scene.
	\item To address this task, we introduce a new, end-to-end, learned probabilistic model that estimates probability distributions for the object class categories and bounding box parameters.
	
	\item We demonstrate that our model captures complex, multi-modal distributions of likely object configurations, which can be sampled to produce diverse hypotheses that have accurate coverage over the ground truth object arrangement. Experiments also demonstrates the superiority of our method against the baselines in terms of accuracy and diversity.
\end{itemize}

\section{Related Work}

\subsubsection{Predicting Human Interactions in Scenes.}
Capturing and modeling interactions between human and scenes has seen impressive progress in recent years, following significant advances in 3D reconstruction and 3D deep learning.
From a visual observation of a scene, interactions and human-object relations are estimated.
Several methods have been proposed for understanding the relations between scene and human poses via object functionality prediction \cite{grabner2011makes,zhu2015understanding,pieropan2013functional,hu2016learning} and affordance analysis \cite{gupta20113d,savva2016pigraphs,sawatzky2017weakly,wang2019geometric,deng20213d,ruiz2018can}.

By parsing the physics and semantics in human interactions, further works have been proposed towards synthesizing static human poses or human body models into 3D scenes \cite{grabner2011makes,gupta20113d,savva2014scenegrok,kim2014shape2pose,savva2016pigraphs,hassan2019resolving,zhang2020place,zhang2020generating,hassan2021populating}. 
These works focus on how to place human avatars into a 3D scene with semantic and physical plausibility (e.g., support or occlusion constraints). 
Various approaches have additionally explored synthesizing dynamic motions for a given scene geometry. 
Early methods retrieve and integrate existing avatar motions from database to make them compatible with scene geometry \cite{lee2002interactive,agrawal2016task,kapadia2016precision,lee2006motion,shum2008interaction}. 
Given a goal pose or a task, more works learn to search for a possible motion path and estimate plausible contact motions \cite{starke2019neural,chao2019learning,corona2020context,merel2020catch,wang2021synthesizing,hassan2021stochastic}. 
These methods explore human-scene interaction understanding by estimating object functionalities or human interactions as poses in a given 3D scene environment.
In contrast, we take a converse perspective, and aim to estimate the 3D scene arrangement from human pose trajectory observations.

\noindent\textbf{Scene Understanding with Human Priors.}
As many environments, particularly indoor scenes, have been designed for people's daily usage, human behavioral priors can be leveraged to additionally reason about 2D or 3D scene observations.
Various methods have been proposed to leverage human context as extra signal towards holistic perception to improve  performance in scene understanding tasks such as semantic segmentation \cite{delaitre2012scene}, layout detection from images \cite{fouhey2012people,shoaib2014estimating}, 3D object labeling \cite{jiang2013hallucinated}, 3D object detection and segmentation \cite{wei2016modeling}, and 3D reconstruction \cite{fowler2017towards,fowler2018human,yi2022mover}.

Additionally, several methods learn joint distributions of human interactions with 3D scenes or RGB video that can be leveraged to re-synthesize the observed scene as an arranged set of synthetic, labeled CAD models \cite{jiang2012learning,fisher2015activity,jiang2015modeling,savva2016pigraphs,monszpart2019imapper}.
Recently, HPS~\cite{guzov2021human} proposed to simultaneously estimate pose trajectory and scene reconstruction from wearable visual and inertial sensors on a person.
We also aim to understand 3D scenes as arrangements of objects, but do not require any labeled interactions nor consider any visual (RGB, RGB-D, etc.) information as input.
The recent approach of Mura et al.~\cite{mura2021walk2map} poses the task of floor plan estimation from 2D human walk trajectories, and proposes to predict occupancy-based floor plans that indicate structure and object footprints, but without object instance distinction and employs a fully-deterministic prediction.
To the best of our knowledge, we introduce the first method to learn 3D object arrangement distributions from only human pose trajectories, without any visual input.

\noindent\textbf{Pose Tracking with IMUs.}
Our method takes the input of human pose trajectories, which is built on the success of motion tracking techniques. 
Seminal work on pose estimation from wearable sensors have demonstrated effective pose estimation from wearable sensors, such as optical markers \cite{chai2005performance,hassan2021stochastic} or IMUs \cite{liu2011realtime,von2017sparse,huang2018deep,kaufmann2021pose,glauser2019interactive}.
Our work is motivated by the capability of reliably estimating human pose from these sensor setups without visual data, from which we aim to learn human-object interaction priors to estimate scene object configurations. 

\section{Method}

From only a human pose trajectory as input, we aim to estimate a distribution of likely object configurations, from which we can sample plausible hypotheses of objects in the scene as sets of class category labels and oriented 3D bounding boxes.
We observe that most human interactions in an environment are targeted towards specific objects, and that general motion behavior is often influenced by the object arrangement in the scene.
We thus aim to discover potential objects that each pose may be interacting with.

\begin{figure*}[!t]
	\centering
	\includegraphics[width=1\linewidth]{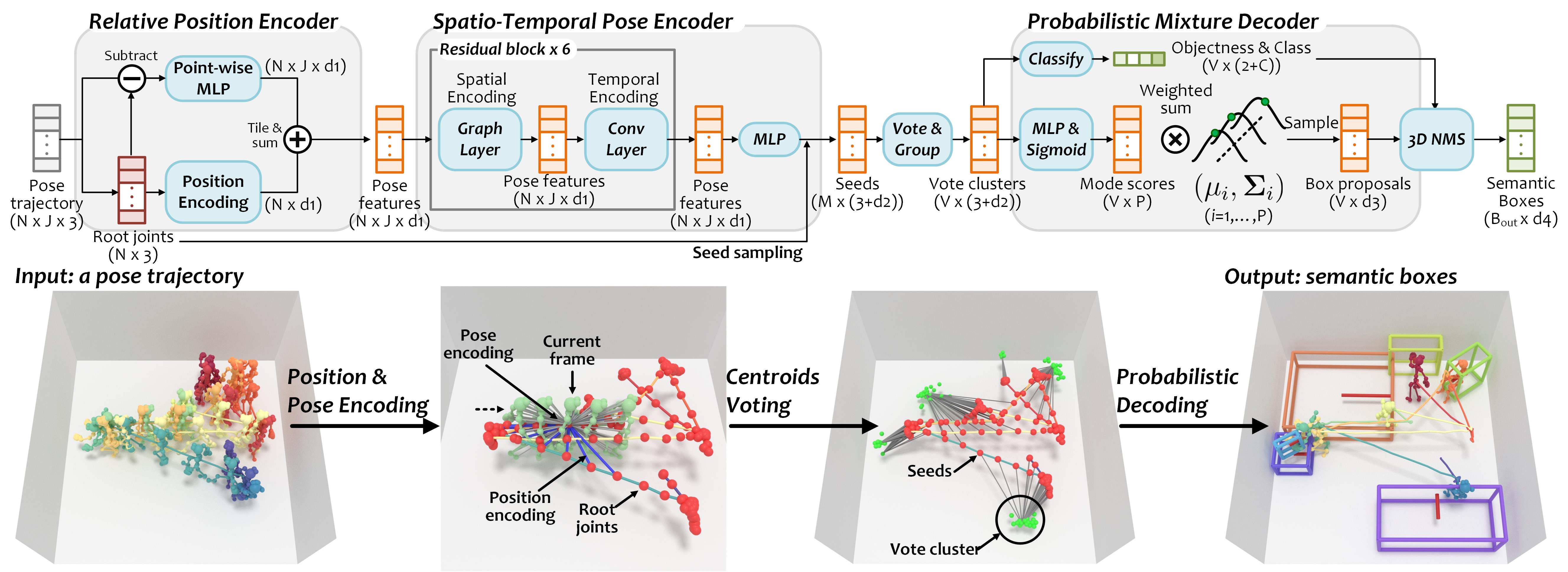}
	\caption{Overview of \OURS{}.
		Given a pose trajectory with $N$ frames and $J$ joints, a position encoder decouples each skeleton frame into a relative position encoding (from its root joint as the hip centroid) and a position-agnostic pose.
		After combining them, a pose encoder learns local pose features from both body joints per skeleton (spatial encoding) and their changes in consecutive frames (temporal encoding).
		Root joints as seeds are then used to vote for the center of a nearby object that each pose is potentially interacting with.
		A probabilistic mixture network learns likely object box distributions, from which object class labels and oriented 3D boxes can be sampled.}
	\label{fig:overview}
\end{figure*}

We first extract meaningful features from the human pose sequence with a \textit{position encoder} to disentangle each frame into a relative position encoding and a position-agnostic pose, as well as a \textit{pose encoder} to learn the local spatio-temporal feature for each pose in consecutive frames.
We then leverage these features to vote for a potential interacting object for each pose.
From these votes, we learn a \textit{probabilistic mixture decoder} to propose box proposals for each object, characterizing likely modes for objectness, class label, and box parameters.
An illustration of our approach is shown in Fig.~\ref{fig:overview}.

\subsection{Relative Position Encoding}
We consider an input pose trajectory with $N$ frames and $J$ joints as the sequence of 3D locations $\bm{T} \in \mathbb{R}^{N\times J\times 3}$. We also denote the root joint of each pose by $\bm{r} \in \mathbb{R}^{N\times 3}$, where the root joint of a pose is the centroid of the joints corresponding to the body hip (for the skeleton configuration, we refer to the supplemental).
To learn informative pose features, we first disentangle for each frame the absolute pose joint coordinates into a relative position encoding $\bm{Q} \in \mathbb{R}^{N\times d_{1}}$ and a position-agnostic pose feature $\bm{P} \in \mathbb{R}^{N\times J \times d_{1}}$, which are formulated as:
\begin{equation}
	\label{eqn:01}
	\begin{aligned}
		\bm{Q} &=\text{Pool}\left[f_{1}\left(\mathcal{N}\left(\bm{r}\right)-\bm{r}\right)\right],\\
		\bm{P} &=f_{2}\left(\bm{T}-\bm{r}\right),\ \ \mathcal{N}\left(\bm{r}\right)\in \mathbb{R}^{N\times k\times 3},
	\end{aligned}
\end{equation}
where $f_{1}(*)$, $f_{2}(*)$ are point-wise MLP layers. $\mathcal{N}\left(\bm{r}\right)$ is the set of $k$ temporal neighbors to each root joint in $\bm{r}$, and $\text{Pool}(*)$ denotes neighbor-wise average pooling.
By broadcast summation, we output $\bm{P}^{r}=\bm{P}+\bm{Q}$ for further spatio-temporal pose encoding.
Understanding relative positions rather than absolute positions helps to provide more generalizable features to understand common pose motion in various object interactions, as these human-object interactions typically occur locally in a scene.

\subsection{Spatio-Temporal Pose Encoding}
The encoding $\bm{P}^{r}$ provides signal for the relative pose trajectory of a person.
We then further encode these features to capture the joint movement to understand local human-object interactions.
That is, from $\bm{P}^{r}$, we learn joint movement in spatio-temporal domain: (1) in the spatial domain, we learn from intra-skeleton joints to capture per-frame pose features; (2) in the temporal domain, we learn from inter-frame relations to perceive each joint's movement.

\begin{figure}[!h]
	\centering
	\includegraphics[width=0.9\textwidth]{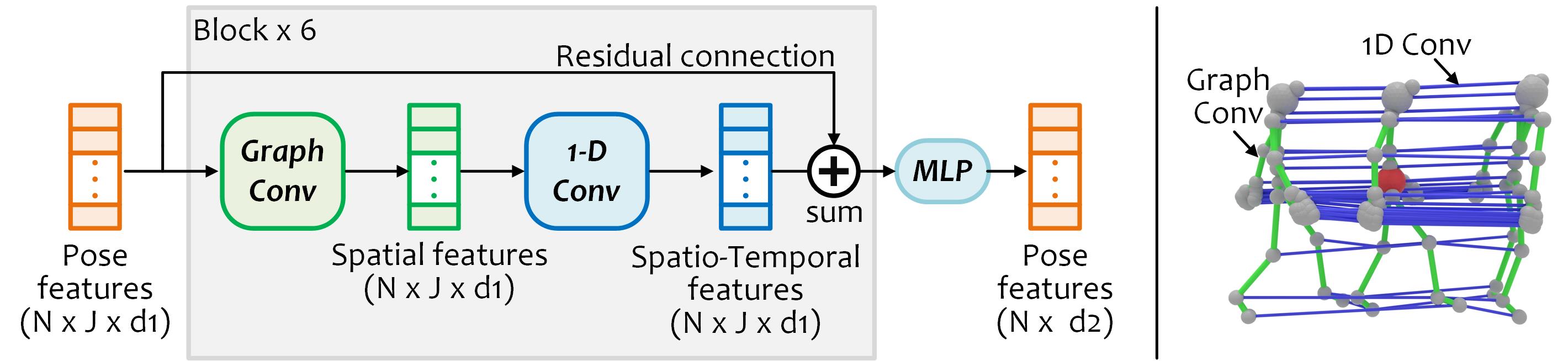}
	\caption{Pose encoding with spatio-temporal convolutions.}
	\label{fig:pose_encoding}
\end{figure}

Inspired by \cite{yan2018spatial} in 2D pose recognition, we first use a graph convolution layer to learn intra-skeleton joint features.
Edges in the graph convolution are constructed following the skeleton bones, which encodes skeleton-wise spatial information.
For each joint, we then use a 1-D convolution layer to capture temporal features from its inter-frame neighbors.
A graph layer and an 1-D convolution layer are linked into a block with a residual connection to process the input $\bm{P}^{r}$ (see Fig.~\ref{fig:pose_encoding}).
By stacking six blocks, we obtain a deeper spatio-temporal pose encoder with a wider receptive field in temporal domain, enabling reasoning over more temporal neighbors for object box estimation.
Finally, we adopt an MLP to process all joints per skeleton to obtain pose features $\bm{P}^{st}\in \mathbb{R}^{N\times d_{2}}$.

\subsection{Locality-Sensitive Voting}
\label{sec:loc_sen_voting}
With pose features $\bm{P}^{st}$, we then learn to vote for all the objects a person could have interacted with in a trajectory (see Fig.~\ref{fig:overview}).
For each pose frame, we predict the center of an object it potentially interacts with.
Since we do not know when interactions begin or end, each pose votes for a potential object interaction.
As human motion in a scene will tend to active interaction with an object or movement to an object, we aim to learn these patterns by encouraging votes for objects close to the respective pose, encouraging locality-based consideration.

For each pose feature $p^{st} \in \bm{P}^{st}$, we use its root joint $r \in \bm{r}$ as a seed location, and vote for an object center by learning the displacement from the seed:
\begin{equation}
	\label{eqn:02}
	\begin{aligned}
		\bm{v} &=\bm{r}_{s} + f_{3}\left(\bm{P}_{s}^{st}\right),\ \ \bm{r}_{s},\bm{v}\in\mathbb{R}^{M\times 3},\\
		\bm{P}^{v} &= \bm{P}_{s}^{st}+f_{4}\left(\bm{P}_{s}^{st}\right),\ \ \bm{P}_{s}^{st},\bm{P}^{v}\in \mathbb{R}^{M\times d_{2}},
	\end{aligned}
\end{equation}
where $\bm{r}_{s}$ are the evenly sampled $M$ seeds from $\bm{r}$; $\bm{P}_{s}^{st}$ are the corresponding pose features of $\bm{r}_{s}$; $f_{3}, f_{4}$ are MLP layers; $\bm{v},\bm{P}^{v}$ denote the vote coordinates and features learned from $\bm{P}_{s}^{st}$. We evenly sample seeds $\bm{r}_{s}$ from $\bm{r}$ to make them cover the whole trajectory and adaptive to sequences with different length.

Since there are several objects in a scene, for each seed in $\bm{r}_{s}$, we vote for the center to the nearest one (see Fig.~\ref{fig:voting}).
The nearest object is both likely to participate in a nearby interaction, and affect motion around the object if not directly participating in an interaction.
This strategy helps to reduce the ambiguities in scene object configuration estimation from a pose trajectory by capturing both direct and indirect effects of object location on pose movement.

For the seeds which vote for the same object, we group their votes to a cluster following \cite{Qi_2019_ICCV}.
This outputs cluster centers $\bm{v}^{c}\in \mathbb{R}^{V\times 3}$ with aggregated cluster feature $\bm{P}^{c}\in \mathbb{R}^{V\times d_{2}}$ where $V$ denotes the number of vote clusters.
We then use the $\bm{P}^{c}$ to decode to distributions that characterize semantic 3D boxes, which is described in Section~\ref{sec:prob_mix_decoder}.
For poses whose root joint is not close to any object during training (beyond a distance threshold $t_d$), we consider them to have little connection with the objects, and do not train them to vote for any object.

\begin{figure}
	\centering
	\includegraphics[width=0.4\textwidth]{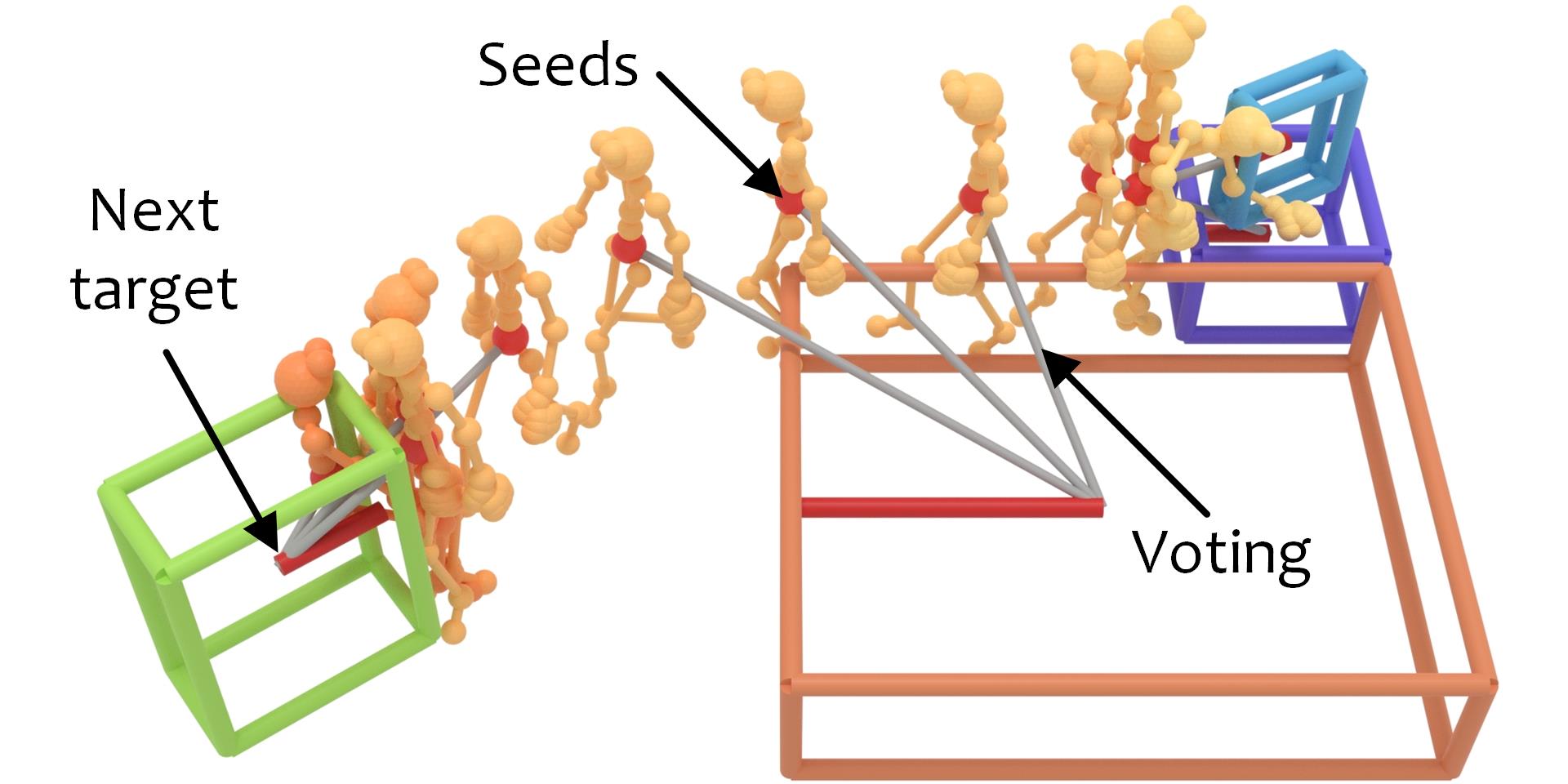}
	\caption{Voting to objects that potentially influence the motion trajectory in approaching the target.
	}
	\label{fig:voting}
\end{figure}

\subsection{Probabilistic Mixture Decoder}
\label{sec:prob_mix_decoder}
We decode vote clusters $(\bm{v}^{c},\bm{P}^{c})$ to propose oriented 3D bounding boxes for each object, along with their class label and objectness score.
Each box is represented by a 3D center $c$, 3D size $s$ and 1D orientation $\theta$, where we represent the size by $\text{log}(s)$ and orientation by $(\text{sin}(\theta),\text{cos}(\theta))$ for regression, similar to \cite{yin2021center}.
Since the nature of our task is inherently ambiguous (e.g., it may be unclear from observing a person sit if they are sitting on a chair or a sofa, or the size of the sofa),  we propose to learn a probabilistic mixture decoder to predict the box centroid, size and orientation with multiple modes, from a vote cluster $v^{c}\in \bm{v}^{c},P^{c}\in \bm{P}^{c}$:

\begin{equation}
	\label{eqn:03}
	\begin{aligned}
		&y_{\tau} = \textstyle\sum_{k=1}^{P} f_{\tau}^{k}\left(P^{c}\right)\cdot y^{k}_{\tau},\ \ \ \ \tau\in \left\{c,s,\theta\right\},\\
		&y^{*}_{c} = v^{c} + y_{c}, \ \ \ \ y_{\tau}^{k}\sim \mathcal{N}(\mu_{\tau}^{k},\,\Sigma_{\tau}^{k}),\ \ \ \  y_{\tau},y^{k}_{\tau}\in\mathbb{R}^{d_{\tau}},
	\end{aligned}
\end{equation}
where $\tau\in\{c,s,\theta\}$ denote the regression targets for center, size, and orientation; $\mathcal{N}(\mu_{\tau}^{k},\,\Sigma_{\tau}^{k})$ is the learned multivariate Gaussian distribution of the $k$-th mode for $\tau$, where $y^{k}_{\tau}$ is sampled from; $P$ is the number of Gaussian distributions (i.e., modes); $f_{\tau}^{k}(*)\in[0,1]$ is the learned score for the $k$-th mode; $y_{\tau}$ is the weighted sum of the samples from all modes, which is the prediction of the center/size/orientation; and  $d_{\tau}$ is their output dimension ($d_{c}$=3, $d_{s}$=3, $d_{\theta}$=2).
Note that the box center $y^{*}_{c}$ is obtained by regressing the offset $y_{c}$ from cluster center $v^{c}$.
We predict the proposal objectness and the probability distribution for class category directly from $P^{c}$, using an MLP.

\noindent\textbf{Multi-modal Prediction.} In Eq.~\ref{eqn:03}, the learnable parameters are $f_{\tau}(*)$ and $(\mu_{\tau},\,\Sigma_{\tau})$.
$f_{\tau}(*)$ is realized with an MLP followed by a sigmoid function, and $(\mu_{\tau},\,\Sigma_{\tau})$ are the learned embeddings shared among all samples.
During training, we sample $y_{\tau}^{k}$ from each mode $\mathcal{N}(\mu_{\tau}^{k},\,\Sigma_{\tau}^{k})$ and predict $y_{\tau}$ using Eq.~\ref{eqn:03}.
To generate diverse and plausible hypotheses during inference, we not only sample $y_{\tau}^{k}$, but also sample various different modes by randomly disregarding mixture elements based on their probabilities $f_{\tau}(*)$.
Then we obtain $y_{\tau}$ as follows:
\begin{equation}
	\label{eqn:04}
	y_{\tau} = \textstyle\sum_{k=1}^{P} I_{\tau}^{k}\cdot y^{k}_{\tau},\ \ I_{\tau}^{k}\sim \text{Bern}(f_{\tau}^{k}),\ \ y_{\tau}^{k}\sim \mathcal{N}(\mu_{\tau}^{k},\,\Sigma_{\tau}^{k})\\,
\end{equation}
where $I_{\tau}^{k}$ is sampled from Bernoulli distribution with probability of $f_{\tau}^{k}(*)$.
We also sample the object classes by the predicted classification probabilities, and discard proposed object boxes with low objectness ($\leq t_o$) after 3D NMS.

We can then generate $N_h$ hypotheses in a scene; each hypothesis is an average of $N_s$ samples of $y_{\tau}$, which empirically strikes a good balance between diversity and accuracy of the set of hypotheses.
To obtain the maximum likelihood prediction, we use $f_{\tau}^{k}(*)$ and the mean value $\mu_{\tau}^{k}$ instead of $I_{\tau}^{k}$ and $y_{\tau}^{k}$ to estimate the boxes with Eq.~\ref{eqn:03}.

\subsection{Loss Function}
The loss consists of classification losses for objectness $\mathcal{L}_{obj}$ and class label $\mathcal{L}_{cls}$, and regression losses for votes $\mathcal{L}_{v}$, box center $\mathcal{L}_{c}$, size $\mathcal{L}_{s}$ and orientation $\mathcal{L}_{\theta}$.

\noindent\textbf{Classification Losses.} Similar to \cite{Qi_2019_ICCV}, $\mathcal{L}_{obj}$ and $\mathcal{L}_{cls}$ are supervised by cross entropy losses, wherein the objectness score is used to classify if a vote cluster center is close to ($\leq 0.3$ m, positive) or far from ($\geq 0.6$ m, negative) the ground truth. Proposals from the clusters with positive objectness are further supervised with box regression losses.

\noindent\textbf{Regression Losses.} We supervise all the predicted votes, box centers, sizes and orientations with a Huber loss. For poses that are located within $d_{p}$ to objects ($d_{p}=1$ m), we use the closest object center to supervise their vote. Votes from those poses that are far from all objects are not considered. For center predictions, we use their nearest ground-truth center to calculate $\mathcal{L}_{c}$. Since box sizes and orientations are predicted from vote clusters, we use the counterpart from the ground-truth box that is nearest to the vote cluster for supervision.
Then the final loss function is
$\mathcal{L}=\textstyle \sum_{\tau} \lambda_{\tau}\mathcal{L}_{\tau}$, where $\mathcal{L}_{\tau} \in \{\mathcal{L}_{obj},\mathcal{L}_{cls},\mathcal{L}_{v},\mathcal{L}_{c},\mathcal{L}_{s},\mathcal{L}_{\theta}\}$ and $\{\lambda_{\tau}\}$ are constant weights that balance the losses.

\section{Experiment Setup}
\noindent\textbf{Datasets.}
To the best of our knowledge, existing 3D human pose trajectory datasets are either with very few sequences ($\leq$ 102)  \cite{savva2016pigraphs,hassan2019resolving,zhang2020generating,cao2020long,yi2022mover}, or without instance annotations \cite{hassan2019resolving,cao2020long,zhang2020generating}, or focused on single objects \cite{hassan2021stochastic}. To this end, we introduce a new large-scale dataset using the simulation platform \textbf{VirtualHome}~\cite{puig2018virtualhome} for the task of estimating multiple scene objects from a human pose trajectory observation.
To demonstrate our applicability to real data, we also evaluate on real human motions from the \textbf{PROX} dataset~\cite{hassan2019resolving,yi2022mover}.

We construct our dataset on VirtualHome~\cite{puig2018virtualhome}, which is built on the Unity3D game engine.
It consists of 29 rooms, with each room containing 88 objects on average; each object is annotated with available interaction types.
VirtualHome allows customization of action scripts and control over humanoid agents to execute a series of complex interactive tasks.
We refer readers to \cite{puig2018virtualhome} for the details of the scene and action types.
In our work, we focus on the interactable objects under 17 common class categories.
In each room, we select up to 10 random static objects to define the scene, and script the agent to interact with each of the objects in a sequential fashion.
For each object, we also select a random interaction type associated with the object class category.
Then we randomly sample 13,913 different sequences with corresponding object boxes to construct the dataset.
During training, we also randomly flip, rotate and translate the scenes and poses for data augmentation. For additional detail about data generation, we refer to the supplemental. For the PROX dataset \cite{hassan2019resolving}, we use its human motions with the 3D instance boxes labeled by \cite{yi2022mover}, which has 46 motion sequences interacting with four object categories (i.e., bed, chair, sofa, table) in 10 rooms. For more details, we refer readers to \cite{hassan2019resolving,yi2022mover}.

\noindent\textbf{Implementation.}
We train \OURS{} end-to-end from scratch with the batch size at 32 on 4 NVIDIA 2080 Ti GPUs for 180 epochs, where Adam is used as the optimizer.
The initial learning rate is at 1e-3 in the first 80 epochs, which is decayed by $0.1\times$ every 40 epochs after that.
The losses are weighted by $\lambda_{obj}$=5, $\lambda_{cls}$=1, \{$\lambda_{v}$,$\lambda_{c}$,$\lambda_{s}$,$\lambda_{\theta}$\}=10 to balance the loss values.
During training, we use pose distance threshold $t_{d}$=1 m.
At inference time, we output box predictions after 3D NMS with an IoU threshold of 0.1.
We use an objectness threshold of $t_{o}$=0.5.
For the layer and data specifications, we refer to the supplemental.

\noindent\textbf{Evaluation.}
We evaluate our task both on our dataset and on PROX. For our dataset, we consider two types of evaluation splits: a sequence-level split $\mathcal{S}_1$ across different interaction sequences, and room-level split $\mathcal{S}_2$ across different rooms as well as interaction sequences.
Note that sequences are trained with and evaluated against only the objects that are interacted with during the input observation, resulting in different variants of each room under different interaction sequences.
For $\mathcal{S}_{1}$, the train/test split ratio is 4:1 over the generated sequences.
$\mathcal{S}_{2}$ is a more challenging setup, with 27 train rooms and 2 test rooms, resulting in 13K/1K sequences.
Since the task is inherently ambiguous, and only a single ground truth configuration of each room is available, we evaluate multi-modal predictions by several metrics:
\emph{mAP@0.5} evaluates the mean average precision with the IoU threshold at 0.5 of the maximum likelihood prediction; \emph{MMD} evaluates the Minimal Matching Distance \cite{achlioptas2018learning} of the best matching prediction with the ground truth out of 10 sampled hypotheses to measure their quality; \emph{TMD} evaluates the Total Mutual Diversity~\cite{wu2020multimodal} to measure the diversity of the 10 hypotheses.
We provide additional detail about MMD and TMD in the supplemental.
For PROX dataset, we split the train/test set by 8:1 considering the very limited number of sequences (46) and use \emph{mAP@0.5} for detection evaluation.

\section{Results and Analysis}
\label{sec:results}

We evaluate our approach on the task of scene object configuration estimation from a pose trajectory observation, in comparison with baselines constructed from state-of-the-art 3D detection and pose understanding methods, as well as an ablation analysis of our multi-modal prediction.

\begin{figure*}[!ht]
	\centering
	\includegraphics[width=\textwidth]
	{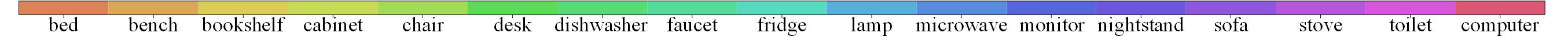}\\
	\begin{subfigure}[t]{0.159\textwidth}
		\includegraphics[width=\textwidth]
		{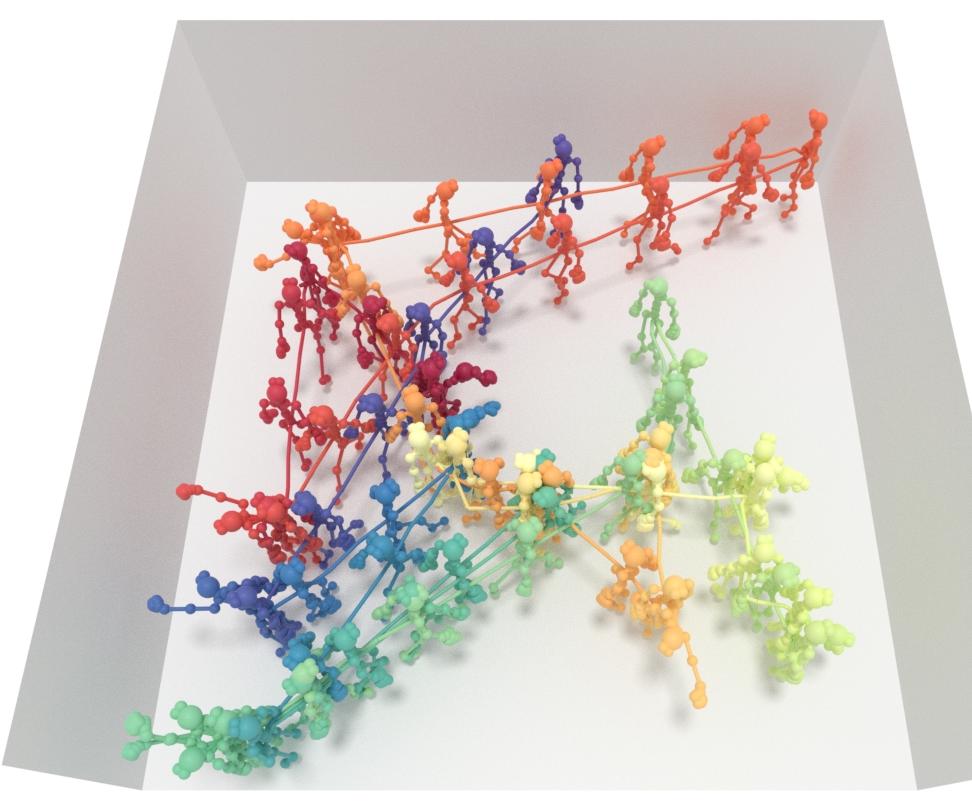}
		\includegraphics[width=\textwidth]
		{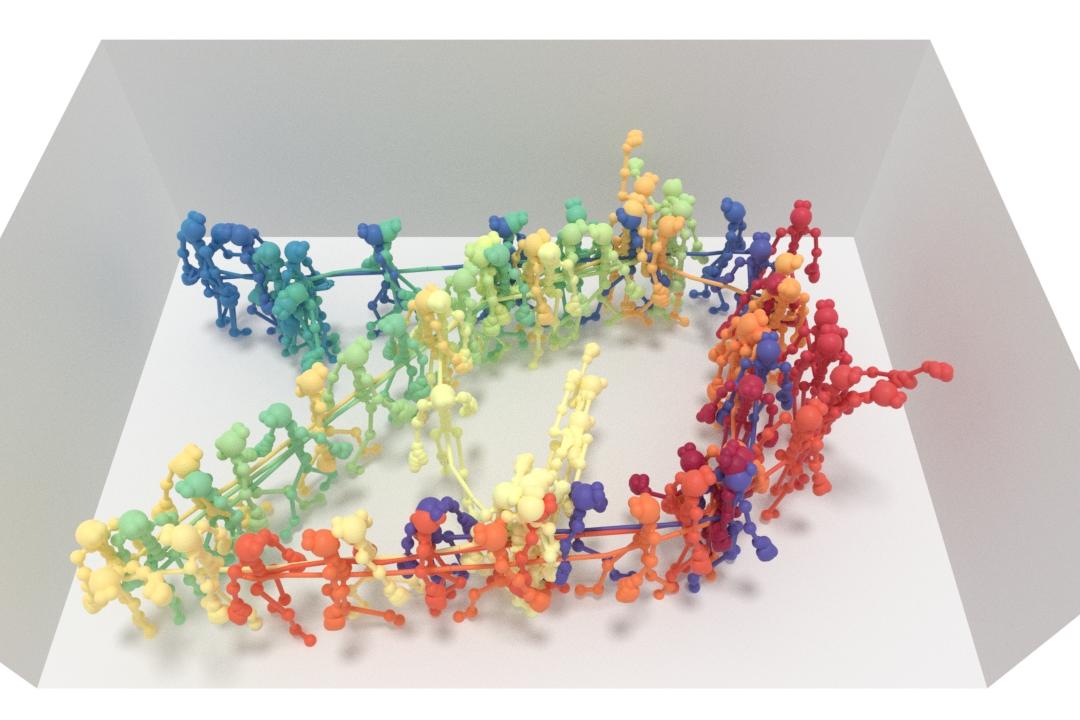}
		\includegraphics[width=\textwidth]
		{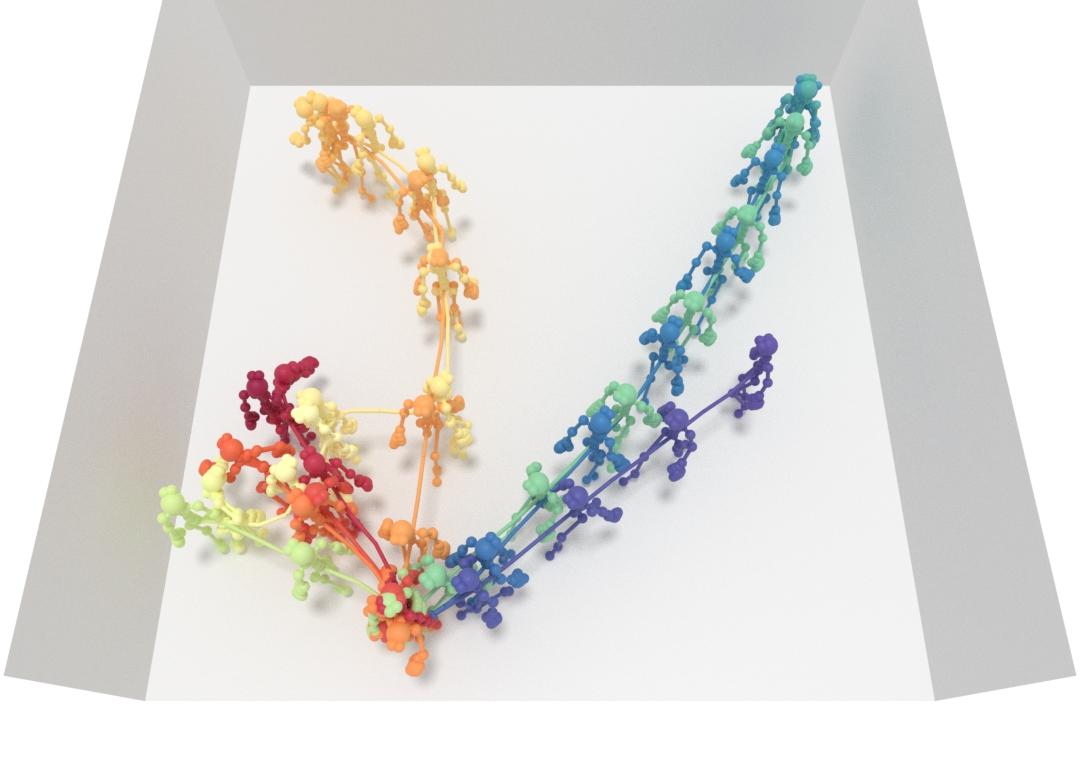}
		\includegraphics[width=\textwidth]
		{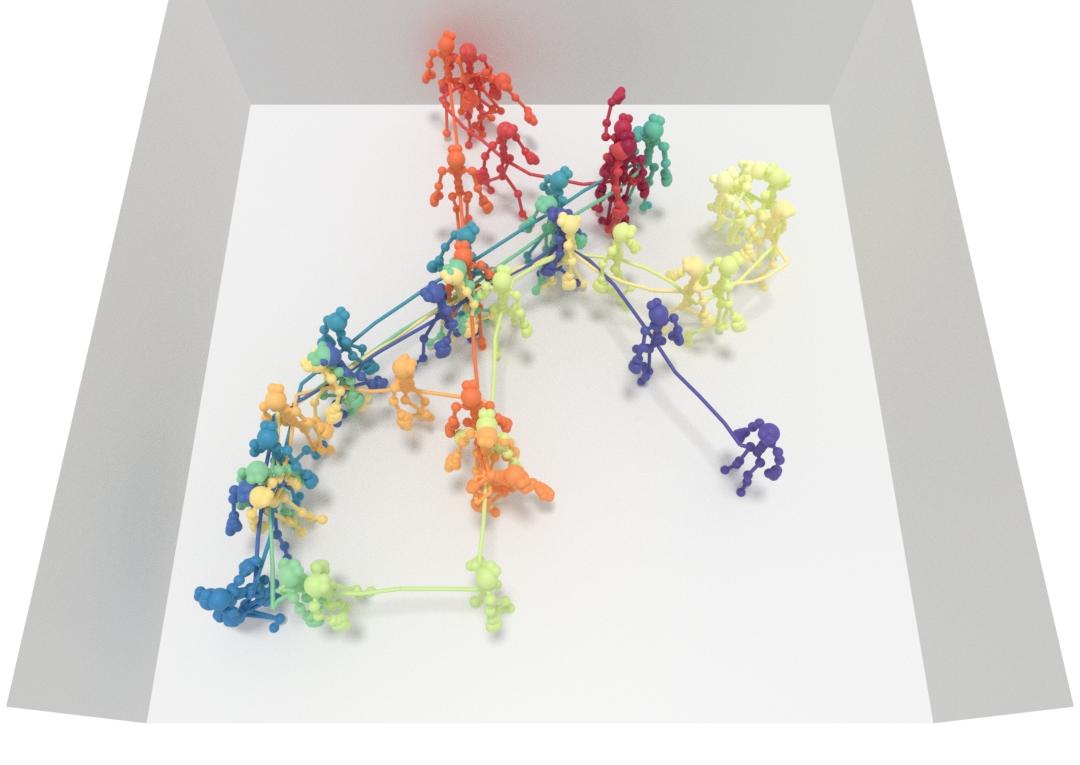}
		\caption{Input}
	\end{subfigure}
	\begin{subfigure}[t]{0.159\textwidth}
		\includegraphics[width=\textwidth]
		{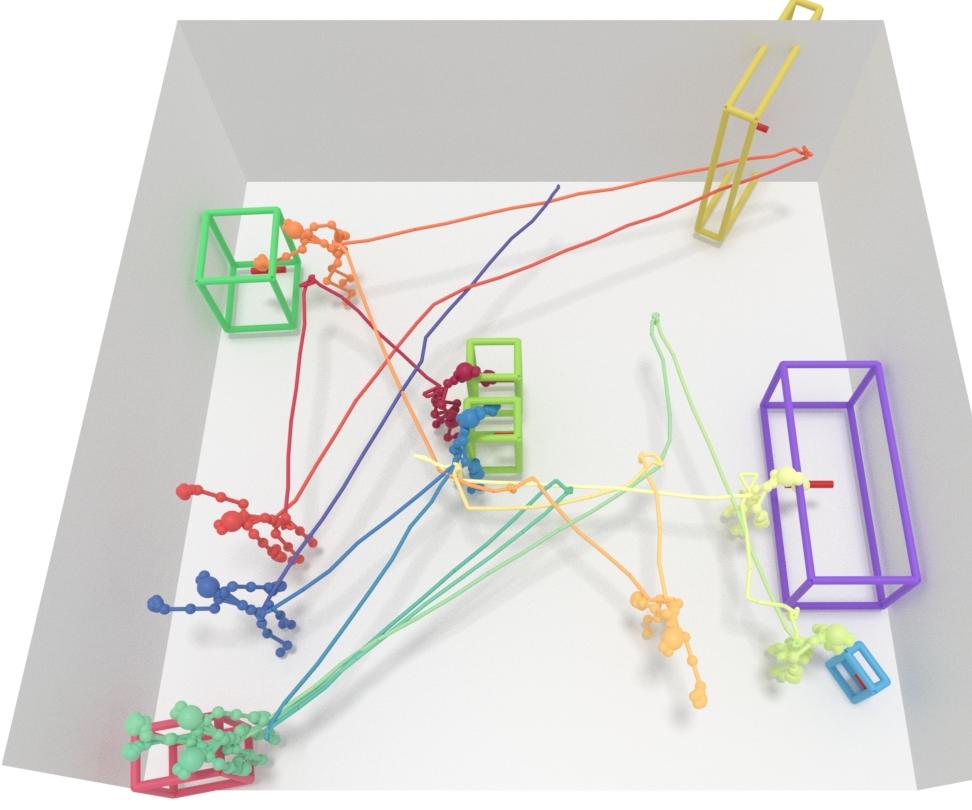}
		\includegraphics[width=\textwidth]
		{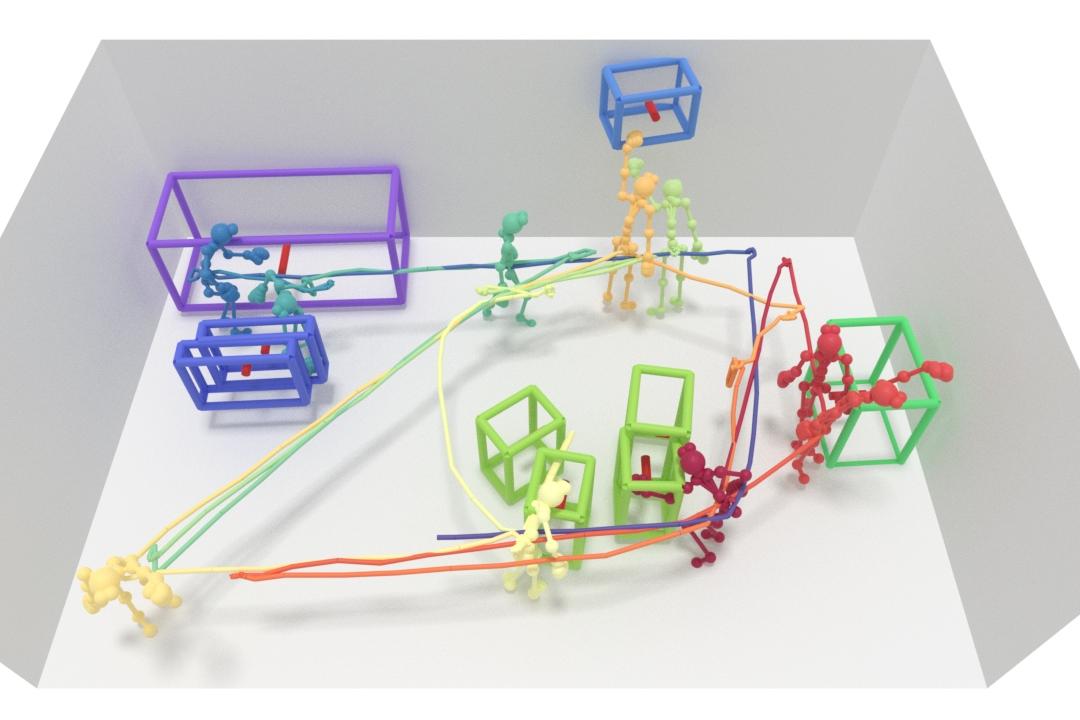}
		\includegraphics[width=\textwidth]
		{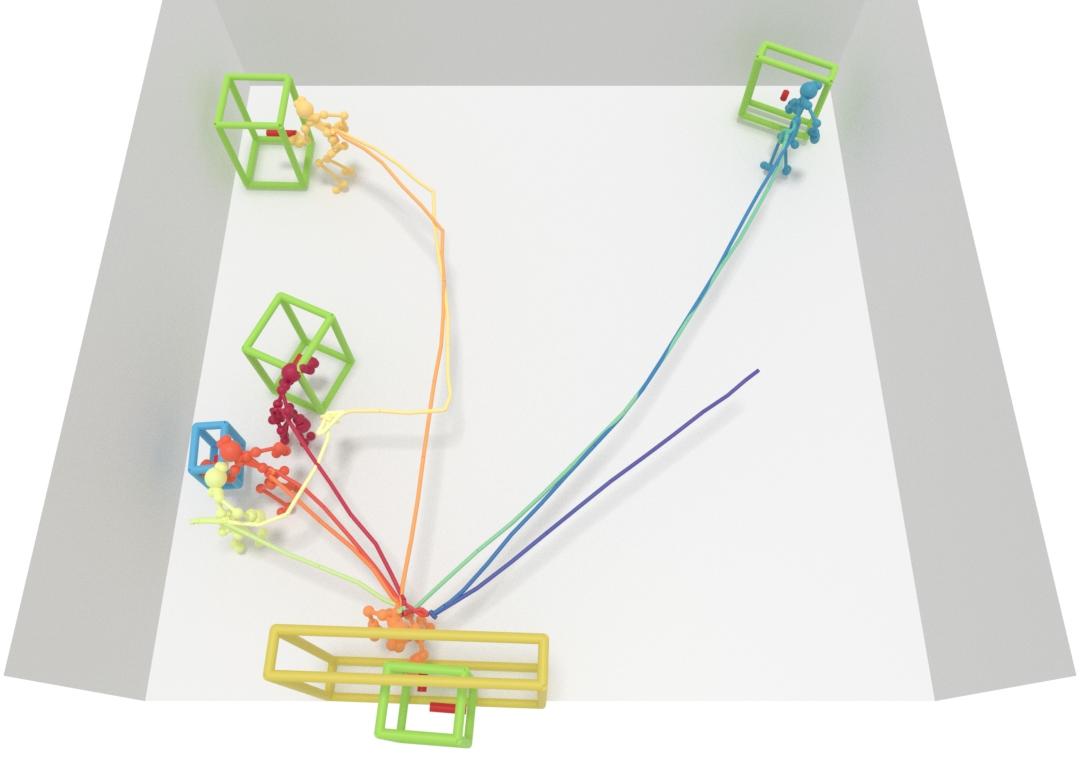}
		\includegraphics[width=\textwidth]
		{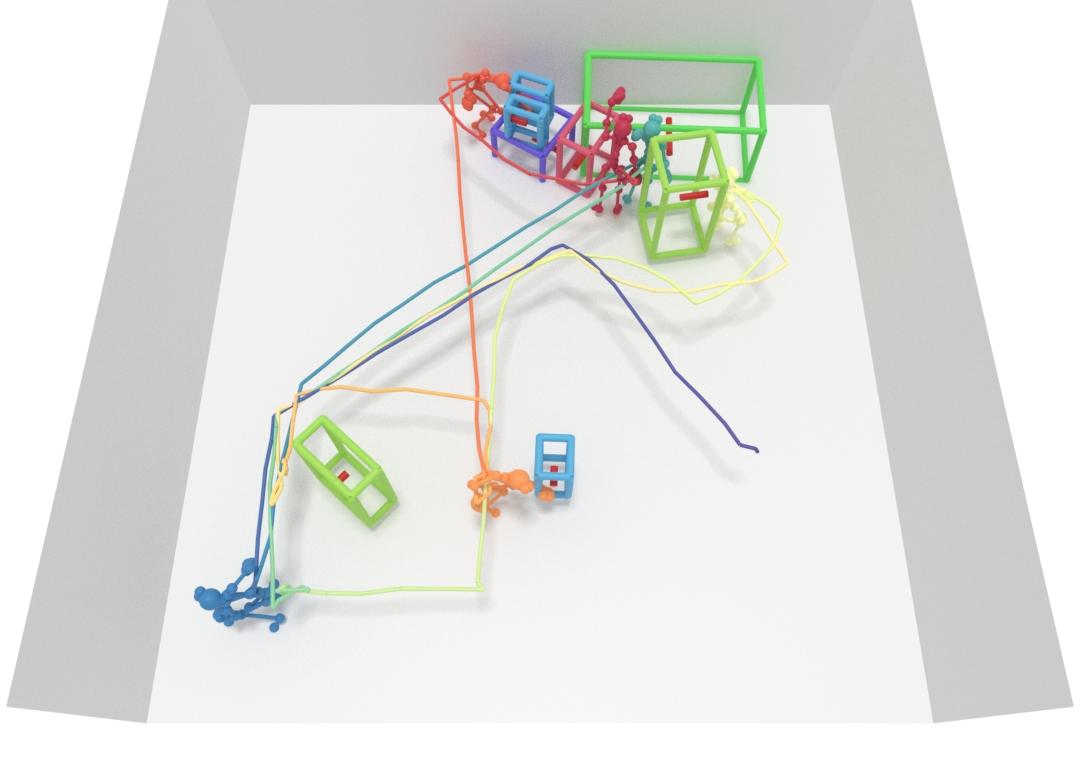}
		\caption{P-Vote.}
	\end{subfigure}
	\begin{subfigure}[t]{0.159\textwidth}
		\includegraphics[width=\textwidth]
		{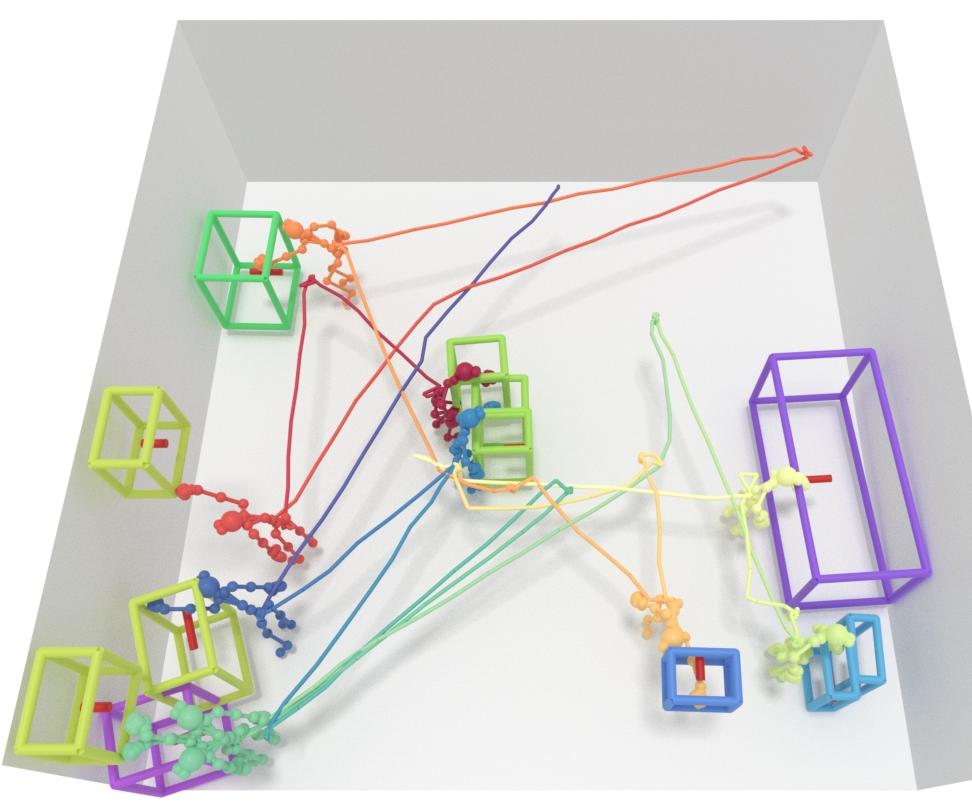}
		\includegraphics[width=\textwidth]
		{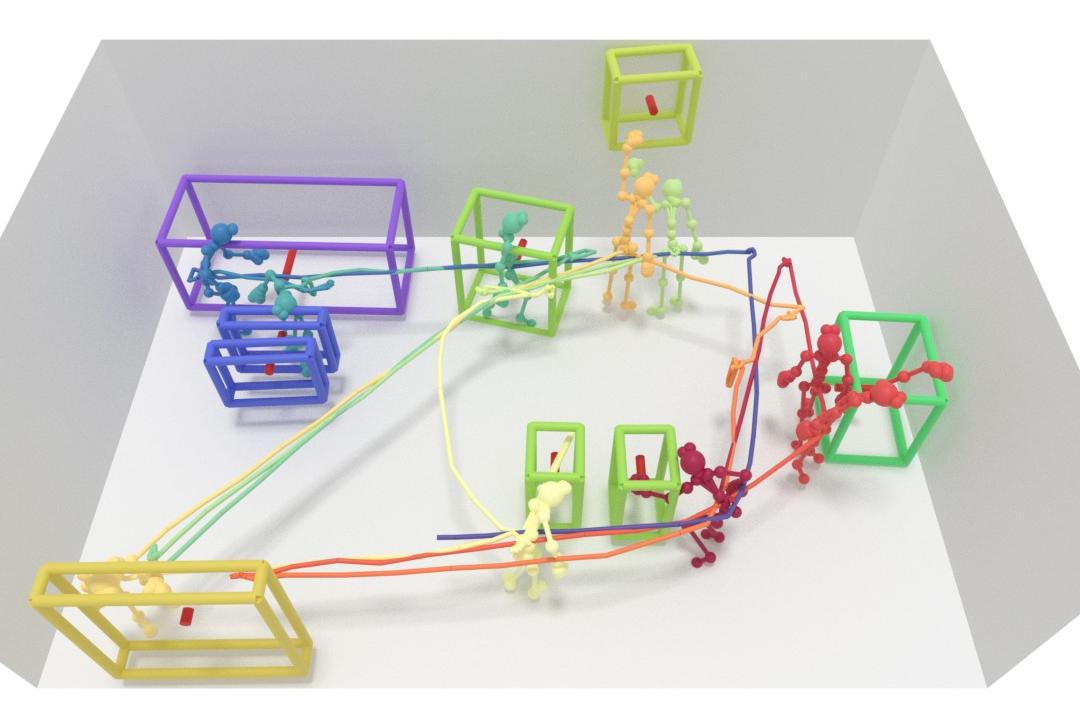}
		\includegraphics[width=\textwidth]
		{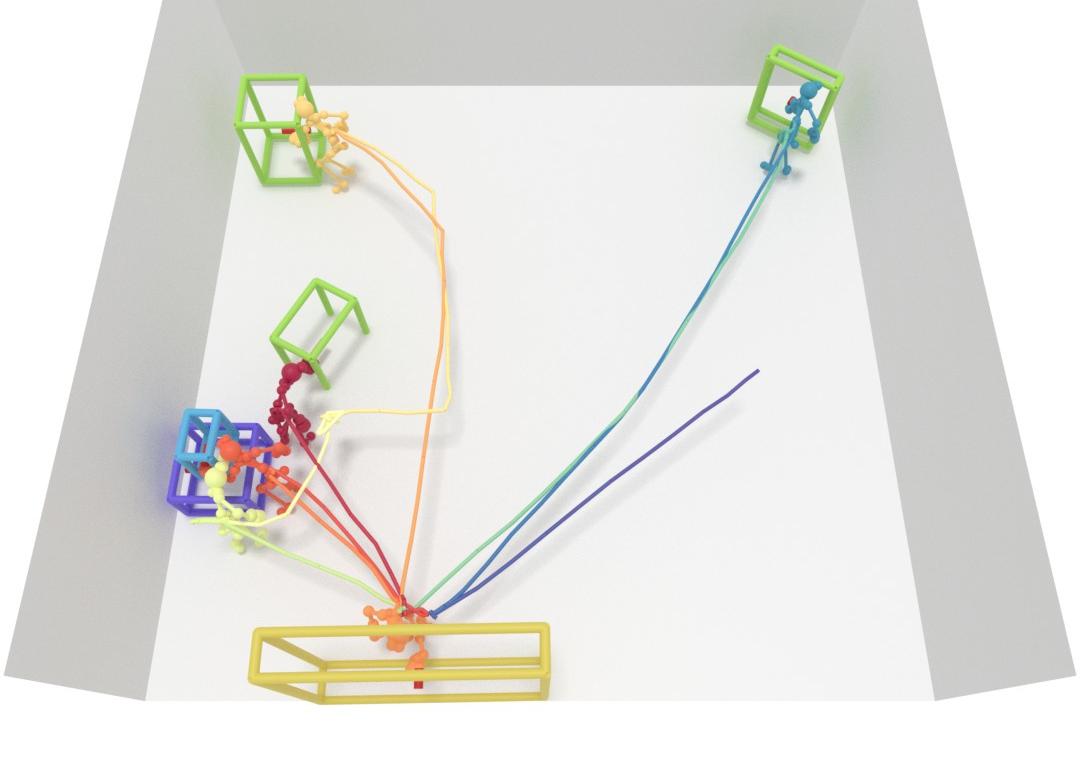}
		\includegraphics[width=\textwidth]
		{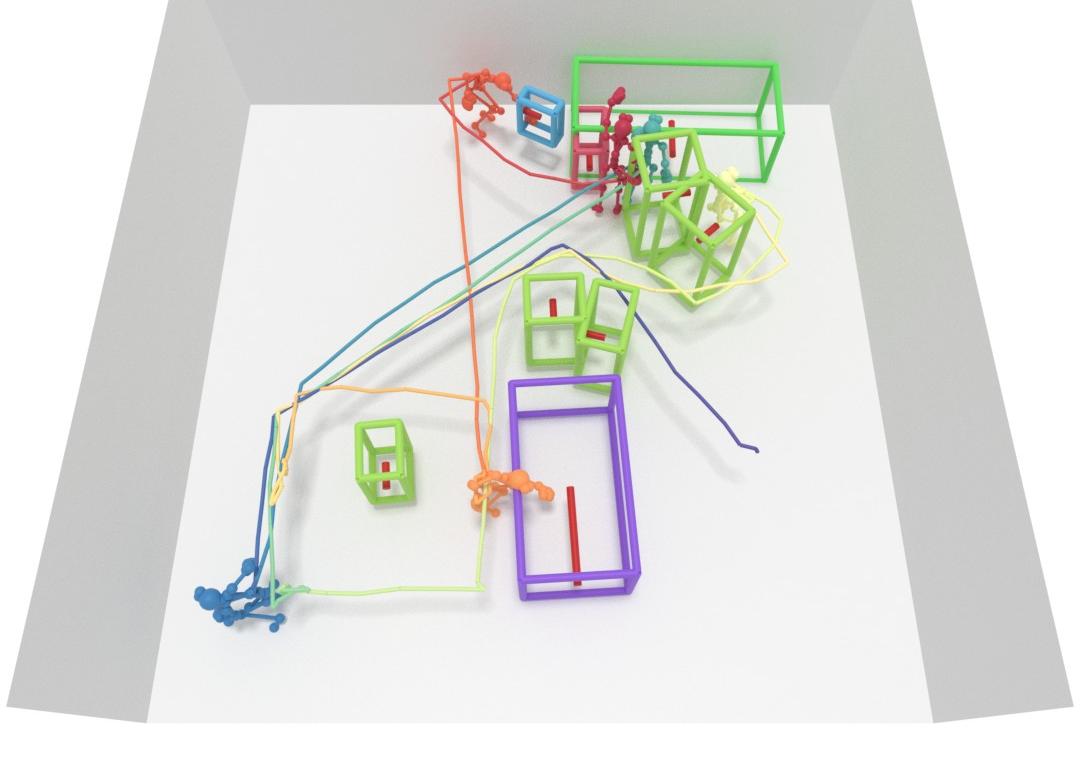}
		\caption{P-VN}
	\end{subfigure}
	\begin{subfigure}[t]{0.159\textwidth}
		\includegraphics[width=\textwidth]
		{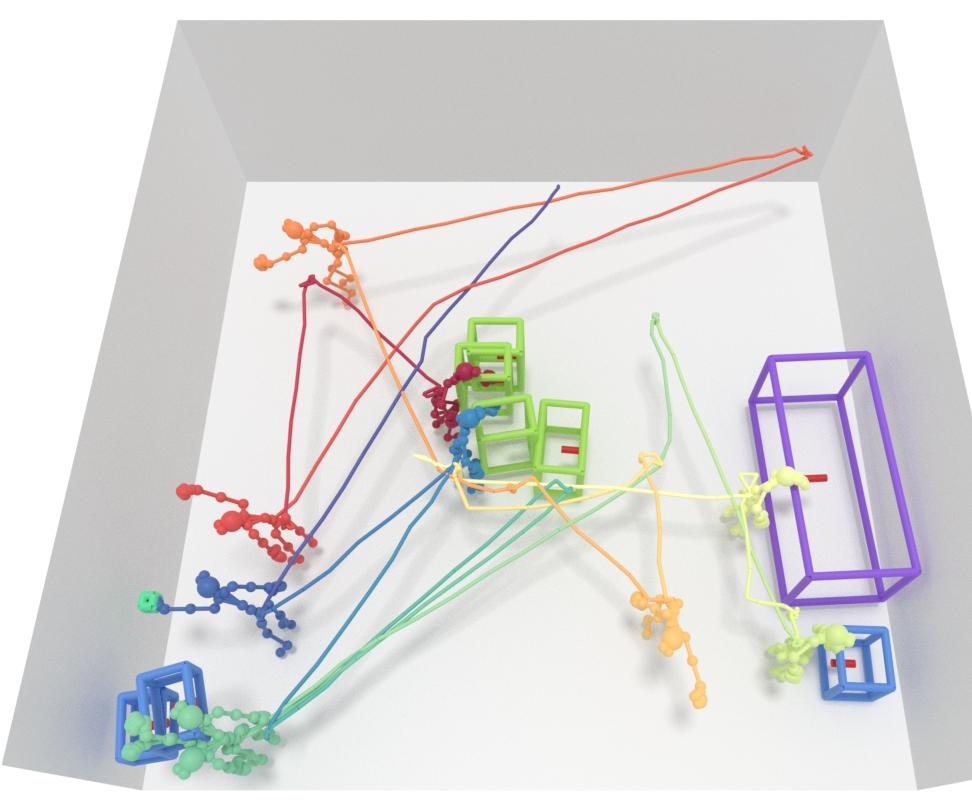}
		\includegraphics[width=\textwidth]
		{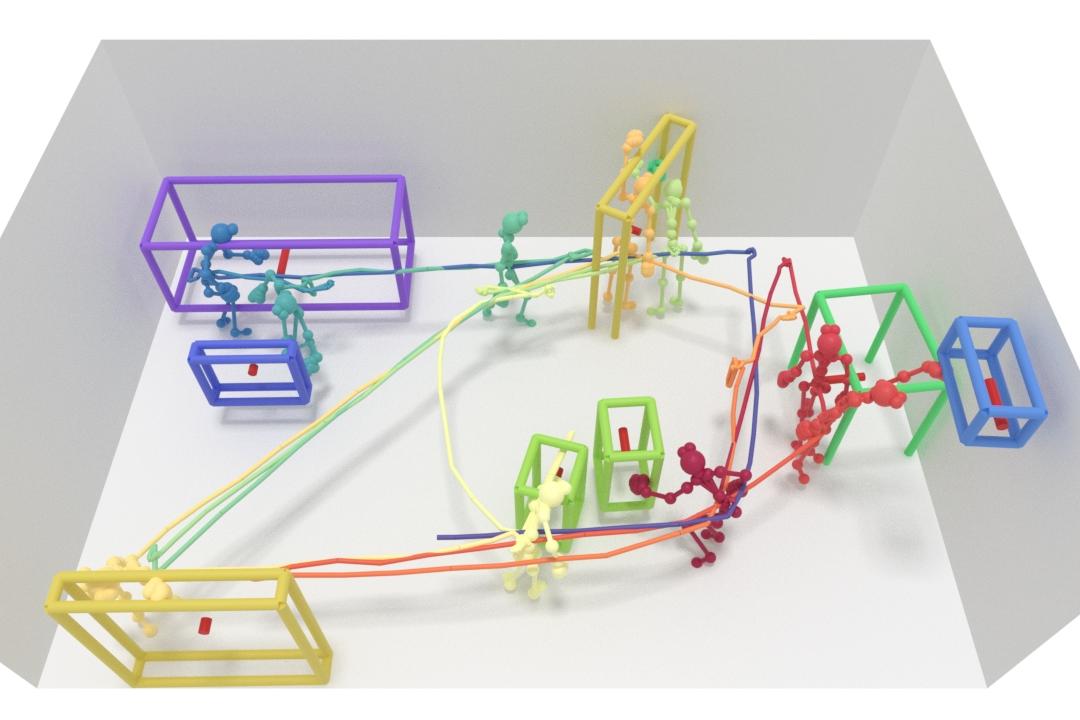}
		\includegraphics[width=\textwidth]
		{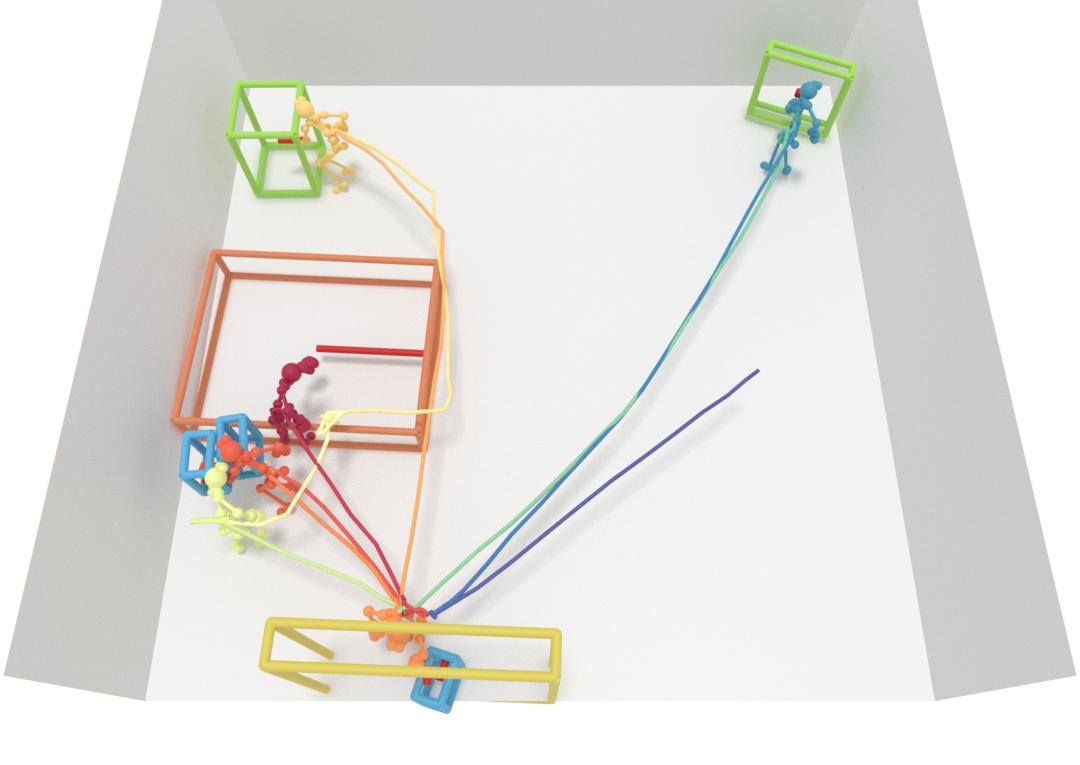}
		\includegraphics[width=\textwidth]
		{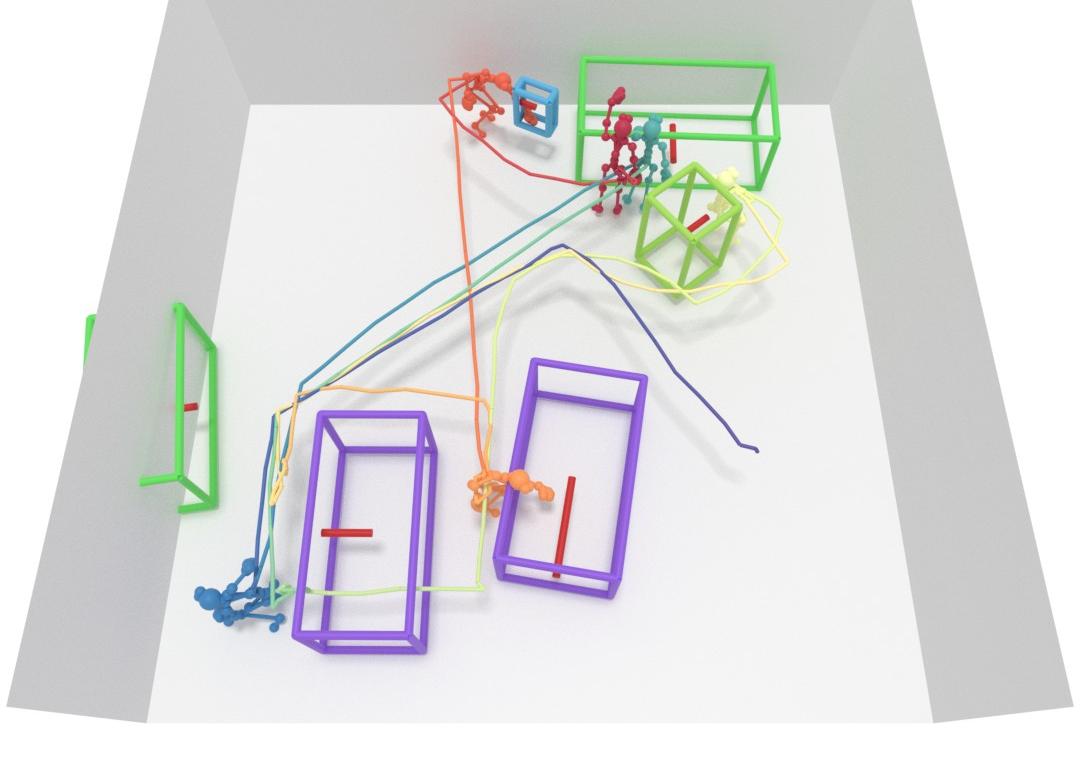}
		\caption{Mo. Attn}
	\end{subfigure}
	\begin{subfigure}[t]{0.159\textwidth}
		\includegraphics[width=\textwidth]
		{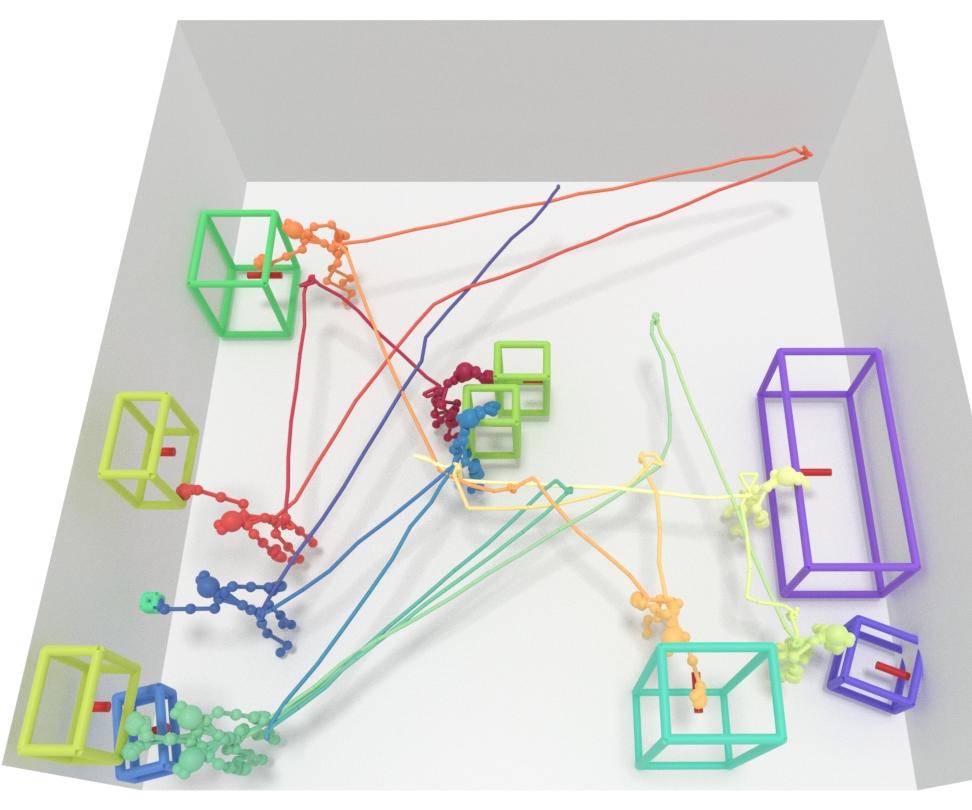}
		\includegraphics[width=\textwidth]
		{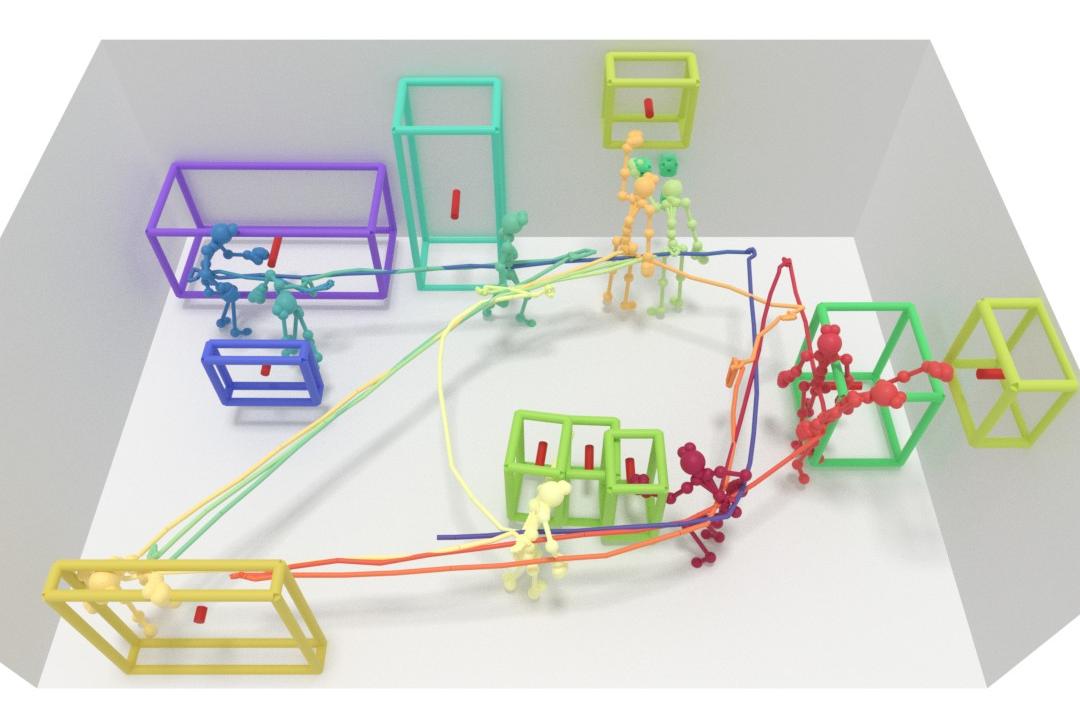}
		\includegraphics[width=\textwidth]
		{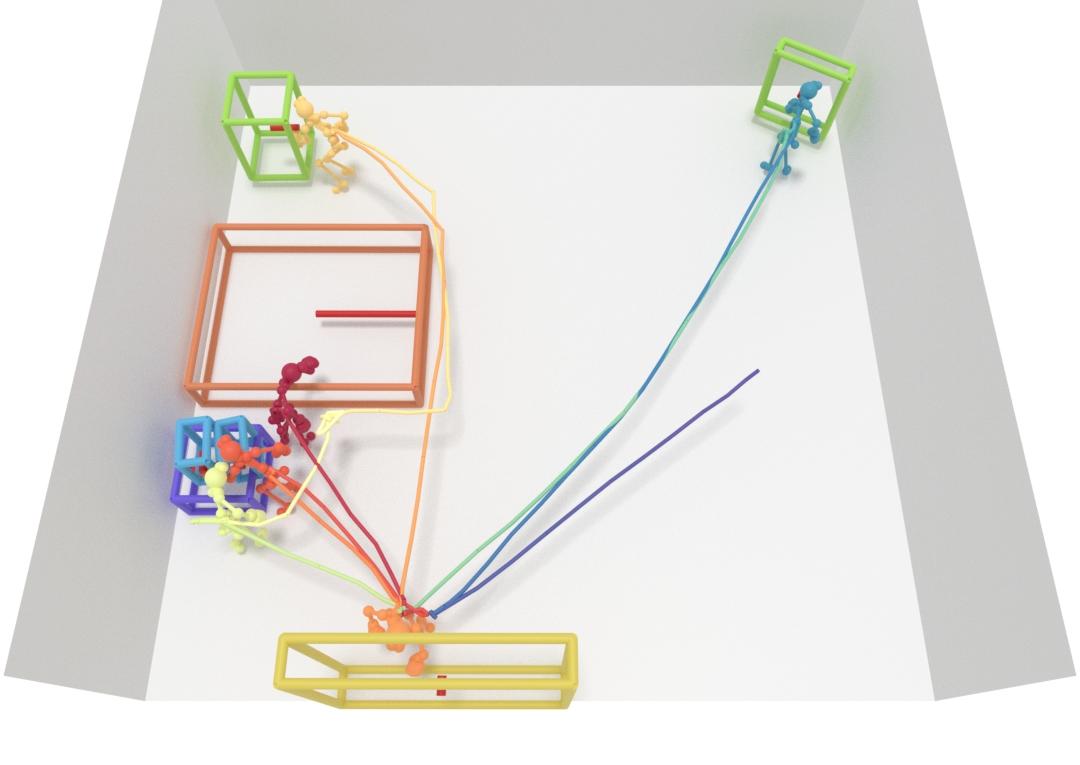}
		\includegraphics[width=\textwidth]
		{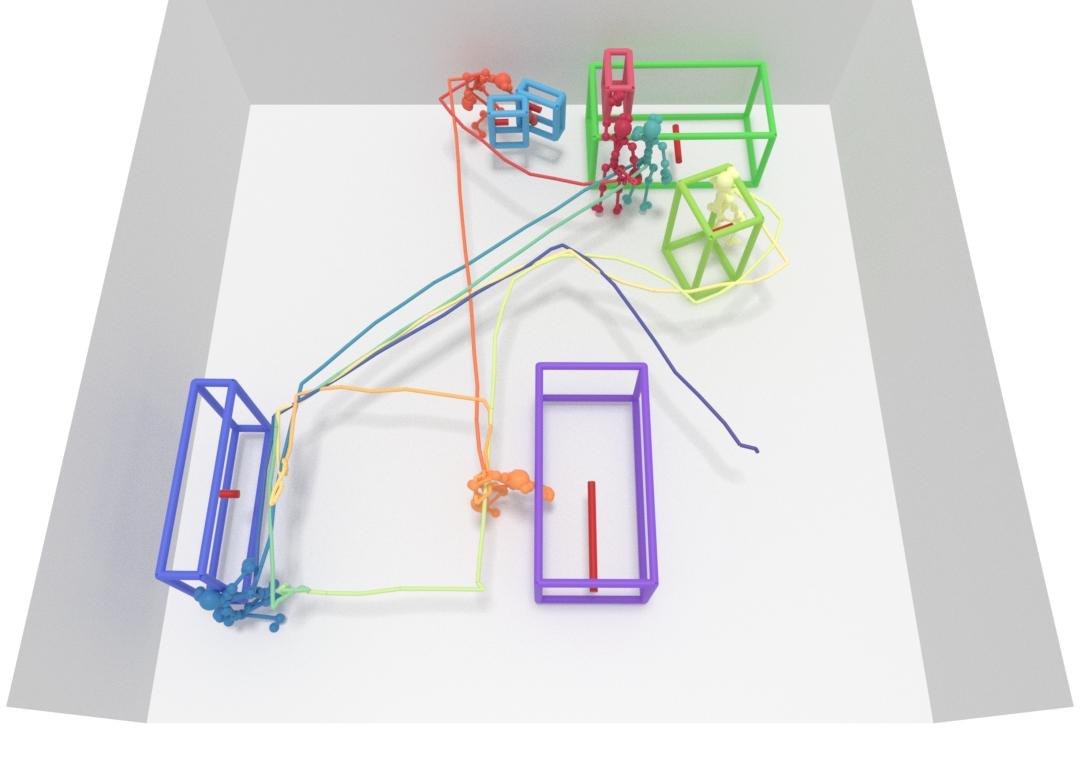}
		\caption{Ours}
	\end{subfigure}
	\begin{subfigure}[t]{0.159\textwidth}
		\includegraphics[width=\textwidth]
		{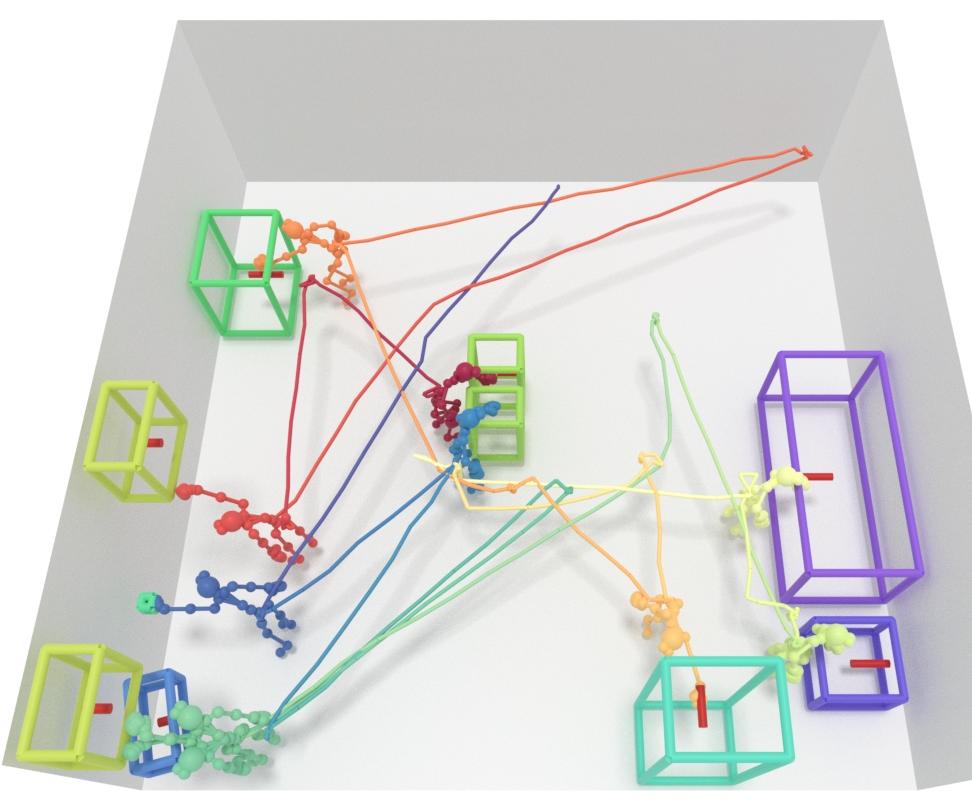}
		\includegraphics[width=\textwidth]
		{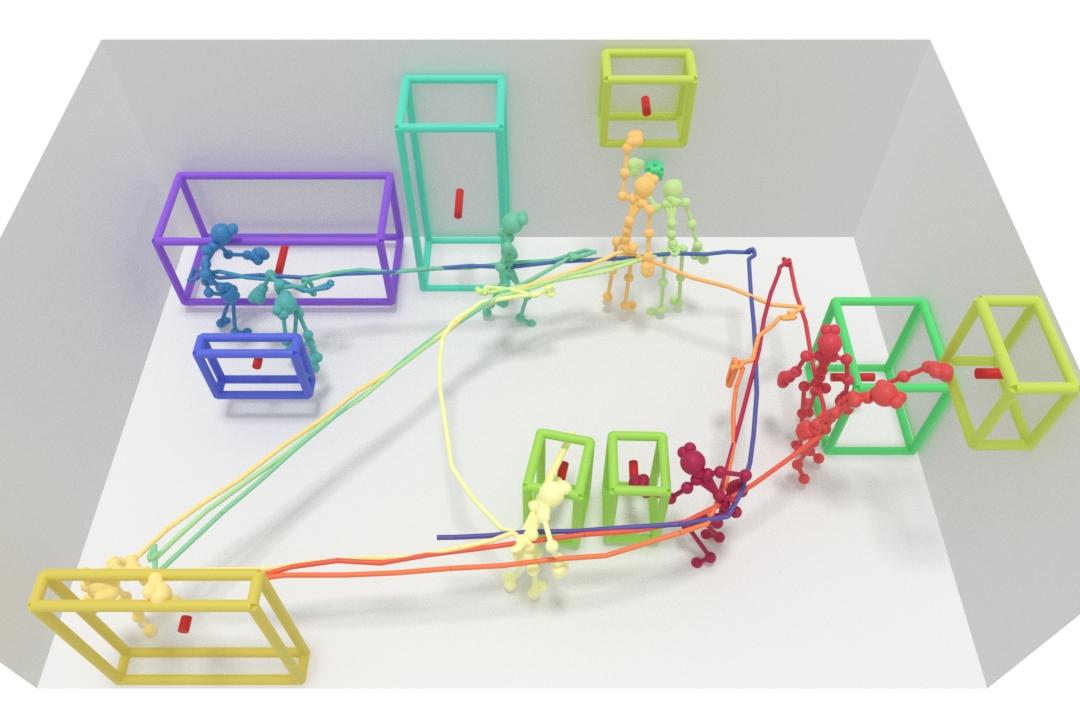}
		\includegraphics[width=\textwidth]
		{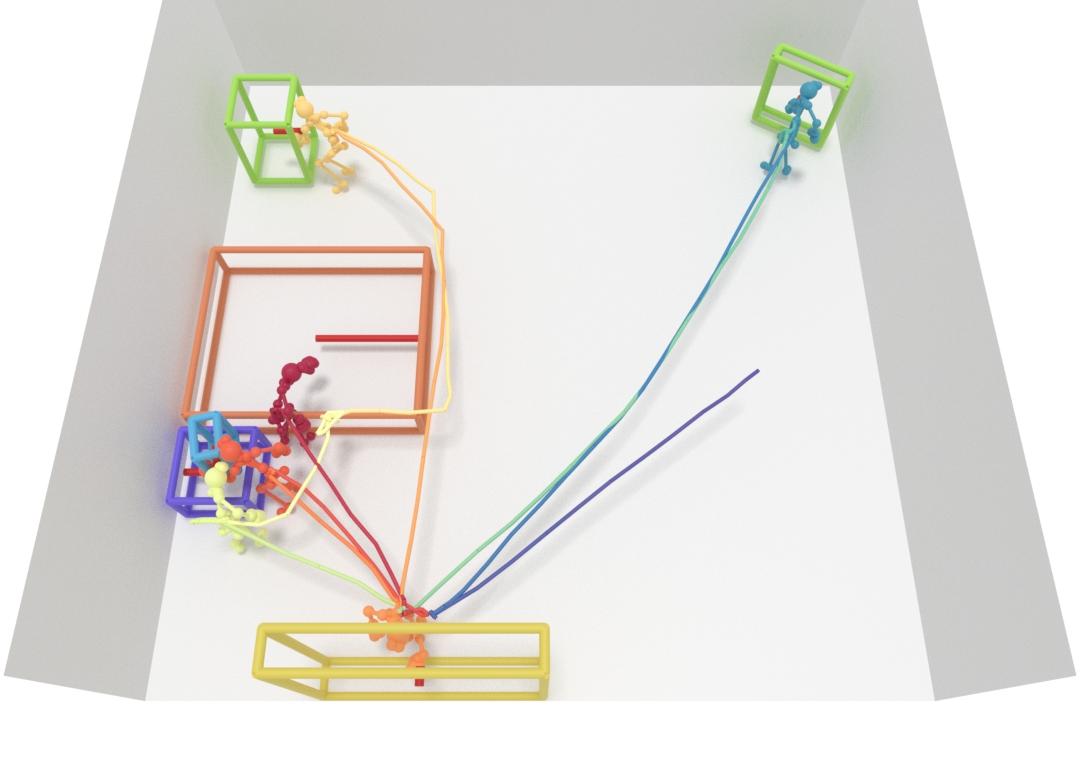}
		\includegraphics[width=\textwidth]
		{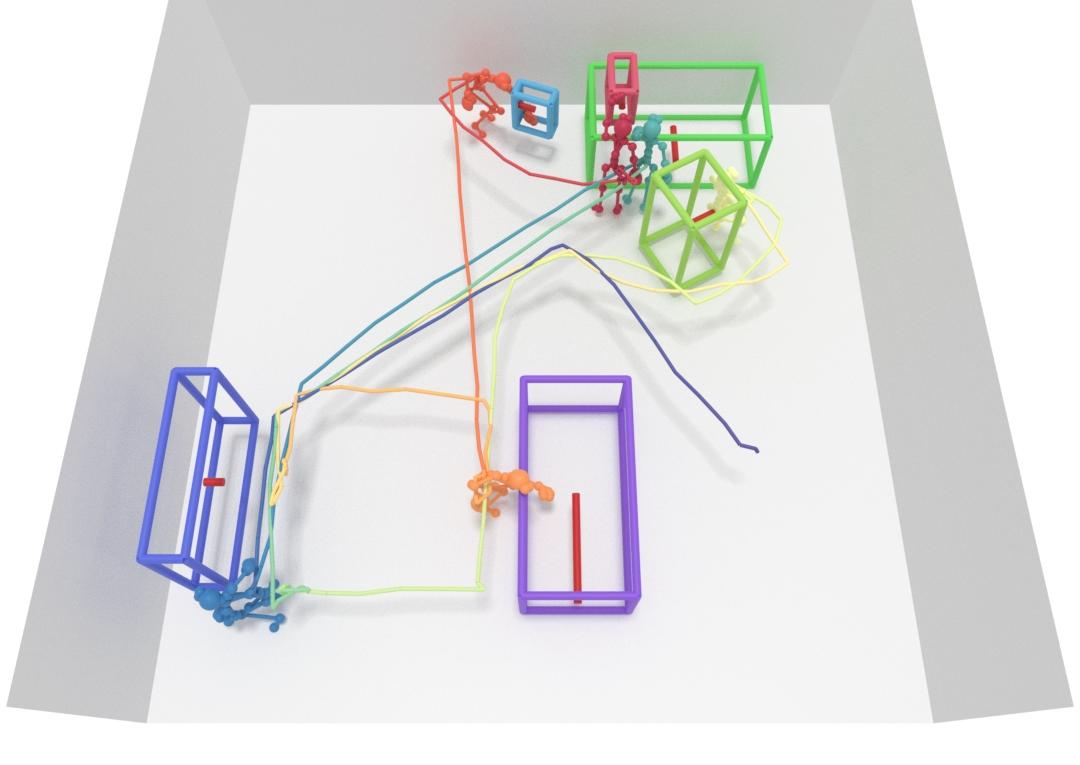}
		\caption{GT}
	\end{subfigure}
	\caption{Qualitative results of object detection from a pose trajectory on the sequence-level split $\mathcal{S}_{1}$ (unseen interaction sequences).}
	\label{fig:quali_comp_s1}
\end{figure*}

\subsection{Baselines}
Since there are no prior works that tackle the task of predicting the object configuration of a scene from solely a 3D pose trajectory, we construct several baselines leveraging state-of-the-art techniques as well as various approaches to estimate multi-modal distributions.
We consider the following baselines: \textbf{1) Pose-VoteNet}~\cite{Qi_2019_ICCV}. Since VoteNet is designed for detection from point clouds, we replace their PointNet++ encoder with our position encoder + MLPs to learn joint features for seeds. \textbf{2) Pose-VN}, Pose-VoteNet based on Vector Neurons \cite{deng2021vn} which replaces MLP layers in Pose-VoteNet with SO(3)-equivariant operators that can capture arbitrary rotations of poses to estimate objects. \textbf{3) Motion Attention} \cite{mao2020history}. Since our task can be also regarded as a sequence-to-sequence problem, we adopt a frame-wise attention encoder to extract repetitive pose patterns in the temporal domain which then inform a VoteNet decoder to regress boxes.
Additionally, we also ablate our probabilistic mixture decoder with other alternatives: \textbf{4) Deterministic \OURS{}} (\OURS{}-D), where we use VoteNet decoder \cite{Qi_2019_ICCV} in our method for box regression to produce deterministic results; \textbf{5) Generative \OURS{}} (\OURS{}-G), where our \OURS{} decoder is designed with a probabilistic generative model \cite{wu2016learning} to decode boxes from a learned latent variable; \textbf{6) Heatmap \OURS{}} (\OURS{}-H), where the box center, size and orientation are discretized into binary heatmaps, and the box regression is converted into a classification task. Detailed architecture specifications for these networks are given in the supplemental material.

\begin{figure}[t]
	\centering
	\begin{subfigure}[t]{0.159\textwidth}
		\includegraphics[width=\textwidth]
		{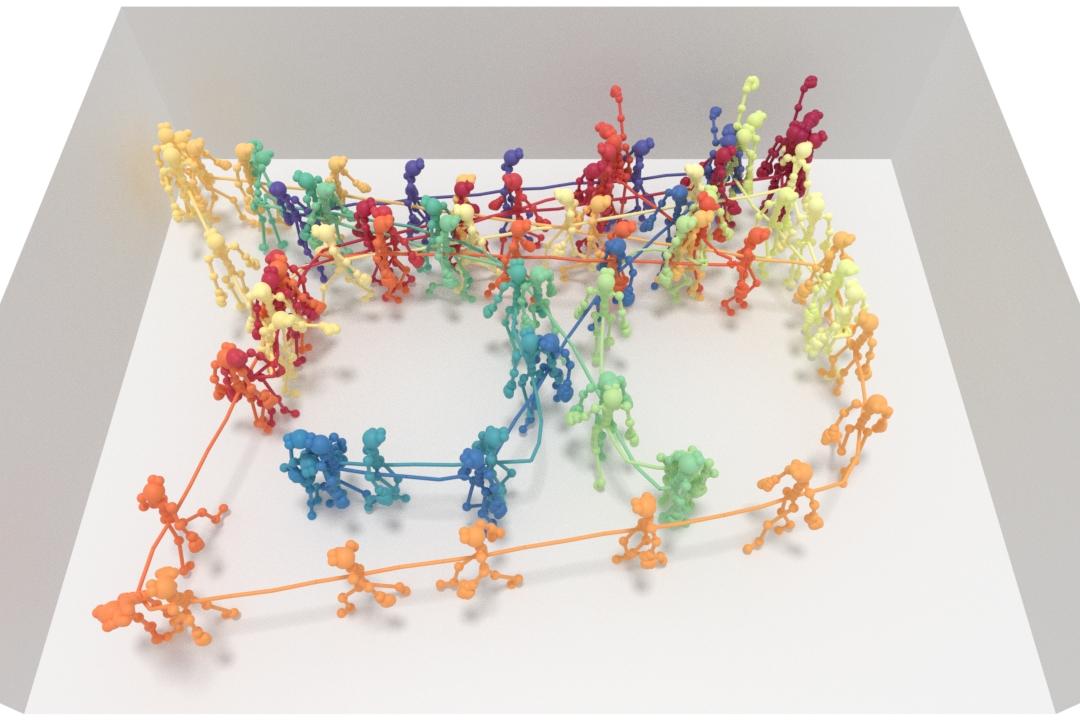}
		\includegraphics[width=\textwidth]
		{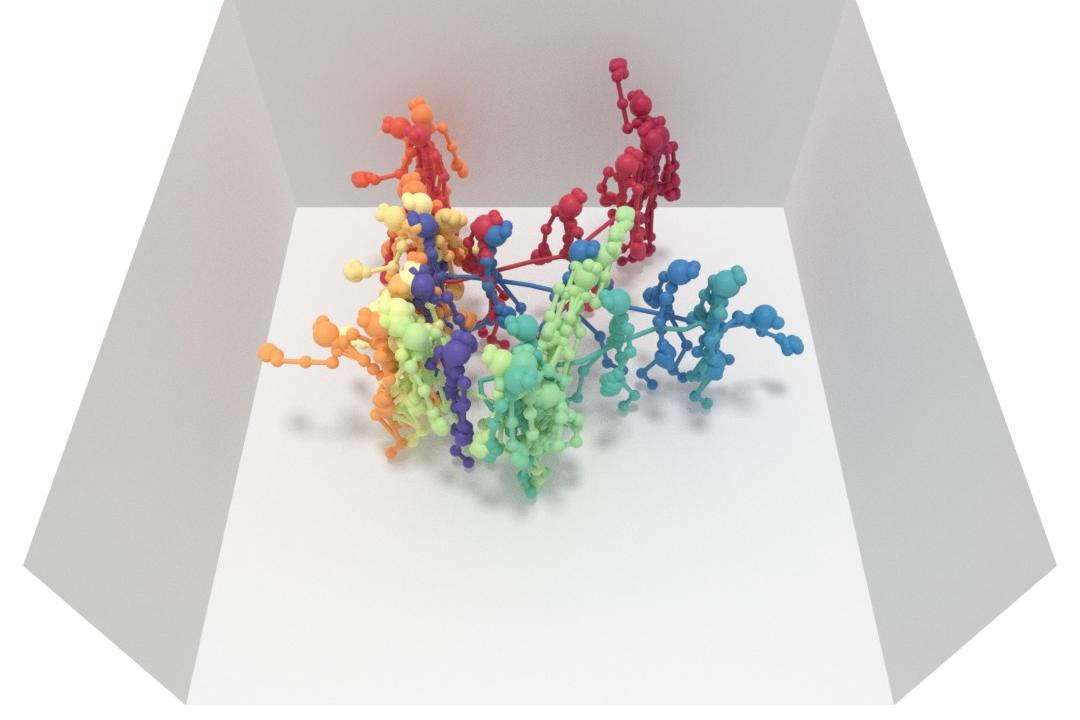}
		\includegraphics[width=\textwidth]
		{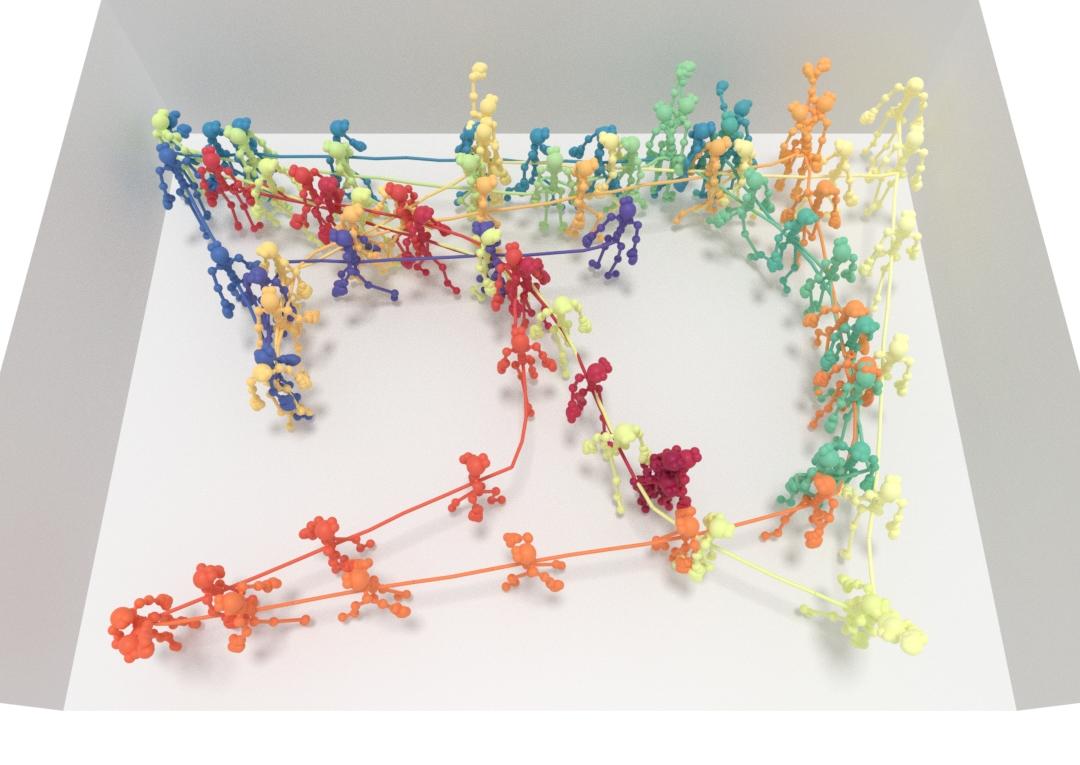}
		\caption{Input}
	\end{subfigure}
	\begin{subfigure}[t]{0.159\textwidth}
		\includegraphics[width=\textwidth]
		{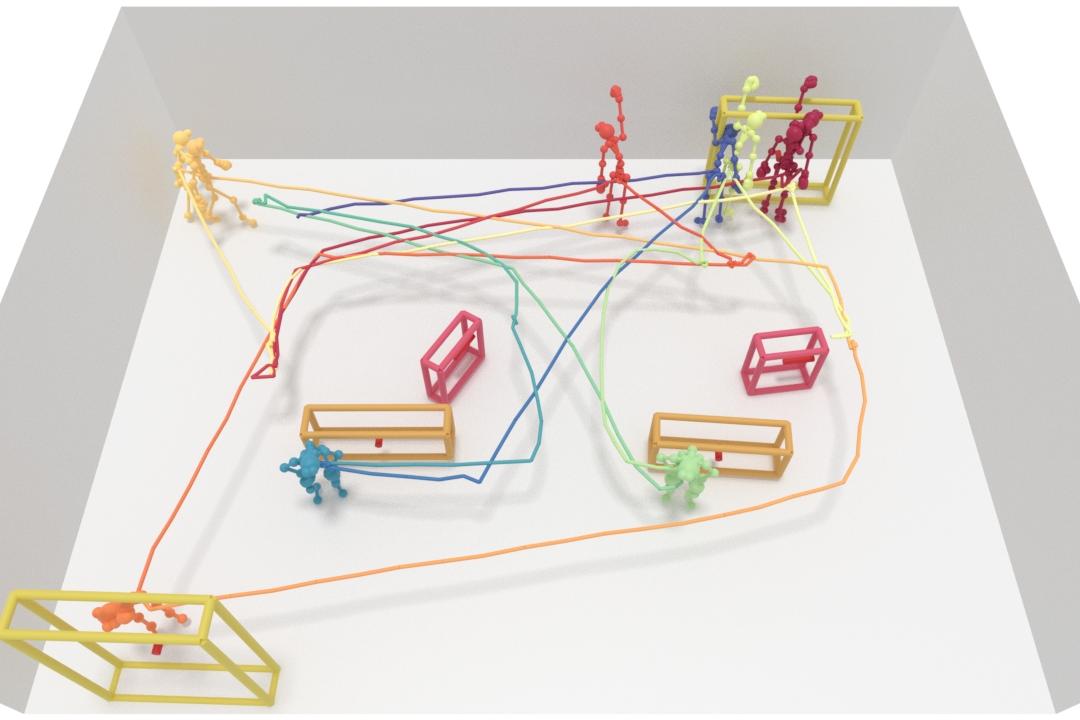}
		\includegraphics[width=\textwidth]
		{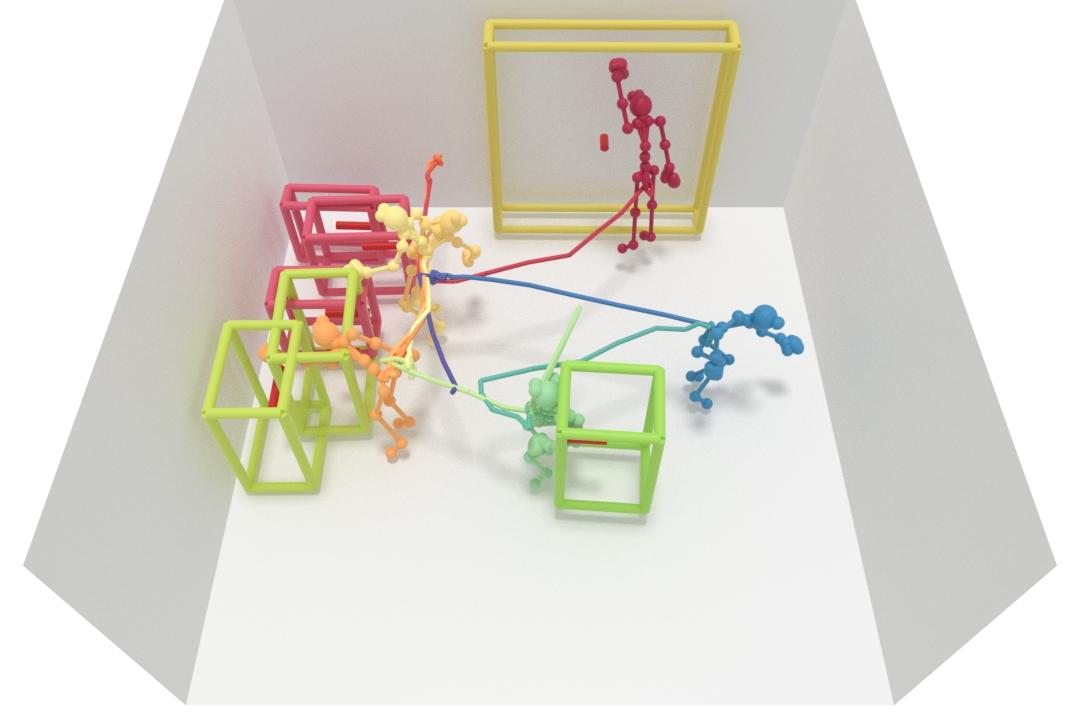}
		\includegraphics[width=\textwidth]
		{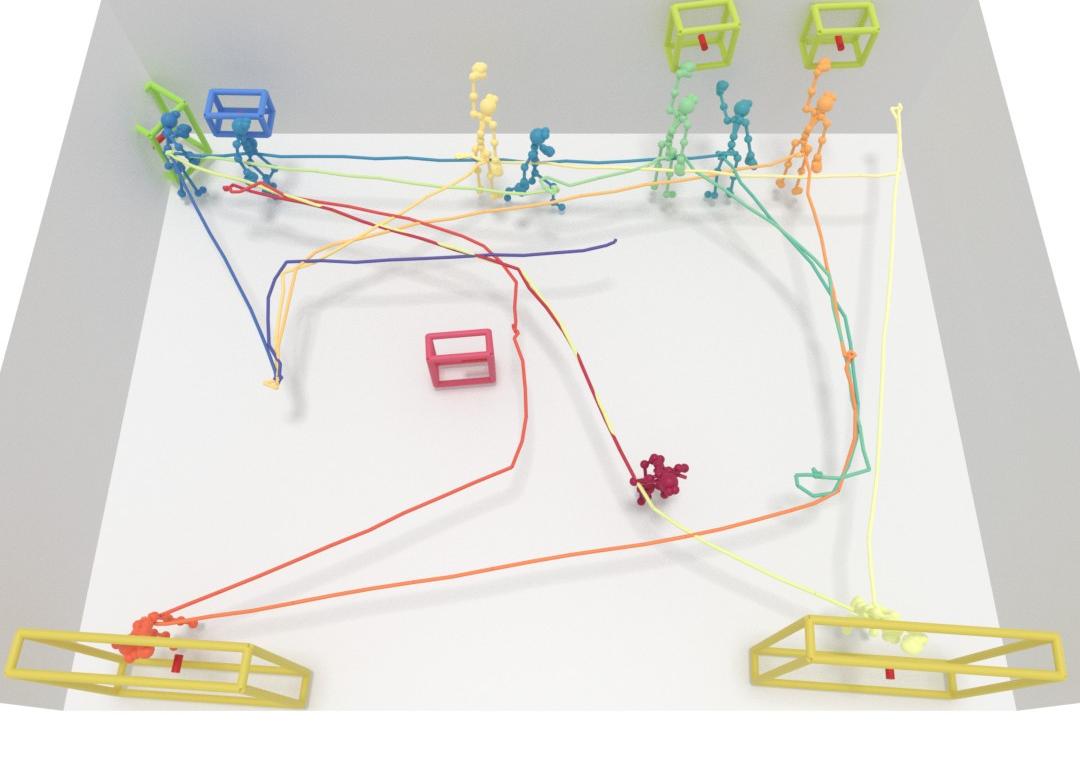}
		\caption{P-Vote.}
	\end{subfigure}
	\begin{subfigure}[t]{0.159\textwidth}
		\includegraphics[width=\textwidth]
		{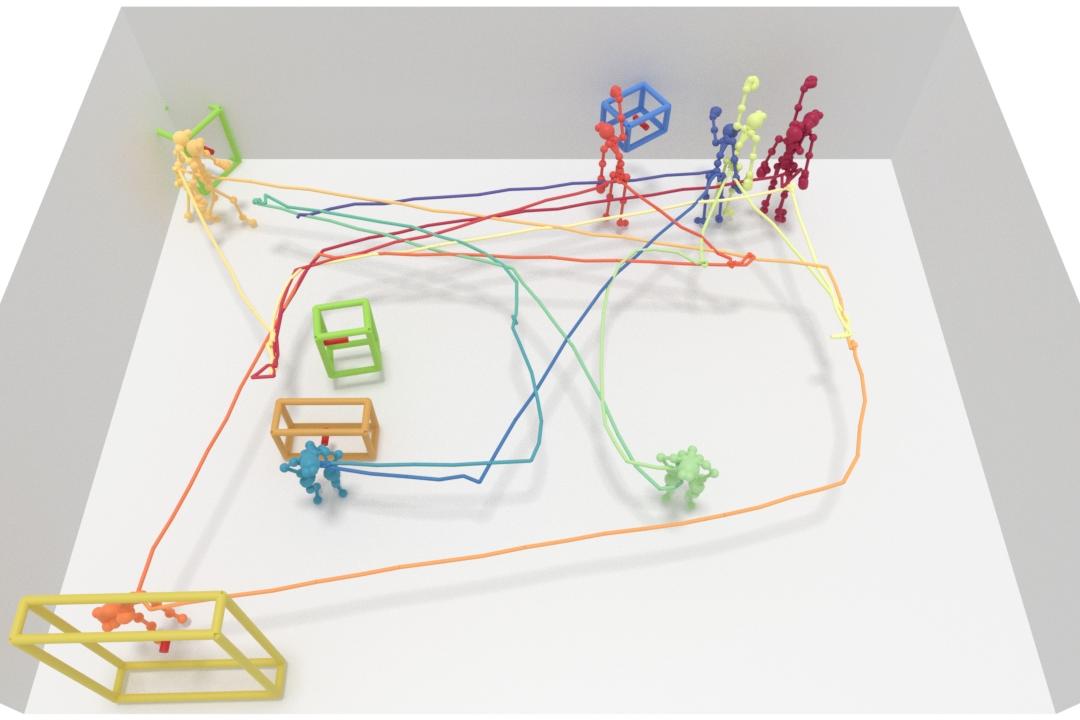}
		\includegraphics[width=\textwidth]
		{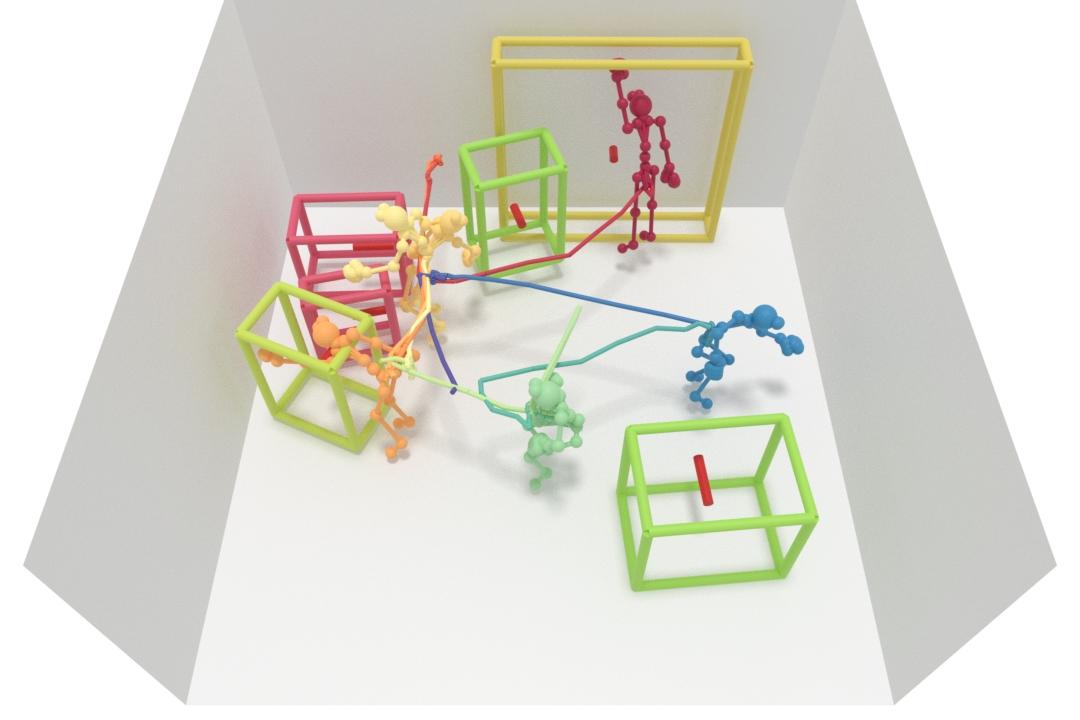}
		\includegraphics[width=\textwidth]
		{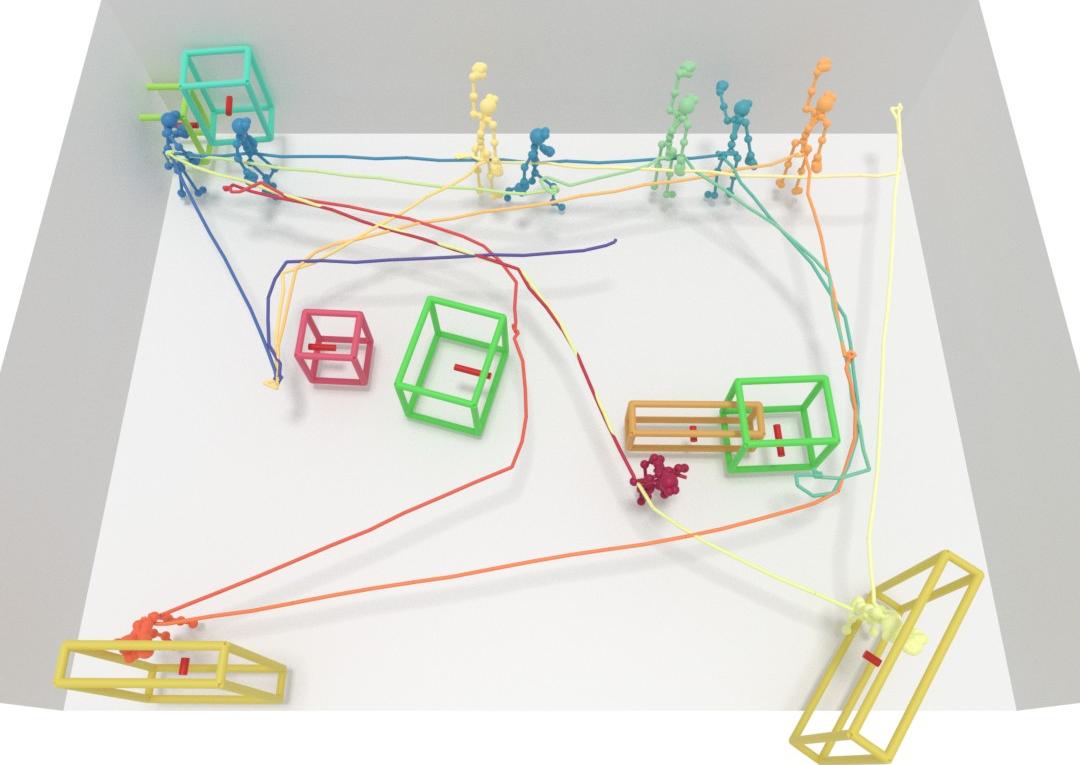}
		\caption{P-VN}
	\end{subfigure}
	\begin{subfigure}[t]{0.159\textwidth}
		\includegraphics[width=\textwidth]
		{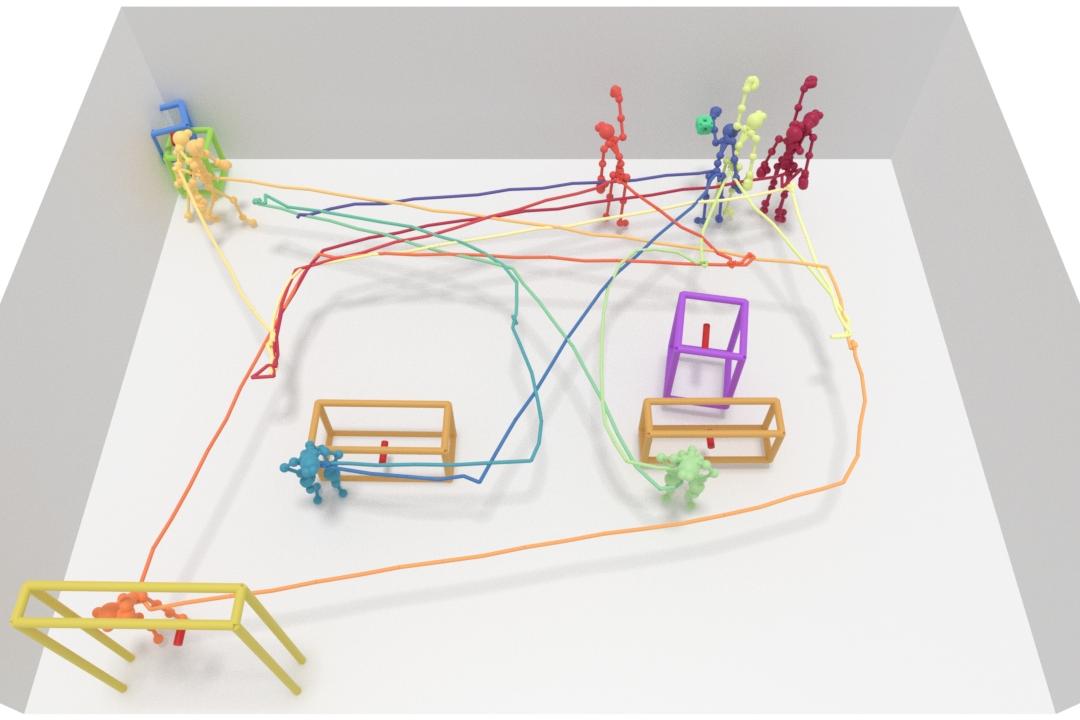}
		\includegraphics[width=\textwidth]
		{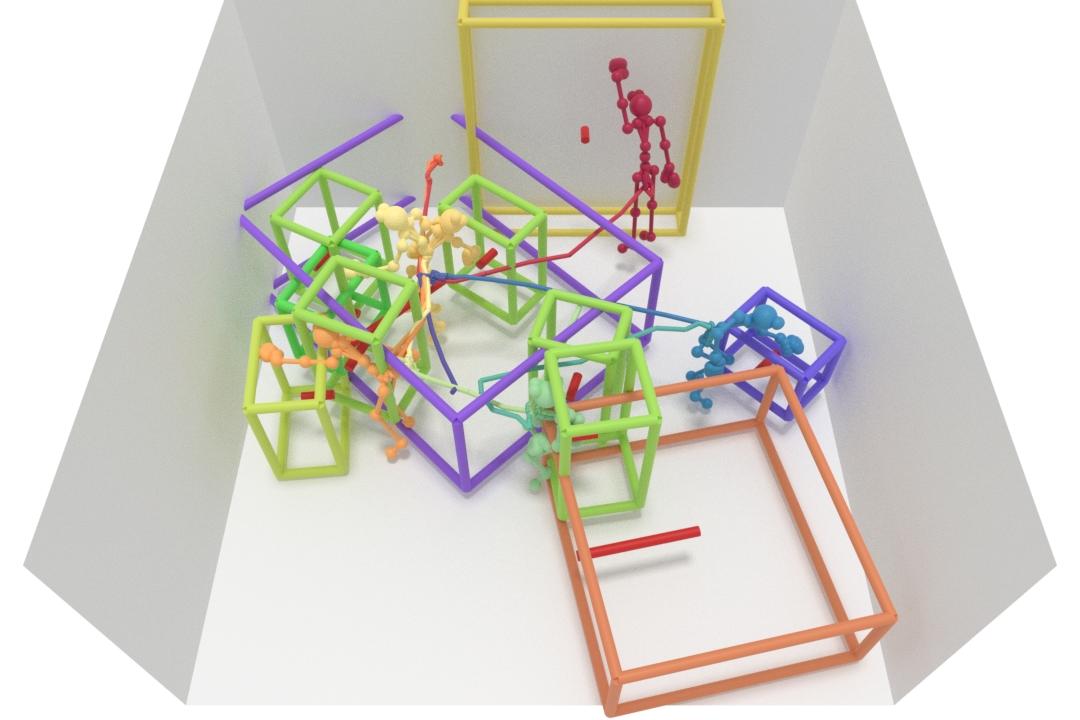}
		\includegraphics[width=\textwidth]
		{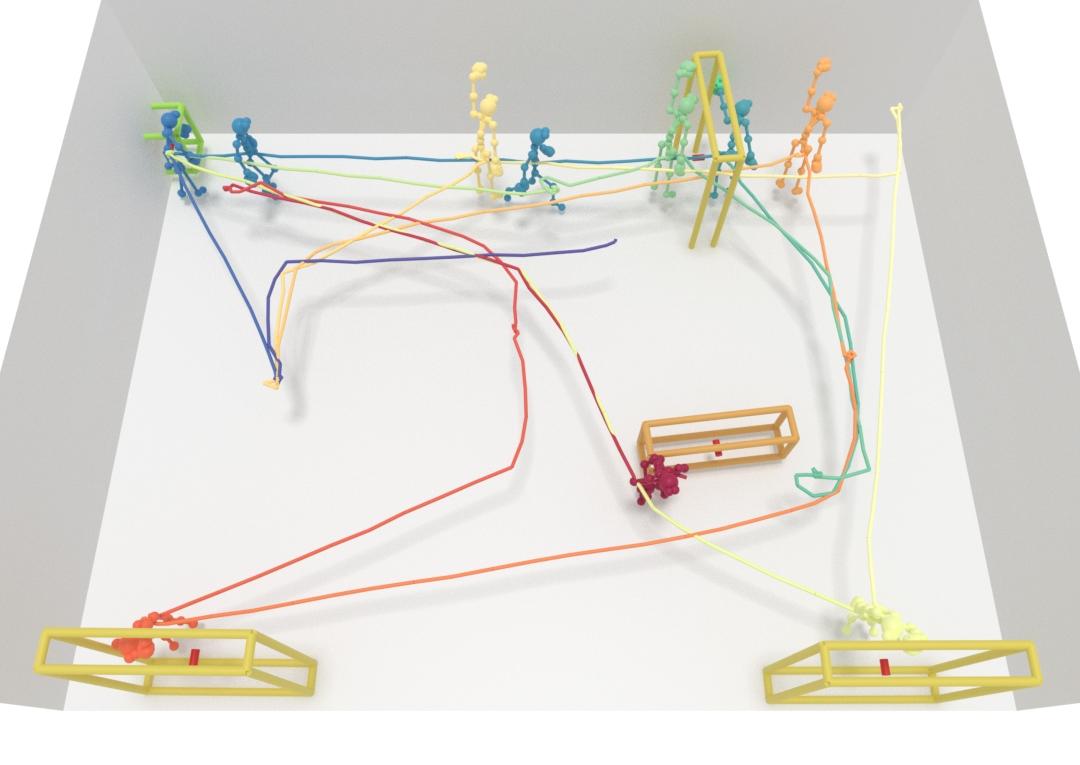}
		\caption{Mo. Attn}
	\end{subfigure}
	\begin{subfigure}[t]{0.159\textwidth}
		\includegraphics[width=\textwidth]
		{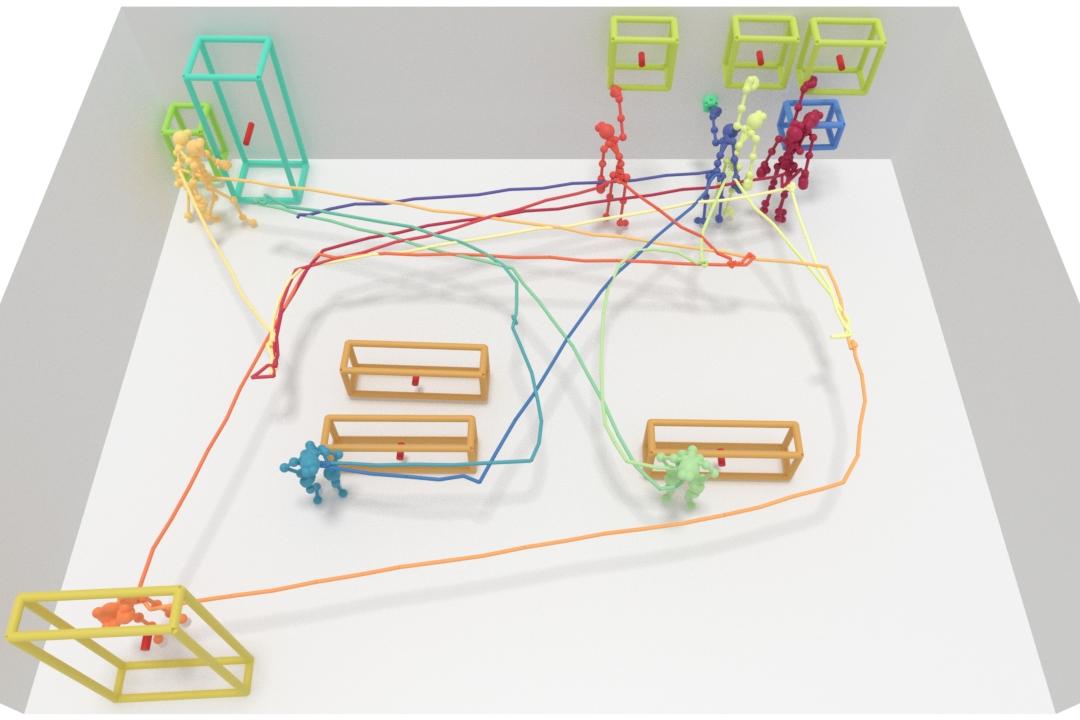}
		\includegraphics[width=\textwidth]
		{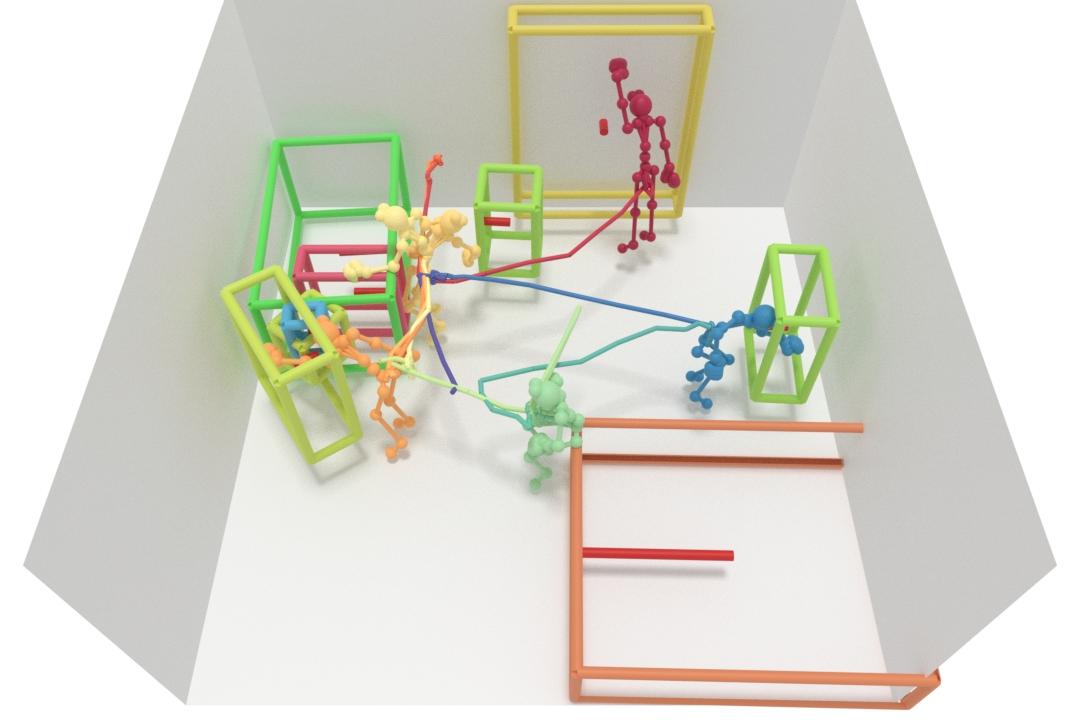}
		\includegraphics[width=\textwidth]
		{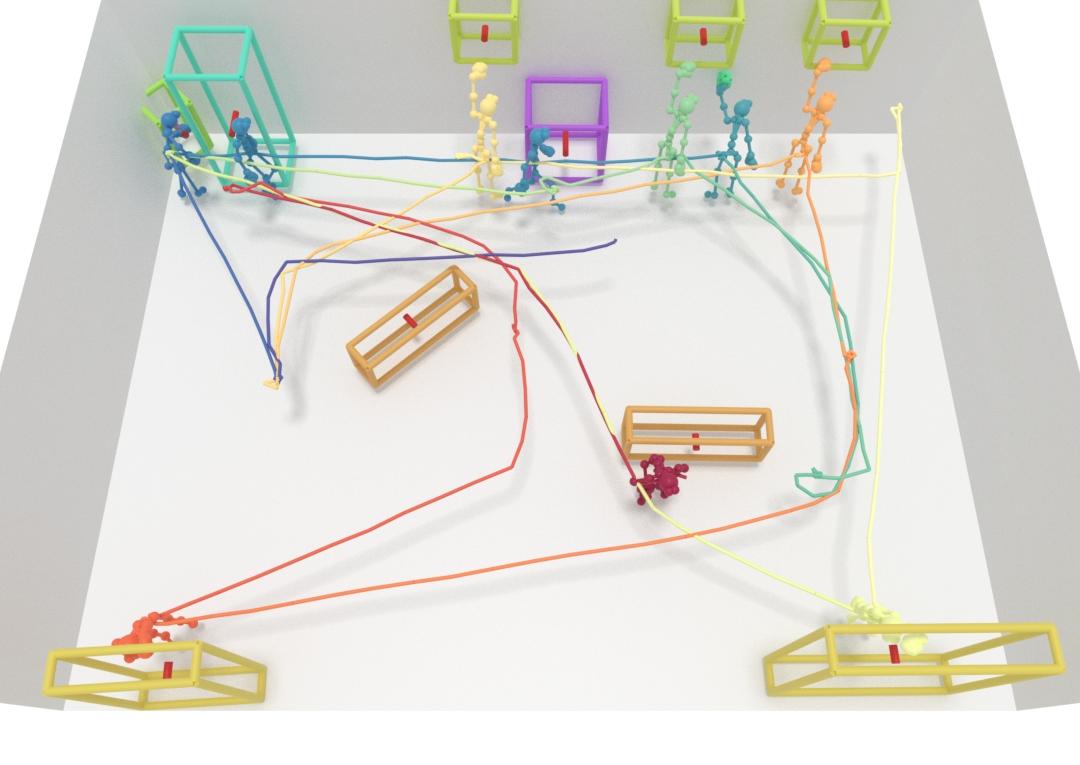}
		\caption{Ours}
	\end{subfigure}
	\begin{subfigure}[t]{0.159\textwidth}
		\includegraphics[width=\textwidth]
		{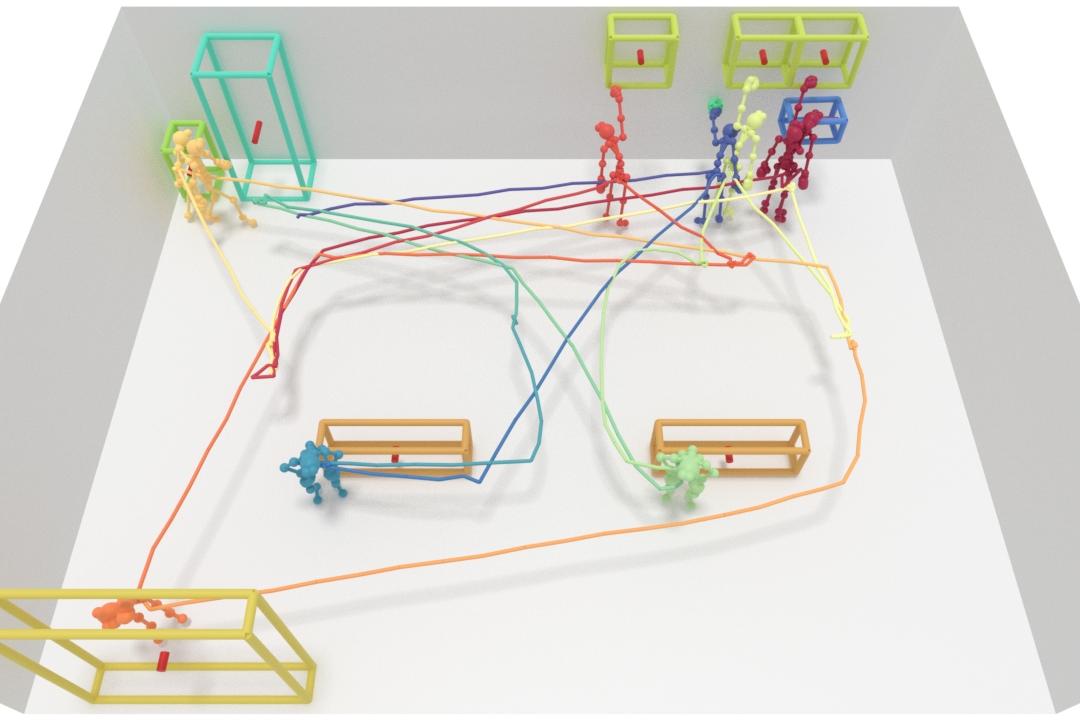}
		\includegraphics[width=\textwidth]
		{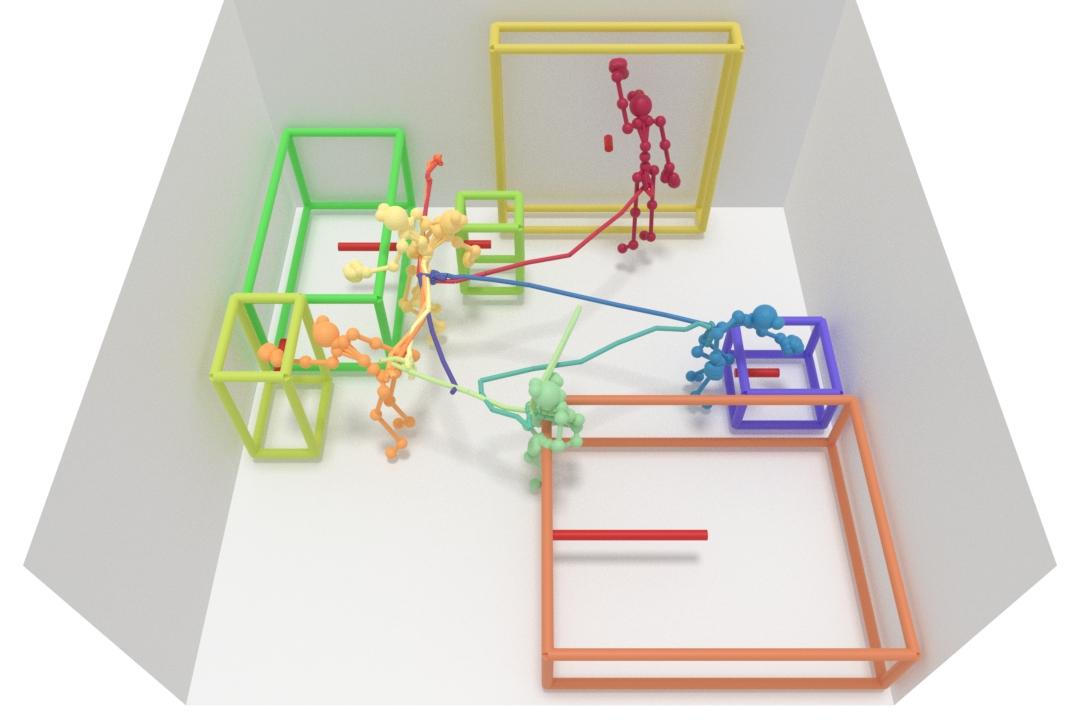}
		\includegraphics[width=\textwidth]
		{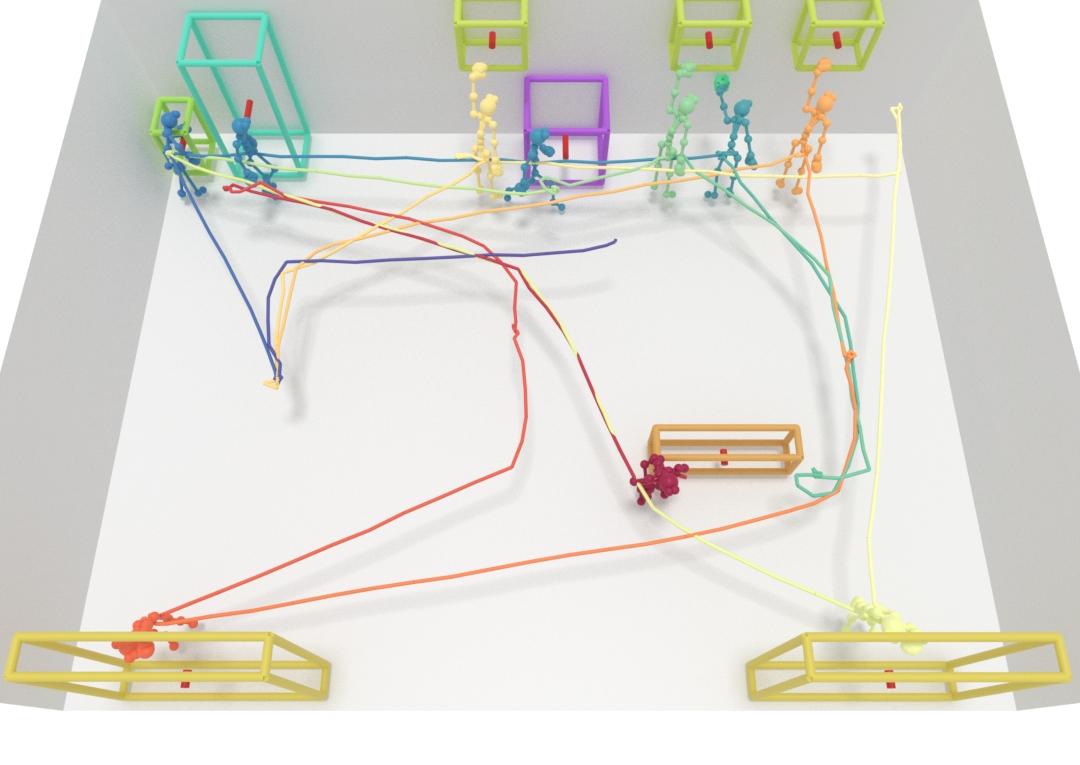}
		\caption{GT}
	\end{subfigure}
	\caption{Qualitative results of object detection from a pose trajectory on the room-level split $\mathcal{S}_{2}$ (unseen interaction sequences and rooms).
	}
	\label{fig:quali_comp_s2}
\end{figure}

\subsection{Qualitative Comparisons}

\noindent\textbf{Comparisons on $\mathcal{S}_{1}$.}
Fig.~\ref{fig:quali_comp_s1} visualizes predictions on the test set of unseen interaction sequences.
Pose-VoteNet struggles to identify the existence of an object, leading to many missed detections, but can estimate reasonable object locations when an object is predicted.
Pose-VN alleviates this problem of under-detection, but struggles to estimate object box sizes (rows 1,3).
These baselines indicate the difficulty in detecting objects without sharing pose features among temporal neighbors. Motion Attention \cite{mao2020history} addresses this by involving global context with inter-frame attention. However, it does not take advantage of the skeleton's spatial layout in continuous frames and struggles to detect the existence of objects (row 1,2). In contrast, our method leverages both target-dependent poses and object occupancy context, that learns the implicit interrelations between poses and objects to infer object boxes, and achieves better estimate of the scene configuration.

\noindent\textbf{Comparisons on $\mathcal{S}_{2}$.} In Fig.~\ref{fig:quali_comp_s2}, we illustrate the qualitative comparisons on the test set of unseen interaction sequences in unknown rooms. In this scenario, most baselines fail to localize objects, while our method can nonetheless produce plausible object layouts.

\begin{figure}[bp]
	\centering
	\begin{subfigure}[t]{0.159\textwidth}
		\includegraphics[width=\textwidth]
		{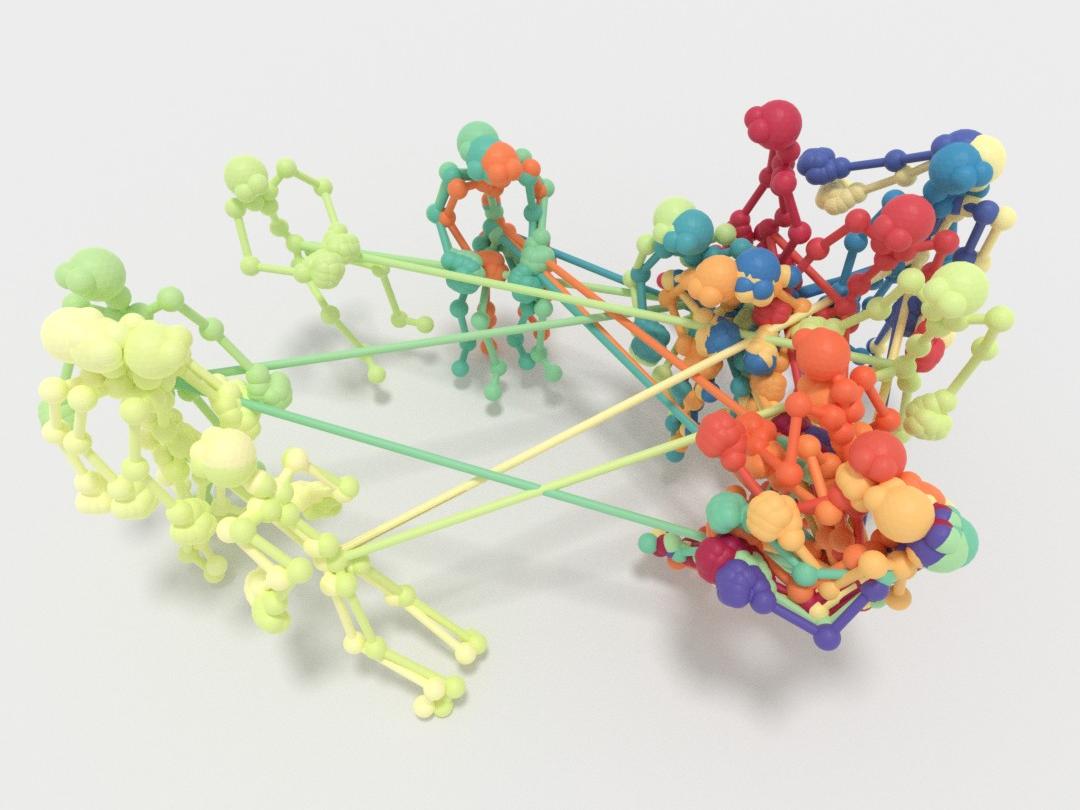}
		\includegraphics[width=\textwidth]
		{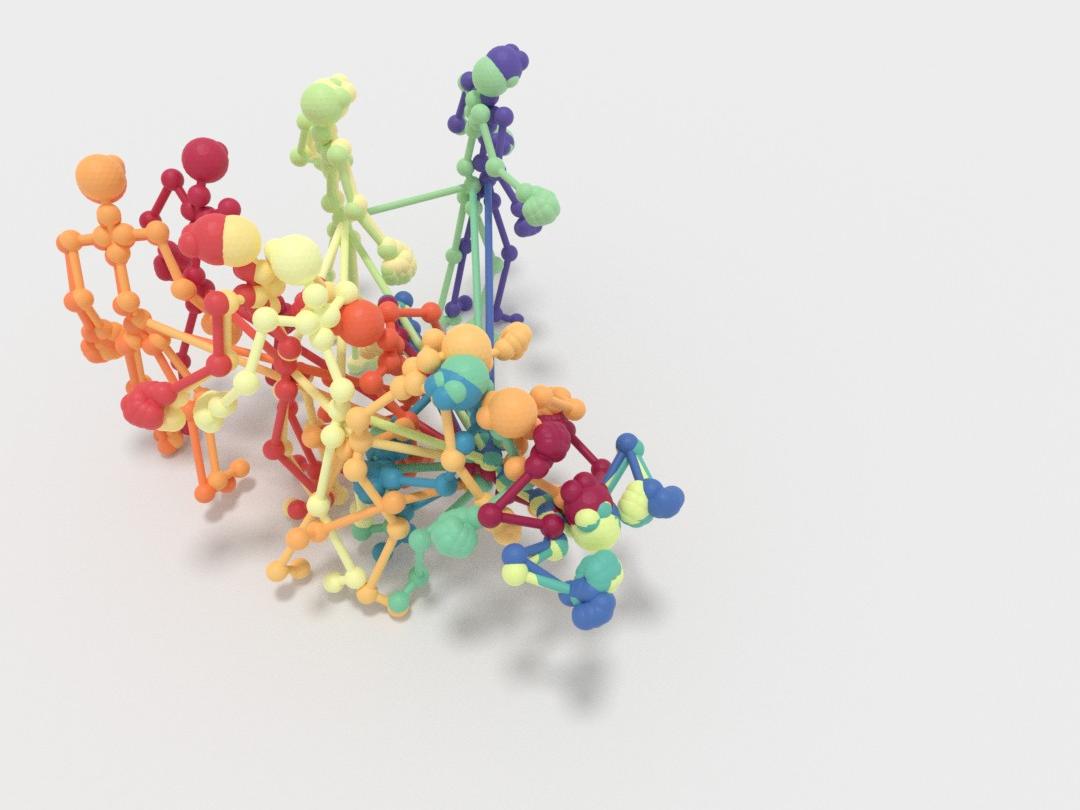}
		\caption{Input}
	\end{subfigure}
	\begin{subfigure}[t]{0.159\textwidth}
		\includegraphics[width=\textwidth]
		{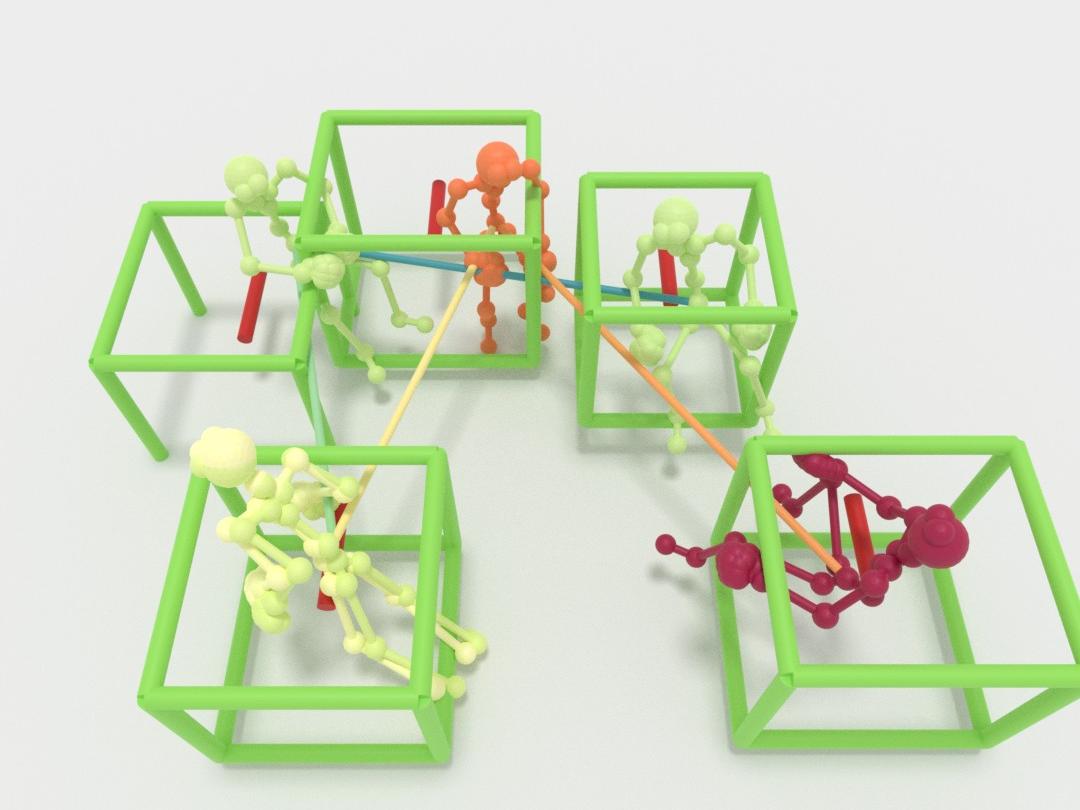}
		\includegraphics[width=\textwidth]
		{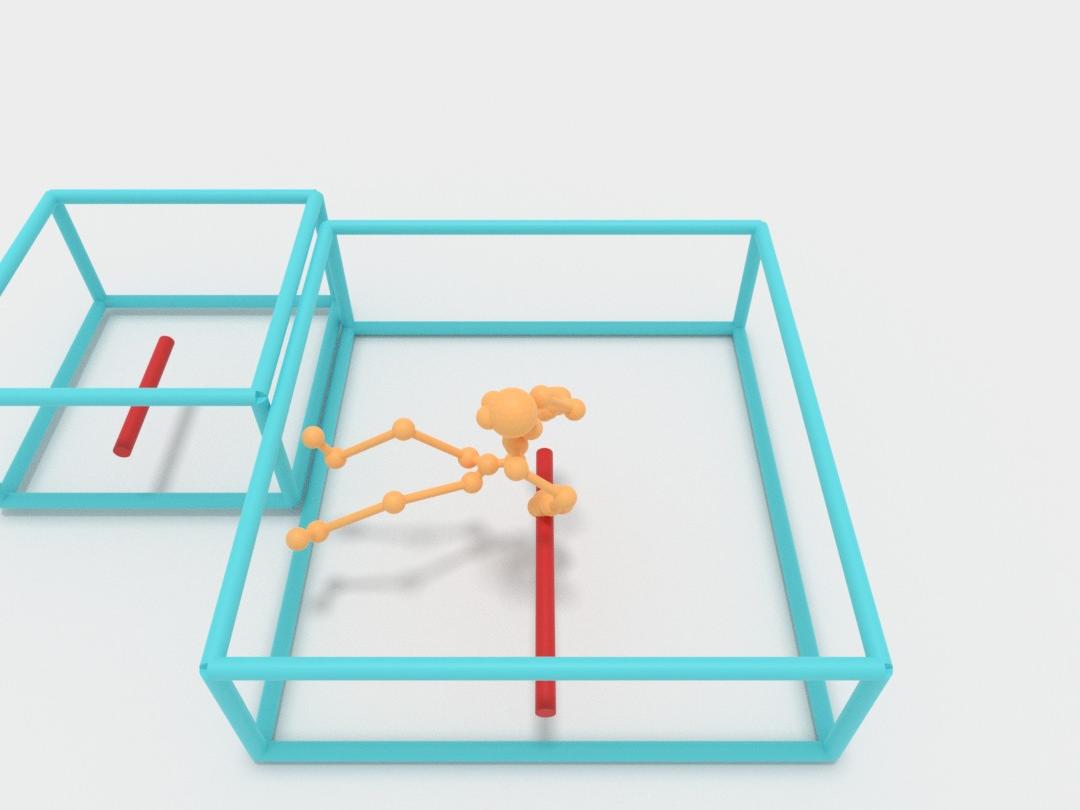}
		\caption{P-Vote.}
	\end{subfigure}
	\begin{subfigure}[t]{0.159\textwidth}
		\includegraphics[width=\textwidth]
		{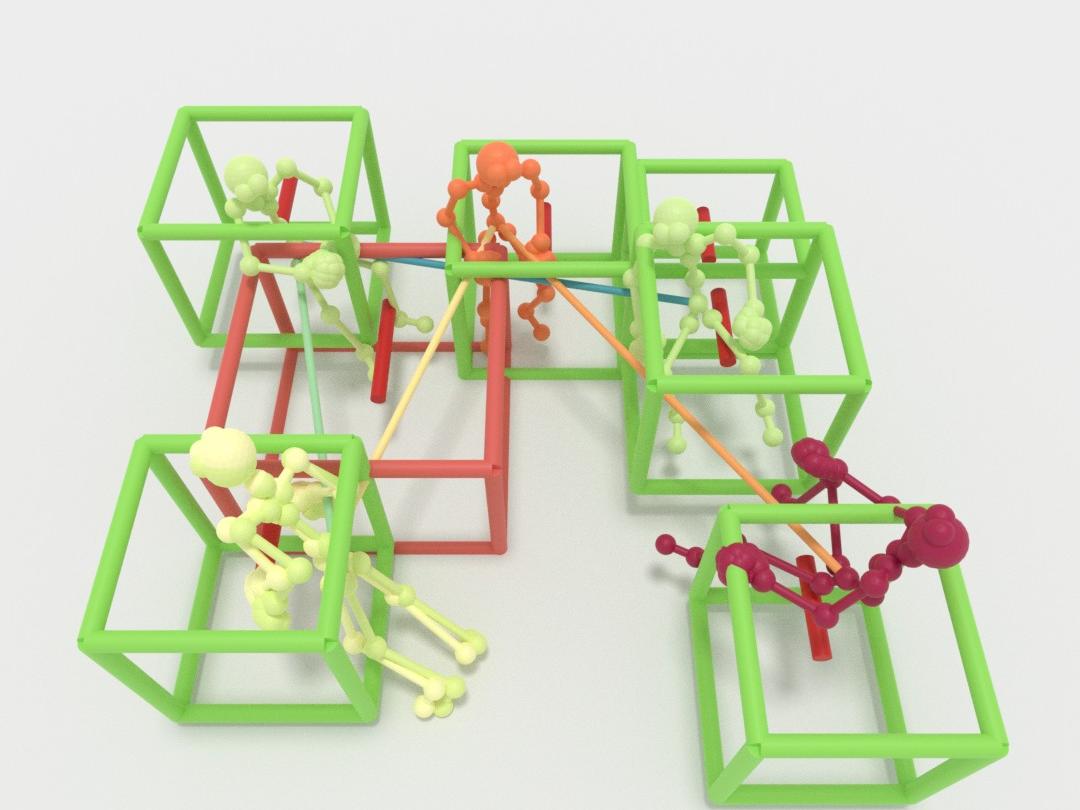}
		\includegraphics[width=\textwidth]
		{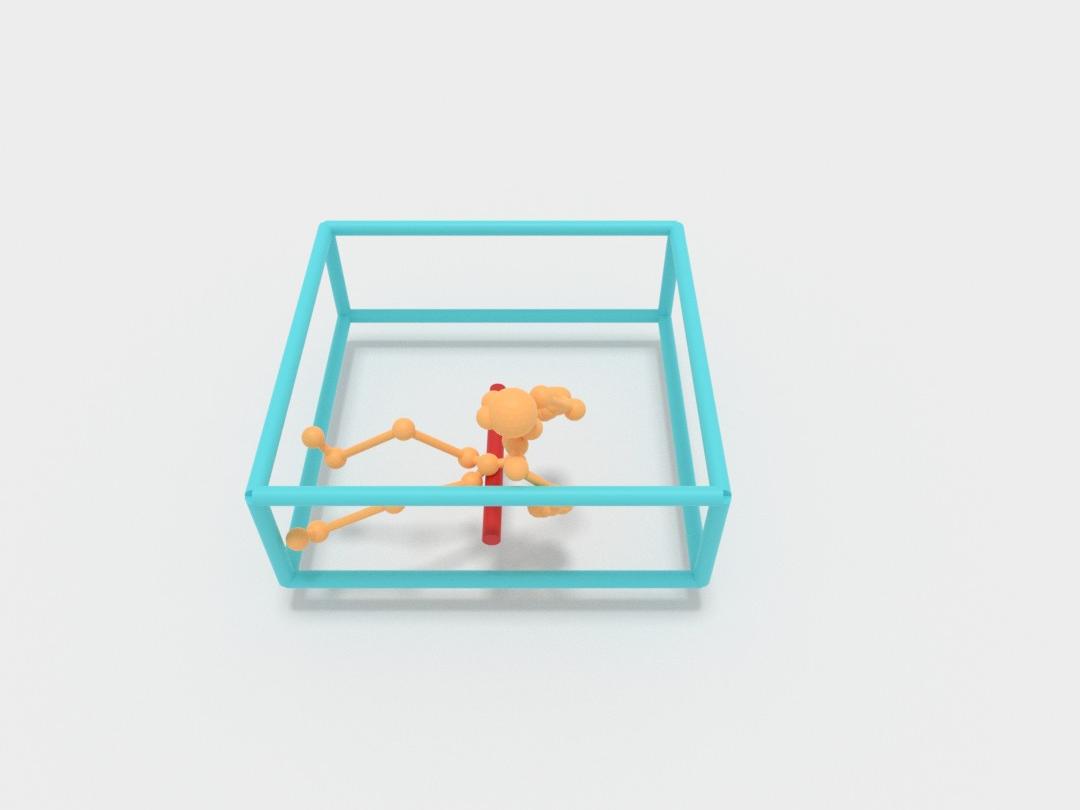}
		\caption{P-VN}
	\end{subfigure}
	\begin{subfigure}[t]{0.159\textwidth}
		\includegraphics[width=\textwidth]
		{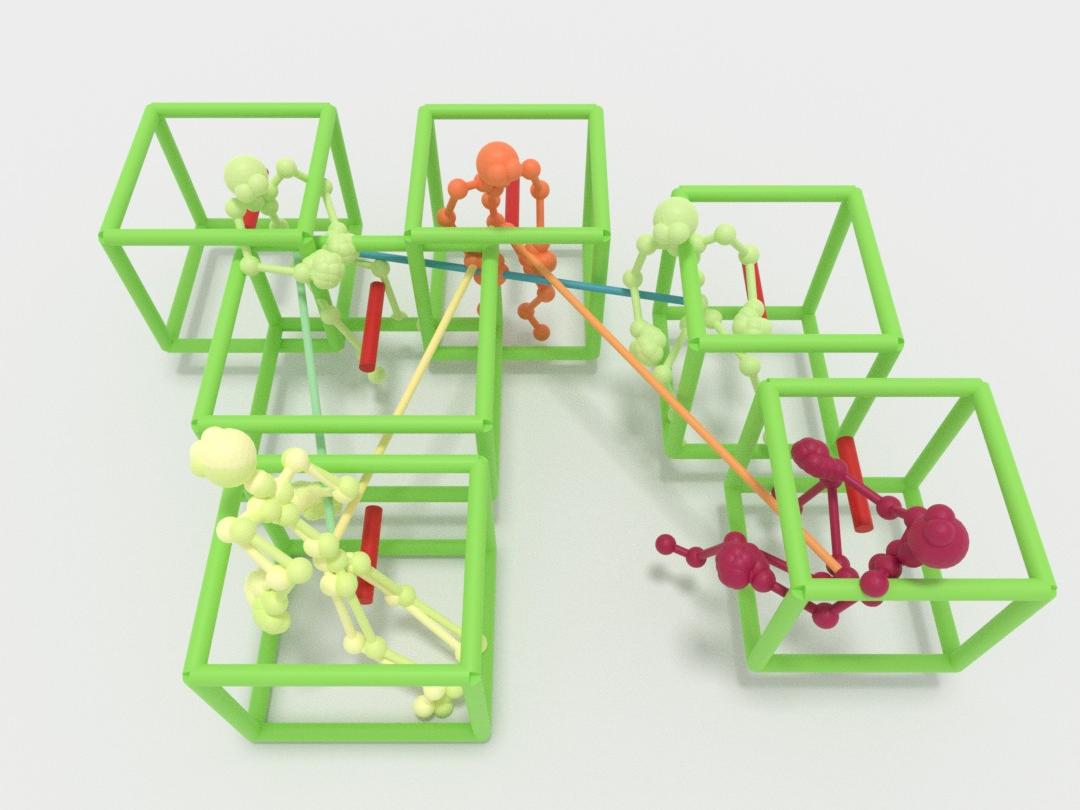}
		\includegraphics[width=\textwidth]
		{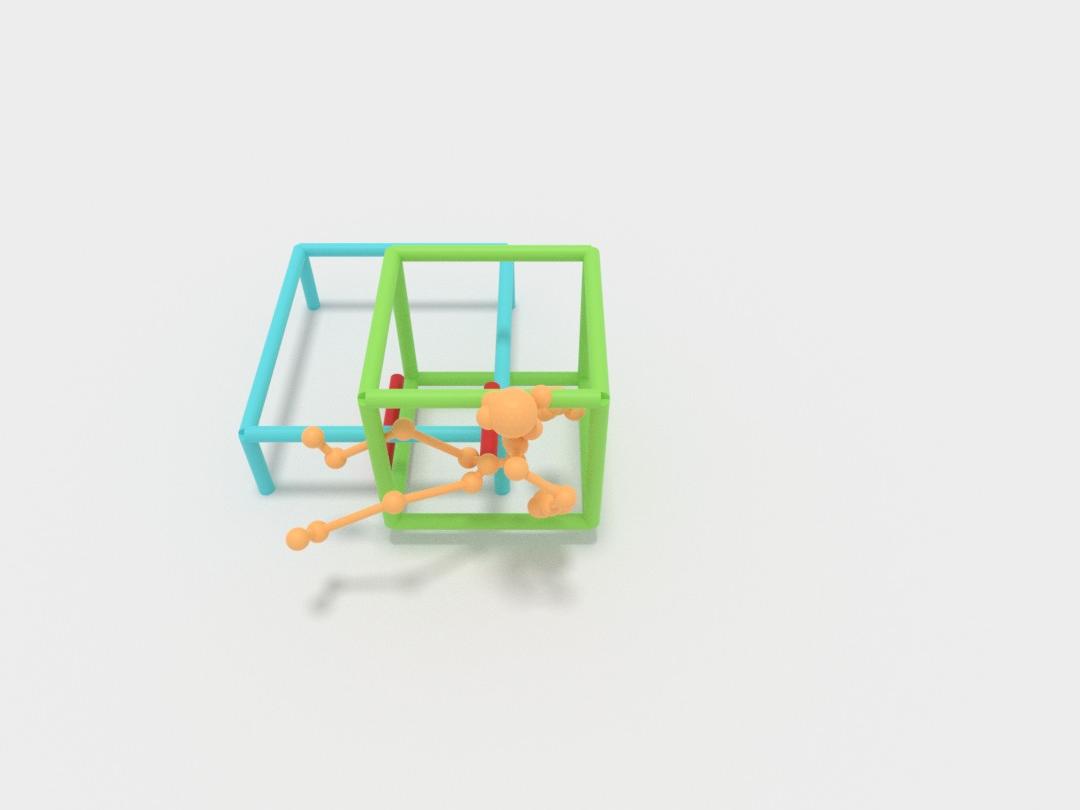}
		\caption{Mo. Attn}
	\end{subfigure}
	\begin{subfigure}[t]{0.159\textwidth}
		\includegraphics[width=\textwidth]
		{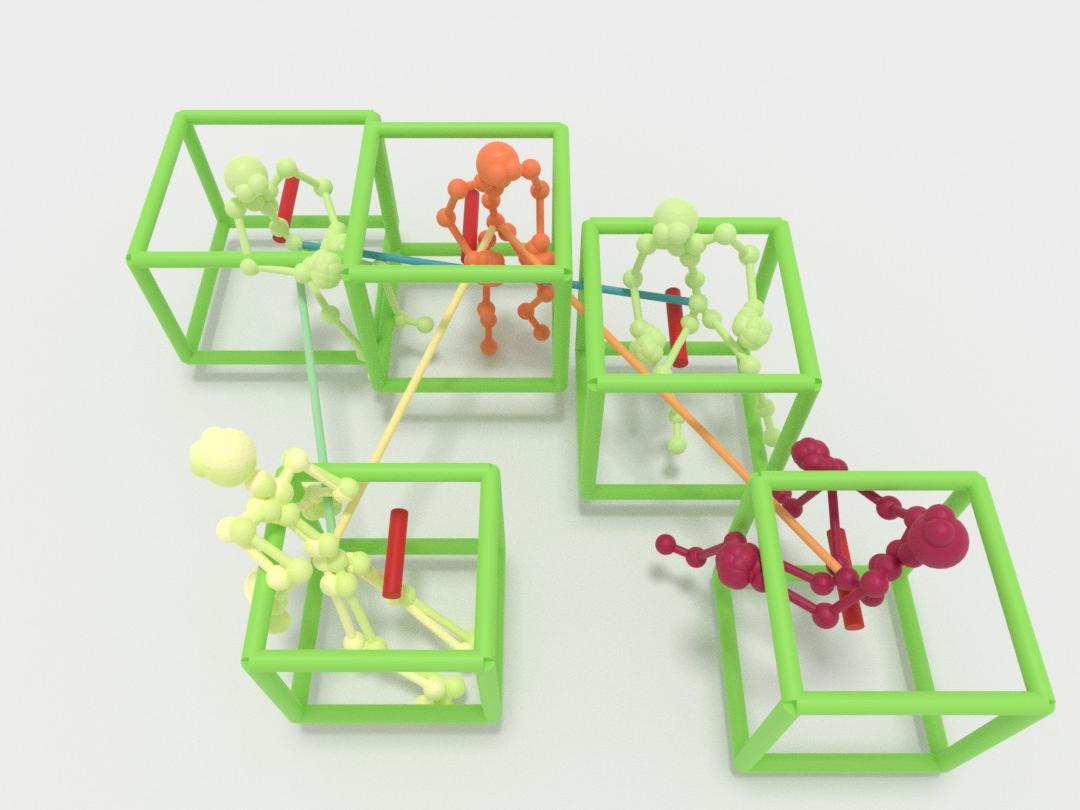}
		\includegraphics[width=\textwidth]
		{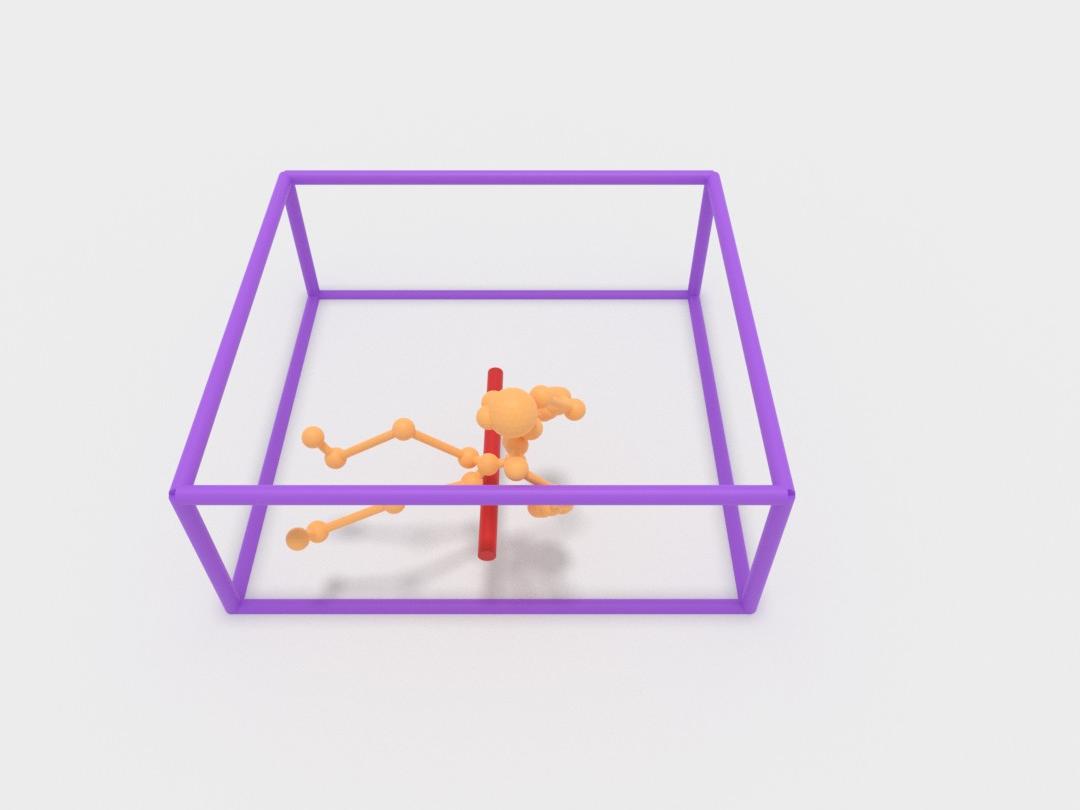}
		\caption{Ours}
	\end{subfigure}
	\begin{subfigure}[t]{0.159\textwidth}
		\includegraphics[width=\textwidth]
		{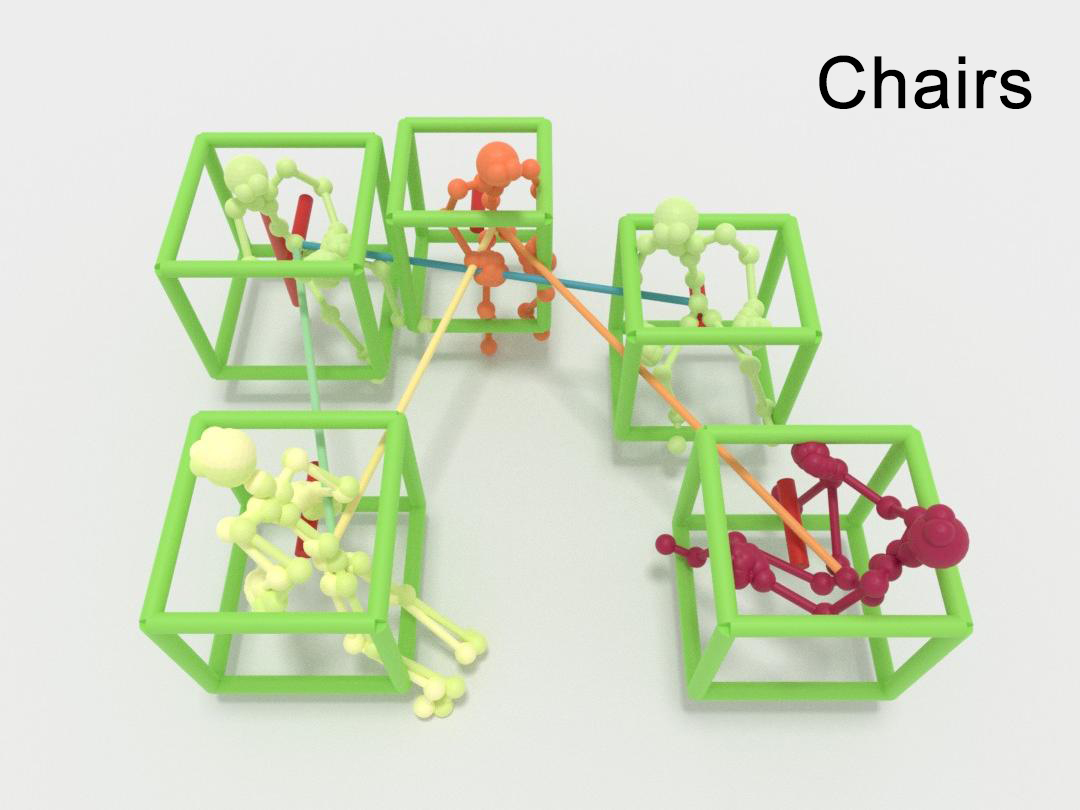}
		\includegraphics[width=\textwidth]
		{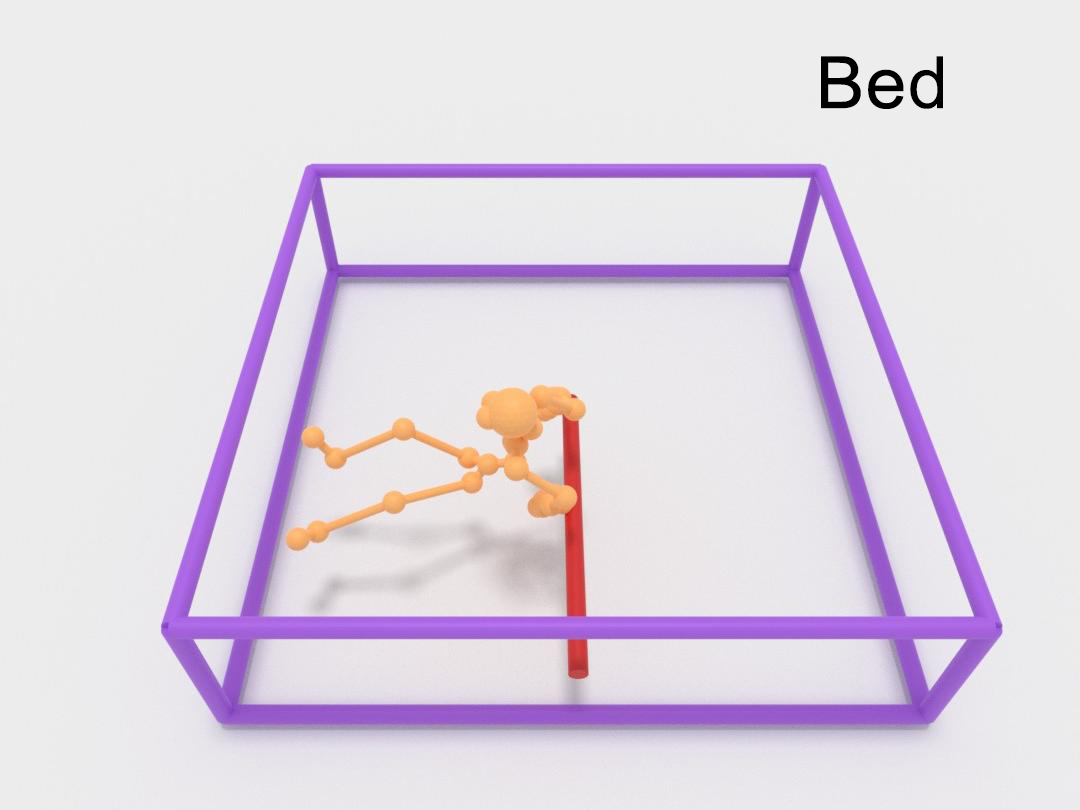}
		\caption{GT}
	\end{subfigure}
	\caption{Qualitative results of object detection on the PROX dataset~\cite{hassan2019resolving,yi2022mover}.
	}
	\label{fig:quali_comp_prox}
\end{figure}

\begin{figure*}[!ht]
	\centering
	\includegraphics[width=\textwidth]
	{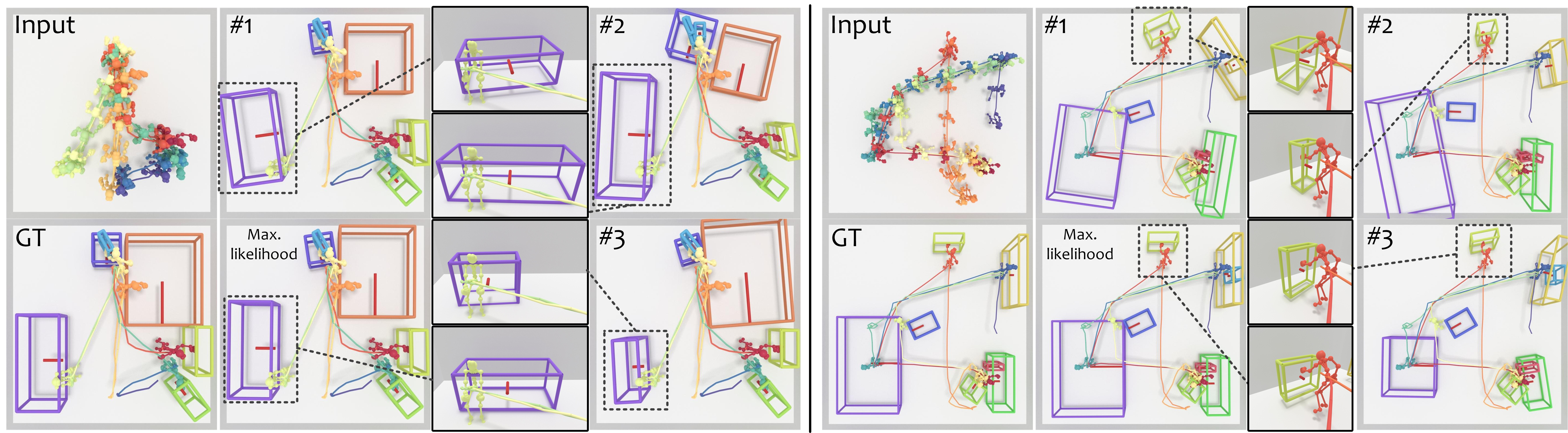}
	\caption{Multi-modal predictions of \OURS{}. By sampling our decoder multiple times, we can obtain different plausible box predictions. Here, we show three randomly sampled hypotheses and the max. likelihood prediction for each input.}
	\label{fig:quali_comp_mm}
\end{figure*}

\noindent\textbf{Multi-modal Predictions.} We visualize various sampled hypotheses from our model $\mathcal{S}_{1}$ in Fig.~\ref{fig:quali_comp_mm}, showing that our method is able to deduce the spatial occupancy of objects from motion trajectories, and enables diverse, plausible estimation of object locations, orientation, and sizes for interactions.

\noindent\textbf{Comparisons on PROX.} We additionally show qualitative results on real motion data from PROX~\cite{hassan2019resolving,yi2022mover} in Fig.\ref{fig:quali_comp_prox}. Since there are only 46 sequences labeled with object boxes, we pretrain each method on our dataset first before finetuning them on PROX. As PROX uses SMPL-X human body model~\cite{pavlakos2019expressive}, we manually map its skeleton joint indexes to ours for transfer learning. The results show that our approach can effectively handle real, noisy pose trajectory inputs.

\subsection{Quantitative Comparisons}
We use mAP@0.5 to measure object detection accuracy, and evaluate the accuracy and diversity of a set of output hypotheses using minimal matching distance (MMD) and total mutual diversity (TMD).

\begin{table*}[!t]
	\centering
	\resizebox{\columnwidth}{!}{\begin{tabular}{|l|c c c c c c c c c c c c|c|}
	\hline
	& bed & bench & cabinet & chair & desk & dishwasher & fridge & lamp  & sofa & stove & toilet & computer & mAP@0.5 \\
	\hline
	\hline
	Pose-VoteNet & 2.90 & 15.00 & 33.14 & 18.77 & 58.52 & 32.14 & 0.00 & \textbf{6.07} & 62.32 & 49.82 & 0.00 & 3.06 & 25.70  \\
	Pose-VN & 20.81 & \textbf{18.13} & 49.76 & 18.68 & 70.92 & 33.56 & 0.00 & 5.60 & 67.24 & 46.76 & 0.00 & 6.11 & 29.90 \\
	Motion Attention & 36.42 & 7.54 & 23.35 & \textbf{19.50} & 77.71 & 15.59 & 17.13 & 2.35 & 78.61 & 51.03 & 14.81 & 5.50 & 28.39  \\
	\hline
	\OURS{}-D & 93.77 & 12.63 & 11.98 & 5.77 & \textbf{95.93} & 61.80 & 73.95 & 0.58 & 88.44 & \textbf{70.42} & 0.00 & 14.17 & 34.91  \\
	\OURS{}-G & 91.69 & 7.56 & 36.61 & 10.05 & 93.47 & \textbf{67.53} & \textbf{77.45} & 1.21 & \textbf{92.97} & 64.86 & 5.56 & 18.67 & 37.48 \\
	\OURS{}-H & 85.84 & 8.04 & 22.04 & 10.91 & 76.08 & 55.20 & 55.15 & 0.00 & 83.92 & 57.60 & 5.00 & 4.33 & 31.41  \\
	\hline
	\OURS{} & \textbf{94.21} & 10.12 & \textbf{54.72} & 8.02 & 93.32 & 56.33 & 59.89 & 3.25 & 90.92 & 57.86 & \textbf{61.11} & \textbf{19.94} & \textbf{42.20}  \\
	\hline
	\end{tabular}}
	\caption{Quantitative evaluation on split $\mathcal{S}_{1}$. For \OURS{}-G, \OURS{}-H and ours, we use the maximum likelihood predictions to calculate mAP scores. The mAP@0.5 is averaged over all 17 classes (see the full table with all classes in the supplementary file).}
	\label{compare:det_compare_s1}
\end{table*}

\noindent\textbf{Detection Accuracy.}
Table~\ref{compare:det_compare_s1} shows a quantitative comparison on split $\mathcal{S}_{1}$, where we observe that Pose-VoteNet and Pose-VN, struggle to recognize some object categories (e.g., bed, fridge and toilet).
By leveraging the inter-frame connections, Motion Attention achieves improved performance in recognizing object classes, but struggles with detecting objectness and predicting object sizes.
In contrast, our position and pose encoder learns both the spatial and temporal signals from motions to estimate likely object locations by leveraging the potential connections between human and objects, with our probabilistic mixture decoder better capturing likely modalities in various challenging ambiguous interactions (e.g., toilet and cabinet). In Table~\ref{compare:multi_mode_compare_s1_s2}, we compare mAP@0.5 scores on split $\mathcal{S}_{2}$, with increased relative improvement in the challenging scenario of scene object configuration estimation in new rooms.

\begin{table}[!t]
	\parbox{.56\linewidth}{
		\centering
		\resizebox{0.56\columnwidth}{!}{
		\begin{tabular}{|l|c c c|c c c|}
			\hline
			& \multicolumn{3}{c|}{$\mathcal{S}_{1}$} & \multicolumn{3}{c|}{$\mathcal{S}_{2}$}\\
			& mAP & MMD & TMD & mAP & MMD & TMD \\
			\hline
			\hline
			Pose-VoteNet & 25.70 & - & - & 10.23 & - & - \\
			Pose-VN & 29.90 & - & - & 6.12 & - & -\\
			Motion Attn. & 28.39 & - & - & 6.53 & - & -\\
			\hline
			\OURS{}-D & 34.91 & - & - & 31.56 & - & -\\
			\OURS{}-G & 37.48 & 37.43 & 1.73 & 31.59 & 31.83 & 1.98\\
			\OURS{}-H & 31.41 & 21.10 & 2.39 & 27.96 & 20.71 & 3.28\\
			\hline
			\OURS{} & \textbf{42.20} & \textbf{38.28} & \textbf{3.34} & \textbf{35.34} & \textbf{32.47} & \textbf{4.35}\\
			\hline
		\end{tabular}}
		\caption{Comparisons on detection accuracy, and multi-modal quality and diversity on $\mathcal{S}_{1}$ and $\mathcal{S}_{2}$. TMD=1 indicates no diversity.}
		\label{compare:multi_mode_compare_s1_s2}
	}
	\hfill
	\parbox{.42\linewidth}{
		\centering
		\resizebox{0.42\columnwidth}{!}{
		\begin{tabular}{|l|c | c |}
			\hline
			& w/o pretrain & w/ pretrain \\
			\hline
			\hline
			Pose-VoteNet & 2.29 & 27.97 \\
			Pose-VN & 1.92 & 9.23 \\
			Motion Attn. & 2.68 & 6.25 \\
			\hline
			P2R-Net & 5.36 & \textbf{31.38} \\
			\hline
		\end{tabular}}
		\caption{Comparisons on detection accuracy (mAP@0.5) using PROX.}
		\label{compare:detection_comp_on_prox}
	}
\end{table}

\noindent\textbf{Quality and Diversity of Multi-modal Predictions.}
We study the multi-modal predictions with our ablation variants \OURS{}-G and \OURS{}-H, and use MMD and TMD to evaluate the quality and diversity of 10 randomly sampled predictions for each method.
From the 10 predictions, MMD records the best detection score (mAP@0.5), and TMD measures the average variance of the 10 semantic boxes per object.
Table~\ref{compare:multi_mode_compare_s1_s2} presents the MMD and TMD scores on $\mathcal{S}_{1}$ and $\mathcal{S}_{2}$ respectively. \OURS{}-G tends to predict with low diversity, as seen in low TMD score (TMD=1 indicates identical samples) and similar mAP@0.5 and MMD scores. \OURS{}-H shows better diversity but with lower accuracy in both of the two splits.
Our final model not only achieves best detection accuracy, it also provides reasonable and diverse object configurations, with a notable performance improvement in the challenging $\mathcal{S}_{2}$ split.

\noindent\textbf{Comparisons on PROX.} We quantitatively compare with baselines on real human motion data from PROX~\cite{hassan2019resolving,yi2022mover} in Table~\ref{compare:detection_comp_on_prox}.
We evaluate our method and baselines with and without pretraining on our synthetic dataset, considering the very limited number (46) of motion sequences in PROX. The results demonstrate that pretraining on our dataset can significantly improve all methods' performance on real data, with our approach outperforming all baselines.

\noindent\textbf{Ablations.}
In Table~\ref{compare:ablation}, we explore the effects of each individual module (relative position encoder, spatio-temporal pose encoder and probabilistic mixture decoder).
We ablate \OURS{} by gradually adding components from the baseline $\mathbf{c}_{0}$:  without relative position encoding (Pstn-Enc), where we use joints' global coordinates relative to the room center; without spatio-temporal pose encoding (Pose-Enc), where we use MLPs to learn pose features from joint coordinates; without probabilistic mixture decoder (P-Dec), where we use two-layer MLPs to regress box parameters.

\begin{table}[!t]
	\centering
	\begin{tabular}{|l|c c c|c|}
		\hline
		$\mathcal{S}_{1}/\mathcal{S}_{2}$ & Pstn-Enc & Pose-Enc & P-Dec & mAP@0.5 \\
		\hline
		\hline
		$\mathbf{c}_{0}$ &-&-&-& 8.71 / 3.12 \\
		$\mathbf{c}_{1}$ &\checkmark&-&-& 34.43 / 19.07 \\
		$\mathbf{c}_{2}$ &\checkmark&\checkmark&-& 39.98 / 29.25 \\
		\hline
		Full &\checkmark&\checkmark&\checkmark& \textbf{42.20} / \textbf{35.34} \\
		\hline
	\end{tabular}
	\caption{Ablation study of our design choices. Note that $\mathbf{c}_{1}$ and $\mathbf{c}_{2}$ are different with Pose-VoteNet and \OURS{-D} since our method parameterizes boxes differently from VoteNet (see Sec.~\ref{sec:prob_mix_decoder})}
	\label{compare:ablation}
\end{table}
From the comparisons, we observe that relative position encoding plays the most significant role. It allows our model to pick up on local pose signal, as many human-object interactions present with strong locality.
The spatio-temporal pose encoder then enhances the pose feature learning, and enables our model to learn the joint changes both in spatial and temporal domains.
This design largely improves our generalization ability, particularly in unseen rooms (from 19.07 to 29.25).
The probabilistic decoder further alleviates the ambiguity of this problem, and combining all the modules together produces the best results.

\noindent\textbf{Limitations.} Although \OURS{} achieves plausible scene estimations from only pose trajectories, it operates on several assumptions that can lead to potential limitations: (1) Objects should be interactable, e.g., our method may not detect objects that do not induce strong pose interactions like mirror or picture;
(2) Interactions occur at close range, e.g., we may struggle to detect a TV from a person switching on it with a remote control.
Additionally, we currently focus on estimating static object boxes.
We believe an interesting avenue for future work is to characterize objects in motion (e.g., due to grabbing) or articulated objects (e.g., laptops).

\section{Conclusion}
We have presented a first exploration to the ill-posed problem of estimating the 3D object configuration in a scene from only a 3D pose trajectory observation of a person interacting with the scene.
Our proposed model \OURS{} leverages spatio-temporal features from the pose trajectory to vote for likely object positions and inform a new probabilistic mixture decoder that captures multi-modal distributions of object box parameters.
We demonstrate that such a probabilistic approach can effectively model likely object configurations in scene, producing plausible object layout hypotheses from an input pose trajectory.
We hope that this establishes a step towards object-based 3D understanding of environments using non-visual signal and opens up new possibilities in leveraging ego-centric motion for 3D perception and understanding.

\section*{Acknowledgments}
This project is funded by the Bavarian State Ministry of Science and the Arts and coordinated by the Bavarian Research Institute for Digital Transformation (bidt), the TUM Institute of Advanced Studies (TUM-IAS), the ERC Starting Grant Scan2CAD (804724), and the German Research Foundation (DFG) Grant Making Machine Learning on Static and Dynamic 3D Data Practical.

\clearpage
%
%
\bibliographystyle{splncs04}
\bibliography{egbib}

\begin{thebibliography}{10}
\providecommand{\url}[1]{\texttt{#1}}
\providecommand{\urlprefix}{URL }
\providecommand{\doi}[1]{https://doi.org/#1}

\bibitem{unity3dskeleton}
Unity documentation: Humanbodybones.
  \url{https://docs.unity3d.com/ScriptReference/HumanBodyBones.html} (03 2020)

\bibitem{vh_homepage}
Virtualhome homepage. \url{http://virtual-home.org/} (11 2021)

\bibitem{achlioptas2018learning}
Achlioptas, P., Diamanti, O., Mitliagkas, I., Guibas, L.: Learning
  representations and generative models for 3d point clouds. In: International
  conference on machine learning. pp. 40--49. PMLR (2018)

\bibitem{agrawal2016task}
Agrawal, S., van~de Panne, M.: Task-based locomotion. ACM Transactions on
  Graphics (TOG)  \textbf{35}(4),  1--11 (2016)

\bibitem{cao2020long}
Cao, Z., Gao, H., Mangalam, K., Cai, Q.Z., Vo, M., Malik, J.: Long-term human
  motion prediction with scene context. In: European Conference on Computer
  Vision. pp. 387--404. Springer (2020)

\bibitem{chai2005performance}
Chai, J., Hodgins, J.K.: Performance animation from low-dimensional control
  signals. In: ACM SIGGRAPH 2005 Papers, pp. 686--696 (2005)

\bibitem{chao2019learning}
Chao, Y.W., Yang, J., Chen, W., Deng, J.: Learning to sit: Synthesizing
  human-chair interactions via hierarchical control. arXiv preprint
  arXiv:1908.07423  (2019)

\bibitem{choi2015robust}
Choi, S., Zhou, Q.Y., Koltun, V.: Robust reconstruction of indoor scenes. In:
  Proceedings of the IEEE Conference on Computer Vision and Pattern
  Recognition. pp. 5556--5565 (2015)

\bibitem{corona2020context}
Corona, E., Pumarola, A., Alenya, G., Moreno-Noguer, F.: Context-aware human
  motion prediction. In: Proceedings of the IEEE/CVF Conference on Computer
  Vision and Pattern Recognition. pp. 6992--7001 (2020)

\bibitem{dahnert2021panoptic}
Dahnert, M., Hou, J., , Nie{\ss}ner, M., Dai, A.: Panoptic 3d scene
  reconstruction from a single rgb image. Proc. Neural Information Processing
  Systems (NeurIPS)  (2021)

\bibitem{dai2017bundlefusion}
Dai, A., Nie{\ss}ner, M., Zollh{\"o}fer, M., Izadi, S., Theobalt, C.:
  Bundlefusion: Real-time globally consistent 3d reconstruction using
  on-the-fly surface reintegration. ACM Transactions on Graphics (ToG)
  \textbf{36}(4), ~1 (2017)

\bibitem{delaitre2012scene}
Delaitre, V., Fouhey, D.F., Laptev, I., Sivic, J., Gupta, A., Efros, A.A.:
  Scene semantics from long-term observation of people. In: European conference
  on computer vision. pp. 284--298. Springer (2012)

\bibitem{deng2021vn}
Deng, C., Litany, O., Duan, Y., Poulenard, A., Tagliasacchi, A., Guibas, L.:
  Vector neurons: a general framework for so(3)-equivariant networks. arXiv
  preprint arXiv:2104.12229  (2021)

\bibitem{deng20213d}
Deng, S., Xu, X., Wu, C., Chen, K., Jia, K.: 3d affordancenet: A benchmark for
  visual object affordance understanding. In: Proceedings of the IEEE/CVF
  Conference on Computer Vision and Pattern Recognition. pp. 1778--1787 (2021)

\bibitem{engel2014lsd}
Engel, J., Sch{\"o}ps, T., Cremers, D.: Lsd-slam: Large-scale direct monocular
  slam. In: European conference on computer vision. pp. 834--849. Springer
  (2014)

\bibitem{engelmann2021points}
Engelmann, F., Rematas, K., Leibe, B., Ferrari, V.: From points to multi-object
  3d reconstruction. In: Proceedings of the IEEE/CVF Conference on Computer
  Vision and Pattern Recognition. pp. 4588--4597 (2021)

\bibitem{fisher2015activity}
Fisher, M., Savva, M., Li, Y., Hanrahan, P., Nie{\ss}ner, M.: Activity-centric
  scene synthesis for functional 3d scene modeling. ACM Transactions on
  Graphics (TOG)  \textbf{34}(6),  1--13 (2015)

\bibitem{fouhey2012people}
Fouhey, D.F., Delaitre, V., Gupta, A., Efros, A.A., Laptev, I., Sivic, J.:
  People watching: Human actions as a cue for single view geometry. In:
  European Conference on Computer Vision. pp. 732--745. Springer (2012)

\bibitem{fowler2017towards}
Fowler, S., Kim, H., Hilton, A.: Towards complete scene reconstruction from
  single-view depth and human motion. In: BMVC (2017)

\bibitem{fowler2018human}
Fowler, S., Kim, H., Hilton, A.: Human-centric scene understanding from single
  view 360 video. In: 2018 International Conference on 3D Vision (3DV). pp.
  334--342. IEEE (2018)

\bibitem{glauser2019interactive}
Glauser, O., Wu, S., Panozzo, D., Hilliges, O., Sorkine-Hornung, O.:
  Interactive hand pose estimation using a stretch-sensing soft glove. ACM
  Transactions on Graphics (TOG)  \textbf{38}(4),  1--15 (2019)

\bibitem{grabner2011makes}
Grabner, H., Gall, J., Van~Gool, L.: What makes a chair a chair? In: CVPR 2011.
  pp. 1529--1536. IEEE (2011)

\bibitem{gupta20113d}
Gupta, A., Satkin, S., Efros, A.A., Hebert, M.: From 3d scene geometry to human
  workspace. In: CVPR 2011. pp. 1961--1968. IEEE (2011)

\bibitem{guzov2021human}
Guzov, V., Mir, A., Sattler, T., Pons-Moll, G.: Human poseitioning system
  (hps): 3d human pose estimation and self-localization in large scenes from
  body-mounted sensors. In: Proceedings of the IEEE/CVF Conference on Computer
  Vision and Pattern Recognition. pp. 4318--4329 (2021)

\bibitem{hassan2021stochastic}
Hassan, M., Ceylan, D., Villegas, R., Saito, J., Yang, J., Zhou, Y., Black,
  M.J.: Stochastic scene-aware motion prediction. In: Proceedings of the
  IEEE/CVF International Conference on Computer Vision. pp. 11374--11384 (2021)

\bibitem{hassan2019resolving}
Hassan, M., Choutas, V., Tzionas, D., Black, M.J.: Resolving 3d human pose
  ambiguities with 3d scene constraints. In: Proceedings of the IEEE/CVF
  International Conference on Computer Vision. pp. 2282--2292 (2019)

\bibitem{hassan2021populating}
Hassan, M., Ghosh, P., Tesch, J., Tzionas, D., Black, M.J.: Populating 3d
  scenes by learning human-scene interaction. In: Proceedings of the IEEE/CVF
  Conference on Computer Vision and Pattern Recognition. pp. 14708--14718
  (2021)

\bibitem{hu2016learning}
Hu, R., van Kaick, O., Wu, B., Huang, H., Shamir, A., Zhang, H.: Learning how
  objects function via co-analysis of interactions. ACM Transactions on
  Graphics (TOG)  \textbf{35}(4),  1--13 (2016)

\bibitem{huang2018holistic}
Huang, S., Qi, S., Zhu, Y., Xiao, Y., Xu, Y., Zhu, S.C.: Holistic 3d scene
  parsing and reconstruction from a single rgb image. In: Proceedings of the
  European Conference on Computer Vision (ECCV). pp. 187--203 (2018)

\bibitem{huang2018deep}
Huang, Y., Kaufmann, M., Aksan, E., Black, M.J., Hilliges, O., Pons-Moll, G.:
  Deep inertial poser: Learning to reconstruct human pose from sparse inertial
  measurements in real time. ACM Transactions on Graphics (TOG)
  \textbf{37}(6),  1--15 (2018)

\bibitem{jiang2013hallucinated}
Jiang, Y., Koppula, H., Saxena, A.: Hallucinated humans as the hidden context
  for labeling 3d scenes. In: Proceedings of the IEEE Conference on Computer
  Vision and Pattern Recognition. pp. 2993--3000 (2013)

\bibitem{jiang2015modeling}
Jiang, Y., Koppula, H.S., Saxena, A.: Modeling 3d environments through hidden
  human context. IEEE transactions on pattern analysis and machine intelligence
   \textbf{38}(10),  2040--2053 (2015)

\bibitem{jiang2012learning}
Jiang, Y., Lim, M., Saxena, A.: Learning object arrangements in 3d scenes using
  human context. arXiv preprint arXiv:1206.6462  (2012)

\bibitem{kapadia2016precision}
Kapadia, M., Xianghao, X., Nitti, M., Kallmann, M., Coros, S., Sumner, R.W.,
  Gross, M.: Precision: Precomputing environment semantics for contact-rich
  character animation. In: Proceedings of the 20th ACM SIGGRAPH Symposium on
  Interactive 3D Graphics and Games. pp. 29--37 (2016)

\bibitem{kaufmann2021pose}
Kaufmann, M., Zhao, Y., Tang, C., Tao, L., Twigg, C., Song, J., Wang, R.,
  Hilliges, O.: Em-pose: 3d human pose estimation from sparse electromagnetic
  trackers. In: Proceedings of the IEEE/CVF International Conference on
  Computer Vision. pp. 11510--11520 (2021)

\bibitem{kim2014shape2pose}
Kim, V.G., Chaudhuri, S., Guibas, L., Funkhouser, T.: Shape2pose: Human-centric
  shape analysis. ACM Transactions on Graphics (TOG)  \textbf{33}(4),  1--12
  (2014)

\bibitem{kuo2020mask2cad}
Kuo, W., Angelova, A., Lin, T.Y., Dai, A.: Mask2cad: 3d shape prediction by
  learning to segment and retrieve. In: Computer Vision--ECCV 2020: 16th
  European Conference, Glasgow, UK, August 23--28, 2020, Proceedings, Part III
  16. pp. 260--277. Springer (2020)

\bibitem{kuo2021patch2cad}
Kuo, W., Angelova, A., Lin, T.Y., Dai, A.: Patch2cad: Patchwise embedding
  learning for in-the-wild shape retrieval from a single image. In: Proceedings
  of the IEEE/CVF International Conference on Computer Vision. pp. 12589--12599
  (2021)

\bibitem{lee2002interactive}
Lee, J., Chai, J., Reitsma, P.S., Hodgins, J.K., Pollard, N.S.: Interactive
  control of avatars animated with human motion data. In: Proceedings of the
  29th annual conference on Computer graphics and interactive techniques. pp.
  491--500 (2002)

\bibitem{lee2006motion}
Lee, K.H., Choi, M.G., Lee, J.: Motion patches: building blocks for virtual
  environments annotated with motion data. In: ACM SIGGRAPH 2006 Papers, pp.
  898--906 (2006)

\bibitem{liu2011realtime}
Liu, H., Wei, X., Chai, J., Ha, I., Rhee, T.: Realtime human motion control
  with a small number of inertial sensors. In: Symposium on interactive 3D
  graphics and games. pp. 133--140 (2011)

\bibitem{mao2020history}
Mao, W., Liu, M., Salzmann, M.: History repeats itself: Human motion prediction
  via motion attention. In: European Conference on Computer Vision. pp.
  474--489. Springer (2020)

\bibitem{merel2020catch}
Merel, J., Tunyasuvunakool, S., Ahuja, A., Tassa, Y., Hasenclever, L., Pham,
  V., Erez, T., Wayne, G., Heess, N.: Catch \& carry: reusable neural
  controllers for vision-guided whole-body tasks. ACM Transactions on Graphics
  (TOG)  \textbf{39}(4),  39--1 (2020)

\bibitem{monszpart2019imapper}
Monszpart, A., Guerrero, P., Ceylan, D., Yumer, E., Mitra, N.J.: imapper:
  interaction-guided scene mapping from monocular videos. ACM Transactions On
  Graphics (TOG)  \textbf{38}(4),  1--15 (2019)

\bibitem{mur2015orb}
Mur-Artal, R., Montiel, J.M.M., Tardos, J.D.: Orb-slam: a versatile and
  accurate monocular slam system. IEEE transactions on robotics
  \textbf{31}(5),  1147--1163 (2015)

\bibitem{mura2021walk2map}
Mura, C., Pajarola, R., Schindler, K., Mitra, N.: Walk2map: Extracting floor
  plans from indoor walk trajectories. In: Computer Graphics Forum. vol.~40,
  pp. 375--388. Wiley Online Library (2021)

\bibitem{newcombe2011kinectfusion}
Newcombe, R.A., Izadi, S., Hilliges, O., Molyneaux, D., Kim, D., Davison, A.J.,
  Kohi, P., Shotton, J., Hodges, S., Fitzgibbon, A.: Kinectfusion: Real-time
  dense surface mapping and tracking. In: 2011 10th IEEE international
  symposium on mixed and augmented reality. pp. 127--136. IEEE (2011)

\bibitem{Nie_2020_CVPR}
Nie, Y., Han, X., Guo, S., Zheng, Y., Chang, J., Zhang, J.J.:
  Total3dunderstanding: Joint layout, object pose and mesh reconstruction for
  indoor scenes from a single image. In: IEEE/CVF Conference on Computer Vision
  and Pattern Recognition (CVPR) (June 2020)

\bibitem{nie2021rfd}
Nie, Y., Hou, J., Han, X., Nie{\ss}ner, M.: Rfd-net: Point scene understanding
  by semantic instance reconstruction. In: Proceedings of the IEEE/CVF
  Conference on Computer Vision and Pattern Recognition. pp. 4608--4618 (2021)

\bibitem{niessner2013real}
Nie{\ss}ner, M., Zollh{\"o}fer, M., Izadi, S., Stamminger, M.: Real-time 3d
  reconstruction at scale using voxel hashing. ACM Transactions on Graphics
  (ToG)  \textbf{32}(6),  1--11 (2013)

\bibitem{pavlakos2019expressive}
Pavlakos, G., Choutas, V., Ghorbani, N., Bolkart, T., Osman, A.A., Tzionas, D.,
  Black, M.J.: Expressive body capture: 3d hands, face, and body from a single
  image. In: Proceedings of the IEEE/CVF conference on computer vision and
  pattern recognition. pp. 10975--10985 (2019)

\bibitem{pieropan2013functional}
Pieropan, A., Ek, C.H., Kjellstr{\"o}m, H.: Functional object descriptors for
  human activity modeling. In: 2013 IEEE International Conference on Robotics
  and Automation. pp. 1282--1289. IEEE (2013)

\bibitem{popov2020corenet}
Popov, S., Bauszat, P., Ferrari, V.: Corenet: Coherent 3d scene reconstruction
  from a single rgb image. In: European Conference on Computer Vision. pp.
  366--383. Springer (2020)

\bibitem{puig2018virtualhome}
Puig, X., Ra, K., Boben, M., Li, J., Wang, T., Fidler, S., Torralba, A.:
  Virtualhome: Simulating household activities via programs. In: Proceedings of
  the IEEE Conference on Computer Vision and Pattern Recognition. pp.
  8494--8502 (2018)

\bibitem{Qi_2019_ICCV}
Qi, C.R., Litany, O., He, K., Guibas, L.J.: Deep hough voting for 3d object
  detection in point clouds. In: Proceedings of the IEEE/CVF International
  Conference on Computer Vision (ICCV) (October 2019)

\bibitem{qian2020associative3d}
Qian, S., Jin, L., Fouhey, D.F.: Associative3d: Volumetric reconstruction from
  sparse views. In: European Conference on Computer Vision. pp. 140--157.
  Springer (2020)

\bibitem{ruiz2018can}
Ruiz, E., Mayol-Cuevas, W.: Where can i do this? geometric affordances from a
  single example with the interaction tensor. In: 2018 IEEE International
  Conference on Robotics and Automation (ICRA). pp. 2192--2199. IEEE (2018)

\bibitem{runz2020frodo}
Runz, M., Li, K., Tang, M., Ma, L., Kong, C., Schmidt, T., Reid, I., Agapito,
  L., Straub, J., Lovegrove, S., et~al.: Frodo: From detections to 3d objects.
  In: Proceedings of the IEEE/CVF Conference on Computer Vision and Pattern
  Recognition. pp. 14720--14729 (2020)

\bibitem{savva2014scenegrok}
Savva, M., Chang, A.X., Hanrahan, P., Fisher, M., Nie{\ss}ner, M.: Scenegrok:
  Inferring action maps in 3d environments. ACM transactions on graphics (TOG)
  \textbf{33}(6),  1--10 (2014)

\bibitem{savva2016pigraphs}
Savva, M., Chang, A.X., Hanrahan, P., Fisher, M., Nie{\ss}ner, M.: Pigraphs:
  learning interaction snapshots from observations. ACM Transactions on
  Graphics (TOG)  \textbf{35}(4),  1--12 (2016)

\bibitem{sawatzky2017weakly}
Sawatzky, J., Srikantha, A., Gall, J.: Weakly supervised affordance detection.
  In: Proceedings of the IEEE Conference on Computer Vision and Pattern
  Recognition. pp. 2795--2804 (2017)

\bibitem{shoaib2014estimating}
Shoaib, M., Yang, M.Y., Rosenhahn, B., Ostermann, J.: Estimating layout of
  cluttered indoor scenes using trajectory-based priors. Image and Vision
  Computing  \textbf{32}(11),  870--883 (2014)

\bibitem{shum2008interaction}
Shum, H.P., Komura, T., Shiraishi, M., Yamazaki, S.: Interaction patches for
  multi-character animation. ACM Transactions on Graphics (TOG)
  \textbf{27}(5), ~1--8 (2008)

\bibitem{starke2019neural}
Starke, S., Zhang, H., Komura, T., Saito, J.: Neural state machine for
  character-scene interactions. ACM Trans. Graph.  \textbf{38}(6),  209--1
  (2019)

\bibitem{von2017sparse}
Von~Marcard, T., Rosenhahn, B., Black, M.J., Pons-Moll, G.: Sparse inertial
  poser: Automatic 3d human pose estimation from sparse imus. In: Computer
  Graphics Forum. vol.~36, pp. 349--360. Wiley Online Library (2017)

\bibitem{wang2021synthesizing}
Wang, J., Xu, H., Xu, J., Liu, S., Wang, X.: Synthesizing long-term 3d human
  motion and interaction in 3d scenes. In: Proceedings of the IEEE/CVF
  Conference on Computer Vision and Pattern Recognition. pp. 9401--9411 (2021)

\bibitem{wang2019geometric}
Wang, Z., Chen, L., Rathore, S., Shin, D., Fowlkes, C.: Geometric pose
  affordance: 3d human pose with scene constraints. arXiv preprint
  arXiv:1905.07718  (2019)

\bibitem{wei2016modeling}
Wei, P., Zhao, Y., Zheng, N., Zhu, S.C.: Modeling 4d human-object interactions
  for joint event segmentation, recognition, and object localization. IEEE
  transactions on pattern analysis and machine intelligence  \textbf{39}(6),
  1165--1179 (2016)

\bibitem{whelan2015elasticfusion}
Whelan, T., Leutenegger, S., Salas-Moreno, R., Glocker, B., Davison, A.:
  Elasticfusion: Dense slam without a pose graph. Robotics: Science and Systems
  (2015)

\bibitem{wu2016learning}
Wu, J., Zhang, C., Xue, T., Freeman, W.T., Tenenbaum, J.B.: Learning a
  probabilistic latent space of object shapes via 3d generative-adversarial
  modeling. In: Proceedings of the 30th International Conference on Neural
  Information Processing Systems. pp. 82--90 (2016)

\bibitem{wu2020multimodal}
Wu, R., Chen, X., Zhuang, Y., Chen, B.: Multimodal shape completion via
  conditional generative adversarial networks. In: Computer Vision--ECCV 2020:
  16th European Conference, Glasgow, UK, August 23--28, 2020, Proceedings, Part
  IV 16. pp. 281--296. Springer (2020)

\bibitem{yan2018spatial}
Yan, S., Xiong, Y., Lin, D.: Spatial temporal graph convolutional networks for
  skeleton-based action recognition. In: Thirty-second AAAI conference on
  artificial intelligence (2018)

\bibitem{yi2022mover}
Yi, H., Huang, C.H.P., Tzionas, D., Kocabas, M., Hassan, M., Tang, S., Thies,
  J., Black, M.J.: Human-aware object placement for visual environment
  reconstruction. In: Computer Vision and Pattern Recognition (CVPR) (2022)

\bibitem{yin2021center}
Yin, T., Zhou, X., Krahenbuhl, P.: Center-based 3d object detection and
  tracking. In: Proceedings of the IEEE/CVF Conference on Computer Vision and
  Pattern Recognition. pp. 11784--11793 (2021)

\bibitem{Zhang_2021_CVPR}
Zhang, C., Cui, Z., Zhang, Y., Zeng, B., Pollefeys, M., Liu, S.: Holistic 3d
  scene understanding from a single image with implicit representation. In:
  Proceedings of the IEEE/CVF Conference on Computer Vision and Pattern
  Recognition (CVPR). pp. 8833--8842 (June 2021)

\bibitem{zhang2020place}
Zhang, S., Zhang, Y., Ma, Q., Black, M.J., Tang, S.: Place: Proximity learning
  of articulation and contact in 3d environments. In: 2020 International
  Conference on 3D Vision (3DV). pp. 642--651. IEEE (2020)

\bibitem{zhang2020generating}
Zhang, Y., Hassan, M., Neumann, H., Black, M.J., Tang, S.: Generating 3d people
  in scenes without people. In: Proceedings of the IEEE/CVF Conference on
  Computer Vision and Pattern Recognition. pp. 6194--6204 (2020)

\bibitem{zhu2015understanding}
Zhu, Y., Zhao, Y., Chun~Zhu, S.: Understanding tools: Task-oriented object
  modeling, learning and recognition. In: Proceedings of the IEEE Conference on
  Computer Vision and Pattern Recognition. pp. 2855--2864 (2015)

\end{thebibliography}

\appendix

\section*{Appendix}
In this supplementary material, we describe network parameters and specifications (Sec.~\ref{sec:network}), details of our data generation and distribution (Sec.~\ref{sec:data}), evaluation metrics in our experiments (Sec.~\ref{sec:metric}), additional quantitative results (Sec.~\ref{sec:quant}), additional qualitative results (Sec.~\ref{sec:quali}), tolerance to noise (Sec.~\ref{sec:tol_noise}),  and qualitative results on other real data (Sec~\ref{sec:realdata}). Our code and data are also attached with this file.

\section{Network Specifications}
\label{sec:network}
We detail the full list of network parameters and specifications in this section. MLP layers used in our network are uniformly denoted by $\text{MLP}[l_{1},l_{2},...,l_{d}]$, where $l_{i}$ is the neuron number in the $i$-th layer. Each layer is followed by a batch normalization and a ReLU layer except the final one. We also report the efficiency and memory usage during inference at the end. Our code will be publicly available.

\subsection{Skeleton Configuration}
In \OURS{}, the input is a pose trajectory with $N$ frames and $J$ joints as the sequence of 3D locations $\bm{T} \in \mathbb{R}^{N\times J\times 3}$, where $N=768$, $J=53$.
For each trajectory, the humanoid agent interacts with up to 10 different objects in a scene, with the frame number varying among sequences (depending on the object interactions). To enable mini-batch training, we uniformly sample $N$ frames per sequence for training.

\begin{figure}[!h]
	\centering
	\includegraphics[width=0.15\textwidth]{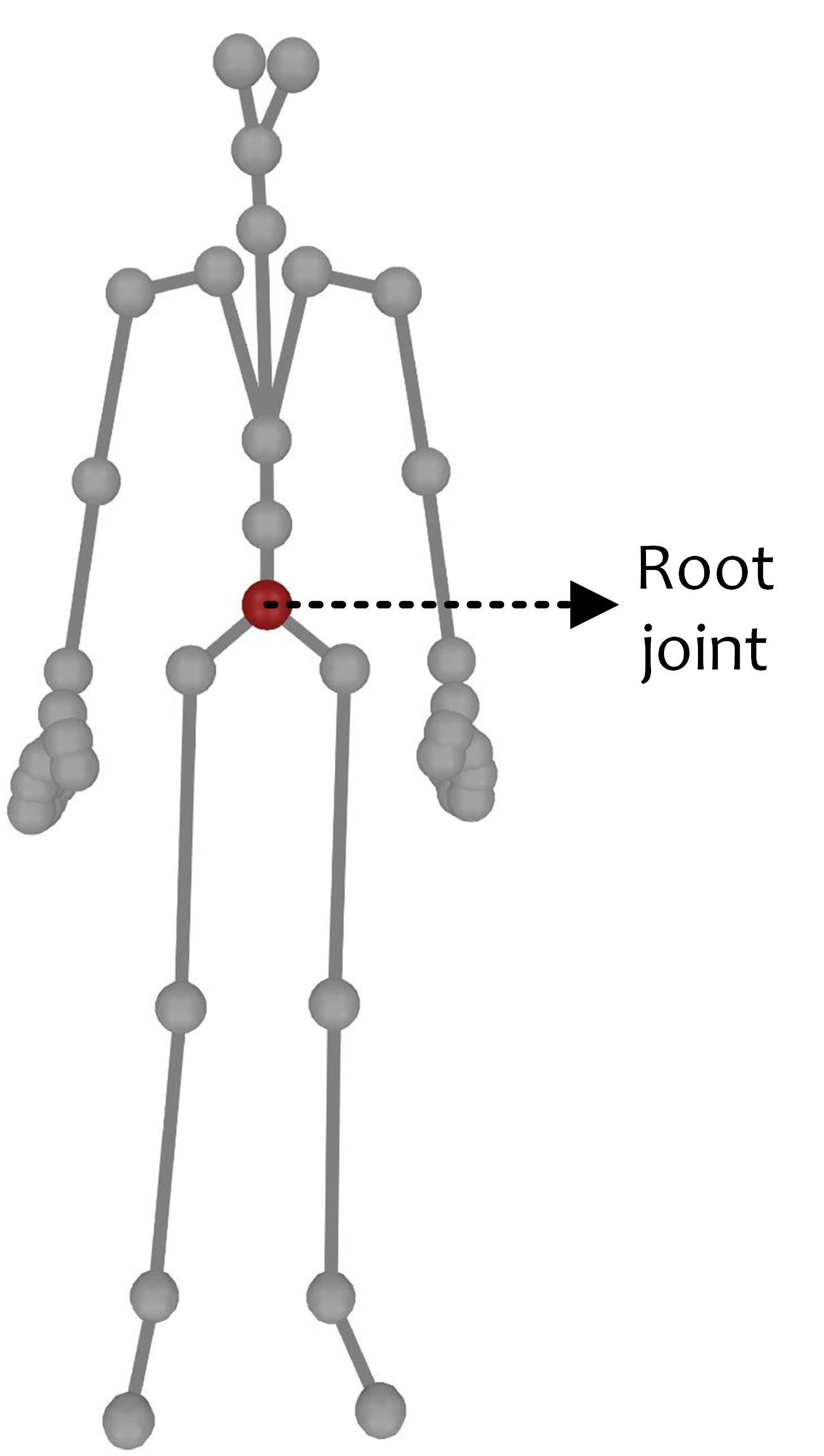}
	\caption{Human skeleton configuration, following VirtualHome~\cite{puig2018virtualhome}.
	}
	\label{fig:skeleton_config}
\end{figure}

For the human skeleton structure, we use the predefined human body model in VirtualHome \cite{puig2018virtualhome}, which uses the Unity3D human body template. We refer readers to \cite{unity3dskeleton} for the detailed definition of the skeleton specifications. The root joint $\bm{r}\in\mathbb{R}^{N\times 3}$ that we use is the centroid of the hips, as illustrated in Figure~\ref{fig:skeleton_config}.

\subsection{Relative Position Encoder}
\begin{figure}[!h]
	\centering
	\includegraphics[width=0.9\textwidth]{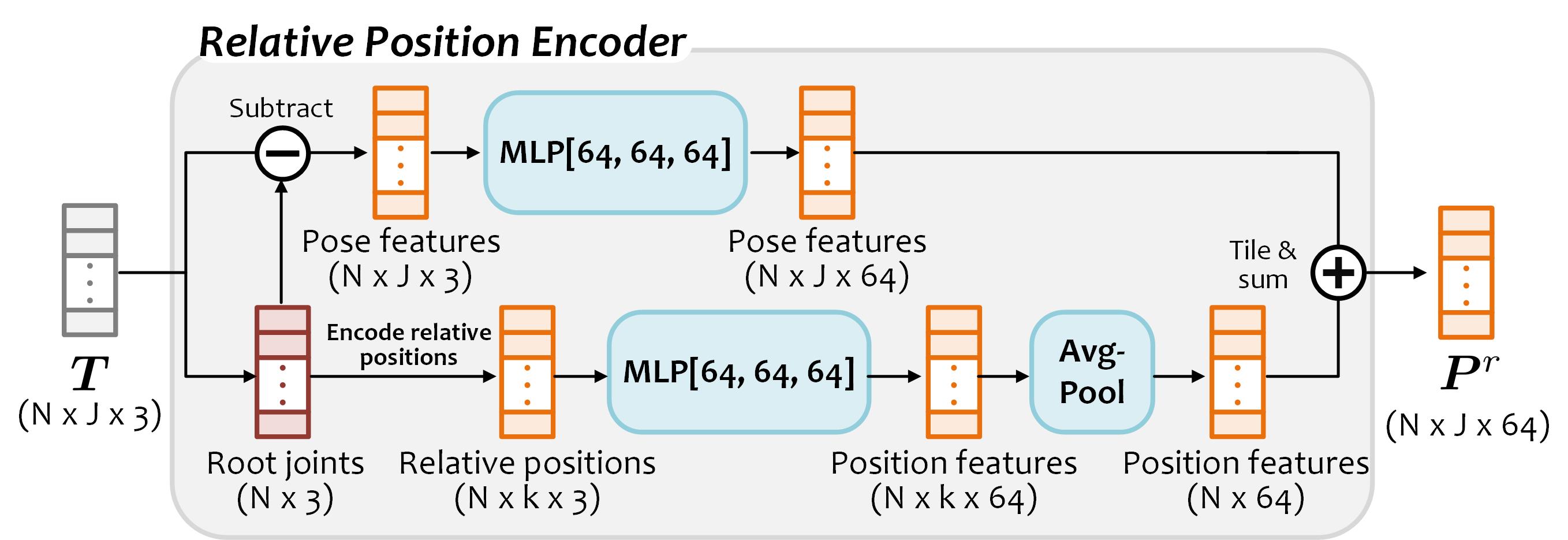}
	\caption{Relative position encoder.
	}
	\label{fig:pstn_encoder}
\end{figure}

We list the layer details of our relative position encoder in Figure~\ref{fig:pstn_encoder}.
In Section~3.1, the output pose feature dimension $d_{1}=64$, the number of temporal neighbors $k=20$.
From the input pose trajectory $\bm{T}$, it outputs the relative pose features $\bm{P}^{r}\in\mathbb{R}^{N\times J\times 64}$ for spatio-temporal encoding.

\subsection{Spatio-Temporal Pose Encoder}
\begin{figure}[!h]
	\centering
	\includegraphics[width=0.8\textwidth]{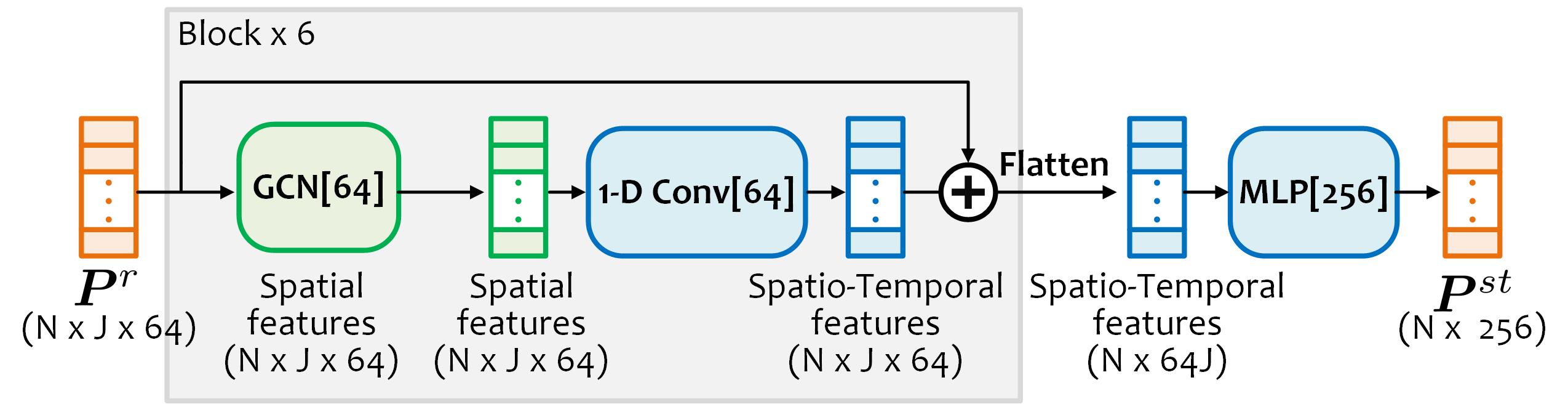}
	\caption{Spatio-temporal pose encoder.
	}
	\label{fig:st_encoder}
\end{figure}

Figure~\ref{fig:st_encoder} illustrates the layer details of our spatio-temporal pose encoder in Section~3.2.
From the input pose feature $\bm{P}^{r}$, we use a graph convolutional layer to learn intra-skeleton joint features.
Edges in the graph convolution are constructed following the skeleton bones (as in Figure~\ref{fig:skeleton_config}), which encodes skeleton-wise spatial information.
We use the method of \cite{yan2018spatial} to build the edge connections from skeleton bones.
With the learned joint features on each skeleton, we then adopt a 1-D convolutional layer (feature dimension at 64, kernel size at 3, padding size at 1) to process joint features in temporal domain.
The kernel size presents its receptive field on neighboring frames.
We connect a graph layer and a 1-D convolutional layer into a block with a residual summation from the input (see Figure~\ref{fig:st_encoder}).
We duplicate the block six times and stack them in a sequence to construct the spatio-temporal pose encoder.
After the spatio-temporal layers, we then flatten all joint features in a skeleton, which results in a $64J$-dimensional feature per pose, followed by an MLP[256] to process each pose and produce the final spatio-temporal pose features $\bm{P}^{st}\in\mathbb{R}^{N\times d_{2}}$. $d_{2}=256$.

\subsection{Locality-Sensitive Voting}
In Section~3.3 of the main paper, we sample $M$ seeds $\bm{r}_{s}$ from root joints $\bm{r}$, where $M=512$.
The $M$ seeds are uniformly sampled along the trajectory of root joints to ensure a even spatial distribution.
$\bm{P}_{s}^{st}$ are the corresponding pose features of $\bm{r}_{s}$, where $\bm{r}_{s}\in\mathbb{R}^{M\times3}$, $\bm{P}_{s}^{st}\in\mathbb{R}^{M\times d_{2}}$, $d_{2}=256$. In Eq.~2 of the paper, we use two MLPs, $f_{3}$ and $f_{4}$, to learn the vote locations and features $(\bm{v},\bm{P}^{v})$ from $(\bm{r}_{s},\bm{P}_{s}^{st})$, where $f_{3}$ and $f_{4}$ share the first two MLP layers and correspondingly predict their targets from the last layer. We illustrate them in Figure~\ref{fig:voting}.

From the 512 votes $(\bm{v},\bm{P}^{v})$, we group them into $V$ clusters following \cite{Qi_2019_ICCV}, which results in $(\bm{v}^{c},\bm{P}^{c})$ cluster centers and features, where $\bm{v}^{c}\in\mathbb{R}^{V\times3}$, $\bm{P}^{c}\in\mathbb{R}^{V\times256}$, $V$=128. For poses whose root joint is not close to any object during training (beyond a distance threshold $t_{d}$=1 m), we do not consider them to vote for any object.

\begin{figure}[!h]
	\centering
	\includegraphics[width=0.65\textwidth]{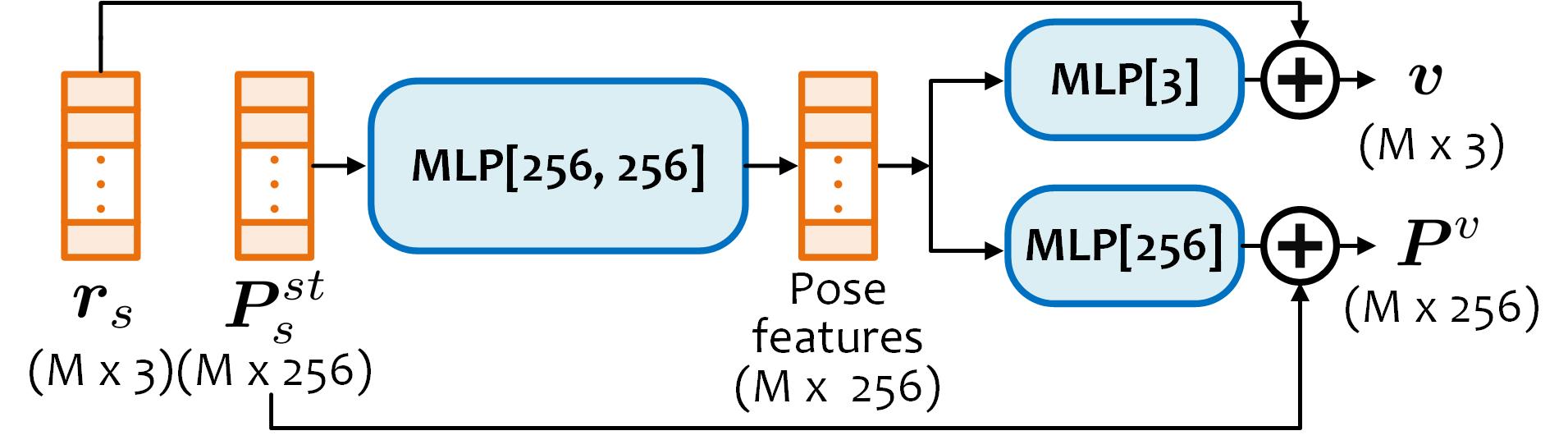}
	\caption{Locality-sensitive voting.
	}
	\label{fig:voting}
\end{figure}

\subsection{Probabilistic Mixture Decoder}
In Section~3.4 of the main paper, we learn multiple Gaussian distributions $\mathcal{N}(\mu_{\tau}^{k},\,\Sigma_{\tau}^{k})$ for each regression target $\tau\in\{c,s,\theta\}$, $k$=1,...,$P$, where $c,s,\theta$ respectively denote the box center, size and orientation; $P$ is the number of distributions ($P$=100). For each distribution, we also learn a mode score $f_{\tau}^{k}(*)\in[0,1]$ as its weight in predicting the target (see Eq.3 in the paper). $v^{c}\in\bm{v}^{c}$, $P^{c}\in\bm{P}^{c}$ represent a cluster center and feature, from which a proposal box is predicted.

The learnable parameters are $\{(\mu_{\tau},\,\Sigma_{\tau})\}$ and $\{f_{\tau}\}$, where $\mu_{\tau}\in\mathbb{R}^{P\times d_{r}}$; $\Sigma_{\tau}\in\mathbb{R}^{P\times d_{r}\times d_{r}}$.
For each target $\tau\in\{c,s,\theta\}$, $\mu_{\tau}$ represents the mean values of the $P$ Gaussian distributions, and $\Sigma_{\tau}$ stores the corresponding covariance matrices.
$\mu^{k}_{\tau}\in\mathbb{R}^{d_{\tau}}$ and $\Sigma^{k}_{\tau}\in\mathbb{R}^{d_{\tau}\times d_{\tau}}$ are the $k$-th item in $\mu_{\tau}$ and $\Sigma_{\tau}$ respectively.
$d_{\tau}$ is the dimension for each target, where $d_{c}$=3; $d_{s}$=3; $d_{\theta}$=2.
Here we consider variables in each Gaussian are independently distributed, resulting in a diagonal covariance matrix. We formulate $\{(\mu_{\tau},\,\Sigma_{\tau})\}$ as learnable embeddings shared among all samples in training and testing.
Since covariance matrix in $\Sigma_{\tau}$ is diagonal and non-negative, we use a exponential function to process it diagonal elements.

$f_{\tau}$ is realized with an MLP layer to predict mode scores for each target, where we use MLP[128,128,128,100] appended with a sigmoid layer.

For the proposal objectness and probability distribution of class category, we predict them directly from $P^{c}$ with MLP[128,128,$d_{obj}$+$d_{cls}$], where $d_{obj}$=2 and $d_{cls}$=17 (the number of object categories), followed by a softmax layer to get their probability scores in [0, 1].

To generate a hypothesis for each object during inference, we sample box parameters with Eq.~4 in our paper. Each hypothesis is an average of $N_{s}$ samples of $y_{\tau}$. $N_{s}$ is also a random number in $[1,100]$.  Additionally, we sample the object class based on the predicted classification probability distribution.

We discard proposed object boxes which have low objectness scores ($\leq t_{o}$) after a 3D non-maximum suppression ($t_{o}=0.5$), which then outputs $N_{h}$ hypotheses in a scene.

\subsection{Efficiency and Memory in Inference}
We train our network with batch size of 32 with 4 NVIDIA RTX 2080 GPUs using PyTorch 1.7.1, and test it with a single GPU. We report the model size, inference timing and allocated GPU memory in a single forward pass.
\begin{table}[!h]
	\centering
	\begin{tabular}{|c c c c c|}
		\hline
		Model size & Avg. time & Peak. time & Avg. memory & Peak memory \\
		\hline
		\hline
		2.04 M & 0.092 s & 0.582 s & 11.64 MB & 260.82 MB \\
		\hline
	\end{tabular}
	\caption{Model size, efficiency and memory usage of \OURS{}.}
	\label{compare:det_compare_s2}
\end{table}

\subsection{Specifications of Baselines}
We explain the network details of each baseline as the following.
\paragraph{Pose-VoteNet} Pose-VoteNet is a variant of VoteNet~\cite{Qi_2019_ICCV} to make it able to vote for object centers from human poses. In our experiments, we replace the PointNet++ in original VoteNet with our position encoder, which produces relative pose feature $\bm{P}^{r}\in\mathbb{R}^{N\times J\times 64}$. We then flatten all joint features for each pose ($\mathbb{R}^{N \times 64J}$) and learn the seed feature with MLP[256, 256, 256]. The coordinates of each seed is located at the root joint, similar with ours. For the remaining structures and loss functions, we keep them consistent with VoteNet.

\paragraph{Pose-VN} To make Pose-VoteNet able to capture rotation information of poses, we augment it by replacing MLP layers in Pose-VoteNet encoder with vector neurons \cite{deng2021vn}. Vector neurons are a set of SO(3)-equivariant operators that capture arbitrary rotations of object poses to estimate objects. For each MLP layer in Pose-VoteNet encoder, we replace it with a `VNLinearLeakyReLU' layer, and the final MLP layer with a `VNLinear' layer, with equal number of parameters. For the details of layer design in vector neurons, we refer readers to \cite{deng2021vn}.

\paragraph{Motion Attention} We replace the MLP layers (i.e., MLP[256, 256, 256]) in Pose-VoteNet encoder with the attention module in \cite{mao2020history} to learn inter-frame pose features in the entire temporal domain. Specifically, for each pose feature in $\bm{P}^{r}$, we use it to query similar features among all frames, which assembles repetitive pose patterns to regress object boxes. For the layer details in motion attention, we refer readers to \cite{mao2020history}.

\paragraph{\OURS{-D}} We replace our probabilistic decoder with the VoteNet decoder \cite{Qi_2019_ICCV} along with their loss functions to produce deterministic results.

\paragraph{\OURS{-G}} We ablate the \OURS{} decoder with a probabilistic generative model \cite{wu2016learning,nie2021rfd}, where we first lean a latent code $\bm{z}\sim \mathcal{N}(\bm{0},\,\bm{1})$ from cluster features $\bm{P}^{c}$. By decoding from the summation of $\bm{z}$ and $P^{c}\in\bm{P}^{c}$, we can predict box parameters in a probabilistic generative way.

\paragraph{\OURS{-H}} We discretize each regression target into a binary heatmap, where box centers are discretized into $10^3$ bins in [-0.3 m, 0.3 m]\textsuperscript{3}, centered at cluster centers $\bm{v}^{c}$; box sizes are discretized into $10^3$ bins in [0.05 m, 3m]\textsuperscript{3}; box orientations are discretized into 12 bins in [-$\pi$, $\pi$]. Then the box regression is converted into a classification task. We train them by cross-entropy losses. In testing, we sample the heatmaps to produce different predictions.

\section{Data Generation and Data Statistics}
\label{sec:data}
We create our dataset using the VirtualHome simulation environment \cite{puig2018virtualhome}, which is built on the Unity3D game engine. It consists of 29 rooms, where each room has 88 objects on average.
Each object is annotated with available interaction types. For the detailed specification of interaction manners for different object categories, we refer to \cite{puig2018virtualhome,vh_homepage}.

VirtualHome allows users to customize action scripts to direct humanoid agents to execute a series of complex interactive
tasks. In our work, we focus on the static, interactable objects under 17 common class categories (i.e., bed, bench, bookshelf, cabinet, chair, desk, dishwasher, faucet, fridge, lamp, microwave, monitor, nightstand, sofa, stove, toilet, computer).

In each room, we select up to 10 random objects in the scene, and script the agent to interact with each of the objects in a sequential fashion. For each object, we also select a random interaction type associated with the object class category. Sequences are trained with and evaluated against only the objects that are interacted with during the input observation, resulting in different variants of each room under different interaction sequences.

\begin{figure}[!t]
	\centering
	\includegraphics[width=0.8\textwidth]{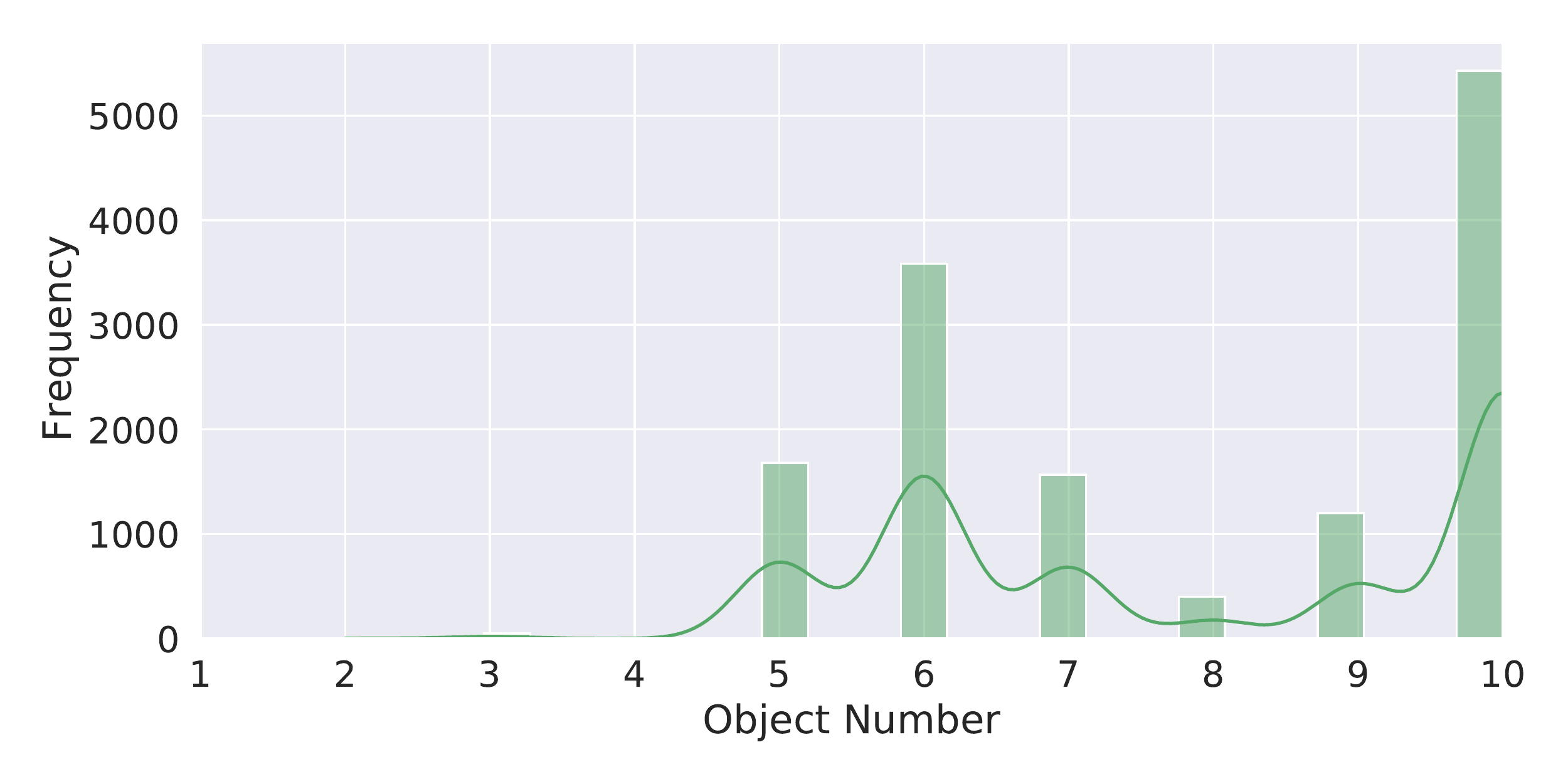}
	\caption{Distribution over number of objects in a pose trajectory.
	}
	\label{fig:obj_num_hist}
\end{figure}

\begin{figure*}[!t]
	\centering
	\includegraphics[width=1\textwidth]{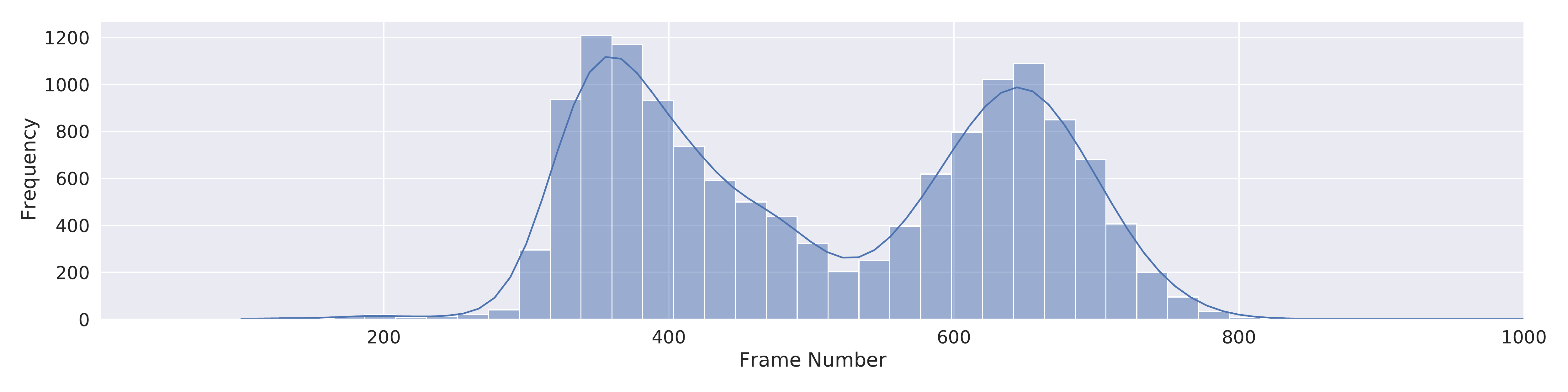}
	\caption{Distribution of frame lengths in pose trajectories.
	}
	\label{fig:frame_num_hist}
\end{figure*}

\begin{figure*}[!t]
	\centering
	\includegraphics[width=1\textwidth]{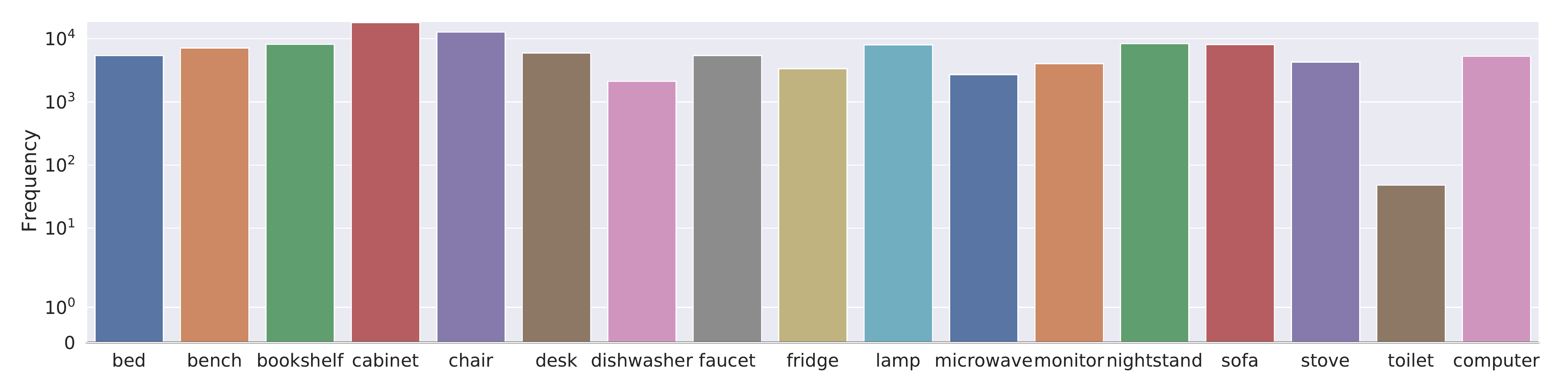}
	\caption{Frequency of object class categories among generated interactions.
	}
	\label{fig:interact_cls_hist}
\end{figure*}

\begin{figure*}[!t]
	\centering
	\includegraphics[width=1\textwidth]{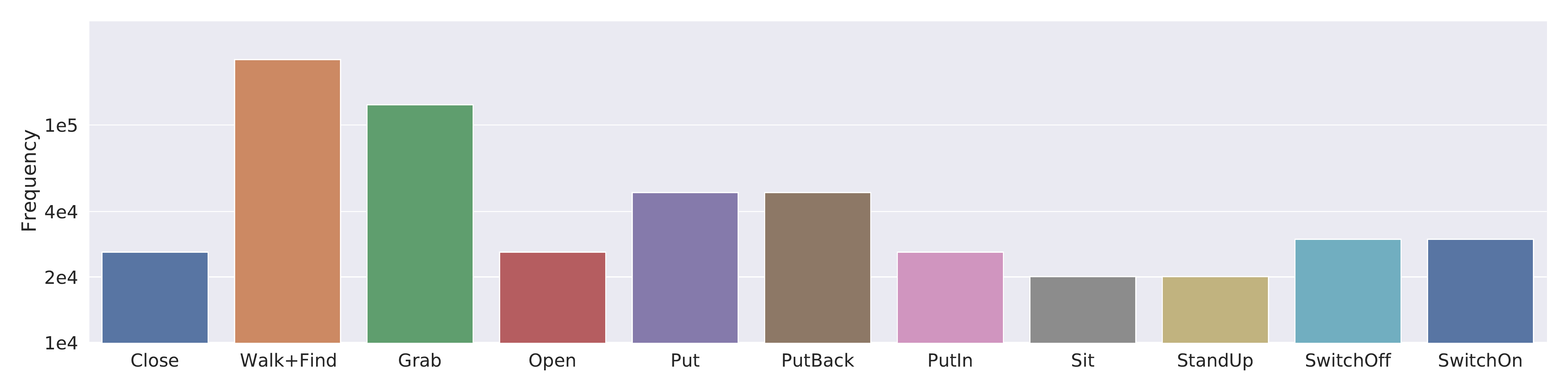}
	\caption{Frequency of interaction types.
	}
	\label{fig:interact_act_hist}
\end{figure*}

Then we randomly sample 13,913 different sequences with corresponding object boxes to construct the dataset. The human pose trajectories are recorded with a frame rate of 5 frames per second. Over the sequences, the average number of objects is 7.86, and the average frame length is 509.34.
The distributions of the frame length and object number in a interaction trajectory are shown in Figure~\ref{fig:frame_num_hist} and Figure~\ref{fig:obj_num_hist}.
The interaction frequency for each object class category is illustrated in Figure~\ref{fig:interact_cls_hist}, and we also list the frequency of each interaction type in Figure~\ref{fig:interact_act_hist}.

\section{Evaluation Metrics}
\label{sec:metric}
In our method, we use mAP@0.5 to evaluate the detection accuracy by comparing our maximum likelihood prediction with the ground-truth. We also use Minimal Matching Distance (MMD) and Total Mutual Diversity (TMD) to respectively evaluate the quality and diversity of our multi-modal predictions. For each input sequence, we sample our probabilistic mixture decoder and produce 10 hypotheses of object arrangements. We adapt MMD from \cite{achlioptas2018learning} to evaluate our task, and measure the best matching mAP@0.5 score with the ground-truth out of the 10 hypotheses. For TMD, we follow \cite{wu2020multimodal} to evaluate the multi-modality of our predictions, formulated as

\begin{equation}
	\label{eqn:01}
	\begin{aligned}
		\text{TMD}(O^{i}) &=[1+\text{Entropy}(\mathcal{C}(O^{i}))]\cdot[1+\text{Div}(\mathcal{B}(O^{i}))],\\
		\text{Div}(\mathcal{B}(O^{i}))
		&=\frac{1}{10}\sum_{p=1}^{10}\sum_{q=1}^{10}\text{Dist}(B_{p}^{i},B_{q}^{i});, \\
		\text{Dist}(B_{p}^{i},B_{q}^{i})
		&=\frac{1}{8}\sum_{j=1}^{8}||P-Q||_{2}, \ \ \ \ P\in B_{p}^{i}, Q\in B_{q}^{i},\\
		\mathcal{C}(O^{i})
		&= \{c_{1}^{i}, c_{2}^{i}, ..., c_{10}^{i}\},\\
		\mathcal{B}(O^{i})
		&= \{B_{1}^{i}, B_{2}^{i}, ..., B_{10}^{i}\}, \ \ \ \ B_{k}^{i}\in\mathbb{R}^{8\times3}.
	\end{aligned}
\end{equation}

In Eq.~\ref{eqn:01}, TMD is defined at the object-level;
for each object $O^{i}$ in a scene, we have 10 hypotheses which are represented by 10 bounding boxes $\mathcal{B}(O^{i})$ with the corresponding 10 predicted class labels $\mathcal{C}(O^{i})$;
$B_{k}^{i}$ is the $k$-th hypothesis in $\mathcal{B}(O^{i})$, which can be represented by a point set with eight box corners, and $c_{k}^{i}$ is the corresponding class label;
$\text{Entropy}(*)$ denotes the Shannon Entropy to measure the variance of class labels;
$\text{Div}(*)$ measures the diversity of predicted bounding boxes among hypotheses, which is defined by the average of distance sum from a hypothesis $B_{p}^{i}$ to all other hypotheses;
$\text{Dist}(*)$ is the average Euclidean distance between pair-wise points from $B_{p}^{i}$ and $B_{q}^{i}$.

$\text{TMD}(O^{i})$=1 if all hypotheses are the same, where $\mathcal{C}(O^{i}))$=0; $\text{Div}(\mathcal{B}(O^{i}))=0$, which indicates no diversity. In Section~5.3 of the main paper, we report the average TMD score over all objects.

\section{Additional Quantitative Results}
\label{sec:quant}
We list the mAP@0.5 scores on all object categories by split $\mathcal{S}_{1}$ and $\mathcal{S}_{2}$ in Table~\ref{compare:det_compare_s1_full} and Table~\ref{compare:det_compare_s2_full} respectively. \OURS{} variants are evaluated by the maximum likelihood predictions to calculate mAP scores. Note that there are fewer categories in the test set of $\mathcal{S}_{2}$.

\begin{table*}[!t]
	\centering
	\resizebox{1\columnwidth}{!}{\begin{tabular}{|l|c c c c c c c c c c c c c c c c c|c|}
	\hline
	& bed & bench & bkshlf & cabnt & chair & desk & dishws & faucet & fridge & lamp & microw & monitor & nstand & sofa & stove & toilet & cmpter & mAP@0.5 \\
	\hline
	\hline
	Pose-VoteNet & 2.90 & 15.00 & 73.79 & 33.14 & 18.77 & 58.52 & 32.14 & 0.00 & 0.00 & 6.07 & 6.49 & 23.55 & 51.30 & 62.32 & 49.82 & 0.00 & 3.06 & 25.70  \\
	Pose-VoteNet + VN & 20.81 & 18.13 & 78.25 & 49.76 & 18.68 & 70.92 & 33.56 & 0.00 & 0.00 & 5.60 & 5.23 & 22.71 & 64.46 & 67.24 & 46.76 & 0.00 & 6.11 & 29.90 \\
	Motion Attention & 36.42 & 7.54 & 66.25 & 23.35 & 19.50 & 77.71 & 15.59 & 0.03 & 17.13 & 2.35 & 6.42 & 29.87 & 30.59 & 78.61 & 51.03 & 14.81 & 5.50 & 28.39  \\
	\hline
	P2R-Net-D & 93.77 & 12.63 & 37.21 & 11.98 & 5.77 & 95.93 & 61.80 & 0.10 & 73.95 & 0.58 & 7.39 & 9.68 & 9.62 & 88.44 & 70.42 & 0.00 & 14.17 & 34.91  \\
	P2R-Net-G & 91.69 & 7.56 & 41.27 & 36.61 & 10.05 & 93.47 & 67.53 & 0.10 & 77.45 & 1.21 & 6.36 & 10.51 & 11.32 & 92.97 & 64.86 & 5.56 & 18.67 & 37.48 \\
	P2R-Net-H & 85.84 & 8.04 & 43.78 & 22.04 & 10.91 & 76.08 & 55.20 & 0.00 & 55.15 & 0.00 & 1.57 & 4.24 & 20.31 & 83.92 & 57.60 & 5.00 & 4.33 & 31.41  \\
	\hline
	Ours & 94.21 & 10.12 & 41.75 & 54.72 & 8.02 & 93.32 & 56.33 & 0.06 & 59.89 & 3.25 & 6.49 & 12.84 & 46.53 & 90.92 & 57.86 & 61.11 & 19.94 & 42.20  \\
	\hline
	\end{tabular}}
	\caption{Quantitative evaluation on split $\mathcal{S}_{1}$.}
	\label{compare:det_compare_s1_full}
\end{table*}

\begin{table*}[!t]
	\centering
	\resizebox{1\columnwidth}{!}{
	\begin{tabular}{|l|c c c c c c c c c c c c |c|}
		\hline
		& bed & bench & bookshelf & cabinet & chair & desk & dishwasher & faucet & fridge & microwave & nightstand & stove &mAP@0.5 \\
		\hline
		\hline
		Pose-VoteNet & 0.00 & 4.48 & 22.84 & 2.30 & 0.56 & 4.05 & 14.64 & 0.00 & 0.00 & 0.61 & 0.00 & 73.33 & 10.23 \\
		Pose-VN & 0.00 & 0.96 & 19.88 & 7.99 & 0.51 & 0.14 & 10.61 & 0.00 & 0.00 & 0.00 & 0.01 & 33.33 & 6.12\\
		Motion Attention & 28.16 & 1.81 & 24.19 & 1.43 & 0.00 & 0.00 & 0.00 & 0.00 & 0.00 & 0.00 & 22.77 & 0.00 & 6.53\\
		\hline
		\OURS{}-D & 39.86 & 25.17 & 26.78 & 3.29 & 0.18 & 56.53 & 76.39 & 0.00 & 86.50 & 30.75 & 0.00 & 33.33 & 31.56\\
		\OURS{}-G & 50.65 & 17.89 & 13.86 & 0.54 & 0.04 & 56.97 & 87.61 & 0.06 & 70.91 & 13.76 & 0.00 & 66.67 & 31.59\\
		\OURS{}-H & 48.22 & 10.43 & 39.12 & 15.12 & 0.02 & 41.26 & 71.18 & 0.00 & 76.10 & 0.69 & 0.01 & 33.33 & 27.96\\
		\hline
		\OURS{} & 81.48 & 12.05 & 17.18 & 6.43 & 0.10 & 72.07 & 100.00 & 0.11 & 63.39 & 4.61 & 0.04 & 66.66 & 35.34\\
		\hline
	\end{tabular}}
	\caption{Quantitative evaluation on split $\mathcal{S}_{2}$. }
	\label{compare:det_compare_s2_full}
\end{table*}

\section{Additional Qualitative Results}
\label{sec:quali}
\begin{figure*}[!ht]
	\centering
	\includegraphics[width=\textwidth]
	{figure/sequence_level_comp/color_palatte.jpg}\\
	\begin{subfigure}[t]{0.31\textwidth}
		\includegraphics[width=\textwidth]
		{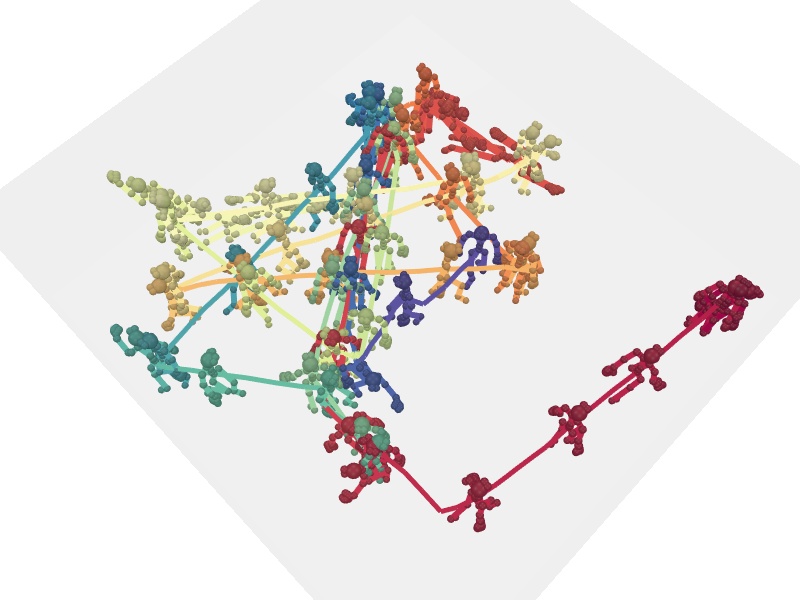}
		\includegraphics[width=\textwidth]
		{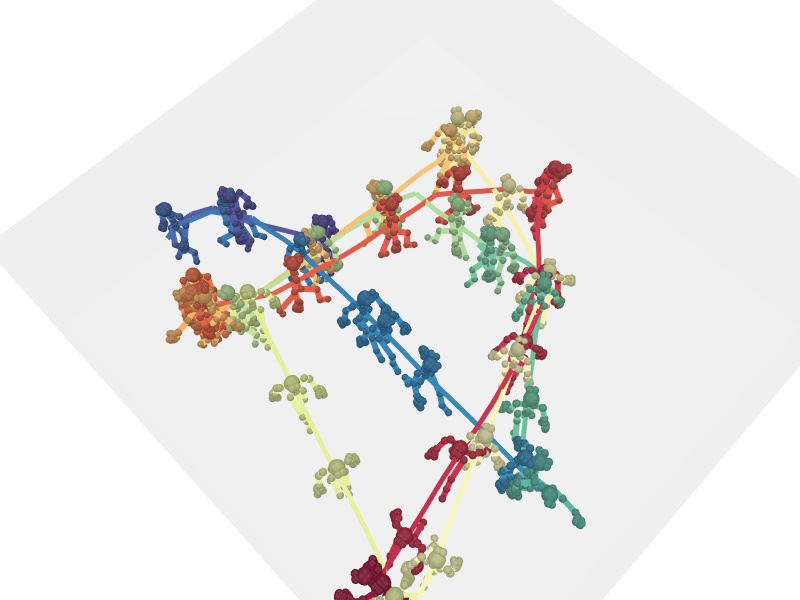}
		\includegraphics[width=\textwidth]
		{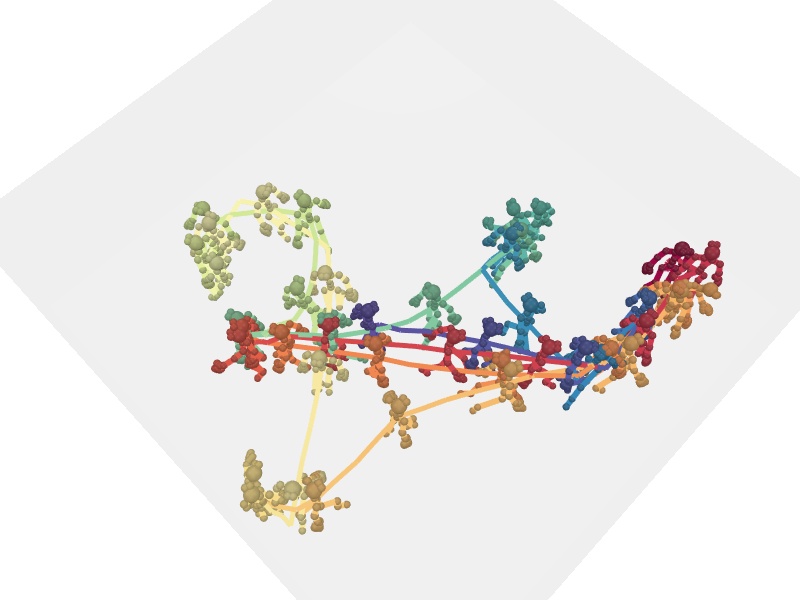}
		\includegraphics[width=\textwidth]
		{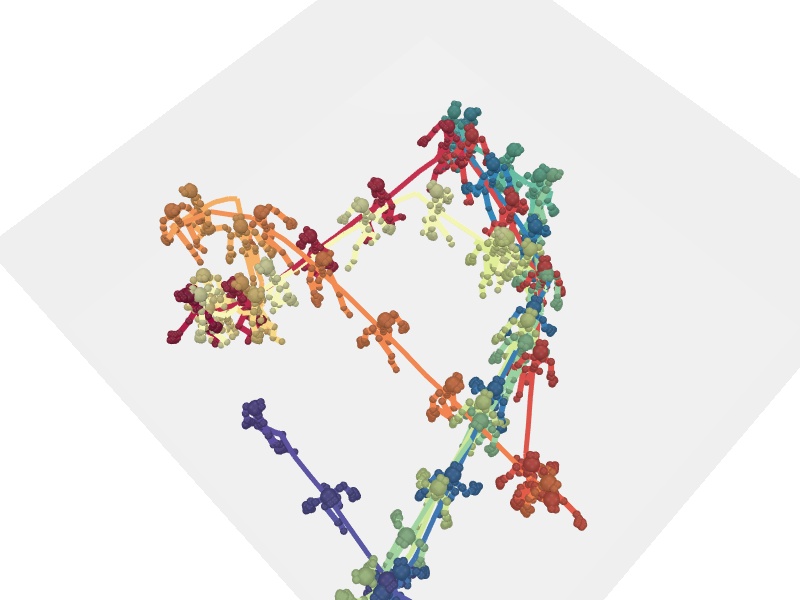}
		\caption{Input}
	\end{subfigure}
	\begin{subfigure}[t]{0.31\textwidth}
		\includegraphics[width=\textwidth]
		{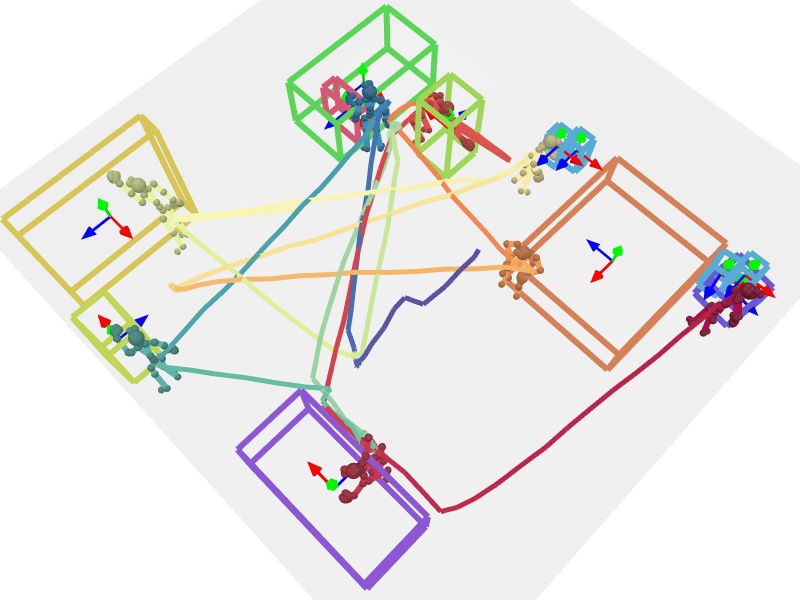}
		\includegraphics[width=\textwidth]
		{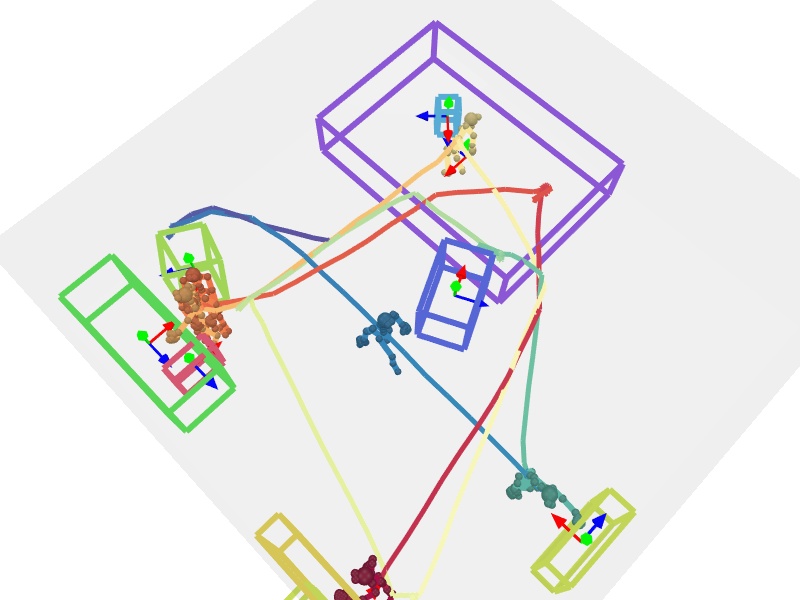}
		\includegraphics[width=\textwidth]
		{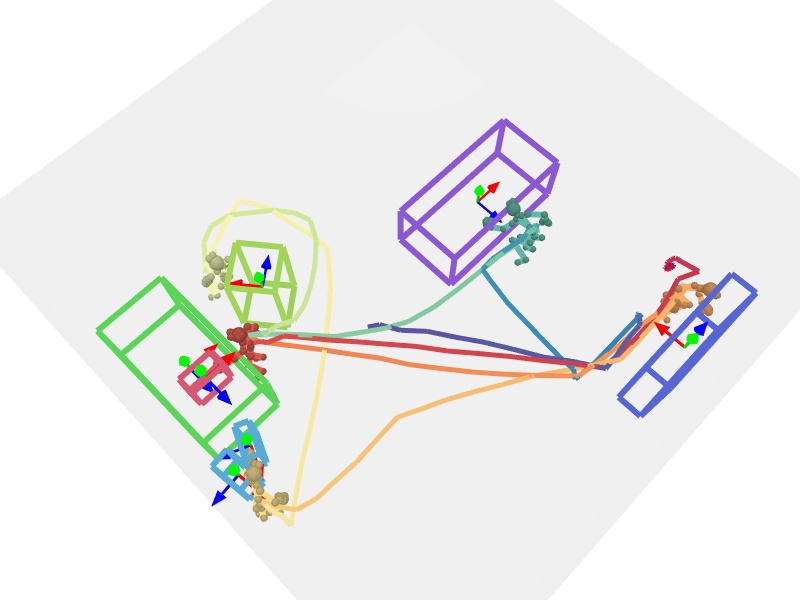}
		\includegraphics[width=\textwidth]
		{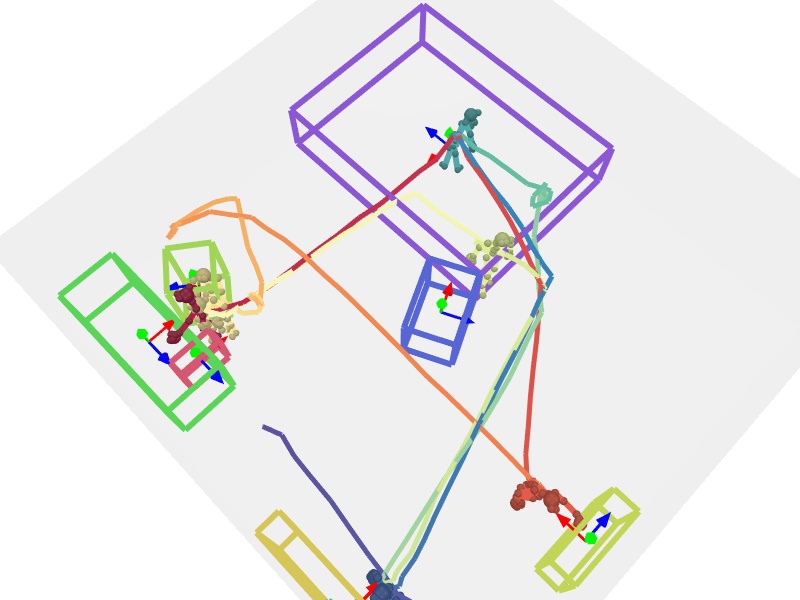}
		\caption{Prediction}
	\end{subfigure}
	\begin{subfigure}[t]{0.31\textwidth}
		\includegraphics[width=\textwidth]
		{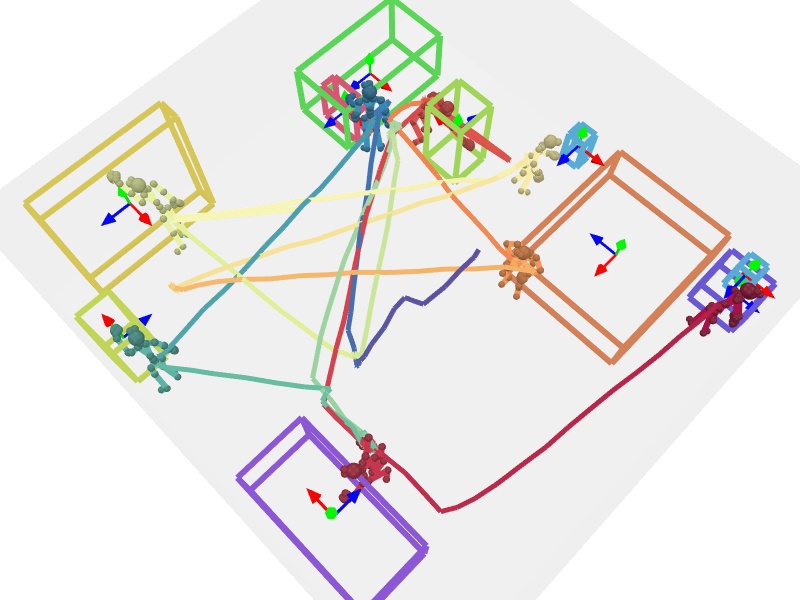}
		\includegraphics[width=\textwidth]
		{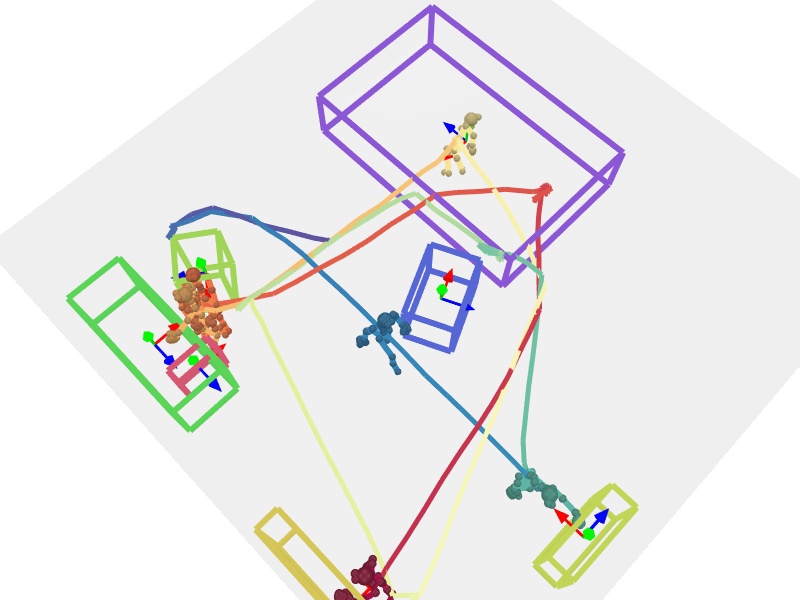}
		\includegraphics[width=\textwidth]
		{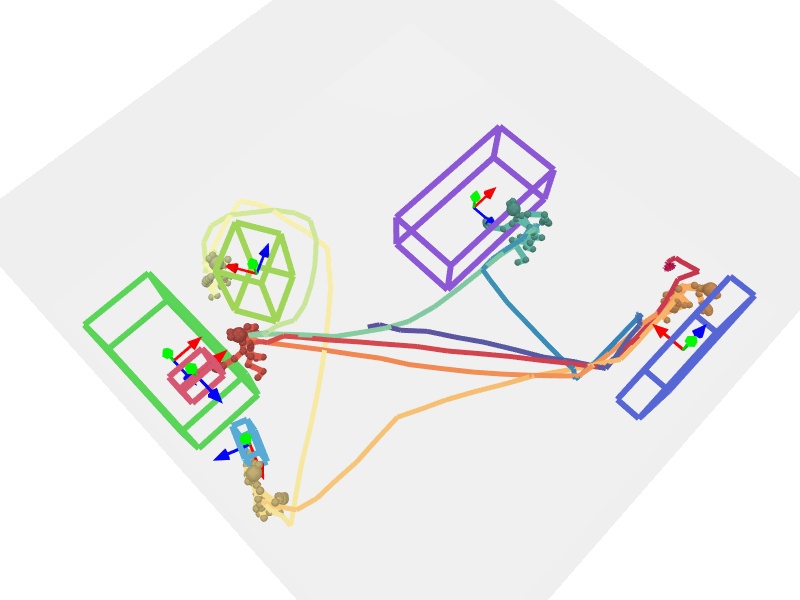}
		\includegraphics[width=\textwidth]
		{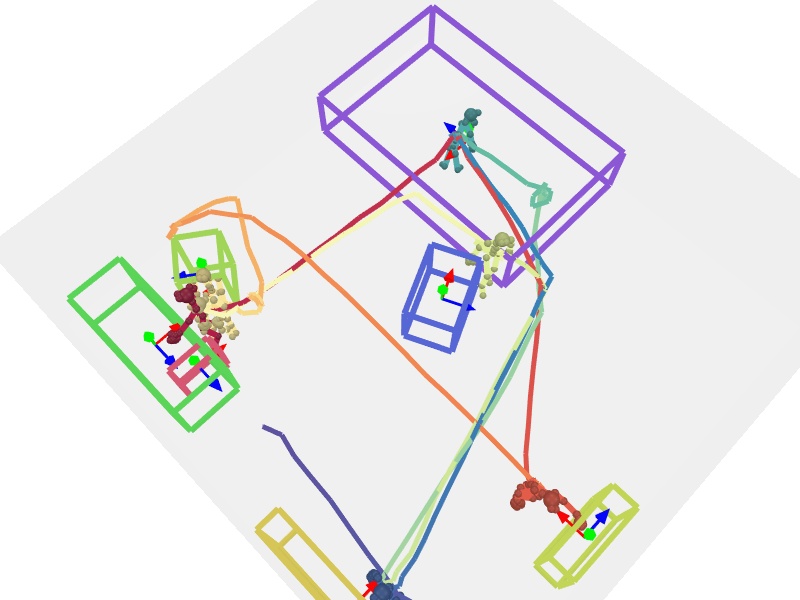}
		\caption{GT}
	\end{subfigure}
	\caption{Additional results on estimating object layouts from a pose trajectory on the sequence-level split $\mathcal{S}_{1}$ (unseen interaction sequences).}
\label{fig:more_on_s1_1}
\end{figure*}

\begin{figure*}[!ht]
	\centering
	\begin{subfigure}[t]{0.32\textwidth}
		\includegraphics[width=\textwidth]
		{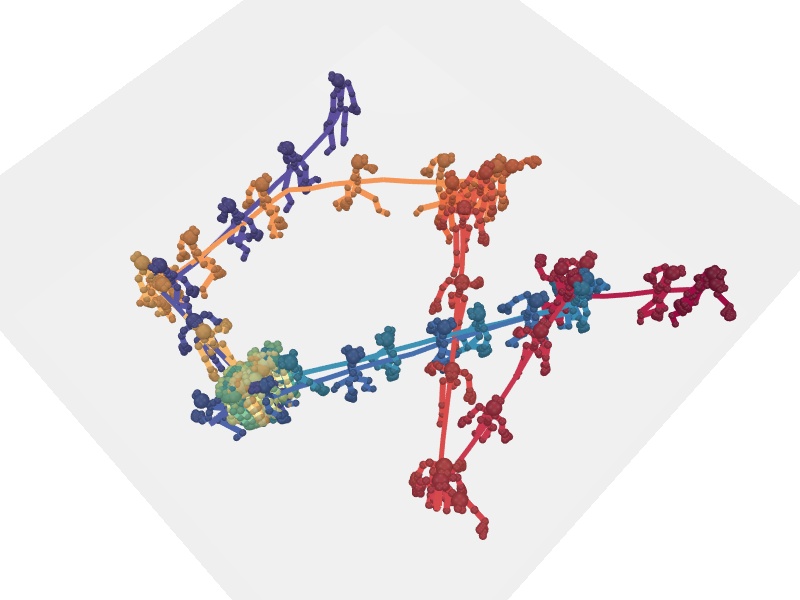}
		\includegraphics[width=\textwidth]
		{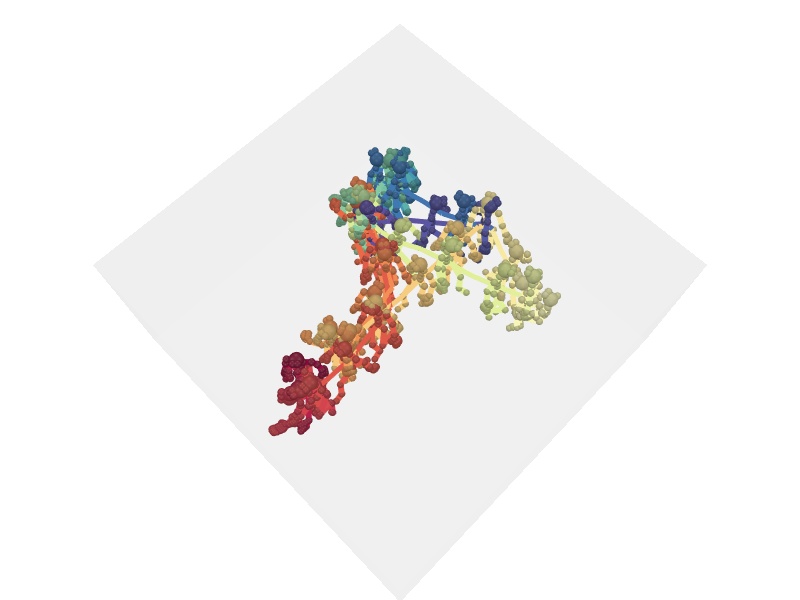}
		\includegraphics[width=\textwidth]
		{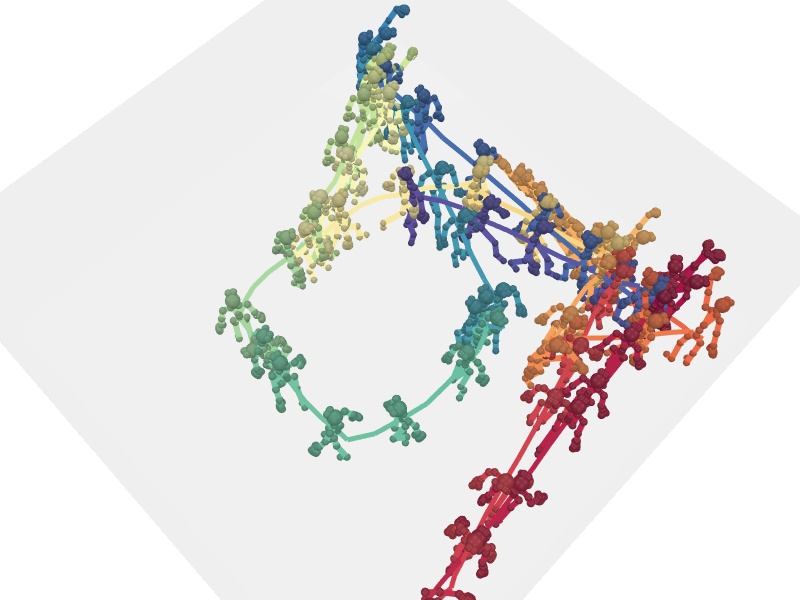}
		\includegraphics[width=\textwidth]
		{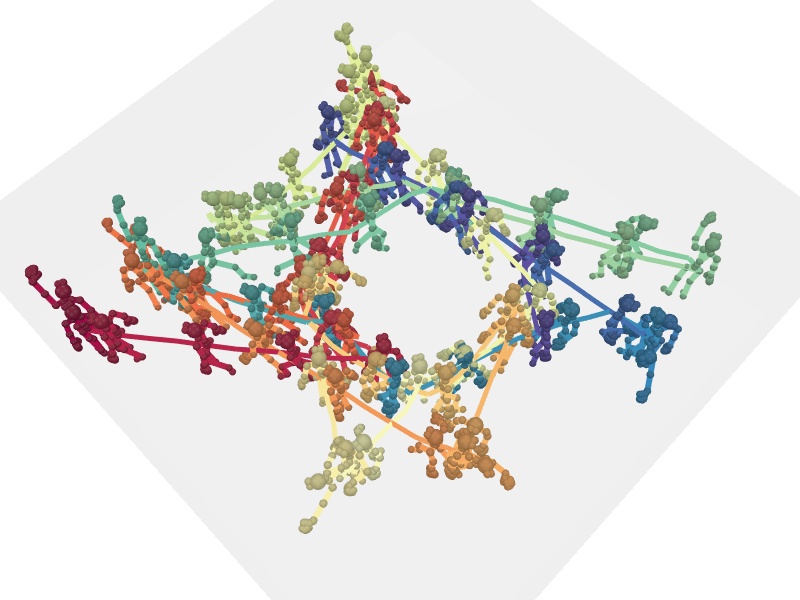}
		\caption{Input}
	\end{subfigure}
	\begin{subfigure}[t]{0.32\textwidth}
		\includegraphics[width=\textwidth]
		{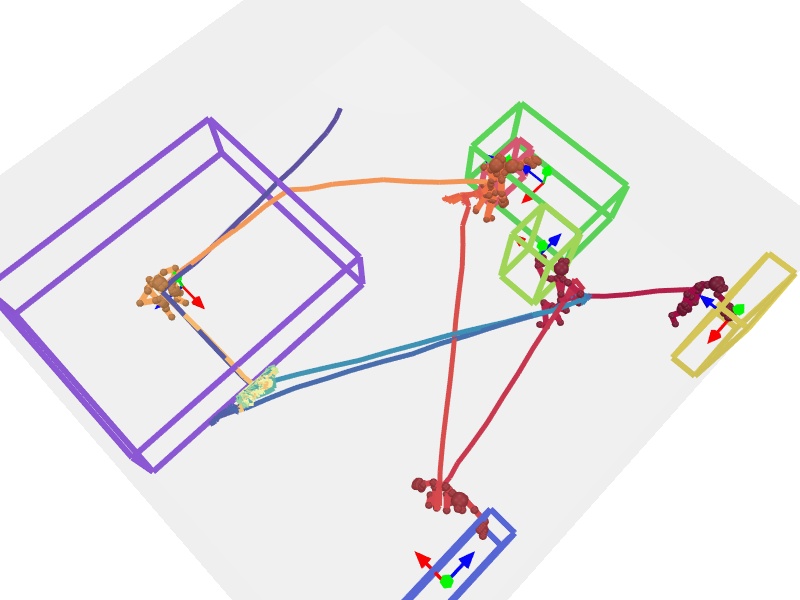}
		\includegraphics[width=\textwidth]
		{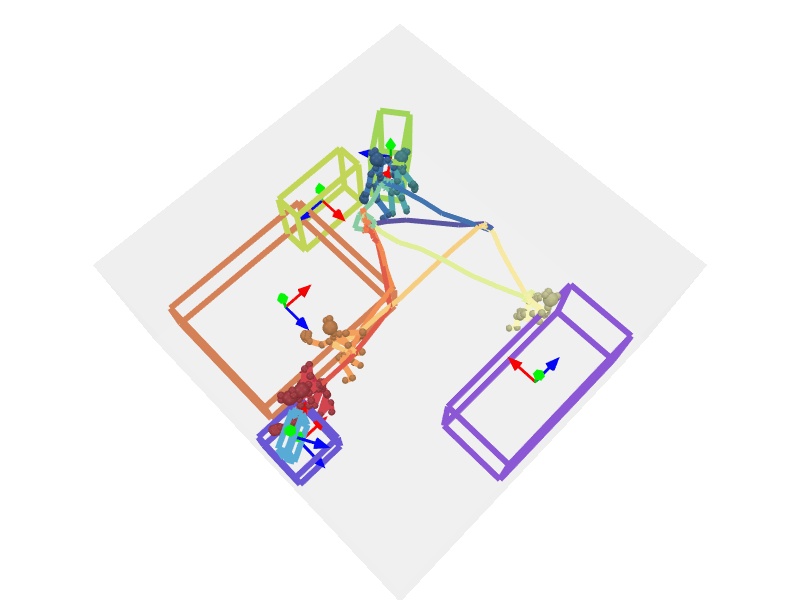}
		\includegraphics[width=\textwidth]
		{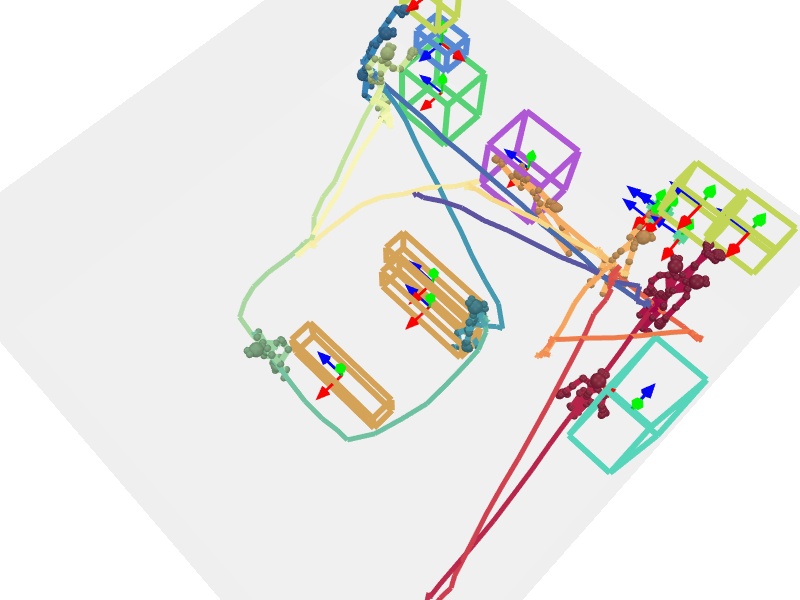}
		\includegraphics[width=\textwidth]
		{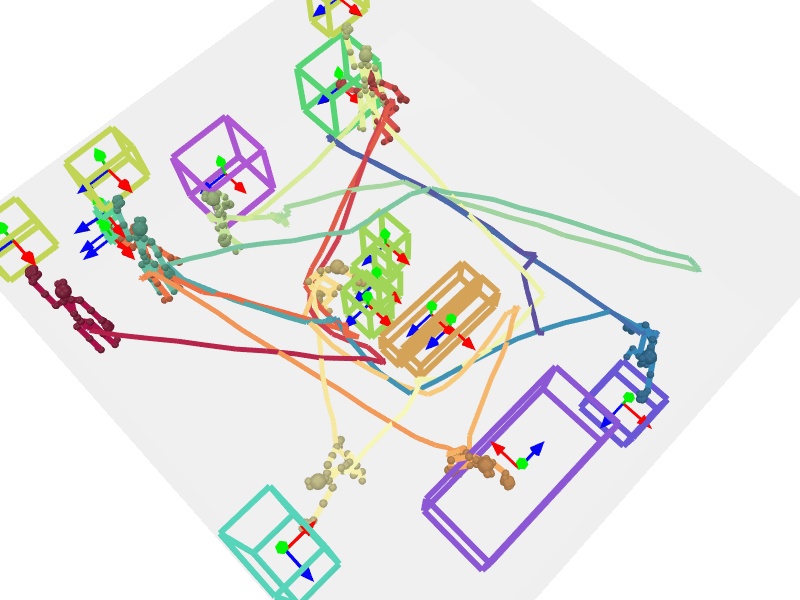}
		\caption{Prediction}
	\end{subfigure}
	\begin{subfigure}[t]{0.32\textwidth}
		\includegraphics[width=\textwidth]
		{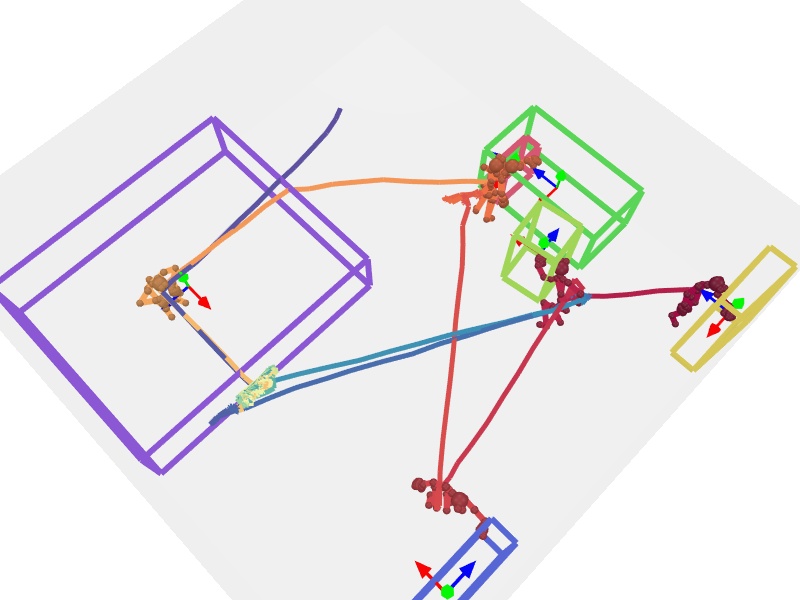}
		\includegraphics[width=\textwidth]
		{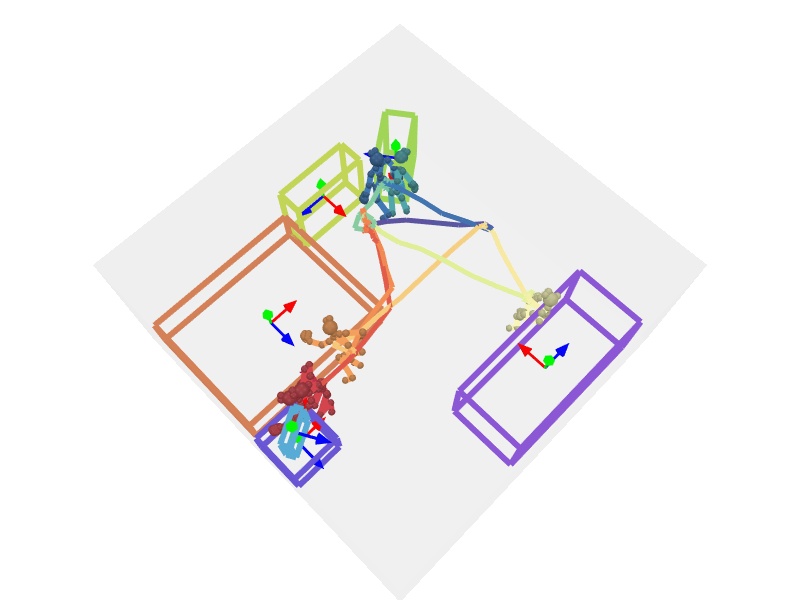}
		\includegraphics[width=\textwidth]
		{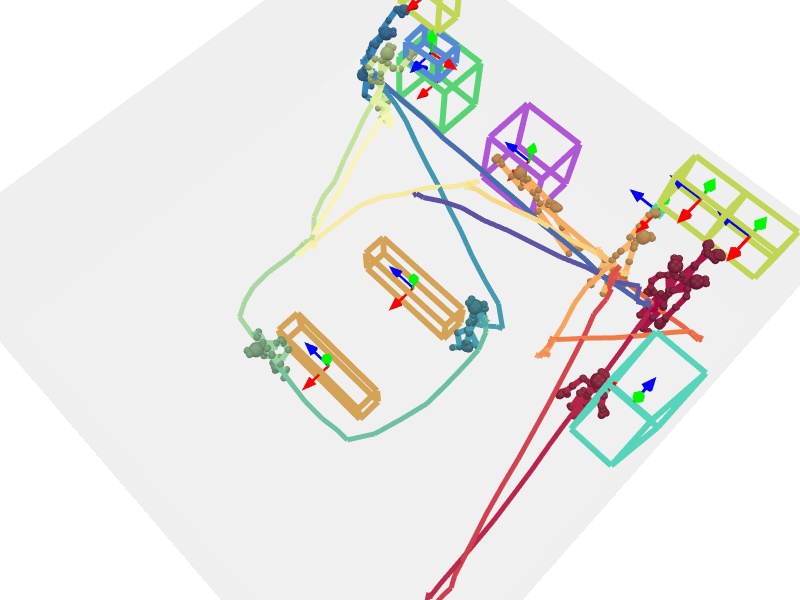}
		\includegraphics[width=\textwidth]
		{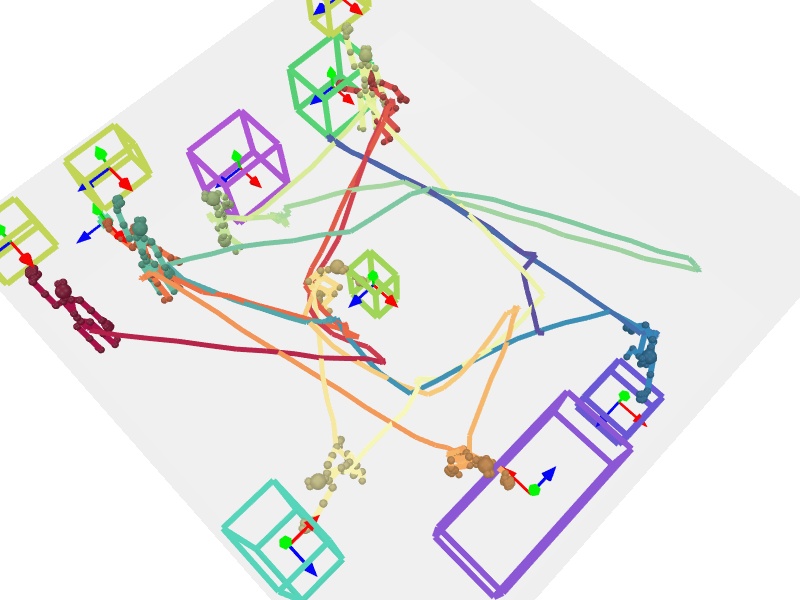}
		\caption{GT}
	\end{subfigure}
	\caption{Additional results on estimating object layouts from a pose trajectory on the sequence-level split $\mathcal{S}_{1}$ (unseen interaction sequences).}
\label{fig:more_on_s1_2}
\end{figure*}

\begin{figure*}[!ht]
	\centering
	\begin{subfigure}[t]{0.32\textwidth}
		\includegraphics[width=\textwidth]
		{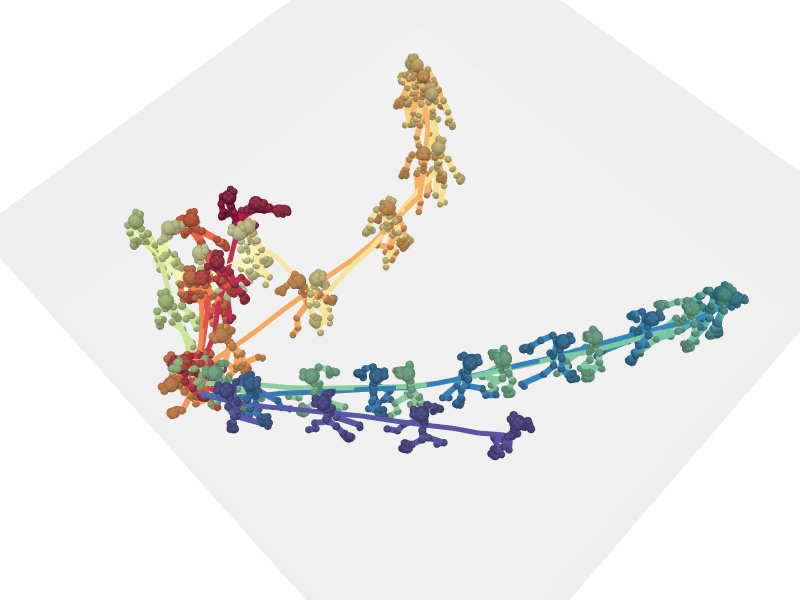}
		\includegraphics[width=\textwidth]
		{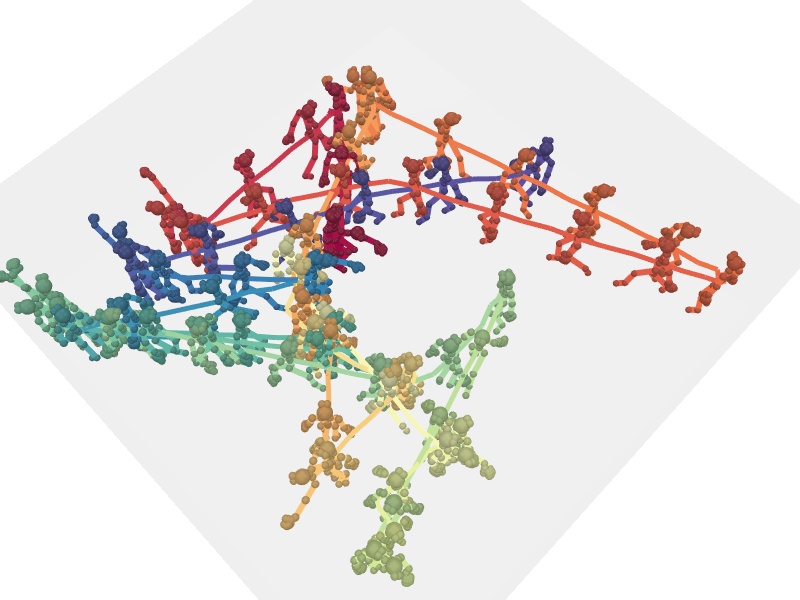}
		\includegraphics[width=\textwidth]
		{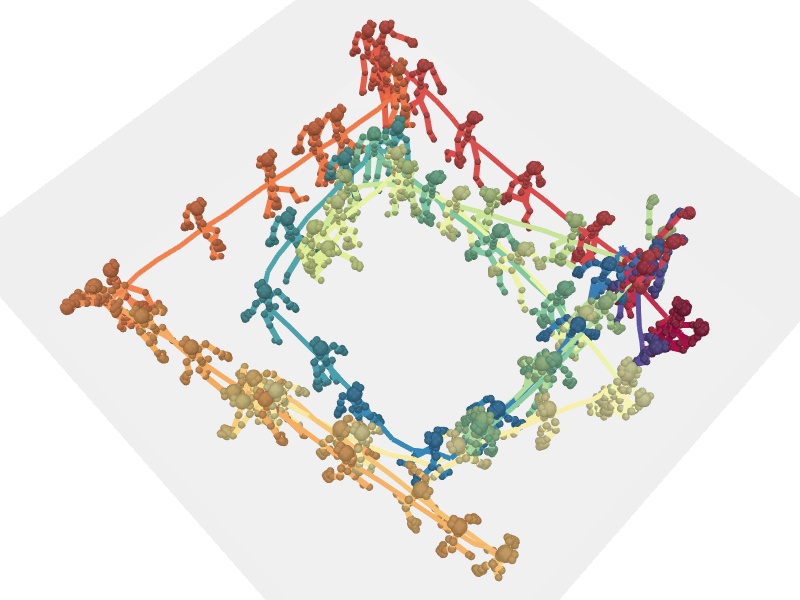}
		\includegraphics[width=\textwidth]
		{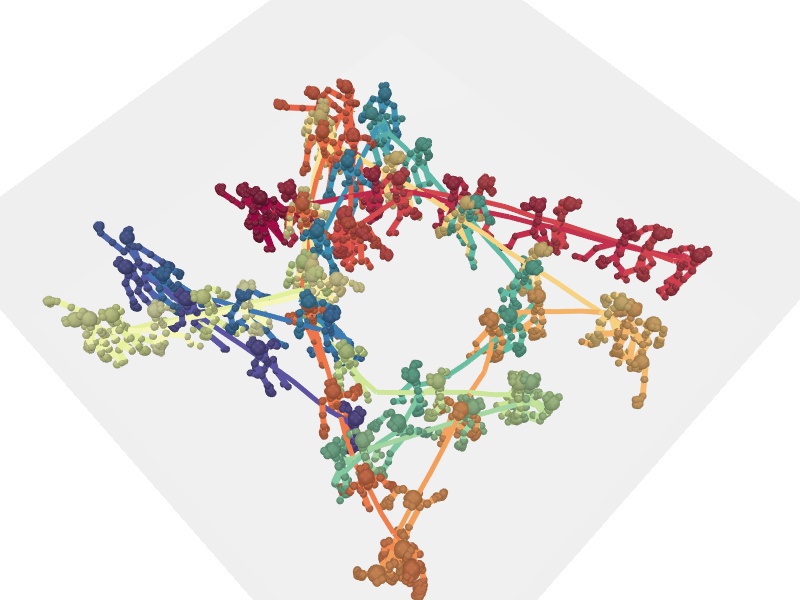}
		\caption{Input}
	\end{subfigure}
	\begin{subfigure}[t]{0.32\textwidth}
		\includegraphics[width=\textwidth]
		{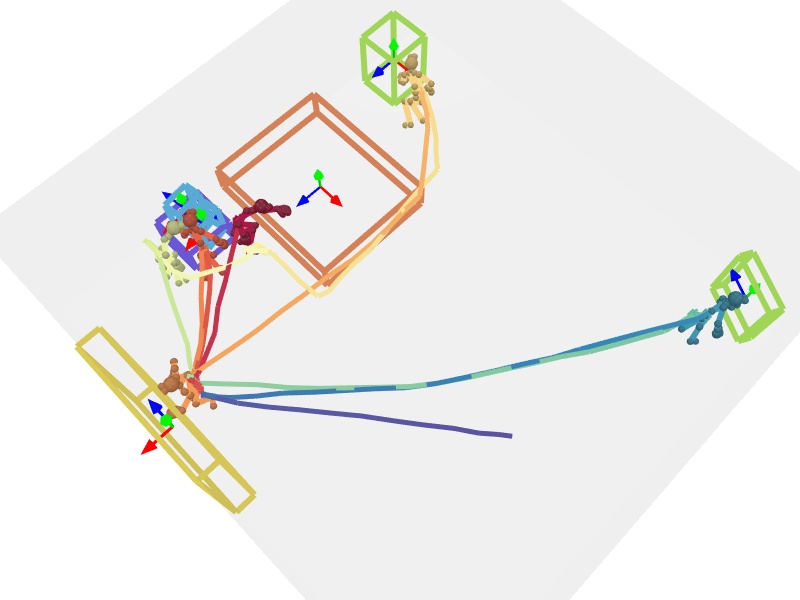}
		\includegraphics[width=\textwidth]
		{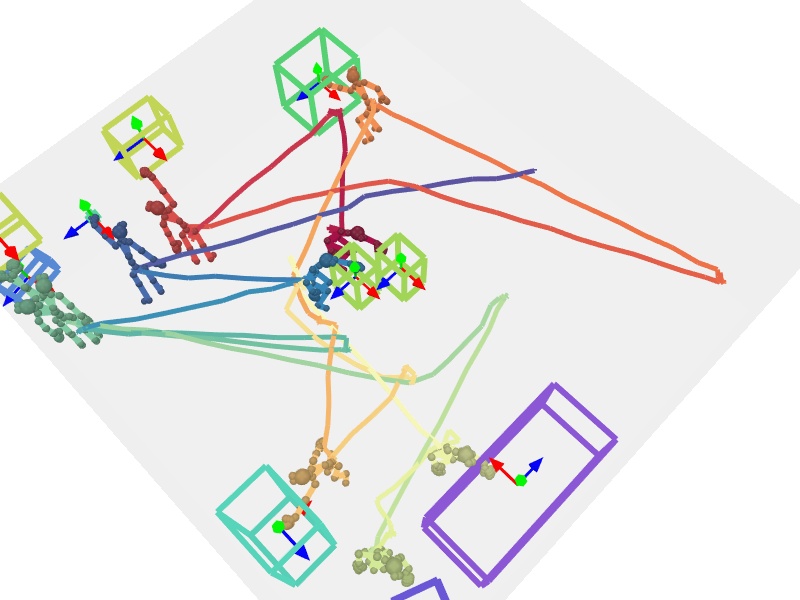}
		\includegraphics[width=\textwidth]
		{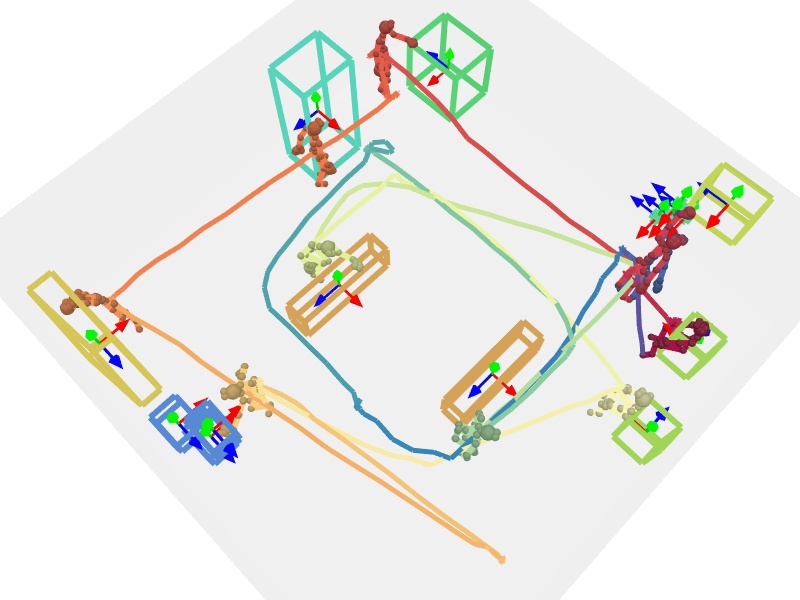}
		\includegraphics[width=\textwidth]
		{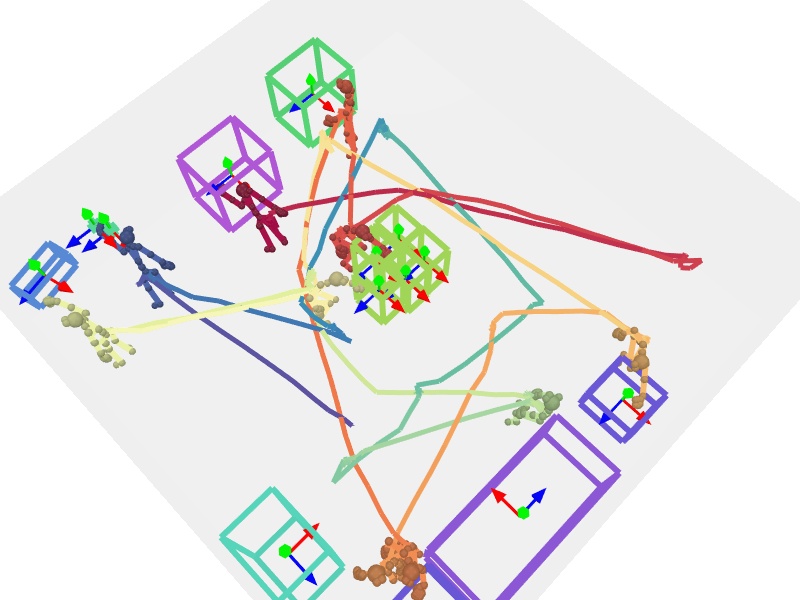}
		\caption{Prediction}
	\end{subfigure}
	\begin{subfigure}[t]{0.32\textwidth}
		\includegraphics[width=\textwidth]
		{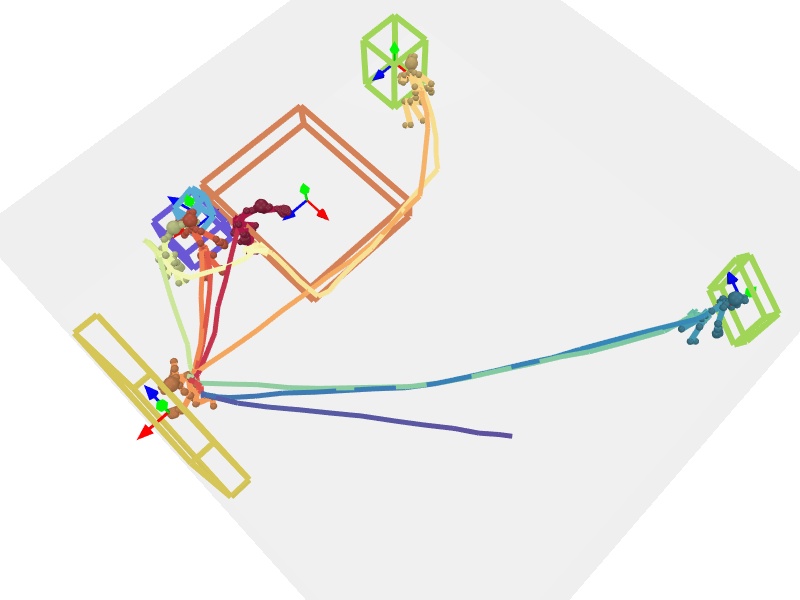}
		\includegraphics[width=\textwidth]
		{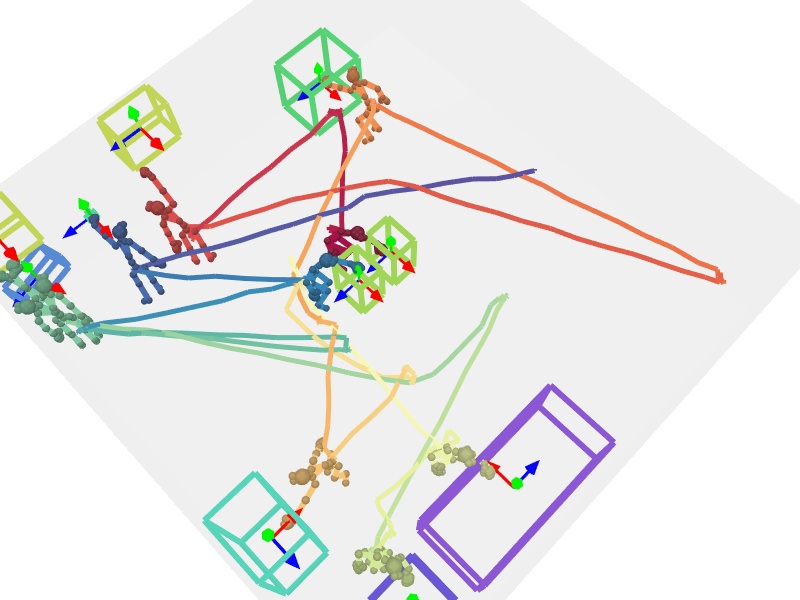}
		\includegraphics[width=\textwidth]
		{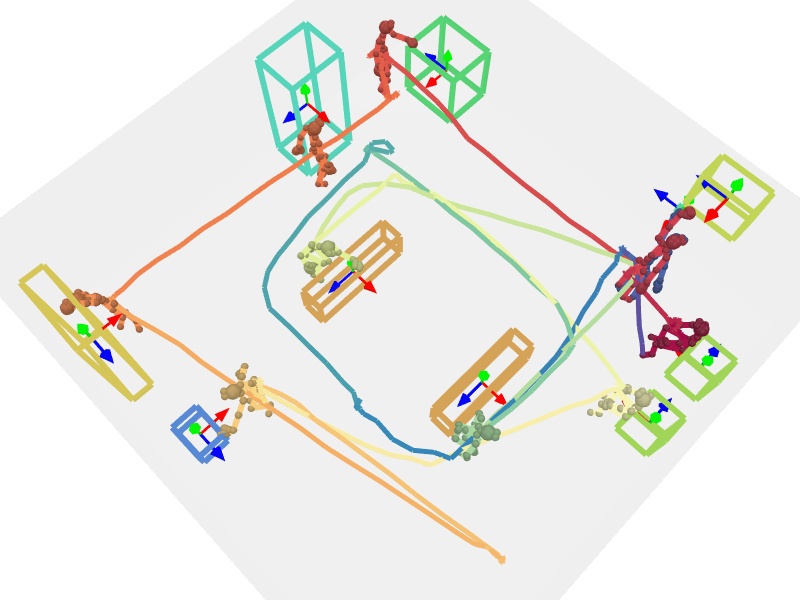}
		\includegraphics[width=\textwidth]
		{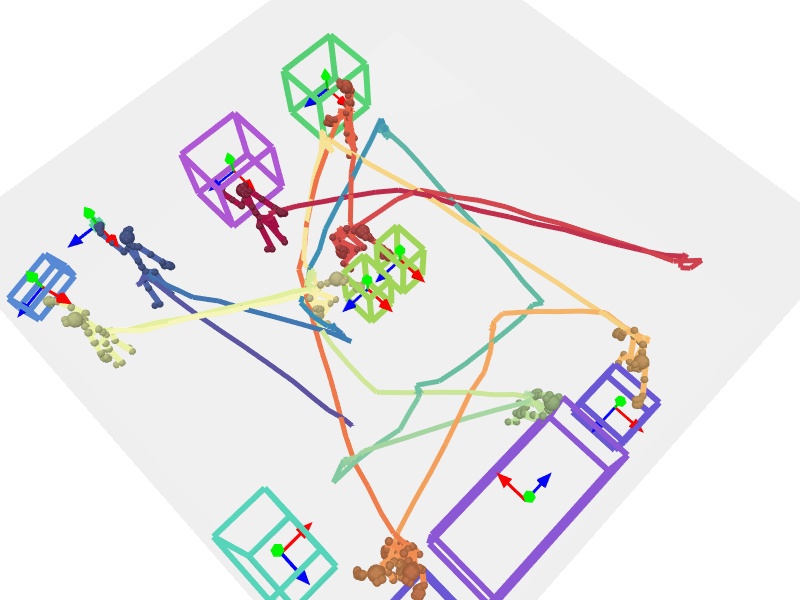}
		\caption{GT}
	\end{subfigure}
	\caption{Additional results on estimating object layouts from a pose trajectory on the sequence-level split $\mathcal{S}_{1}$ (unseen interaction sequences).}
	\label{fig:more_on_s1_3}
\end{figure*}

\begin{figure*}[!ht]
	\centering
	\begin{subfigure}[t]{0.32\textwidth}
		\includegraphics[width=\textwidth]
		{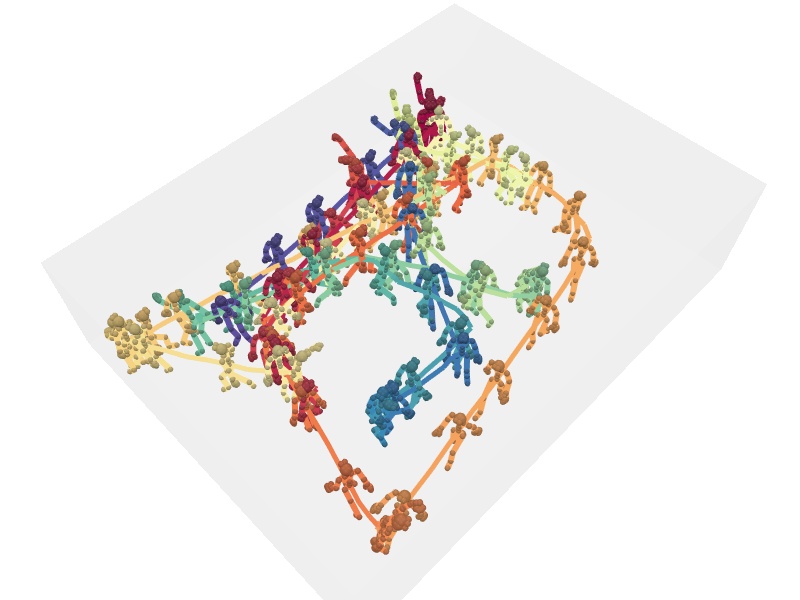}
		\includegraphics[width=\textwidth]
		{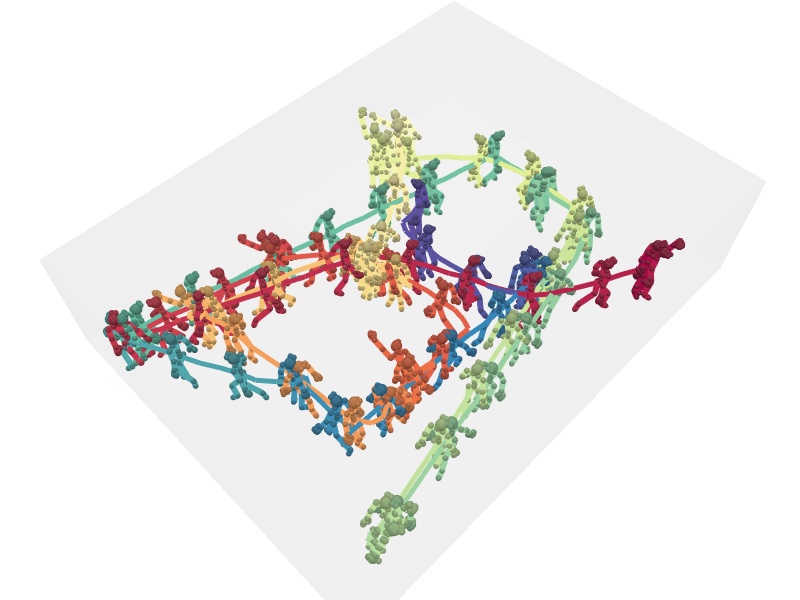}
		\includegraphics[width=\textwidth]
		{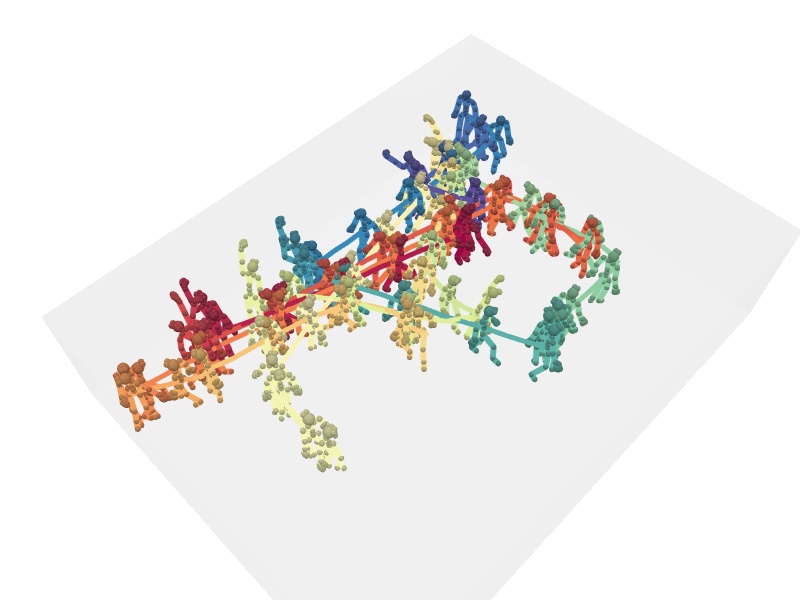}
		\includegraphics[width=\textwidth]
		{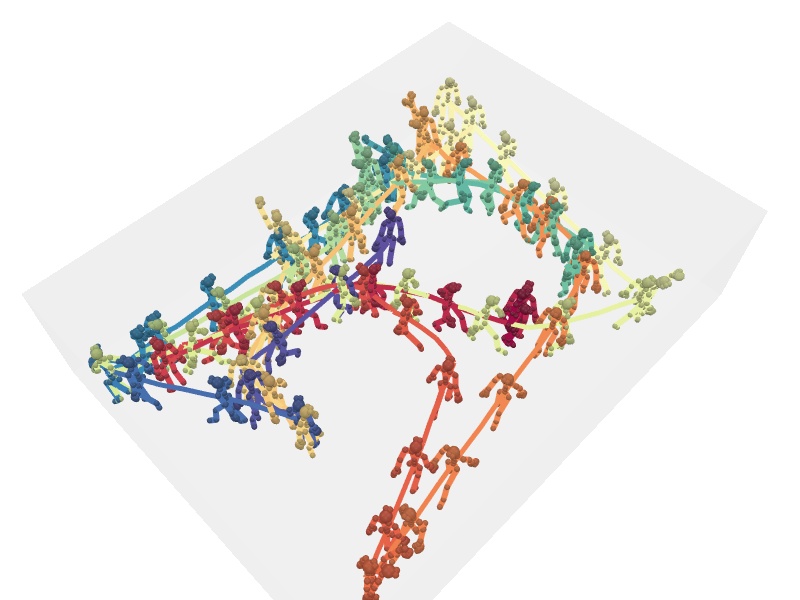}
		\includegraphics[width=\textwidth]
		{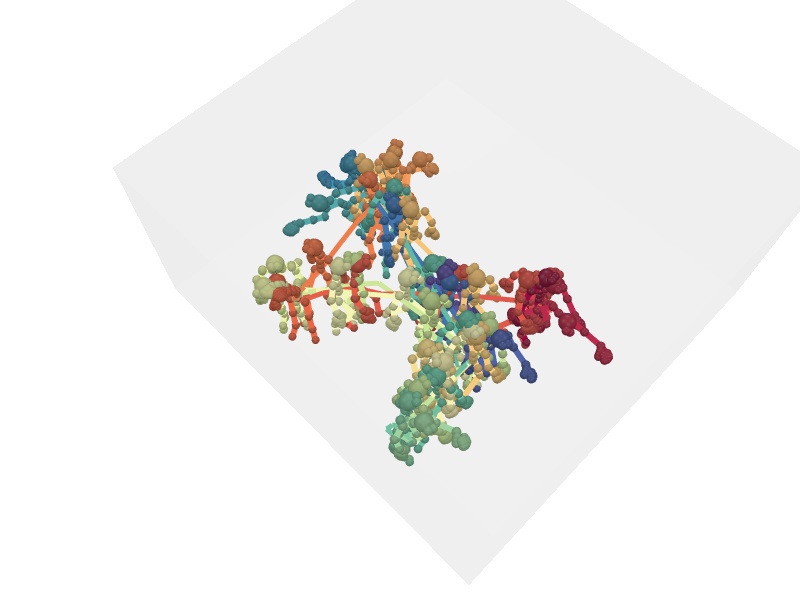}
		\caption{Input}
	\end{subfigure}
	\begin{subfigure}[t]{0.32\textwidth}
		\includegraphics[width=\textwidth]
		{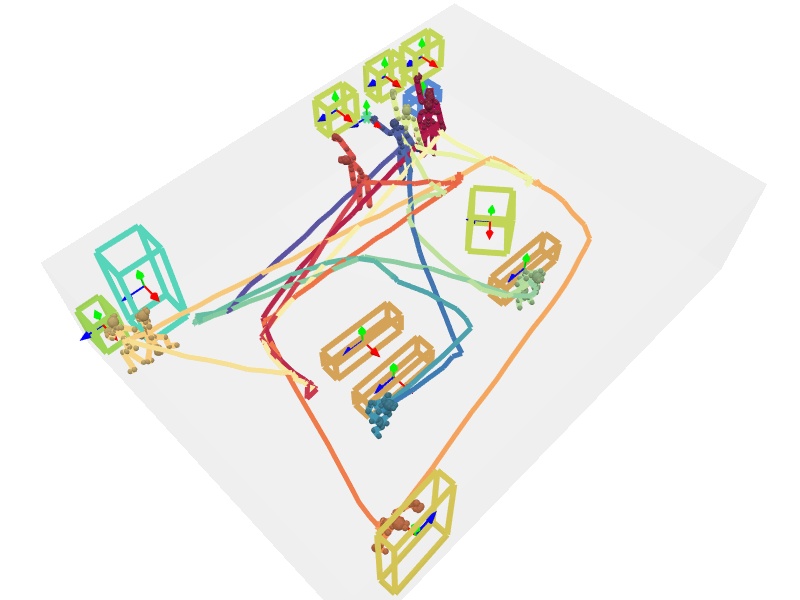}
		\includegraphics[width=\textwidth]
		{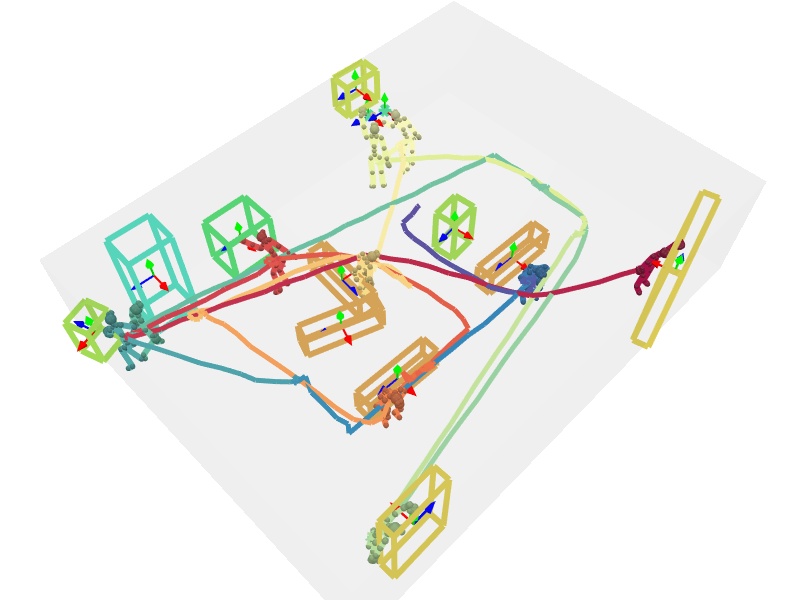}
		\includegraphics[width=\textwidth]
		{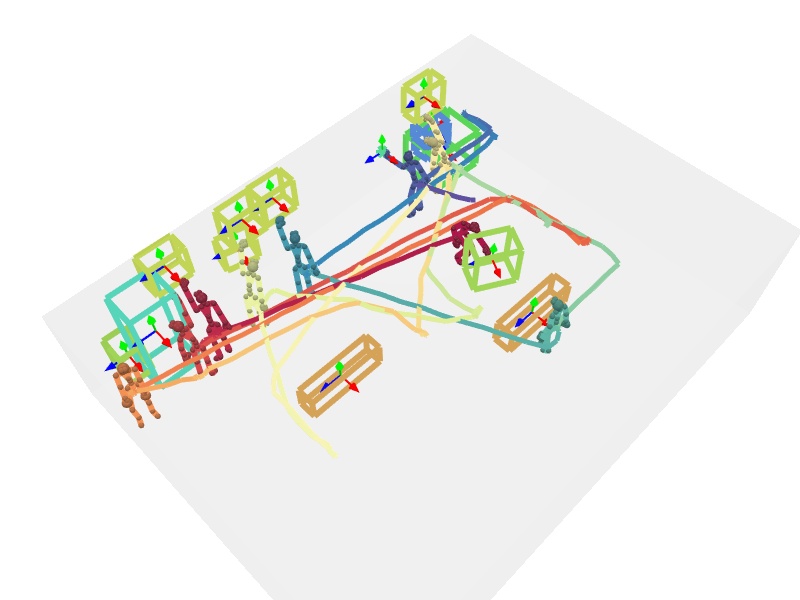}
		\includegraphics[width=\textwidth]
		{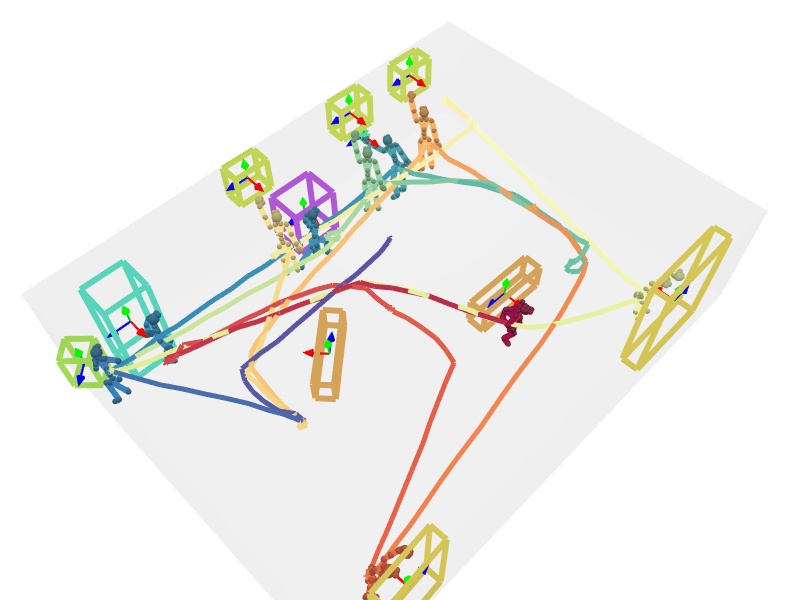}
		\includegraphics[width=\textwidth]
		{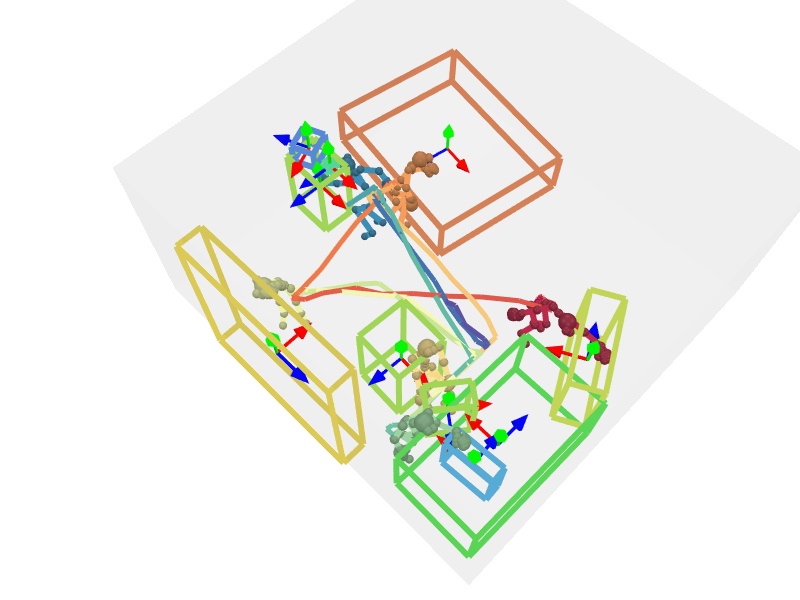}
		\caption{Prediction}
	\end{subfigure}
	\begin{subfigure}[t]{0.32\textwidth}
		\includegraphics[width=\textwidth]
		{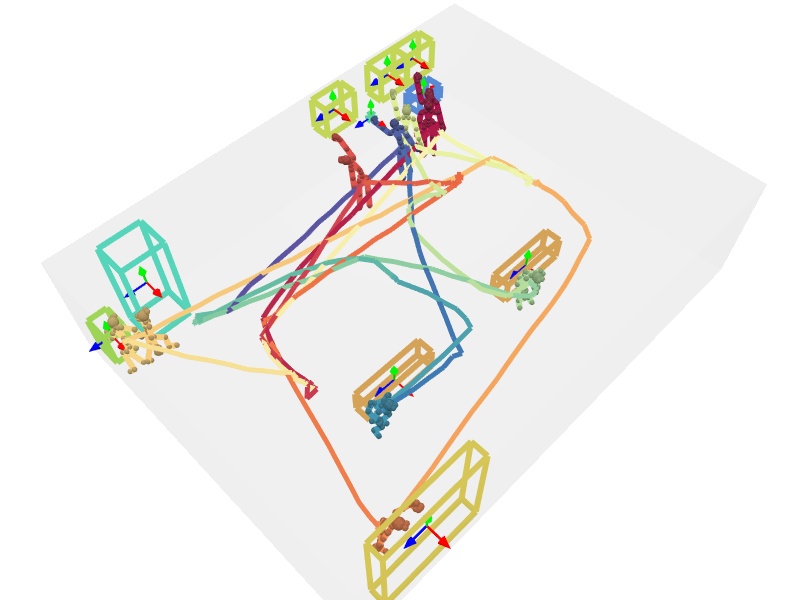}
		\includegraphics[width=\textwidth]
		{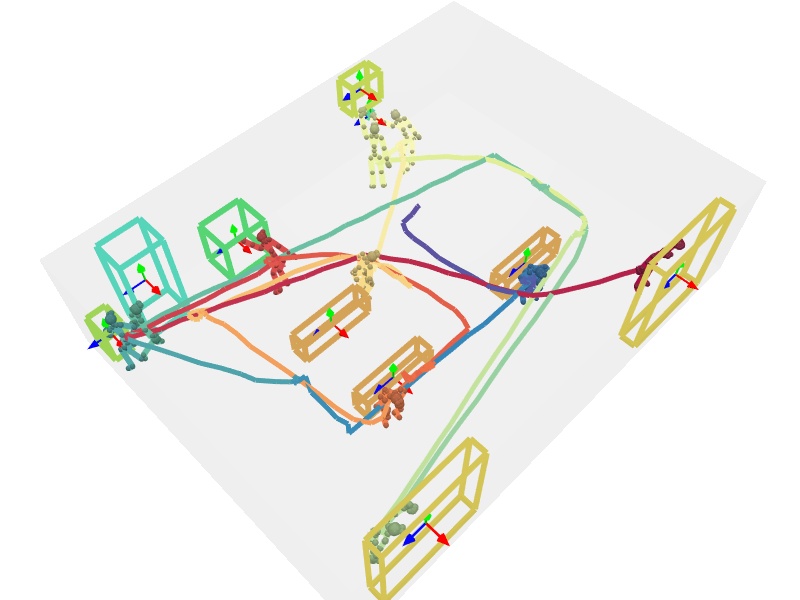}
		\includegraphics[width=\textwidth]
		{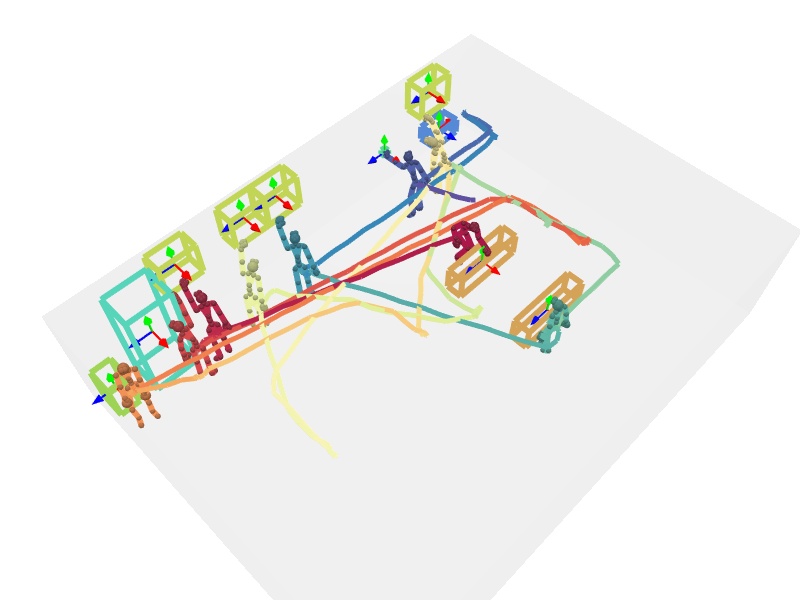}
		\includegraphics[width=\textwidth]
		{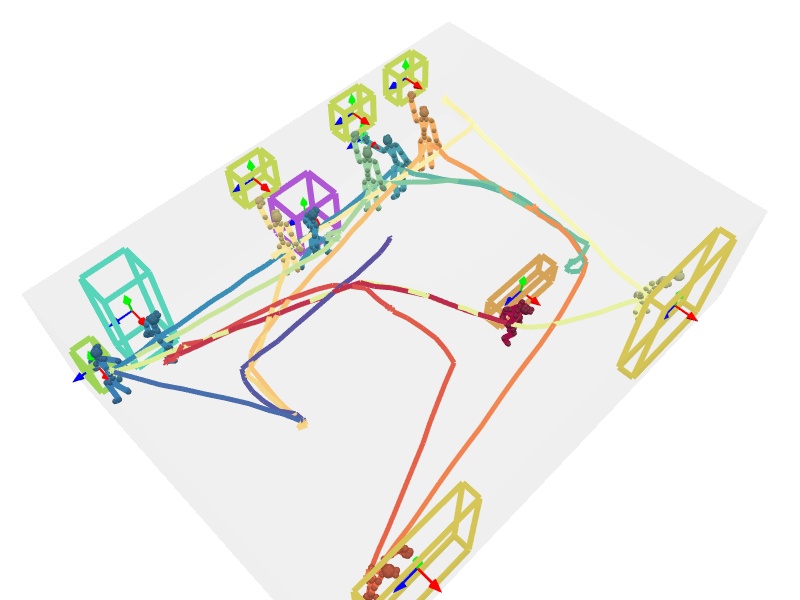}
		\includegraphics[width=\textwidth]
		{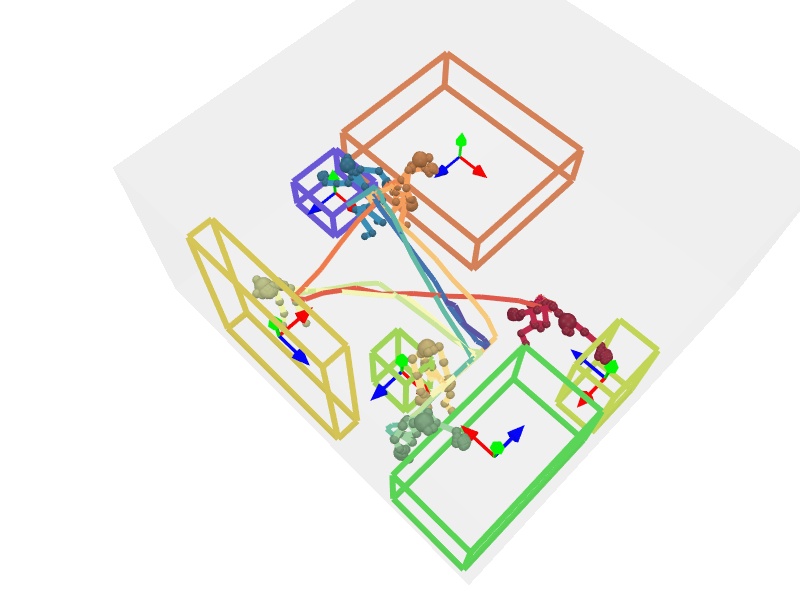}
		\caption{GT}
	\end{subfigure}
	\caption{Additional results on estimating object layouts from a pose trajectory on the room-level split $\mathcal{S}_{2}$ (unseen interaction sequences and rooms).}
	\label{fig:more_on_s2}
\end{figure*}

\begin{figure*}[!t]
	\centering
	\begin{subfigure}[t]{0.16\textwidth}
		\includegraphics[width=\textwidth]
		{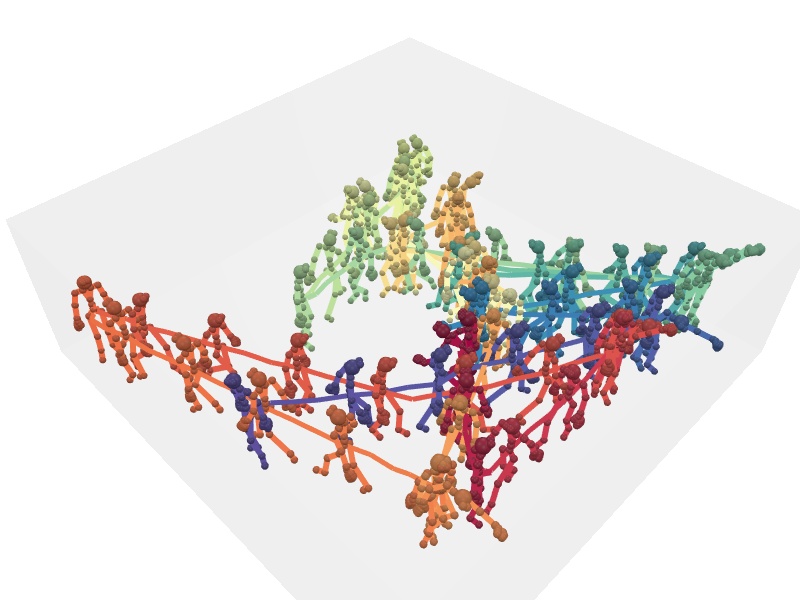}
		\includegraphics[width=\textwidth]
		{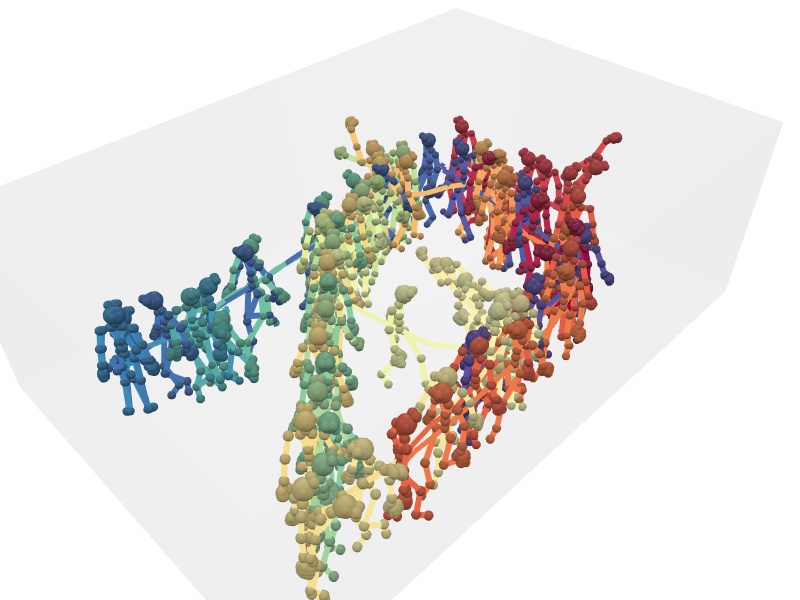}
		\includegraphics[width=\textwidth]
		{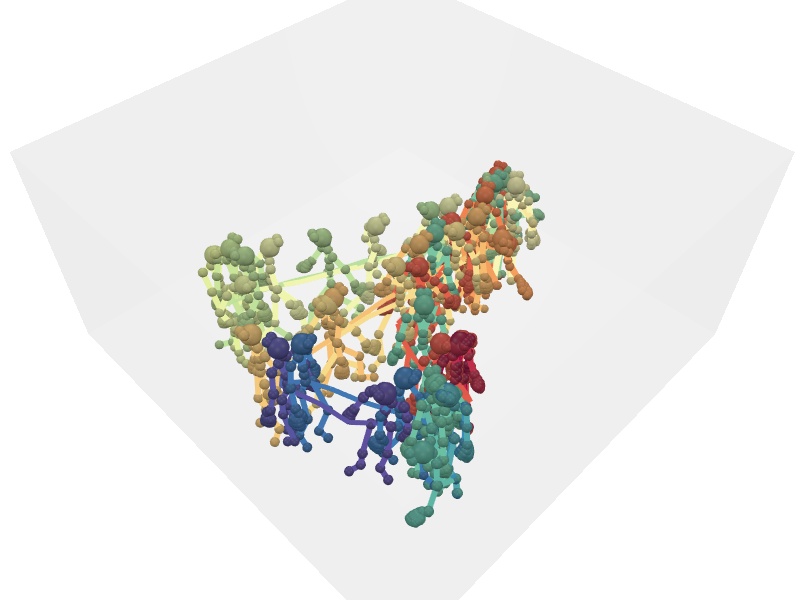}
		\includegraphics[width=\textwidth]
		{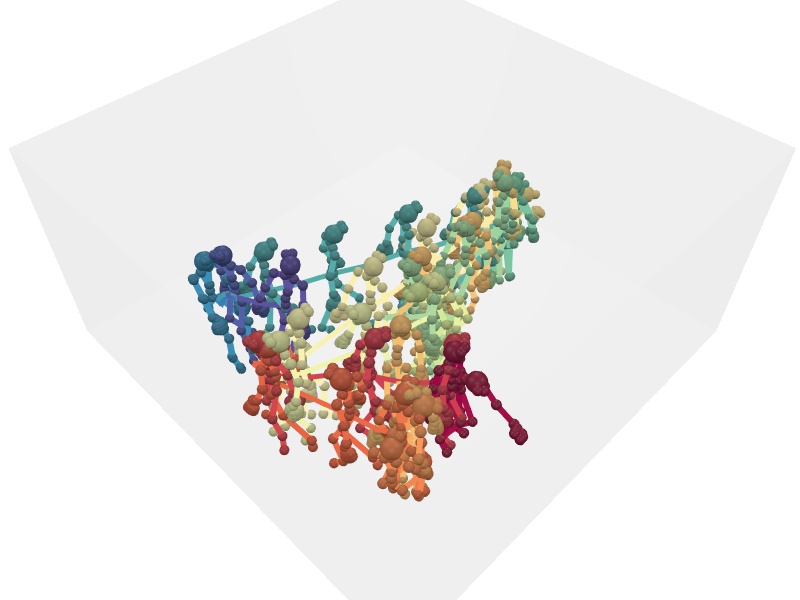}
		\includegraphics[width=\textwidth]
		{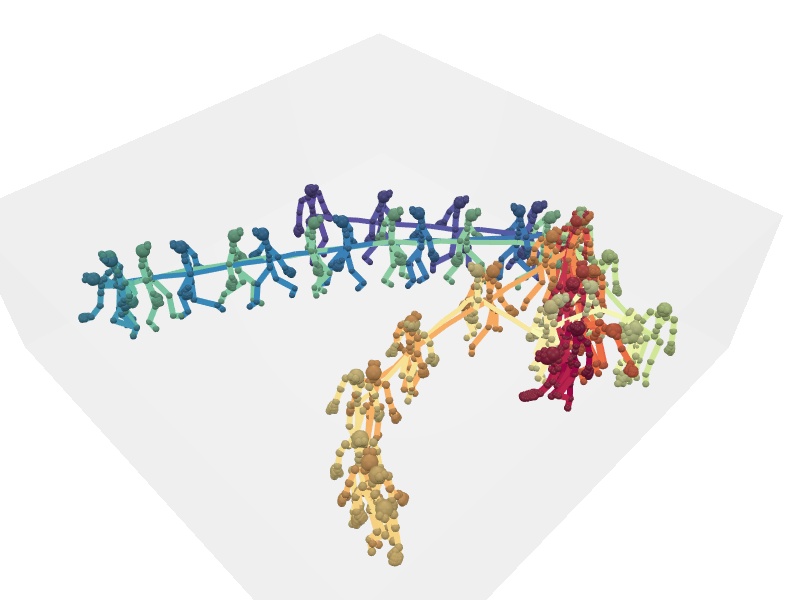}
		\includegraphics[width=\textwidth]
		{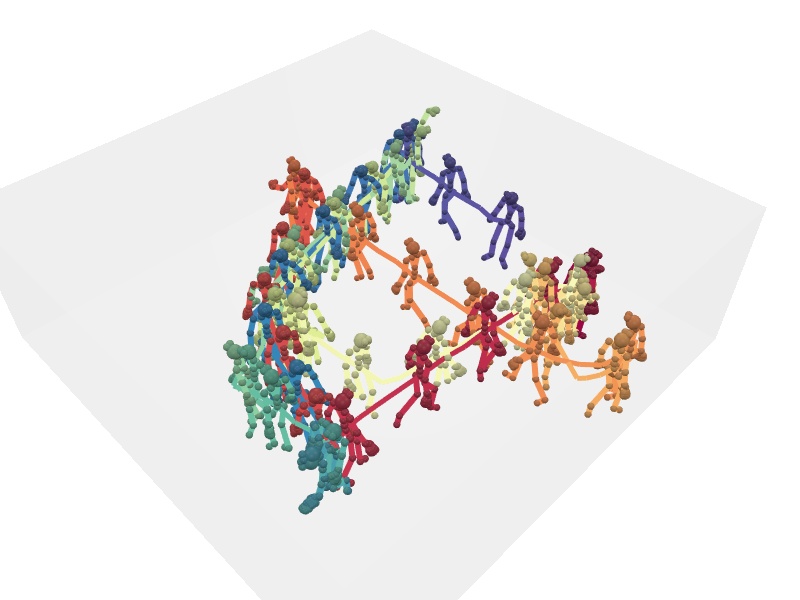}
		\caption{Input}
	\end{subfigure}
	\begin{subfigure}[t]{0.16\textwidth}
		\includegraphics[width=\textwidth]
		{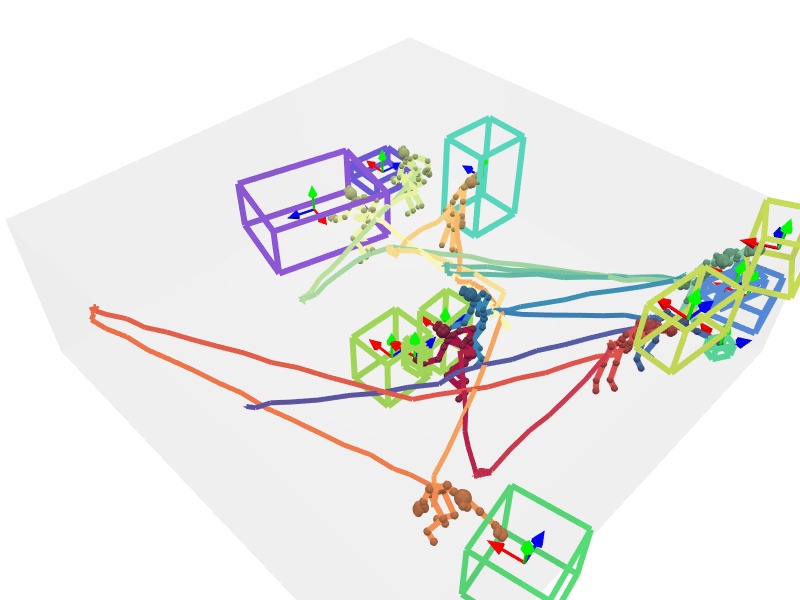}
		\includegraphics[width=\textwidth]
		{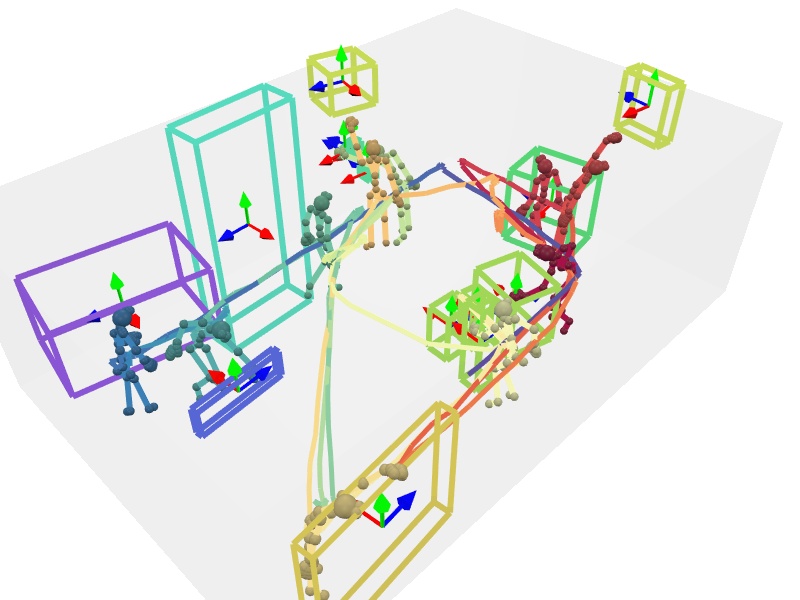}
		\includegraphics[width=\textwidth]
		{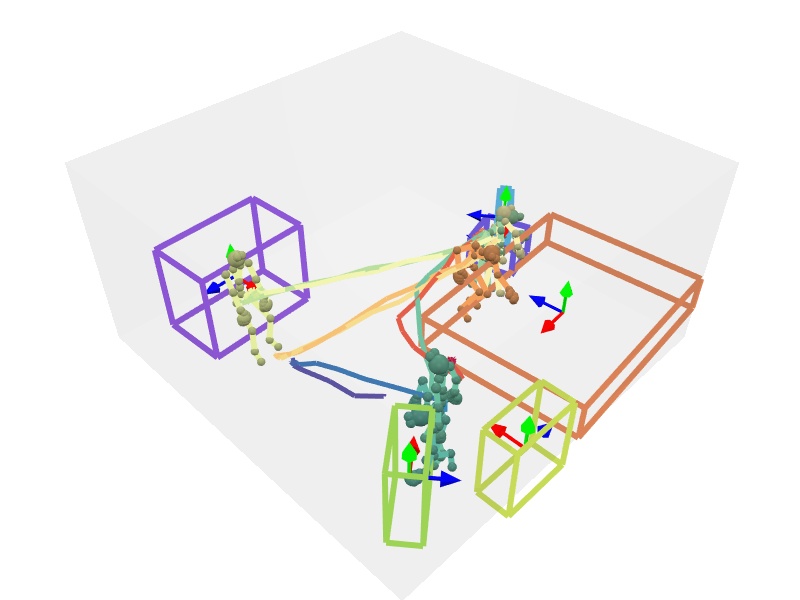}
		\includegraphics[width=\textwidth]
		{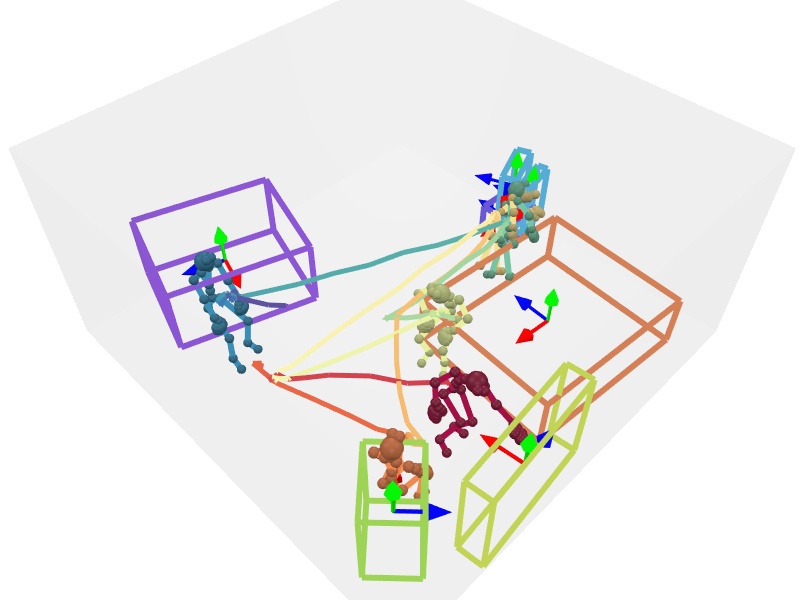}
		\includegraphics[width=\textwidth]
		{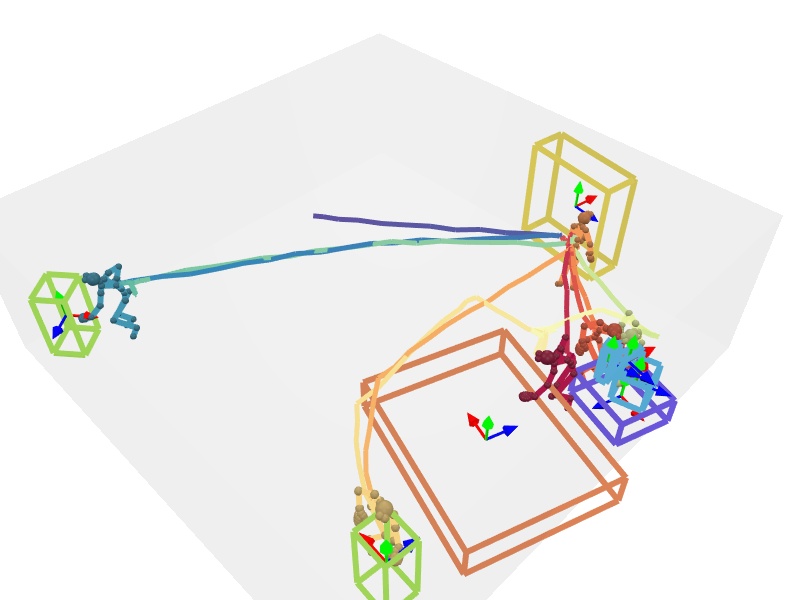}
		\includegraphics[width=\textwidth]
		{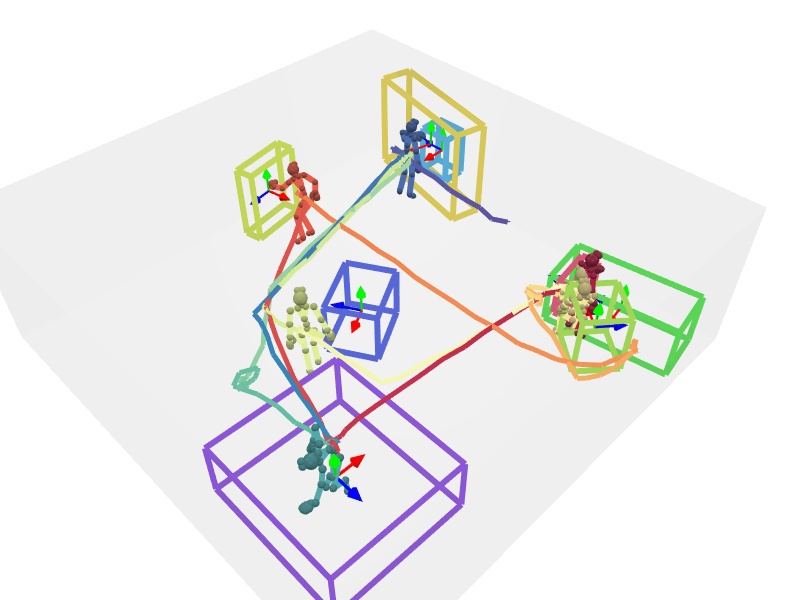}
		\caption{Sample 1}
	\end{subfigure}
	\begin{subfigure}[t]{0.16\textwidth}
		\includegraphics[width=\textwidth]
		{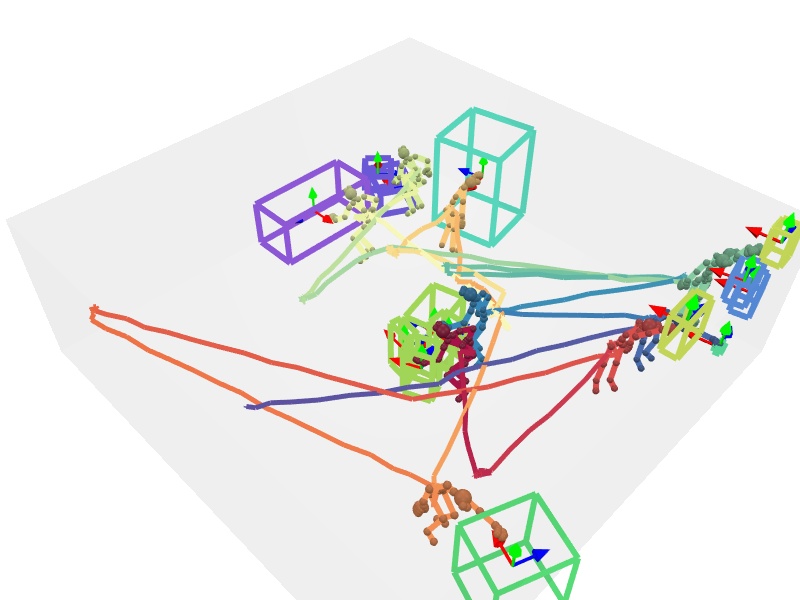}
		\includegraphics[width=\textwidth]
		{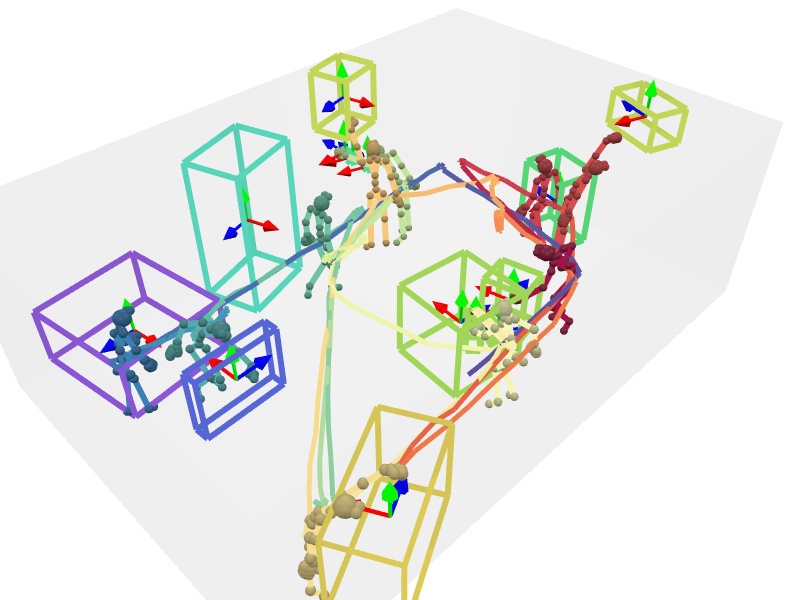}
		\includegraphics[width=\textwidth]
		{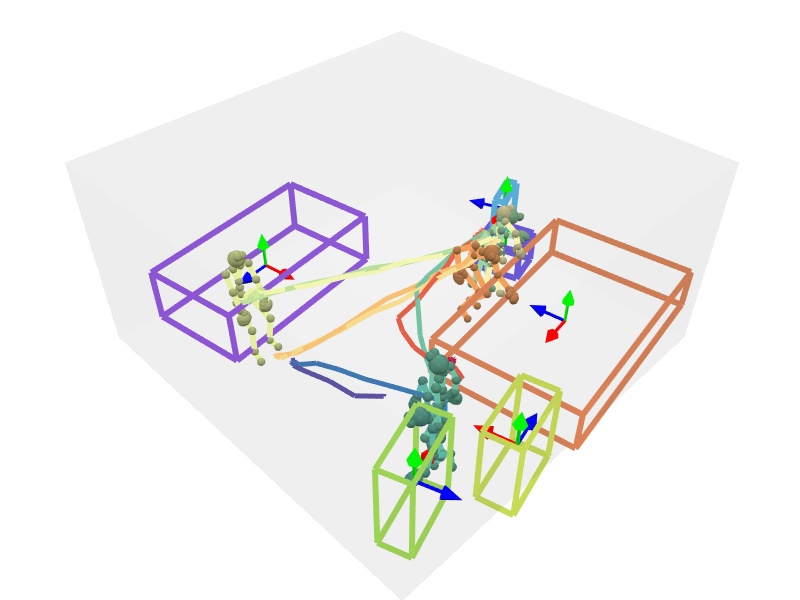}
		\includegraphics[width=\textwidth]
		{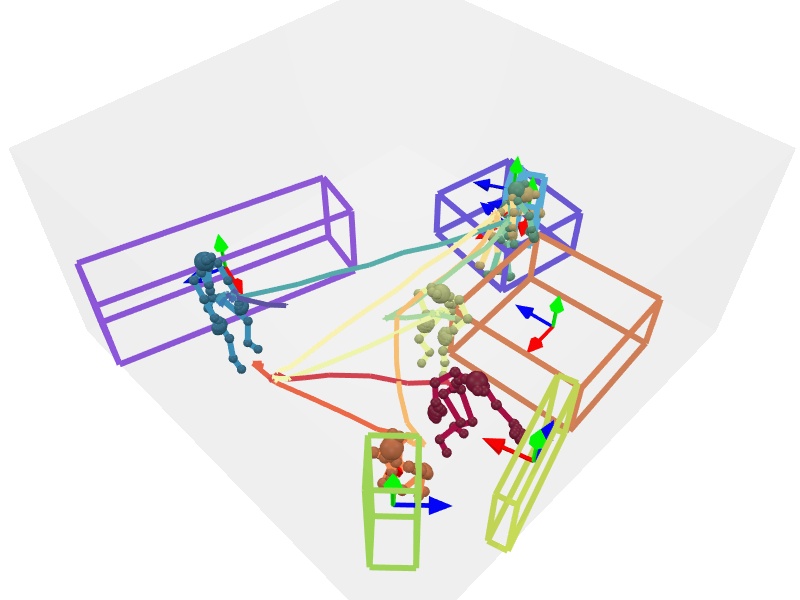}
		\includegraphics[width=\textwidth]
		{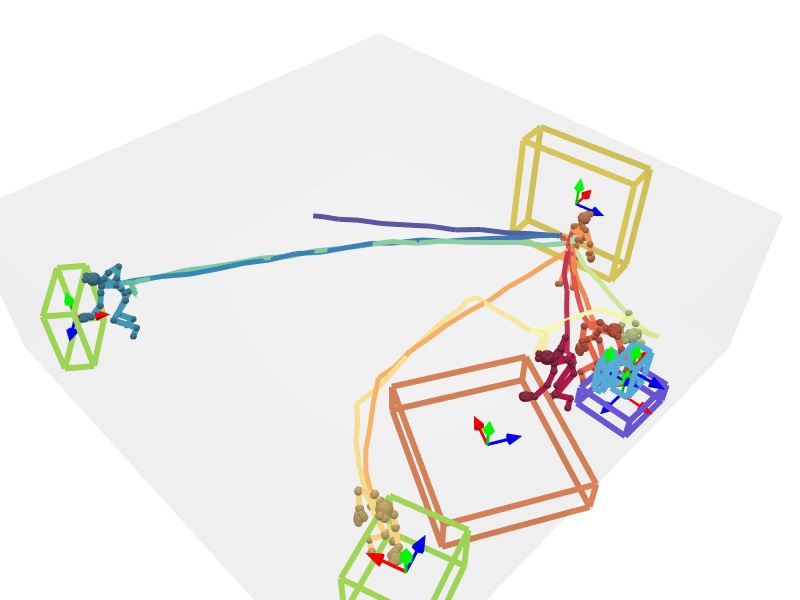}
		\includegraphics[width=\textwidth]
		{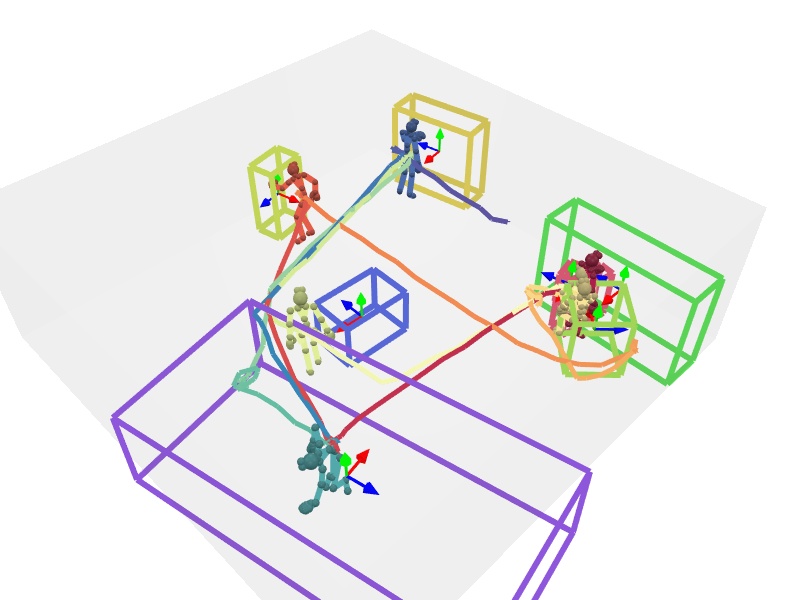}
		\caption{Sample 2}
	\end{subfigure}
	\begin{subfigure}[t]{0.16\textwidth}
		\includegraphics[width=\textwidth]
		{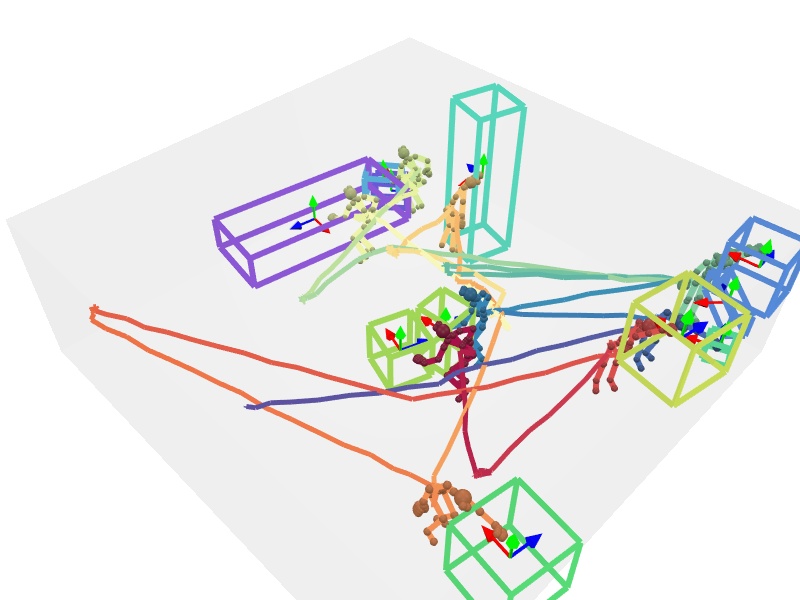}
		\includegraphics[width=\textwidth]
		{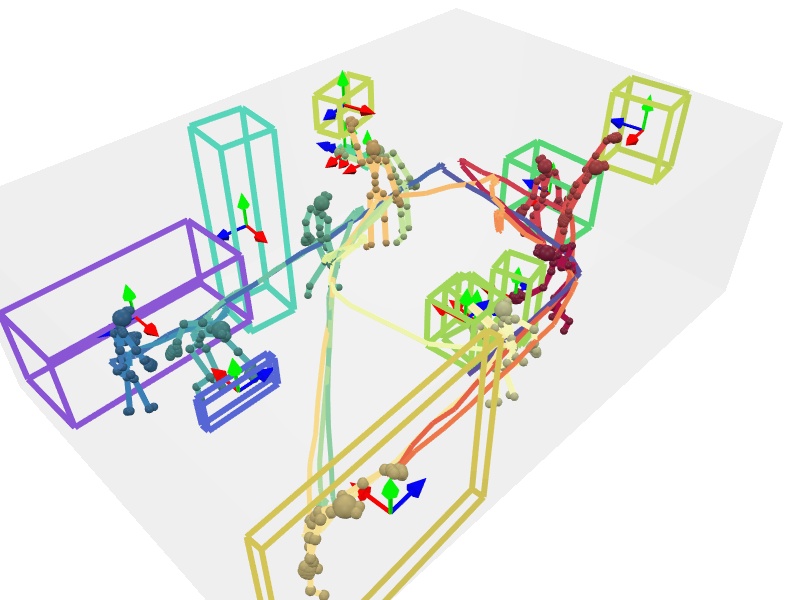}
		\includegraphics[width=\textwidth]
		{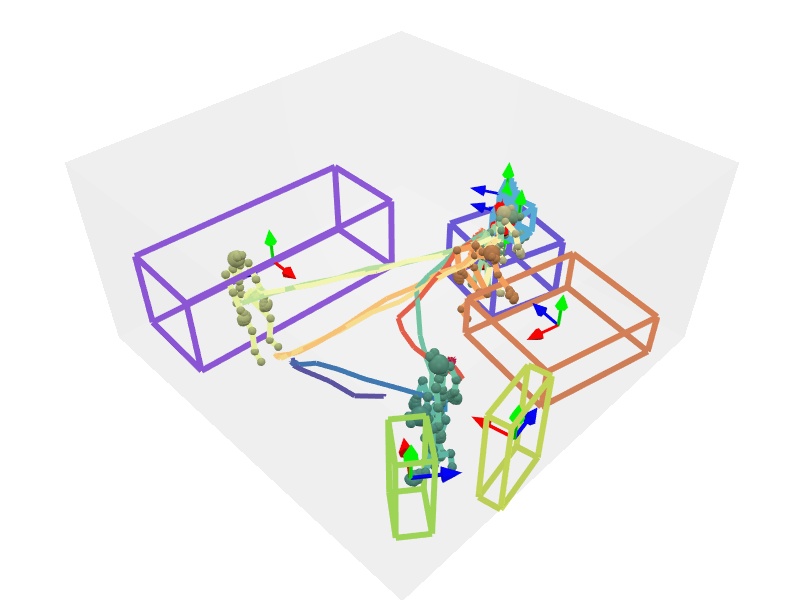}
		\includegraphics[width=\textwidth]
		{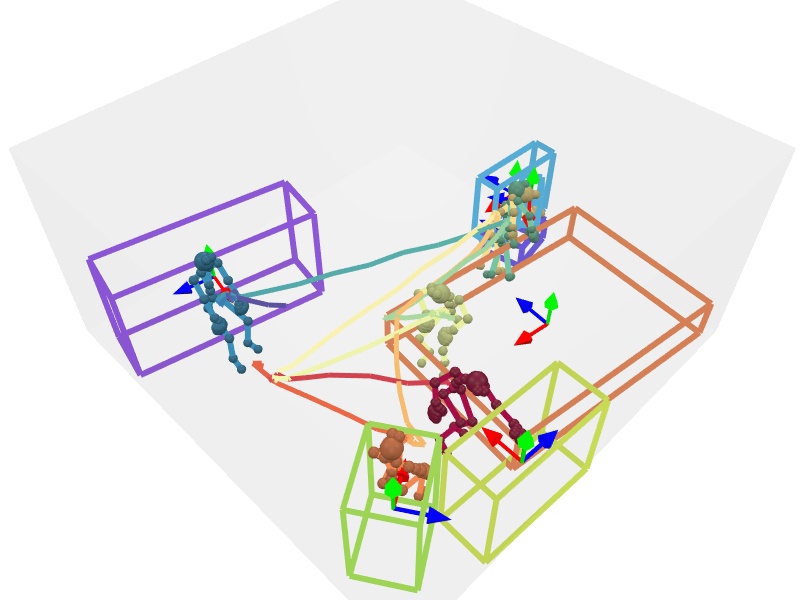}
		\includegraphics[width=\textwidth]
		{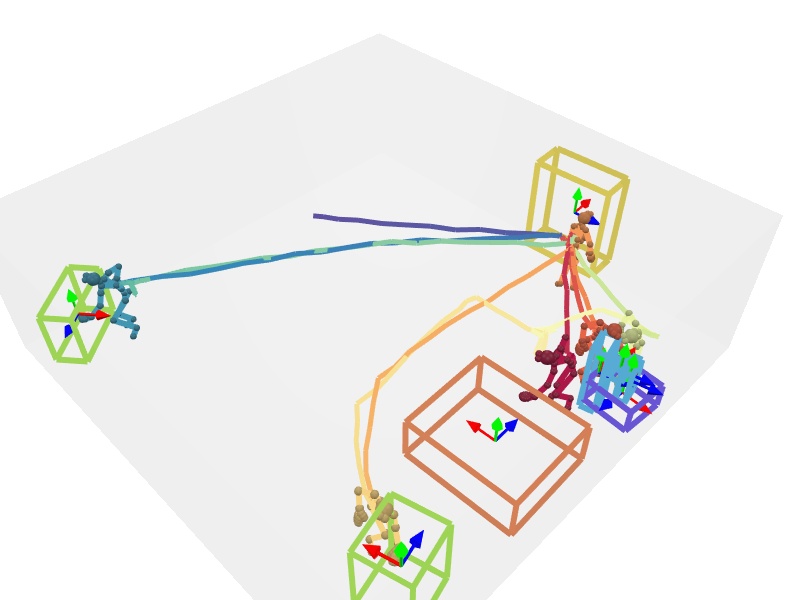}
		\includegraphics[width=\textwidth]
		{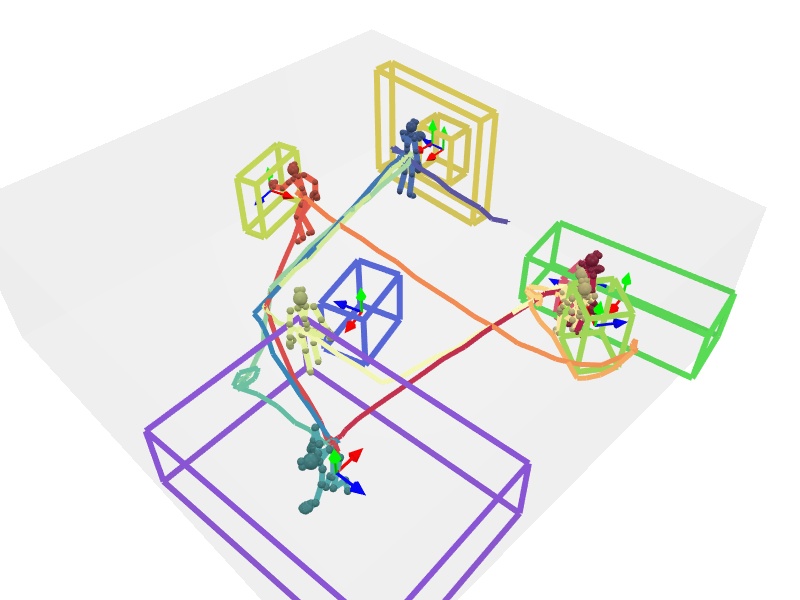}
		\caption{Sample 3}
	\end{subfigure}
	\begin{subfigure}[t]{0.16\textwidth}
		\includegraphics[width=\textwidth]
		{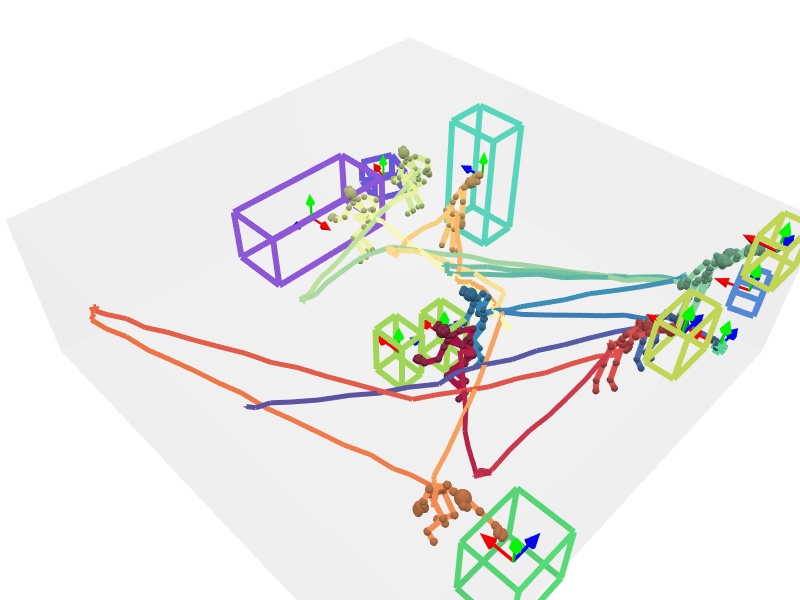}
		\includegraphics[width=\textwidth]
		{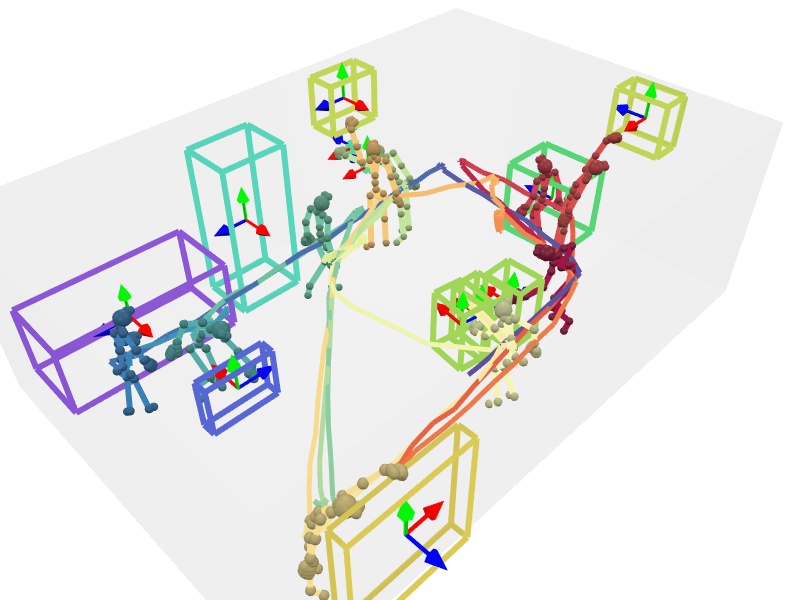}
		\includegraphics[width=\textwidth]
		{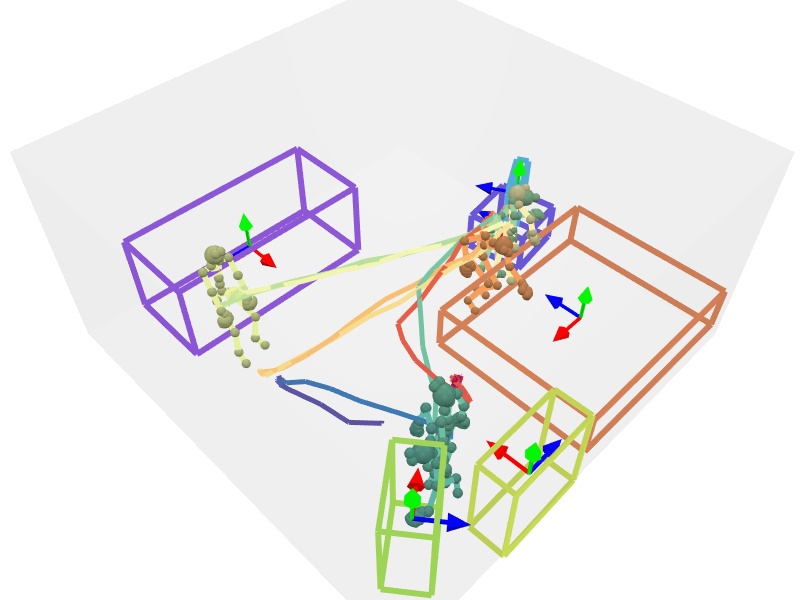}
		\includegraphics[width=\textwidth]
		{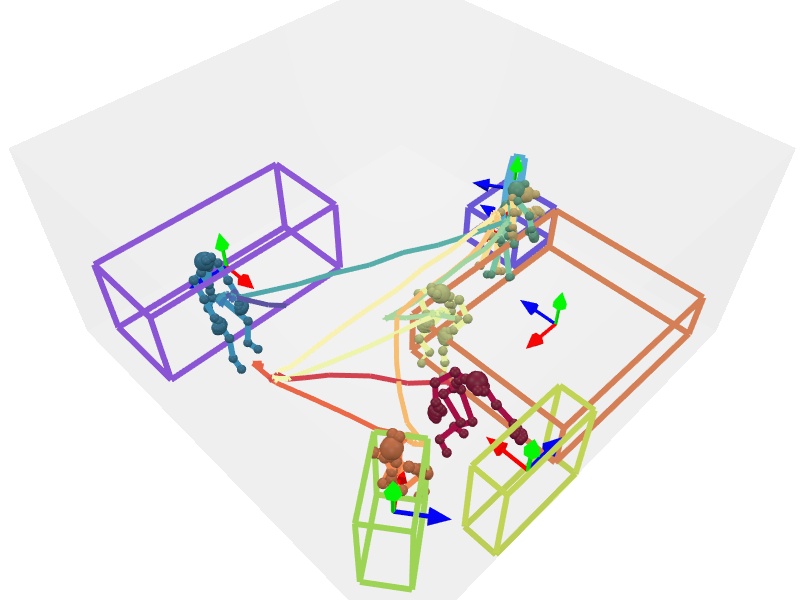}
		\includegraphics[width=\textwidth]
		{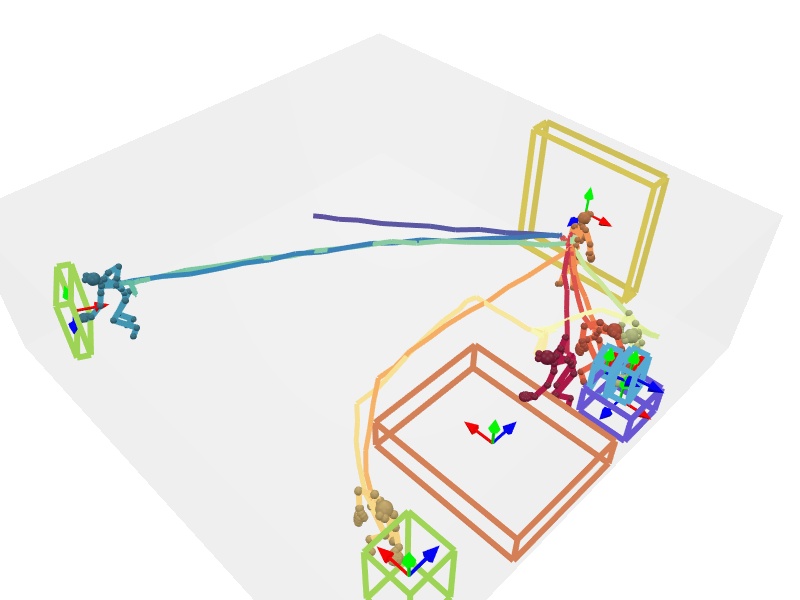}
		\includegraphics[width=\textwidth]
		{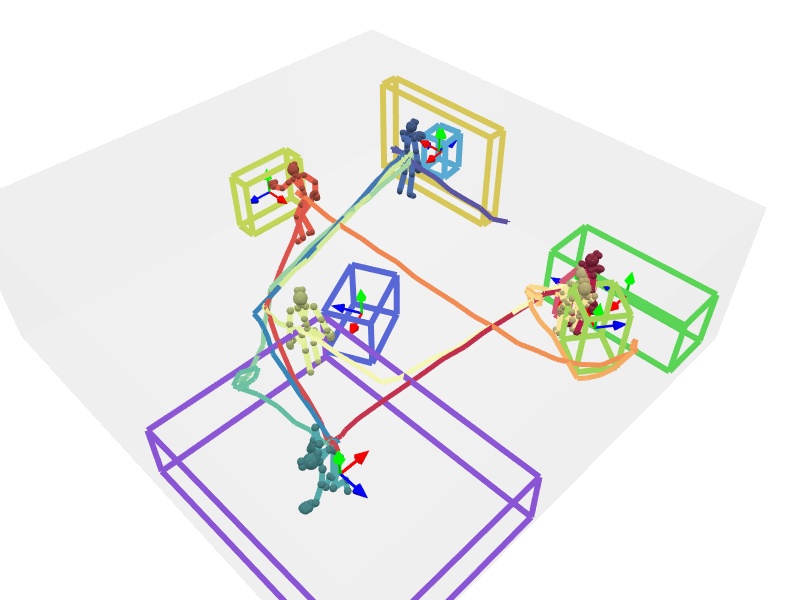}
		\caption{Max. likelihood prediction}
	\end{subfigure}
	\begin{subfigure}[t]{0.16\textwidth}
		\includegraphics[width=\textwidth]
		{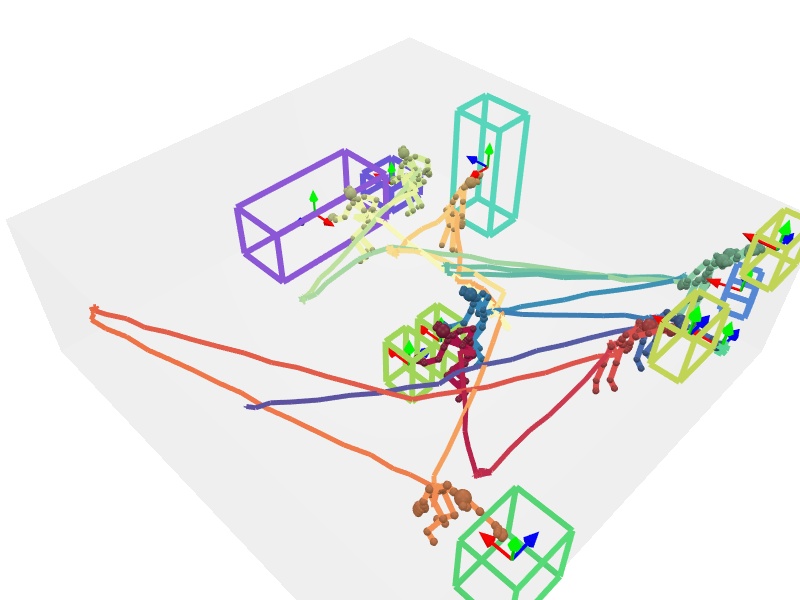}
		\includegraphics[width=\textwidth]
		{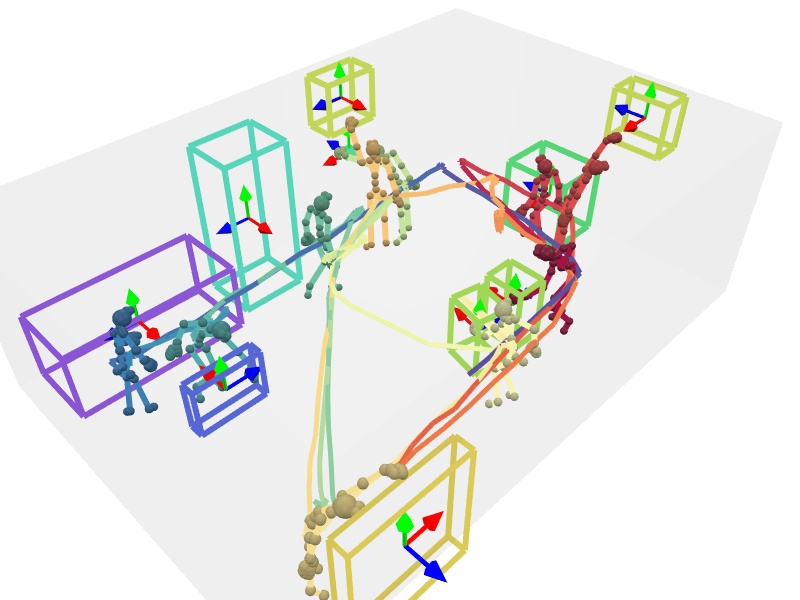}
		\includegraphics[width=\textwidth]
		{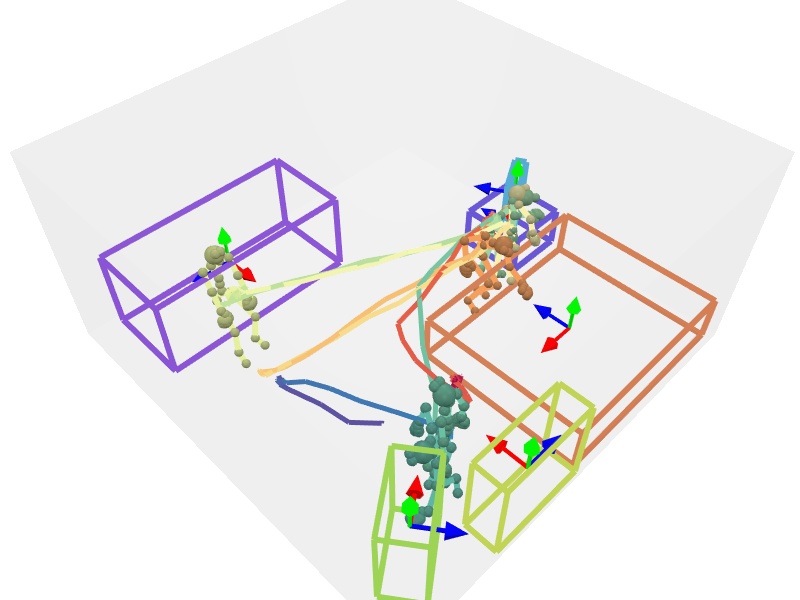}
		\includegraphics[width=\textwidth]
		{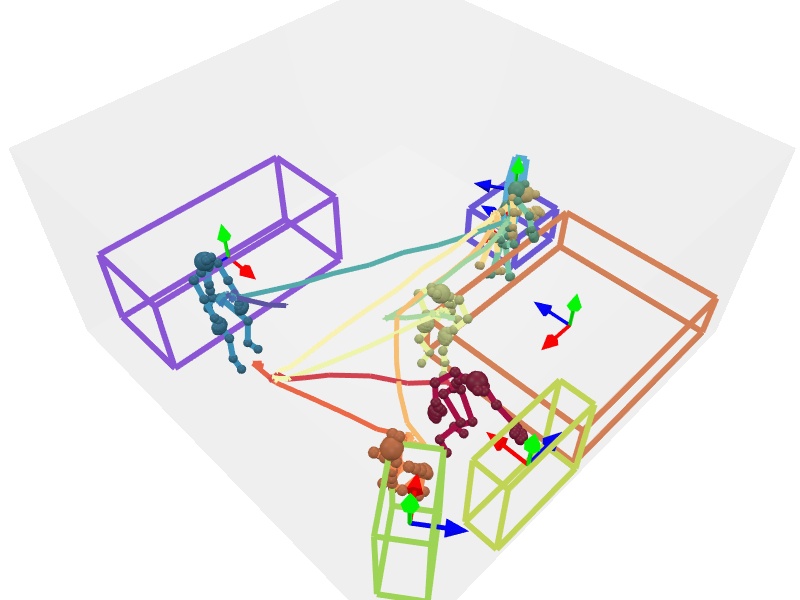}
		\includegraphics[width=\textwidth]
		{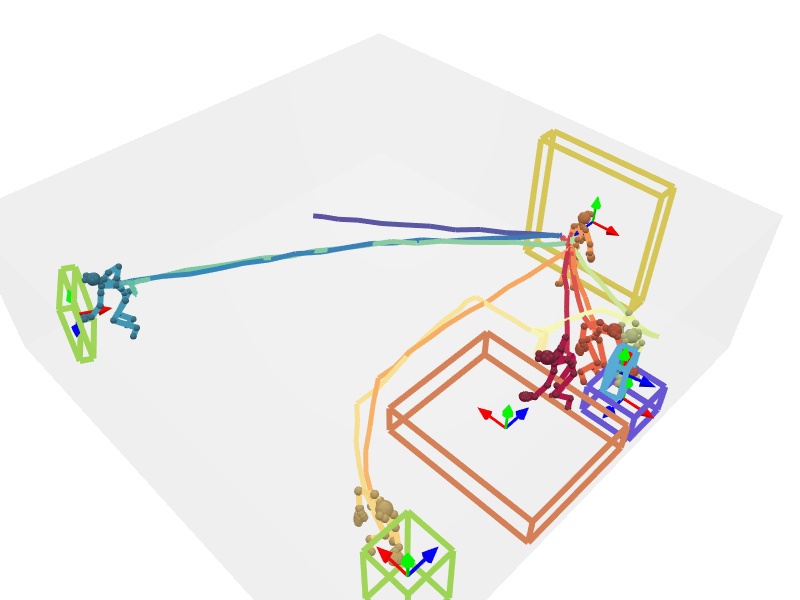}
		\includegraphics[width=\textwidth]
		{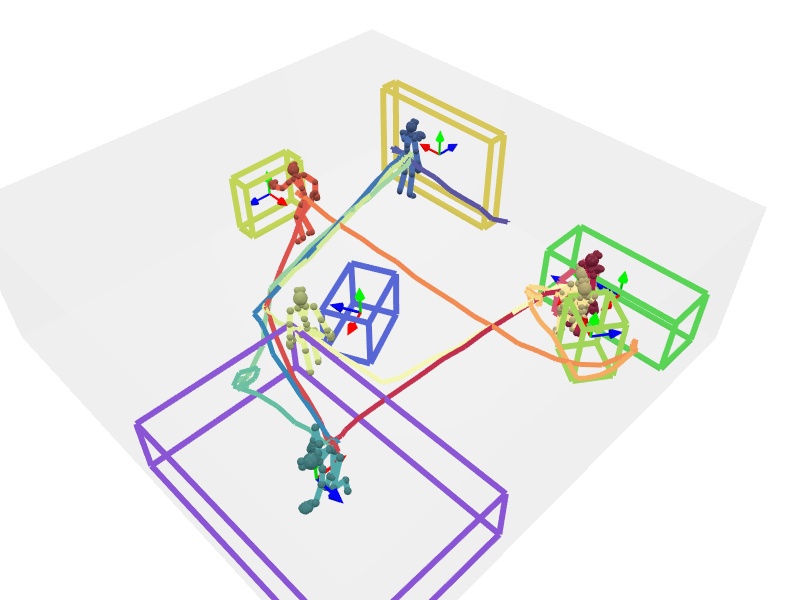}
		\caption{GT}
	\end{subfigure}
	\caption{Additional multi-modal predictions of \OURS{}. By sampling our probabilistic decoder multiple times, we can obtain various different plausible box predictions.}
	\label{fig:multi_modes_s1}
\end{figure*}

We show additional qualitative results on test splits $\mathcal{S}_{1}$ and $\mathcal{S}_{2}$ in Figures~\ref{fig:more_on_s1_1}-\ref{fig:more_on_s1_3} and \ref{fig:more_on_s2}, respectively. We additionally visualize various multi-modal predictions from our model on $\mathcal{S}_{1}$ in Figure~\ref{fig:multi_modes_s1}.

\section{Tolerance to noise}
\label{sec:tol_noise}
\begin{figure}[!ht]
	\centering
	\begin{subfigure}[t]{0.23\textwidth}
		\includegraphics[width=\textwidth]
		{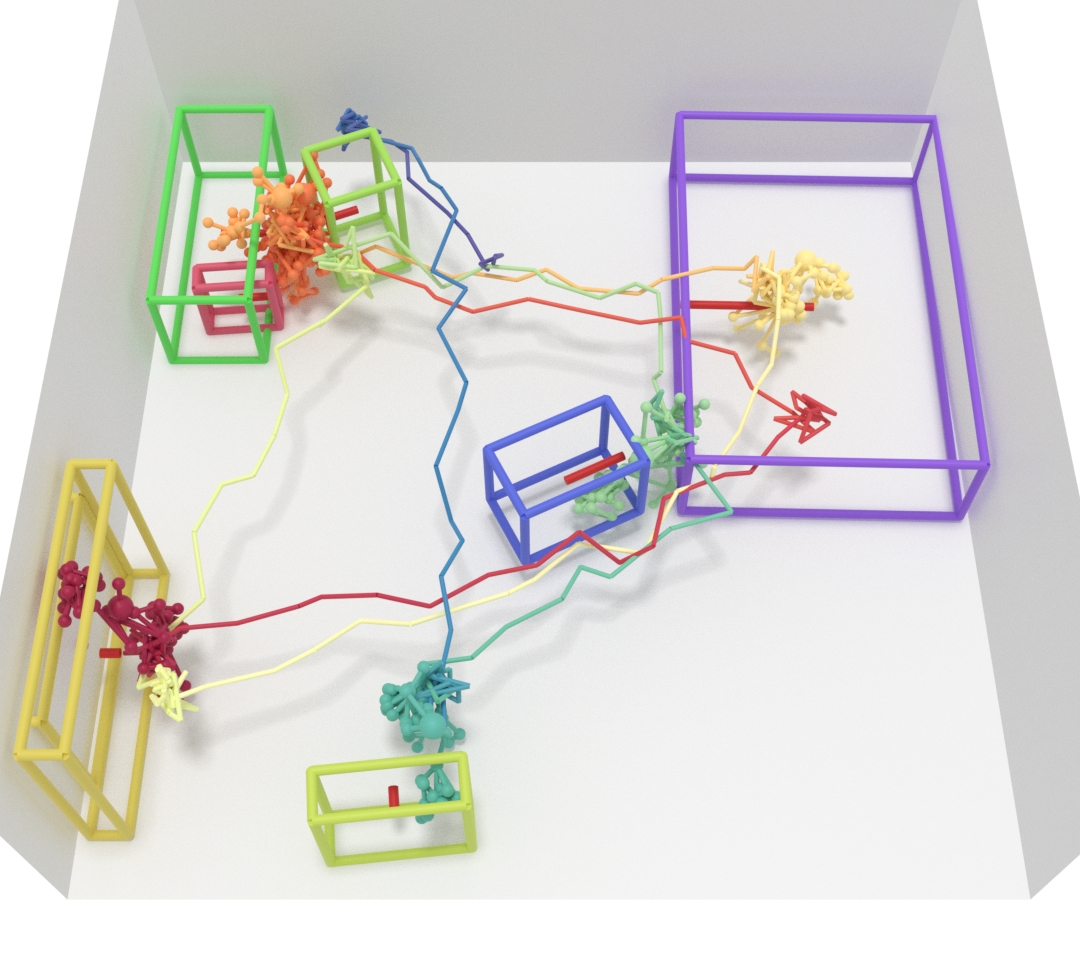}
		\caption{Prediction}
	\end{subfigure}
	\begin{subfigure}[t]{0.23\textwidth}
		\includegraphics[width=\textwidth]
		{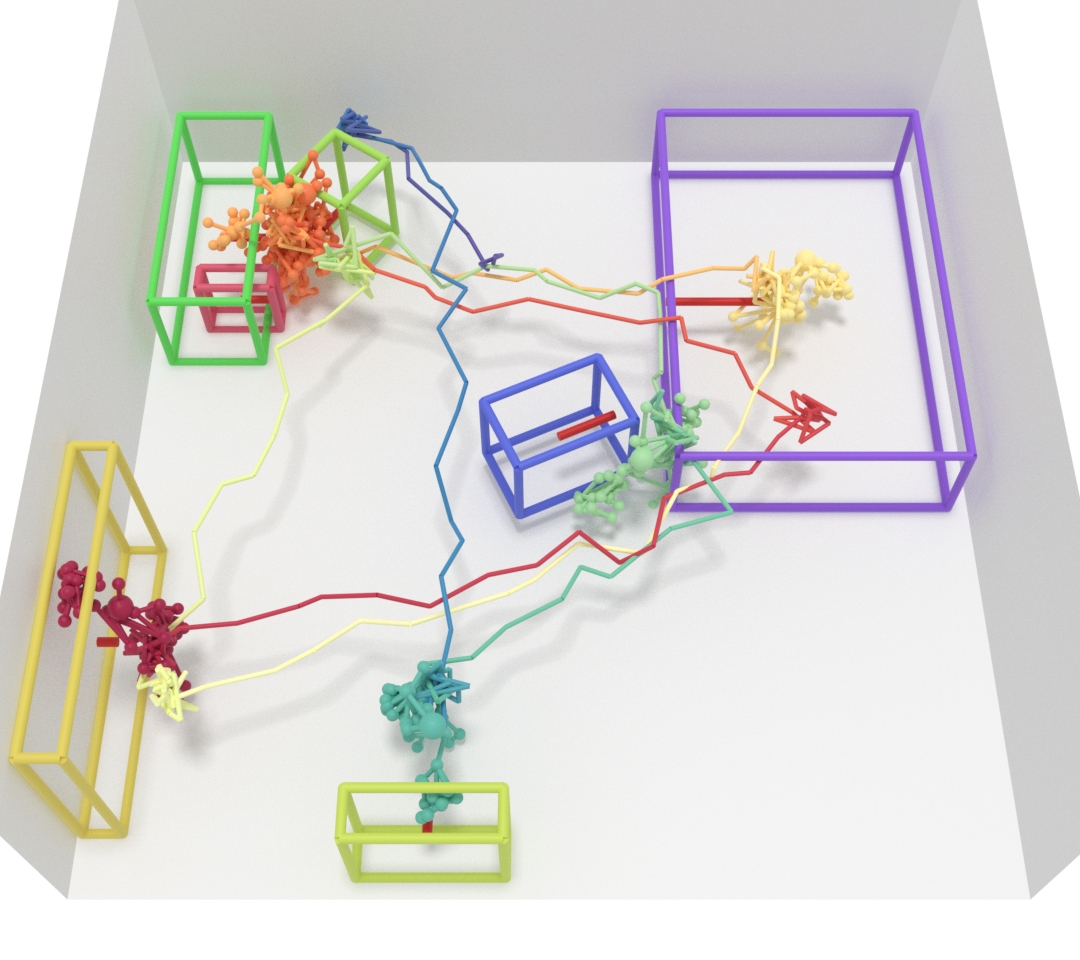}
		\caption{GT}
	\end{subfigure}
	\rulesep
	\begin{subfigure}[t]{0.23\textwidth}
		\includegraphics[width=\textwidth]
		{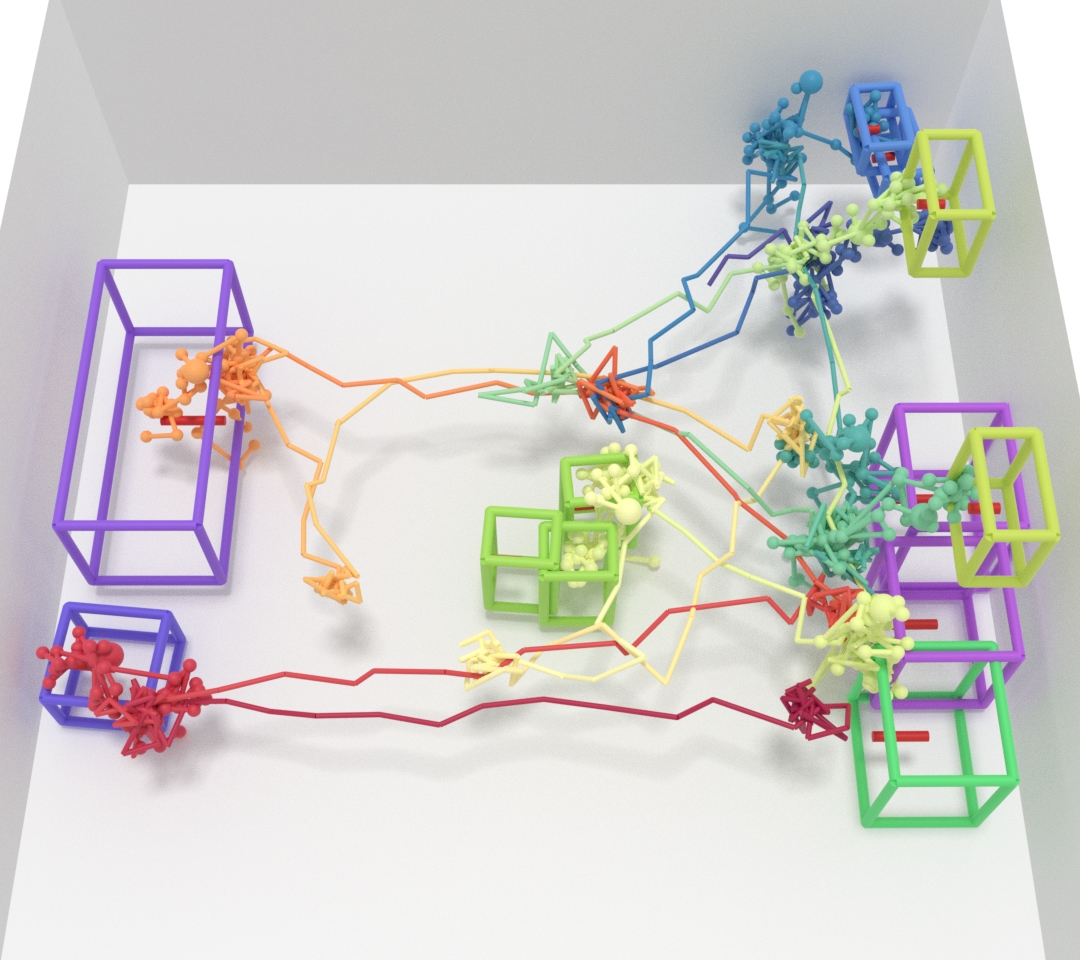}
		\caption{Prediction}
	\end{subfigure}
	\begin{subfigure}[t]{0.23\textwidth}
		\includegraphics[width=\textwidth]
		{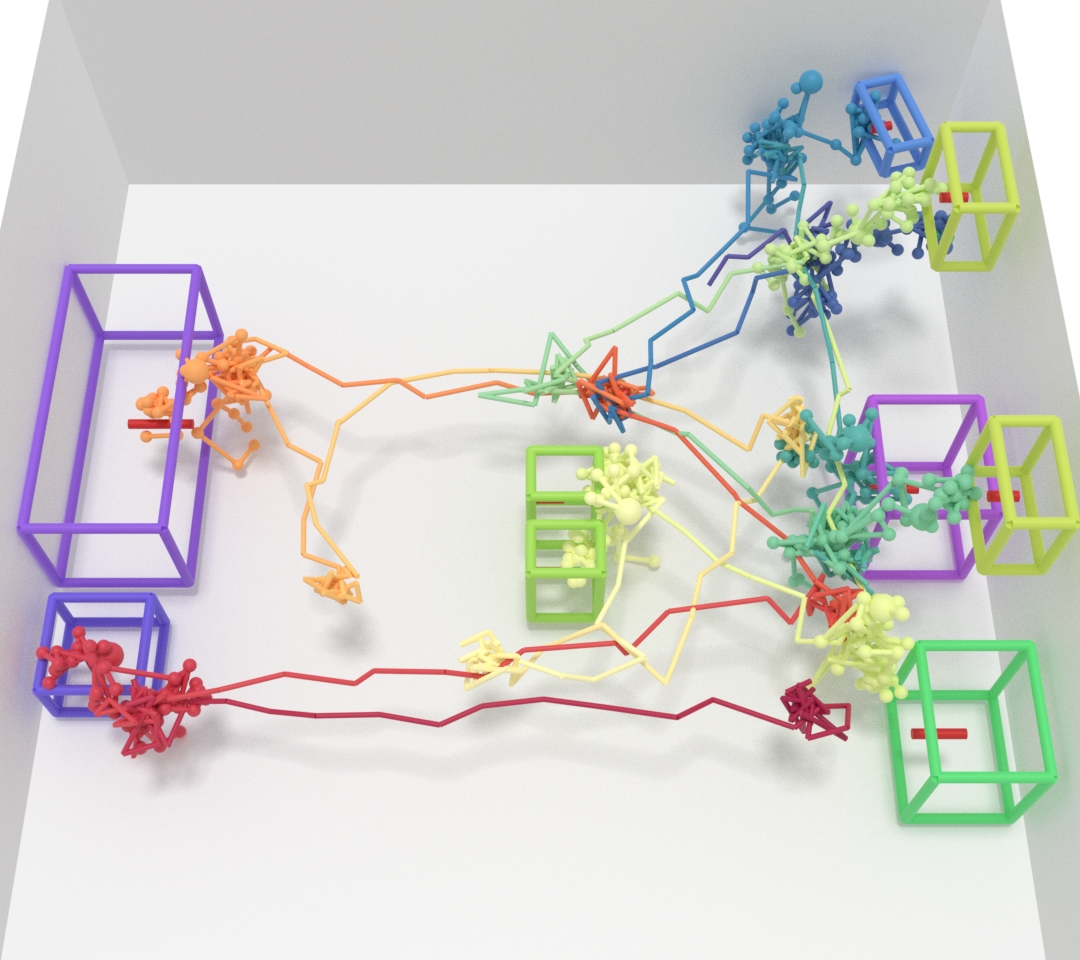}
		\caption{GT}
	\end{subfigure}
	\vspace{-0.2cm}
	\caption{Predictions on noisy inputs (std=10 cm).}
	\label{fig:quali_comp_noises}
\end{figure}

As real data often contain noise, we additionally study the effect of Gaussian noise (std=5 cm) onto the xyz coordinates of all joints in training and testing with our dataset.
Table~\ref{compare:tolerance_to_noise} shows the effect of different noise levels in evaluation. We also visualize some sampled predictions under the noise level at 10$\sigma$ in Figure~\ref{fig:quali_comp_noises}, where our method presents compelling tolerance for very noisy inputs.

\begin{table}[!h]
	\centering
		\begin{tabular}{|c|c|c|c|c|c|}
		\hline
		Noise level & $\sigma$ & $2\sigma$ & $3\sigma$ & $5\sigma$ & $10\sigma$ \\
		\hline
		\hline
		$\mathcal{S}_{1}$ & 38.58 & 38.77 & 38.36 & 37.16 & 31.71 \\
		$\mathcal{S}_{2}$ & 27.16 & 26.13 & 27.72 & 29.18 & 26.18\\
		\hline
	\end{tabular}
	\caption{mAP@0.5 under varying levels of noise ($\sigma$=1 cm).}
	\label{compare:tolerance_to_noise}
\end{table}

\section{Qualitative Results on Other Real Data}
\label{sec:realdata}
\begin{figure*}[!ht]
	\centering
	\begin{subfigure}[t]{0.3\textwidth}
		\includegraphics[width=\textwidth]
		{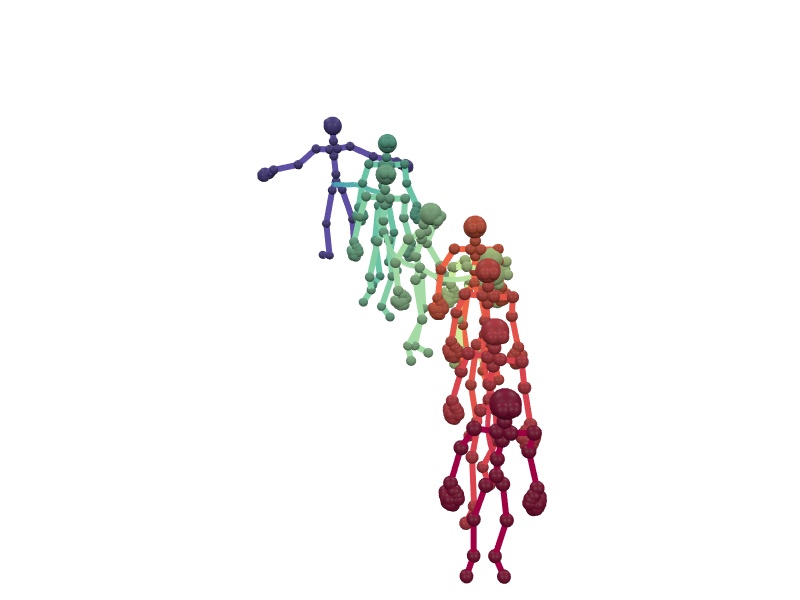}
		\includegraphics[width=\textwidth]
		{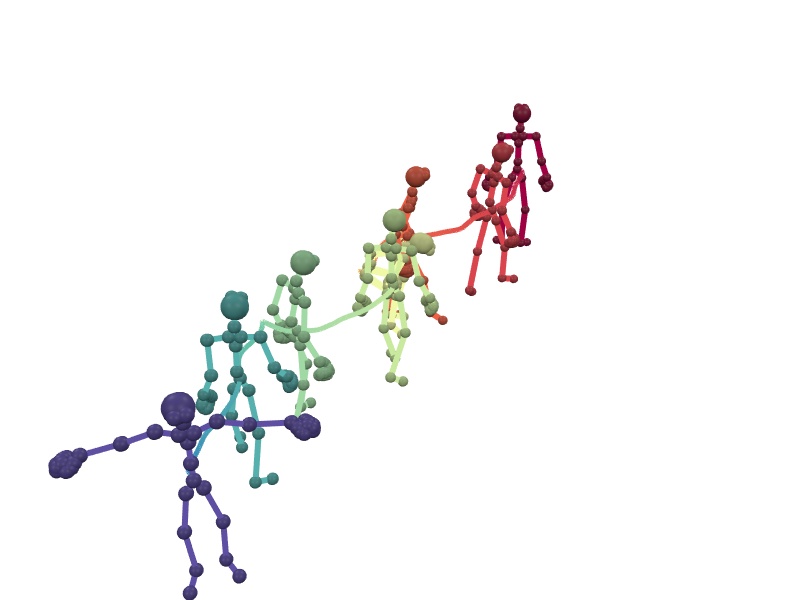}
		\includegraphics[width=\textwidth]
		{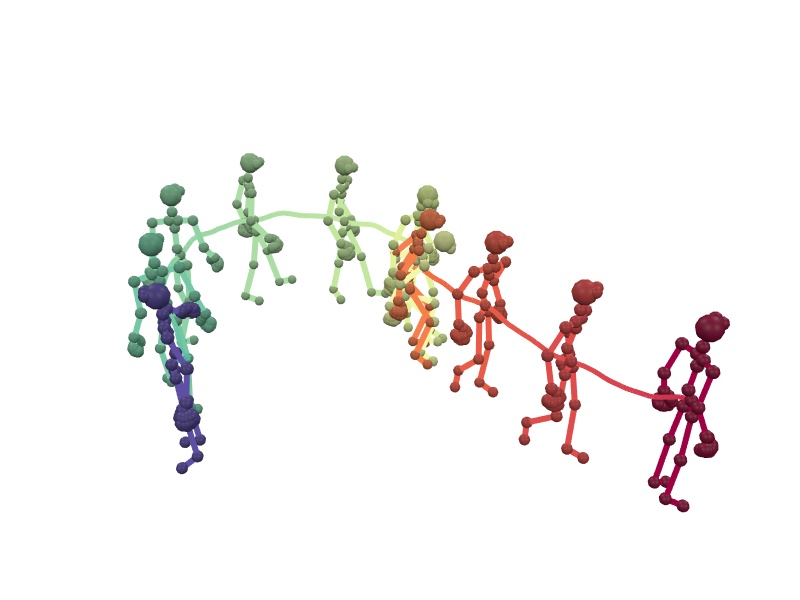}
		\includegraphics[width=\textwidth]
		{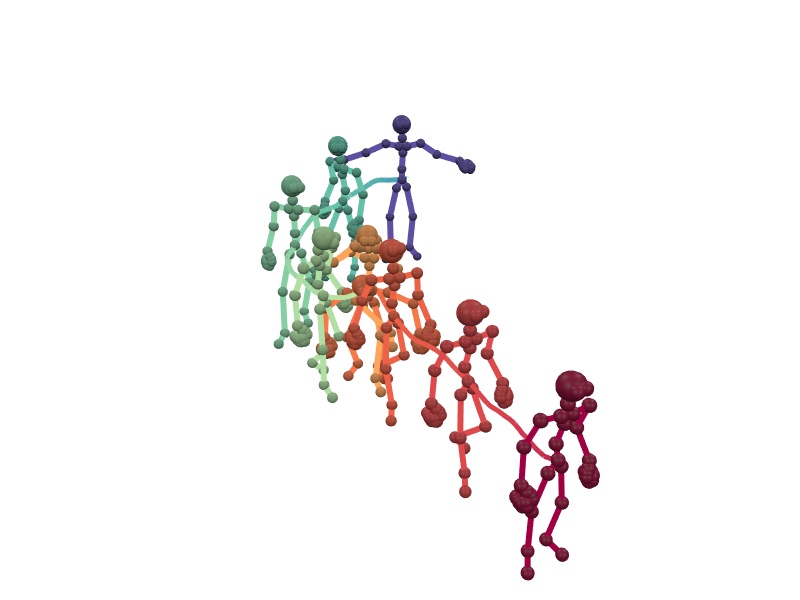}
		\includegraphics[width=\textwidth]
		{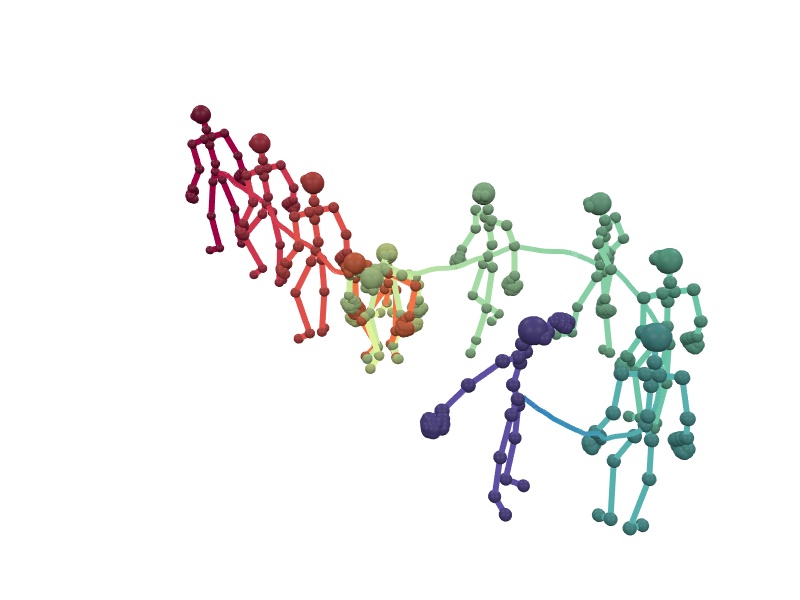}
		\includegraphics[width=\textwidth]
		{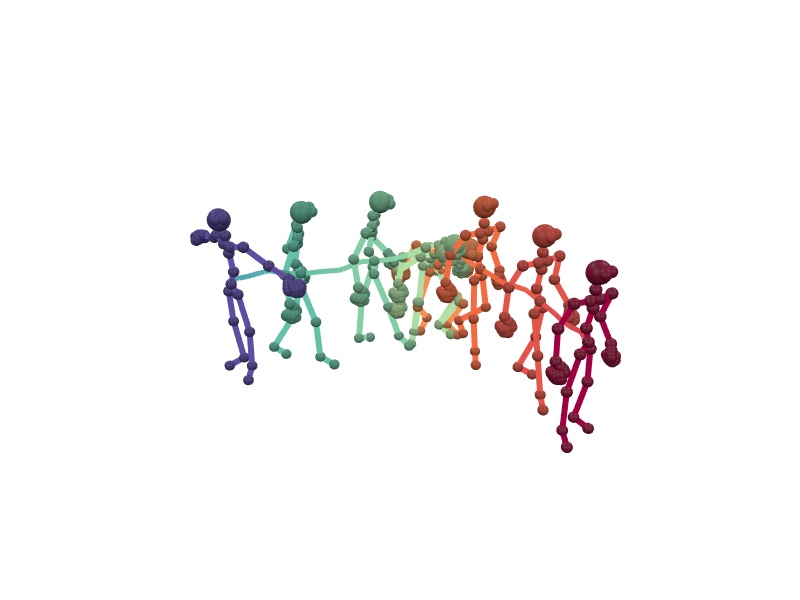}
		\caption{Input}
	\end{subfigure}
	\begin{subfigure}[t]{0.3\textwidth}
		\includegraphics[width=\textwidth]
		{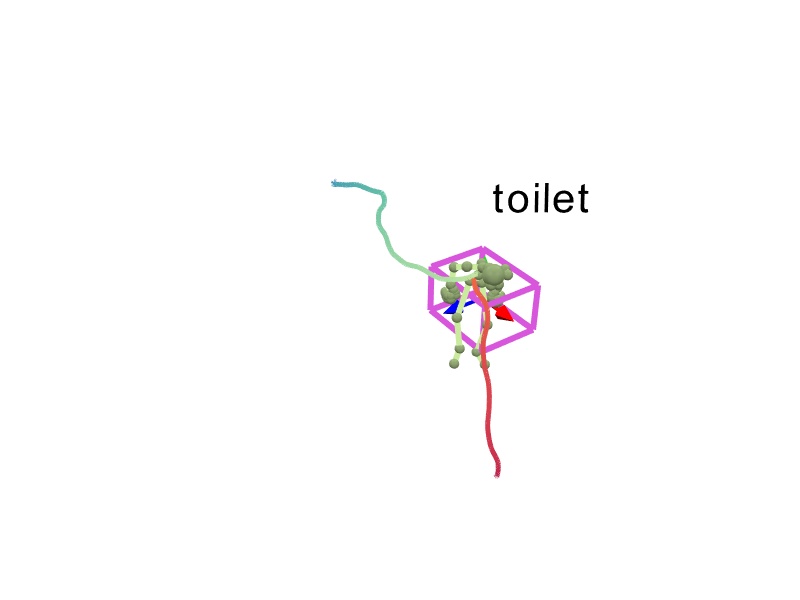}
		\includegraphics[width=\textwidth]
		{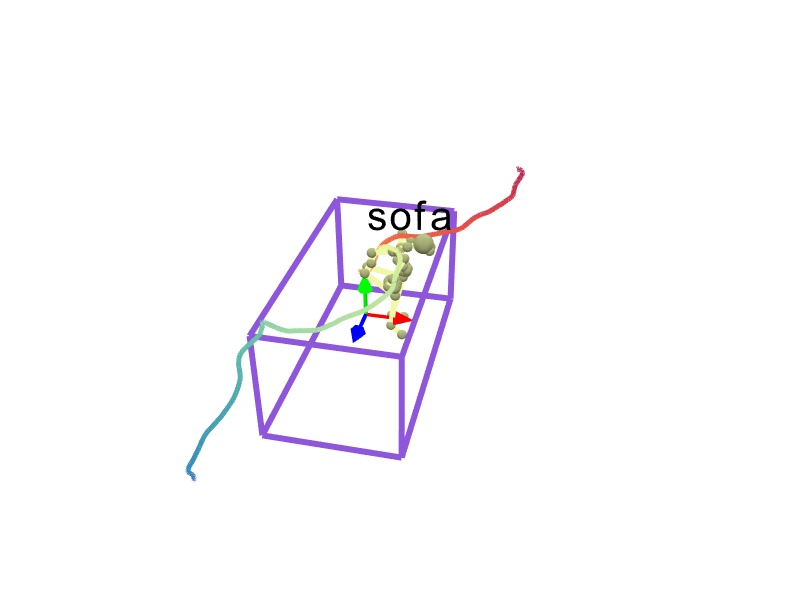}
		\includegraphics[width=\textwidth]
		{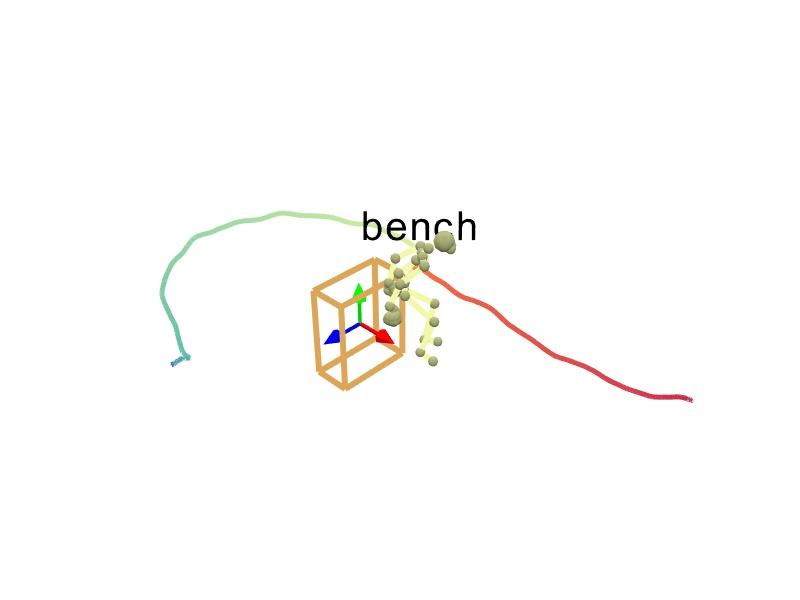}
		\includegraphics[width=\textwidth]
		{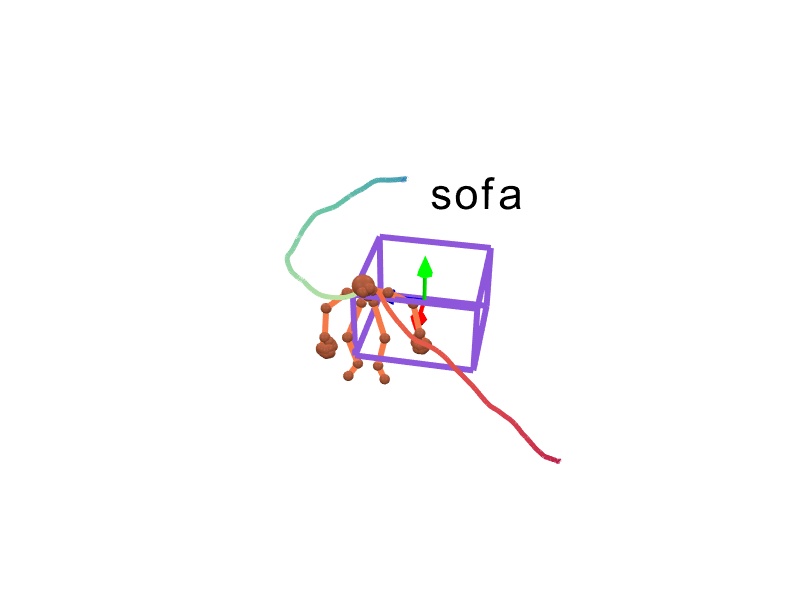}
		\includegraphics[width=\textwidth]
		{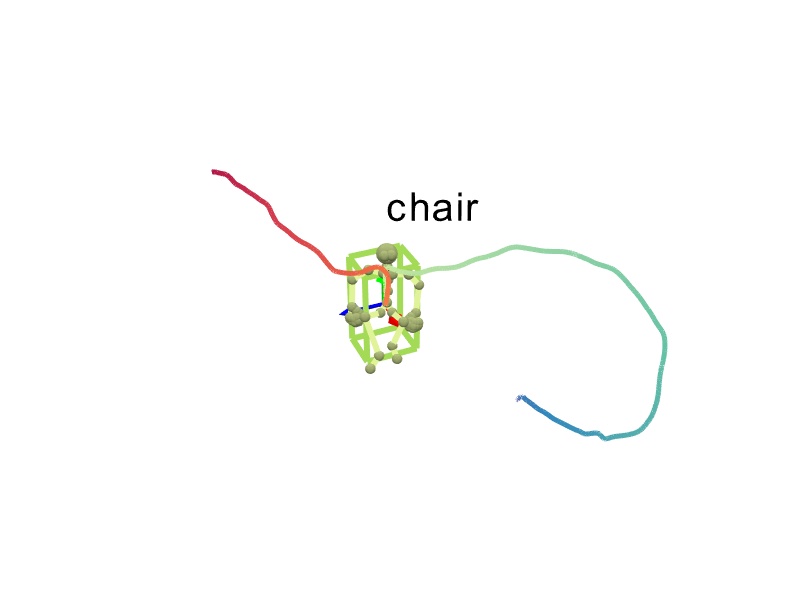}
		\includegraphics[width=\textwidth]
		{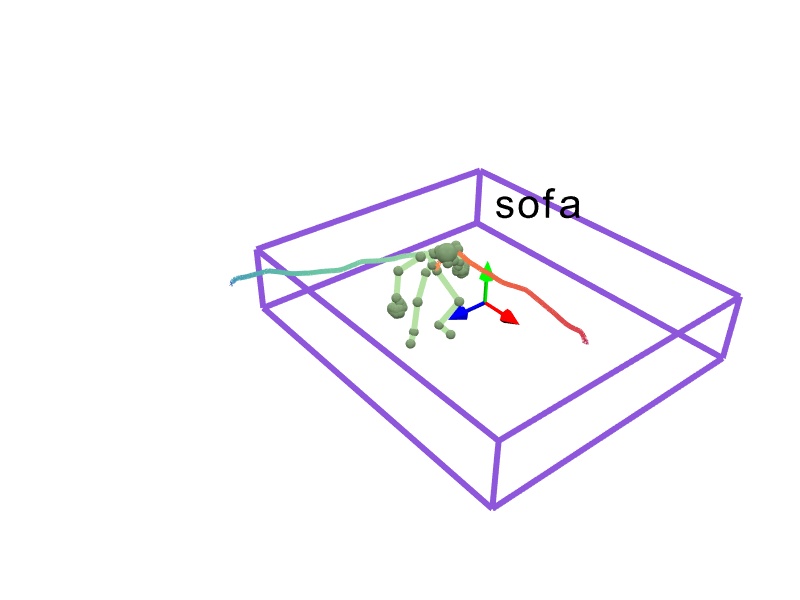}
		\caption{Multi-modal sample}
	\end{subfigure}
	\begin{subfigure}[t]{0.3\textwidth}
		\includegraphics[width=\textwidth]
		{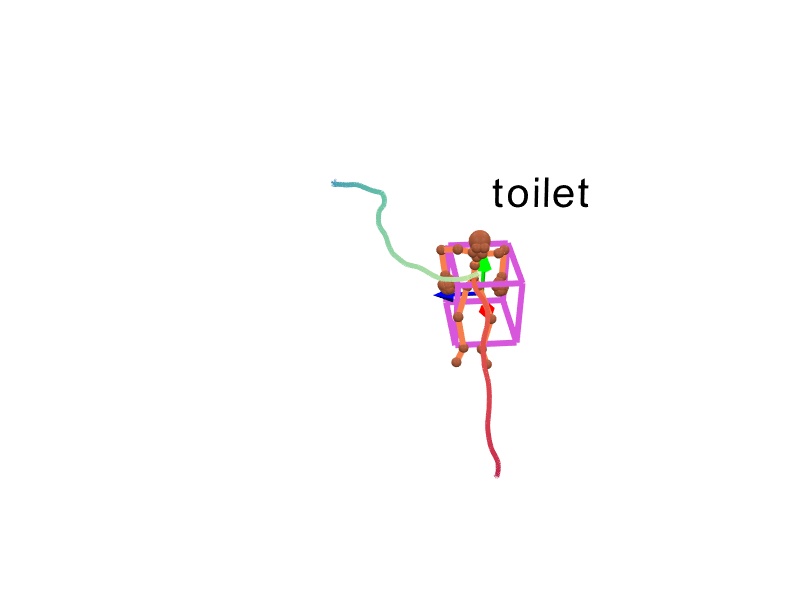}
		\includegraphics[width=\textwidth]
		{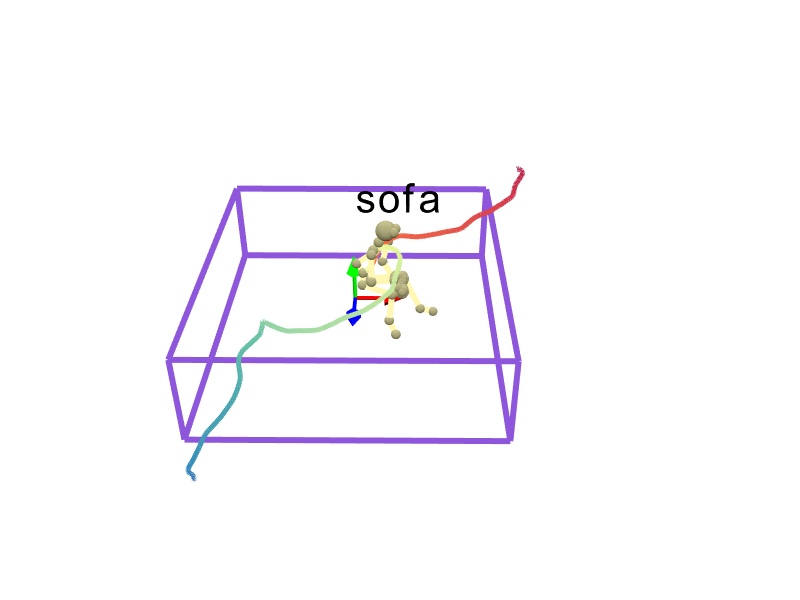}
		\includegraphics[width=\textwidth]
		{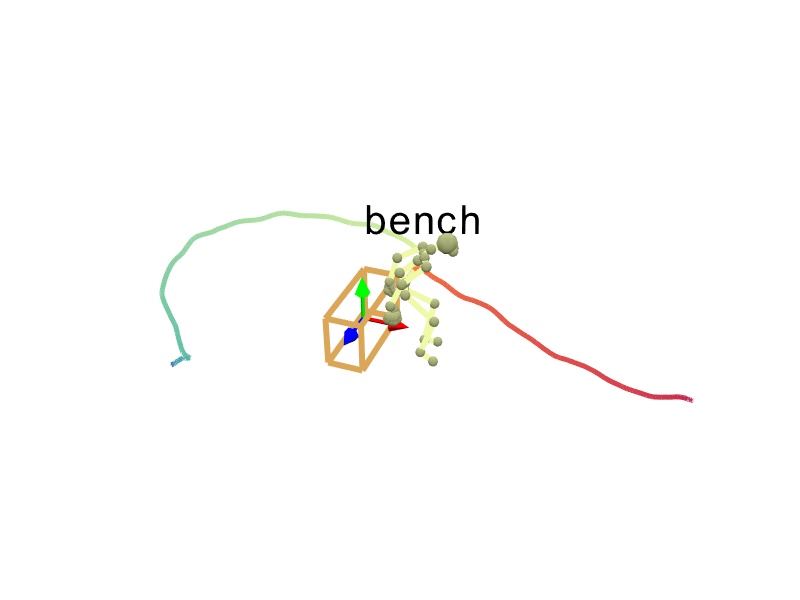}
		\includegraphics[width=\textwidth]
		{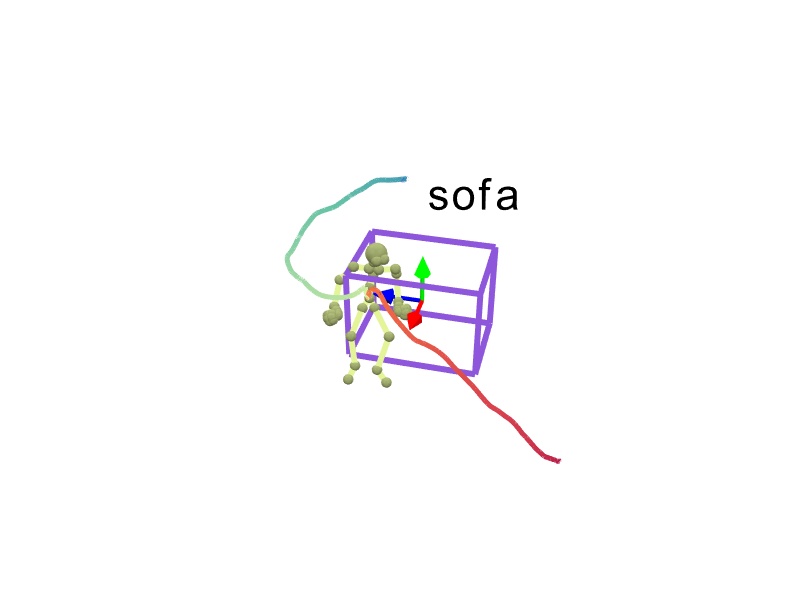}
		\includegraphics[width=\textwidth]
		{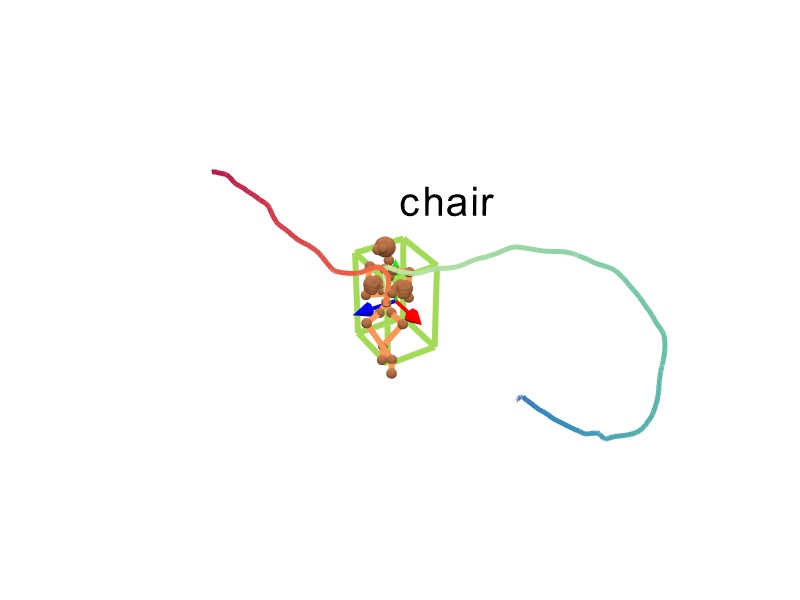}
		\includegraphics[width=\textwidth]
		{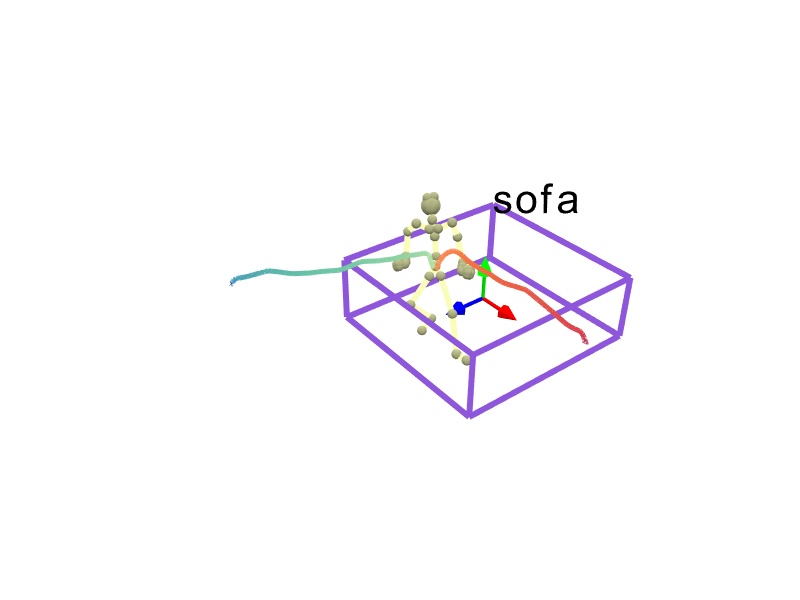}
		\caption{Max. likelihood prediction}
	\end{subfigure}
	\caption{Multi-modal predictions on the real human pose trajectory input of \cite{hassan2021stochastic}.}
	\label{fig:results_on_real_data}
\end{figure*}

Besides the experiments on VirtualHome~\cite{puig2018virtualhome} and PROX~\cite{hassan2019resolving,yi2022mover}, we additionally qualitatively evaluate \OURS{} by training it on our dataset, and apply it to the real-world human pose trajectory data provided by \cite{hassan2021stochastic} describing human interactions with single objects. The qualitative results are illustrated in Figure~\ref{fig:results_on_real_data}, where we see that our method still can provide plausible object explanations from natural and diverse real human poses.

\end{document}